\documentclass
{thesis}

\usepackage[T1]{fontenc} 
\usepackage{qtree}       
\qtreecenterfalse        
\usepackage{rotating}    
\usepackage{amssymb}     
\usepackage{epsfig}      
\usepackage{verbatim}    

\newcommand{\majority}{\mathop{\rm majority}}
\newcommand{\argmax}[1]{\mathop{\rm argmax}\limits_{#1}}
\newcommand{\argmin}[1]{\mathop{\rm argmin}\limits_{#1}}

\newtheorem{lemma}{Lemma}[chapter]

\newenvironment{proof}
{\noindent{\bf Proof:}\hspace*{1em}}
{\hfill$\blacksquare$\medskip}

\newcounter{parse}[chapter]
\renewcommand{\theparse}{\thechapter.\arabic{parse}}
\newenvironment{parse}
{\bigskip\noindent\refstepcounter{parse}\hfill}
{\hfill(\theparse)\bigskip}

\newcounter{algorithm}[chapter]
\renewcommand{\thealgorithm}{\thechapter.\arabic{algorithm}}
\newenvironment{algorithm}[2]
{
\noindent
\begin{minipage}{\linewidth}
\refstepcounter{algorithm}
\vspace{1em}\noindent
{\bf Algorithm \thealgorithm: #1 \hrulefill}

\noindent#2
\begin{enumerate}}
{
\end{enumerate}\hrulefill\vspace{1em}
\end{minipage}
}

\renewcommand{\Re}{\mathbb{R}} 

\begin{document}
\title{EXPLOITING DIVERSITY \\
FOR NATURAL LANGUAGE PARSING}

\author{John Charles Henderson}

\doctorphilosophy

\dissertation

\copyrightnotice

\degreeyear{1999}
\degreemonth{August}

\advisor{Eric Brill}
\readers{Steven Salzberg \\
  David Yarowsky}

\maketitle

\begin{frontmatter}

\begin{abstract}

Accurate linguistic annotation is a core requirement of natural
language processing systems.  The demand for accuracy in the
face of rapid prototyping constraints and numerous target languages
has led to the employment of machine learning methods for developing
linguistic annotation systems.

The popularity of applying machine learning methods to computational
linguistics problems has given rise to a large supply of trainable
natural language processing systems.  Most problems of interest have
an array of off-the-shelf products or downloadable code implementing
solutions using various techniques.  In situations where these
solutions are developed independently, it is observed that their
errors tend to be independently distributed.  In this thesis we
discuss approaches for capitalizing on this situation in a sample
problem domain, Penn Treebank-style parsing.

The machine learning community provides us with techniques for
combining outputs of classifiers, but parser output is more structured
and interdependent than classifications.  To overcome this, two novel
strategies for combining parsers are used: learning to control a
switch between parsers and constructing a hybrid parse from multiple
parsers' outputs.  In this thesis we give supervised and unsupervised
techniques for each of these strategies as well as performance and
robustness results from evaluation of the techniques.

One shortcoming of combining off-the-shelf parsers is that the parsers
are not developed with the intention to perform well on complementary
data or to compensate for each others' weaknesses.  The individual
parsers are globally optimized.  We present two techniques for
producing an ensemble of parsers in such a way that their outputs can
be constructively combined.  All of the ensemble members will be
created using the same underlying parser induction algorithm, and the
method for producing complementary parsers is only loosely coupled to
that algorithm.

\end{abstract}


\begin{dedication}\centering\Large\vspace*{\fill}
Dedicated to
\\
good parents,
\\
{\bf Kathleen} and {\bf Daniel}
\vspace*{\fill}\end{dedication}

\begin{acknowledgement}

  
  No one flourishes in isolation and I am another example to support
  the claim.  There are many people I should thank, and tracing back
  all of the paths of inspiration, motivation, and support that
  facilitate a thesis is impossible.  Here is an overview of the major
  supporting members of the cast:

  My readers, Steven Salzberg and David Yarowsky, have given me very
  useful suggestions and comments about my thesis research.  They have
  each contributed immensely to my education.
  
  My friends at Hopkins have given me plenty of arguments and feedback
  concerning this thesis and other things, and I especially thank them
  for tolerating my discussions concerning ground beef superhighways
  and antiseptic qualities of Diet Coke.  I was always in good company
  at Hopkins.
  
  My advisor, Eric Brill, provided me with both the liberty to pursue
  research that I found interesting, and the constraints I required to
  complete my thesis research.  He made me feel valuable to the
  Hopkins NLP lab and the larger research community without treating
  me as just another information worker.  I could not have asked for a
  better advisor.
 
 
  My parents gifted me with a love of learning by teaching me through
  play and letting me disassemble most household items.  All along,
  my entire family prepared me to pursue research.
  
  Katya, my beloved wife, has developed a capacity for patience that I
  would have never believed possible.  Her love and support have eased
  many desperate moments in my graduate career.  Without her
  encouragement, my days would have been dim and my nights dismal.


\end{acknowledgement}


\tableofcontents
\listoftables
\listoffigures

\end{frontmatter}


\chapter{Corpus-based Natural Language Processing}

Computers do not understand human languages.  They can store and
search instances of linguistic data, as long as the search keys are
patterns which are very simple and similar to the data.  In this
respect, though, they are no more than advanced books or tape
recorders.  The massive quantity of human knowledge can be preserved
in this way, but not extended.  It can be inspected, but not
summarized.

The accelerating growth of the quantity of knowledge possessed by the
human race has been of concern for more than half a century
\cite{bush:memex}.  The concern has been whether our archival media
can keep pace with that rate of growth.  At this point, however, it
appears that the problem is understood and solvable with current
tools.  The World Wide Web has quickly become the de facto repository
for knowledge.

A problem of equal concern has been looming over the horizon, and did
not require our attention until its predecessor was solved.  At some
point in our future the temporally finite nature of human life will
restrict what inferences can be made from the wealth of knowledge.
The time will come when adding a piece of scientific knowledge via
deduction or experimentation will require more examination of the
repository of knowledge, and more time spent in deduction and
experimentation than a single human has the ability to give.

Whether humans can develop a social system for passing incomplete
deductions for others to continue is an open question.  At this
point it seems plausible that every deduction that has been made can
be attributed to some individual.

We have just presented two motivating reasons for producing systems
that are able to understand human languages, in the guise of a single
reason.  To clarify:
\begin{itemize}
\item Computers that understand language can better search through and
  summarize the existing wealth of human knowledge.
\item Artifacts with the ability to inference about concepts
  expressible in human languages can be given arbitrarily long
  lifetimes.  They will be able to make additions to the repository of
  human knowledge without restriction.  In the near term, they will be
  able to double-check the repository for consistency, validate
  new scientific claims, and suggest lines of research that have not
  yet been explored.
\end{itemize}

The problem of reasoning about natural language concepts is far beyond
the scope of a thesis.
  The CYC project attempted to solve this problem in only
eleven years starting in 1984, and they continue to work on it today
\cite{lenat95:cyc}.  
The computational linguistics and natural
language processing community is attempting to move toward the
solution to this problem by modeling progressively more complex
linguistic phenomena.  The high-level goal is to produce a model that
can infer underlying semantics given only surface realizations, the
observable pieces of a language.

There are many techniques that have been used to build these models.
Many people (probably every budding computer scientist) have tried to build
these systems by hand using introspection as their guide.  
Repeated experimentation has shown us that with a few outstanding
exceptions the resulting systems suffer from at least one of three
different maladies.  They either cover too little of the
phenomena present in the real world,
are opaque enough to require human intervention for interpretation,
or are trivially inadequate for use in real world tasks.
There are two simple possible reasons for this: people are unable to
inspect the internal workings of their language machinery, or they are
bad at generalizing or expressing their knowledge in a way that
ensures they can cover novel events that make up many cases in natural
language.

In this thesis one of our main goals is to provide a better technique
for creating natural language processing systems that outperform
independently developed state of the art systems.  In this chapter
will discuss experimental techniques, define some terms, and reflect
upon the current state of the art for natural language processing
system development.

\section{Data-driven Language Acquisition}

Natural language processing started out as people building processing
systems completely manually.  That approach proved too difficult, or
too cost-intensive for repeated application to other languages, as
well as for modeling changes in a single language.  The speed of
modern computers allows more of the burden of system creation to be
placed on a machine.  Recently, and inspired by successful machine
learning systems, there has been a movement to create more natural
language processing systems using inductive techniques from the
machine learning community.

Data-driven approaches to natural language processing require a strict
experimental setup.  One of the reasons for this is that machines,
unlike humans, are very good at memorizing phenomena.  Iteratively
working on an algorithm using a single set of data for both learning
and evaluation can result in a language processing system that has
memorized many of the specific features of that particular set.  The
system is then useless for working with language found outside of that
set.  To avoid this problem, experimenters partition their data into a
training set and a test set before beginning any experiments.  The
training set is used for developing a system, and the test set for
evaluating it.  Furthermore, to avoid a directed search on the test
set, a further partitioning of the training set is often used for
evaluation during system development.

\subsection{Supervised v. Unsupervised}

Most data-driven induction algorithms presented by the machine
learning community are \emph{supervised} techniques.  They are given a
set of training data to study that is labelled both with inputs the
resulting system is expected to handle and the correct classification
or structural annotation associated with those inputs.

In contrast to the supervised learning algorithms, there exist
induction techniques that are completely unsupervised.  They utilize
data to arrive at their predictions, but they are not given the
correct annotations of what they are to predict for a corpus.  Often,
they are not given any annotation for the predicted phenomenon.
Instead, they attempt to discover the correct hidden structure by
utilizing principles and beliefs about the general nature of language.
Examples of these algorithms include the many variants of the EM
algorithm including Baum-Welch \cite{baumwelch} and PCFG induction
\cite{lariyoung91:pcfg}; there is also an unsupervised version of
Brill's part of speech tagger \cite{brill:unsuptag}.  The Baum-Welch
algorithm has been very successful in speech recognition.

Recently there has been a great deal of interest in the development of
unsupervised systems because of their cost-effectiveness.  Few people
argue that unsupervised methods can surpass supervised methods when
the corpora are the same, but when the cost of annotating data is very
expensive relative to computing power (as it is now), the potential
savings can outweigh the performance hit.  This is especially true in
cases where there is an abundance of unannotated data, the reference
corpus is noisy, or the task is only vaguely defined.  The recent ACL
Workshop on Unsupervised Learning in Natural Language Processing was
organized around this topic \cite{unsupwkshp99}.

It is important to realize that unsupervised methods are still
data-driven, even though they are not looking at annotated data.  They
induce some model using training data and some intuition on the part
of the experimenter about the nature of the phenomenon they are
addressing, and they evaluate against the annotations of a test set
that are not seen during training.

In this thesis we will be presenting both supervised and unsupervised
algorithms for some of the tasks we address.

\subsubsection{Partially Unsupervised}

Many algorithms utilize both a small amount of labelled data and a
large amount of data that has no associated annotation.  These
algorithms are called \emph{partially unsupervised} because only the
small amount of data that is labelled provides supervision for a
learner.  The rest of the data helps the learner characterize the
nature of the unlabelled input it is expected to process.

Some successful examples of partially unsupervised algorithms for
natural language processing include Pereira and Schabes's technique
for grammar induction from a partially-bracketed corpus
\cite{pereiraschabes:pcfg}, Yarowsky's technique for word sense
disambiguation \cite{yarowsky95:unsupwsd}, Engelson and Dagan's
\cite{engelson96:tagunsup} as well as Brill's \cite{brill:unsuptag}
techniques for part of speech tagging, and David Lewis's text
categorization technique \cite{lewis94:uncertaintysampling}.

Pereira and Schabes extended the PCFG induction technique of Baker
\cite{baker79:pcfg} to utilize data that had been annotated by a
human.  They results are inconclusive on real world data, but the
technique is interesting, and they show both theoretically and by
simulation on an artificial task that it is sound.

The success of Yarowsky's algorithm has been recently explained by
Blum and Mitchell \cite{blum98:cotraining} who give a general
technique for using unlabelled data together with labelled data in a
batch-style processing fashion.  The main requirement for this
technique to work is the existence of separate views of the data, each
of which is sufficient for predicting the phenomenon in question.
Collins and Singer give more evidence of this technique's value by
applying it with success to named entity classification
\cite{collins99:unsupNE}.

Engelson and Dagan's and David Lewis's algorithms are very similar and
both trace their roots back to the Cohn et al. algorithm for active
learning \cite{cohn94:activeml}.  This technique differs from
Yarowsky's in that it requires interactive annotation.  The labeller
(a human or automated data collection system) is told which samples to
annotate by the machine learning algorithm.  Generally, the labeler is
asked to annotate those samples about which the machine is least
confident in its current prediction.  This interaction between person
and machine is known as a \emph{mixed-initiative} approach to
annotation \cite{day97:mixedinit}.

Charniak's parser has been tested in a partially unsupervised method
in the most straightforward example of the concept
\cite{charniak:parsing}.  After developing a parser in a supervised
manner, he parsed 40 million words of previously unparsed text and
re-estimated his parameters using the result as a training corpus.
This is reminiscent of the general expectation-maximization technique,
and gave him a slight, but significant improvement in accuracy on a
separate test set. Golding and Roth performed a similar study for
context-sensitive spelling correction \cite{goldingroth:cssc}.  They
showed that, consistent with intuition, the extra data these
techniques exploit allows them to dominate the performance of
supervised training alone.

\subsection{Parametric v. Non-parametric}

Parametric techniques require the setting of parameters based on
intuition or data.  All statistical approaches to natural language
processing are parametric.  They use the statistics they collect from
corpora to set parameters in their models.

In contrast, non-parametric techniques satisfy constraints on the
data or solve some optimization based on input from problem instance
only.  They do not have parameters that are learned or set by humans.
Purely non-parametric techniques are rare.  This is not a division
between symbolic and probabilistic systems, as the parameters in many
symbolic systems are hidden in the structure of the symbolic system.
There is typically a hierarchy of rules involved in the system, and we
can view the hierarchy as a set of parameters that are learned.  Also,
the particular rules that are chosen to participate in the system are
chosen as nonzero parameter values from the set of all possible rules.
In short, the difference is that non-parametric techniques do not
require \emph{any} training data.

A good example of a non-parametric algorithm is Hobbs's algorithm for
anaphora resolution \cite{hobbs76:anaphora,hobbs78:anaphora}.
Although it leaves the analytic procedure for comparing person,
number, and gender unspecified, it operates entirely on the input
parse trees aside from those requirements, soliciting no knowledge
from a training corpus.

Non-parametric techniques rarely perform as well as parametric
techniques, because natural language is idiosyncratic.  For most
tasks, there are concepts that require inspection of real data in
order to be observed and learned.

In this work we will describe non-parametric algorithms for switching
between parsers.  Some of the algorithms given are competitive with
their parametric counterparts.

\subsection{Corpora}

There is a wealth of corpora available for automated learning systems
in natural language processing, and more corpora become available each
year.  Some of the more richly annotated sources of text include are
described below.

\begin{itemize}
\item
  
  The Brown Corpus \cite{browncorpus} is a collection of various
  genres and sources of written text including fiction and non-fiction
  such as news stories.  The text is annotated with part of speech
  tags.

\item
  
  The University of Pennsylvania's Wall Street Journal Treebank
  (version II) \cite{penntreebank} is a collection of several corpora.
  Three years of the Wall Street Journal, about 1 million words of
  text, is annotated with part of speech tags as well as phrase
  bracketing structure.  Another 40 million words are annotated with
  part of speech information, but no parse trees.

\item
  
  The SUSANNE Corpus \cite{susannecorpus} was the side-effect of a
  project aimed at standardizing annotation schemes and producing an
  annotation scheme capable of completely describing linguistic
  phenomena found in text.  It contains high-quality phrase bracketing
  information and more for a 130,000-word subset of the Brown Corpus.

\item 
  
  The British National Corpus looks like a promising source of
  annotated data.  It is the result of a recent corpus collection
  program that was completed in 1996.  As such, it may be the corpus
  that contains the most recent English documents.  It does not
  contain phrase annotations, but its 100 million words are each
  tagged with a part of speech tag chosen from 61 categories.  Most of
  the tagging was automated, however, so its utility for machine
  language learning may be a bit suspect.  We cannot say more about
  this corpus, because it is currently unavailable outside of the EU.

\item
  
  The Prague Dependency Treebank \cite{hajic98:ctreebank} is about
  500,000 words in size.  Czech is representative of many Slavic
  languages in that there is considerable liberties in word ordering
  allowed.  The corpus is annotated in dependency style, with links
  from words to the heads of the syntactic constructions that dominate
  them.  The morphological tagging for Czech is very rich when
  compared to English, and the treebank is fully annotated in this
  respect as well.

\item 
  
  It is to be expected that the technological advances that depended
  on the various English treebank projects will be desired in many
  non-English-speaking countries.  Treebank projects are starting to
  spring up in many countries.  Among many, there is a German corpus
  of newspaper articles underway \cite{brants99:gtreebank}, and plans
  for a corpus of Turkish \cite{oflazer99:ttreebank}.

\end{itemize}

In this work we describe experiments performed on the Penn Treebank.

\subsection{Tasks of Interest}

There are many tasks that the natural language processing community
has identified as interesting, and potentially addressable using
data-driven approaches.  Here are some of them, listed in an
approximate order of increasing complexity.

\begin{itemize}

\item {\bf Part of speech tagging}
  
  One of the most straightforward tasks, part of speech tagging
  involves giving the part of speech tag for each word.  For example,
  if the sentence 
  
  {\tt She ate the juicy apple.} 
  
  is an input, the corresponding output is
  
  {\tt She/pronoun ate/verb the/determiner
    juicy/adjective apple/noun.}
  
  \noindent There is not complete agreement on
  what the set of possible tags should be.  Many natural language
  processing problems can be theoretically reduced to this
  one~\footnote{For an excellent example of this, see Ramshaw's
    formulation of noun phrase bracketing as a tagging problem
    \cite{ramshaw99:basenp}.}, so algorithms for automatically
  creating part of speech taggers are valuable.  Also, many tasks that
  produce higher order linguistic annotation rely on a good part of
  speech tagger as a component system.  Collins's parser, for example,
  requires part of speech tags from Ratnaparkhi's {\tt MXPOST}
  program.
  
  Currently, English POS tagging can be performed with an accuracy
  equal to tagging 97\% of the words correctly \cite{brill95:tagging,
  adwait96:tagging, halteren98:combine, brillwu:combine}.

\item {\bf Word sense disambiguation}
  
  The sentence 

{\tt He drew a line on a piece of paper while he stood
    in line for the movie.} 

\noindent demonstrates word sense ambiguity.  The two
  instances of the word {\tt line} have different meanings, and
  those senses are immediately evident to the human reader.
  Some difficulty remains in the practical evaluation of WSD systems.
  Typically a small set of words are selected for annotation, and a
  partitioning of their senses is agreed upon by a committee.
  Instances of those words in a large corpus are annotated, and
  systems are compared on their performance on those words.  The
  limited set of words and the arbitrary partitioning of senses is of
  concern to some \cite{wilks98:wsd}, but it led to rapid
  progress on the task \cite{yarowsky95:unsupwsd}.

\item {\bf Parsing}
  
  Parsing involves marking a sentence with its phrase structure.  We
  treat it in more detail in Section \ref{section:corpusnlp:parsing}.

\item {\bf Anaphora Resolution}
  
  Determining which noun phrase a particular pronoun refers to is part
  of the anaphora resolution problem.  The best anaphora resolution
  algorithms rely on parse trees as their input.  That dependency and
  the lack of available automated parsing systems that achieve high
  accuracy has hindered some progress in solving this task.  Most
  groups working on the problem have annotated proprietary data, or
  developed proprietary unsupervised algorithms for the task.  A
  recent attempt to bring the task to a more quantitatively comparable
  state suggests that anaphora resolution can be performed with an
  accuracy of approximately 70-70\% \cite{tetreault99:anaphora}.


\item {\bf Coreference}
  
  Once anaphora problems have been solved, the question of which noun
  phrases in a document are talking about the same real world object
  arises.  This is the coreference task, finding which set of phrases
  all refer to the same real-world (aside from the document) concept
  or entity.  In a civil war document it may be necessary to determine
  that {\tt Lincoln}, {\tt Abraham Lincoln}, {\tt President Lincoln}
  and {The President} are all referring to the same person, who is not
  the same as {\tt Lincoln, Nebraska} (if it had existed at the time).
  There are ambiguity problems here as well.  Consider {\tt Lincoln's
    Address}: there are instances in which it refers to a speech that
  he gave, and others in which it refers to the place that he lived.
  Various approaches to this task have been addressed in the Message
  Understanding Conferences (MUCs), with MUC-6 being the first time it
  was evaluated as a separate task \cite{muc6}.


\item {\bf Machine Translation}
  
  The goal of machine translation is to produce a
  document in language B that preserves the meaning of a given
  document in language A.  Machine translation is difficult to
  evaluate in an empirical setting because there are no agreed upon
  best or even canonical translations for most sentences.  While there
  are many translation systems in circulation, a few of the more
  recent and prominent ones that use parse trees are starting to
  develop formal evaluation techniques \cite{mt:threeheads,
  mooneyparse1997, llenglishkorean}.  

Although there are many available translation systems for translating
between Western languages, those systems do not perform well on
spontaneous speech, nor do they offer much insight into how to perform
MT between Chinese and English, for example.  The best available
systems were created manually, and rely on the relatively similar word
order of the languages they address as well as high availability of
cognates.


\end{itemize}

These are just some of the tasks that are being actively pursued by
researchers.  This is a field littered with a wide variety of problems
and tasks. 



\section{Parsing}
\label{section:corpusnlp:parsing}

In this thesis we will be focusing on parsing.  Parsing is the task of
delimiting phrases of a sentence and describing the relations between
them.  The parser is given an unmarked sentence and it is required to
perform these annotations.  The task is a crucial step in the chain
that characterizes linguistic phenomena.  It corresponds to
determining the syntactic structure of a sentence.

The particular form of parsing we will be working on is the type
represented in the Penn Treebank.  In their annotation, which is an
amalgam of many grammatical formalisms, properly nested sections of
text are delimited by brackets and identified by labels.  Because they
are properly nested, the bracketings can be viewed as representing a
projective parse tree over the sentence, where there is a unique path
from each word to the root of the tree.  Part of speech tags are the
preterminal nodes in this tree, and every word has a part of speech
tag associated with it.  In parser evaluations, part of speech tagging
is treated as a separate task, so those nodes are treated differently
from the rest of the tree (generally ignored).

The purpose of parsing is to remove as much ambiguity in a sentence
that can be determined by syntax as possible.  For example, the
sentence

{\tt She saw the boy on the hill with binoculars.}

\noindent should be interpreted differently in different contexts.
The representation of the particular interpretation intended is
available in the parse tree.  We will explain this with an example in
Penn Treebank form.

\begin{parse}
\hspace{-1in}
  \Tree
[.S
[.NP [.N She ] ]
[.VP [.V saw ]
[.NP  
[.NP [.Det the ] [.N boy ] ]
[.PP [.P on ]
[.NP [.Det the ] [.N hill ] ] ] ] ]
[.PP [.P with ]
[.NP [.N binoculars ] ] ] ]
  \label{parse:intro:binoculars:preferred}
\end{parse}

In Parse \ref{parse:intro:binoculars:preferred} the girl has the
binoculars and the boy is on the hill.  Since {\tt with binoculars} is
not underneath the verb phrase, it is modifying the verb phrase and
telling us how the girl did the seeing.

\begin{parse}
\hspace{-1in}
  \Tree
[.S
[.NP [.N She ] ]
[.VP [.V saw ]
[.NP [.NP [.Det the ] [.N boy ] ]
[.PP [.P on ]
[.NP [.Det the ] [.N hill ] ] ] 
[.PP [.P with ]
[.NP [.N binoculars ] ] ] ] ] ]
  \label{parse:intro:binoculars:his}
\end{parse}

In Parse \ref{parse:intro:binoculars:his} the boy is on the hill and
has the binoculars.  The prepositional phrase {\tt with binoculars}
has moved inside of the verb phrase to describe the boy.

\begin{parse}
\hspace{-1in}
  \Tree
[.S
[.NP She ] 
[.VP [.V saw ]
[.NP [.Det the ] [.N boy ] ]
[.PP [.P on ]
[.NP [.Det the ] [.N hill ] ] ] 
[.PP [.P with ]
[.NP [.N binoculars ] ] ] ] ]
  \label{parse:intro:binoculars:hisonhill}
\end{parse}

In Parse \ref{parse:intro:binoculars:hisonhill} the girl is on the
hill and has the binoculars.  Both of the prepositional phrases have
moved out of the noun phrase that describes the boy.  This
interpretation shows one of the idiosyncrasies of the Penn Treebank:

\begin{parse}
\hspace{-1in}
  \Tree
[.S
[.NP She ] 
[.VP [.V saw ]
[.NP [.Det the ] [.N boy ] ]
[.PP [.P on ]
[.NP [.NP [.Det the ] [.N hill ] ]
[.PP [.P with ]
[.NP [.N binoculars ] ] ] ] ] ] ]
  \label{parse:intro:binoculars:hills}
\end{parse}

Finally, in the somewhat absurd Parse
\ref{parse:intro:binoculars:hills}, the hill has the binoculars.  This
example shows that there are parse trees that can be interpreted, but
which are unreasonable.  The reason we disagree with that parse is
that we do not think hills can have binoculars.  That is a semantic,
not syntactic constraint.

Choosing between these potential interpretations for the sentence is
the task of the parser.

As a technical note, even though we have removed the punctuation,
these complete trees are still burdensome to read.  To remedy this, we
can abbreviate them as seen below.  We have removed the preterminals
(part-of-speech tags) and collapsed some of the phrases denoted by
triangles.  Parse \ref{parse:intro:binoculars:preferred:abbrv} is an
abbreviated version of Parse \ref{parse:intro:binoculars:preferred}
and Parse \ref{parse:intro:binoculars:his:abbrv} is the abbreviation
of Parse \ref{parse:intro:binoculars:his}.  The bottom-most
constituent in Parse \ref{parse:intro:binoculars:his} is now ambiguous
(the hill could come equipped with binoculars), but when we make the
abbreviation in this manner the ambiguity we overshadow will not be
the one we are trying to highlight.

\begin{parse}
  \Tree
[.S
[.NP She ] 
[.VP saw 
\qroof{the boy on  the hill}.NP
\qroof{with  binoculars}.PP ] ]
  \label{parse:intro:binoculars:preferred:abbrv}
\end{parse}

\begin{parse}
  \Tree
[.S
[.NP She ] 
[.VP saw 
\qroof{the boy on the hill with binoculars}.NP ] ]
  \label{parse:intro:binoculars:his:abbrv}
\end{parse}

\subsection{Parsing Technology}

There is a long line of research in parsing.  We will focus on the
work that was designed specifically for the natural language
processing task.

The earliest work on corpus-based automatic parser induction dates to
Black et al. \cite{black91:parsemetric} who describe the metrics that
are still used for measuring parser performance.  Around the same time,
Pereira and Schabes produced some experimental results on PCFG-style
parser induction \cite{pereiraschabes:pcfg}.

Early work on parsing using the Penn Treebank was done by Magerman
\cite{magerman95:parsing}, Brill \cite{brill96:parsing}, and Collins
\cite{collins:parsing96}.  Magerman's system controlled a
left-to-right parser using a decision tree.  Brill's system used
automatically-learned rules for transforming initially poor parse
trees into better ones.  Vilain and Day \cite{vilain96:parsing}
produced a faster version of the transformation-based parser.
Collins's work was one of the first successful PCFG head-passing
grammar-based systems for this task.

More recently, Ratnaparkhi \cite{adwait:parsing}, Charniak
\cite{charniak:parsing}, and Collins \cite{collins:parsing97} have
each independently developed statistical parsers using the same
training and testing split of the Penn Treebank.  Collins and Charniak
both use a head-passing PCFG as the basis of their models, although
the features they use for their models are different.  Ratnaparkhi
uses a maximum entropy classifier to control a machine that
iteratively builds and prunes a parse tree from the bottom up.  We
will discuss their parsers more in Chapter \ref{chapter:combining}.

Hermjakob and Mooney created a parser trained on only 1000 sentences
which performs with state-of-the art accuracy \cite{mooneyparse1997}.
The training set was very small because the model has very many
parameters and the search algorithm used for developing the parser is
slow.

Goodman's work \cite{goodman98:phd} develops some formal approaches to
defining parsing systems and shows how to create parsers that directly
maximize some given performance metrics.  He gives separate automated
parser induction algorithms that directly maximize recall and an
approximation of precision.  Also, he points out that there is a basic
incompatibility between parsing with the goal of getting sentences
correct and parsing with the goal of getting constituents correct.
The two metrics have the same maximum point, namely when everything is
parsed correctly, but in practice there is a tradeoff involved in
maximizing them independently.  Goodman also provides practical
techniques for parsing with large vocabularies and large grammars.  He
presents experiments involving multi-pass pruning algorithm to parse
in the face of computational time and space constraints.


Johnson has studied the effect that the idiosyncrasies of tree
representations has on the quality achievable by parser induction
algorithms \cite{johnson98:treerep}.  The Penn Treebank (version II) is 
idiosyncratic in that it represents verb phrase adjunction with a flat
tree structure.  Johnson describes techniques for producing a more
informative representation for modeling with a PCFG.  He furthermore
shows theoretically as well as experimentally that performing simple
invertible tree transformations on the Treebank produces a corpus that
better facilitates automatically inducing PCFG-style parsers.


The parsing community has recently had a large improvement in accuracy
while suffering from a loss in speed.  Caraballo and Charniak address
this issue by finding a good heuristic for searching for a good parse
in a PCFG-style parser \cite{caraballo98:parsingfom}.

Chelba and Jelinek have created an online parser which operates in a
left-to-right manner like a pushdown automaton in order to better
perform language modeling for speech recognition
\cite{chelba98:structlm}. They use a maximum likelihood technique to
learn the controlling automaton for a shift-reduce parser.  Recently
it has been shown that this parsing architecture is not entirely
equivalent to PCFG parsing, although both formalisms can learn the
same set of probability distributions over
strings\cite{abney99:ppdavpcfg}.


With the recent successes in parsing English text, the parsing task
has been ``ported'' to other languages including Czech
\cite{collins99:czechparse} and Japanese \cite{boostingparsing98}.
Each of these languages has required a redesign or modification of the
task.  They each operate in a dependency representation.  Each word
(or chunk) is annotated with an arrow directed toward the word that it
syntactically supports.  In Czech this is required because the word
order is much more liberal than in English.  In Japanese, each phrase
(bunsetsu) is guaranteed to modify a phrase that comes before it, but
not necessarily the most recent phrase.  As we described earlier, both
of these parsing tasks are supported by treebank efforts, as well.


\subsection{Why Parsing?}

The parsing task is of interest to theoreticians and computational
linguists, but it also has applications in many real-world problems.
Like most natural language processing systems, it is a component that
is meant to be inserted into a larger application.

\subsubsection{Grammar Checking}

The original purpose of parsing was to determine if sentences conform
to a grammar.  It has progressed quite a bit since then, but this task
has become important with the widespread use of word-processing
software.  Statistical parses that will always give a most likely
parse for a sentence can still be used as grammar checkers by
thresholding the score for a sentence to determine its acceptance, or
highlighting sections of the parse that have particularly low scores.

\subsubsection{Machine Translation}

For many translation tasks, especially translating between languages
with differing word order, parsing is a crucial step.  There is a
strong belief that once words and small phrases can be translated,
transformations on the parse of a sentence can be used to rearrange
large portions of text to make it conform to the expected ordering.

The TINA parsing system \cite{seneff:tina} is used in a Korean-English
machine translation system \cite{llenglishkorean}, and Hermjakob and
Mooney's parser was designed to be closely coupled with a translation
system \cite{mooneyparse1997}.




\subsubsection{Embedded Applications}

There are some tasks which require parsing as a precursor to further
processing.  Moving up the linguistic chain from syntax to semantics,
we see that many tasks involving semantics tend to require
high-quality syntactic structure representations as input.

\begin{itemize}
\item {\bf Prepositional Phrase Attachment}
  
  This task \cite{brill94:pp-attach, collins95:pp-attach,
    ratnaparkhi94:pp-attach, merlo97:multiplepp} attempts to fix some
  of the mistakes created by parsers.  The examples we gave in Parses
  \ref{parse:intro:binoculars:preferred} through
  \ref{parse:intro:binoculars:hills} vary in how the prepositional
  phrases are attached.  Parsers based on context-free grammars are
  not as accurate at these attachment decisions as they should be, and
  so this task is often worked on separately.

\item {\bf Anaphora Resolution}

  Hobbs's algorithm for anaphora resolution requires a parse tree in
  order to decide how to search among candidate noun phrases as it
  searches for the antecedent for a pronoun
  \cite{hobbs76:anaphora, hobbs78:anaphora}.

\item {\bf Summarization}
  
  Recently, automated summarization systems have begun to use
  statistical parsers to determine large chunks of text that are
  repeated, or which can be removed in order to make the text
  syntactically more concise \cite{barzilay99:summary,
    mani99:improvsumm}.

\end{itemize}

\subsubsection{Similar Problems in Different Domains}

Problems similar in structure to parsing arise in other fields, and we
expect to see many problems that theoretically reduce to parsing arise
as well.  For example, Miller and Viola hierarchically segment images
of mathematical expressions \cite{miller98:eqnparse} in order to
recover the expression tree that they represent, and work has been
done in the field of computational biology focusing on hierarchically
determining the physical structure of molecules that are created from
sequences of RNA \cite{sakakibara94:tRNAscfg,grate95:tRNAscfg}.  It is
possible that advances in parsing technology as applied to natural
language processing can be useful in these other fields.


\chapter{Combining Independent Hypotheses}

The recent rapid onset of data-driven approaches to natural language
processing has provided the community with many systems addressing
each task.  There are natural language processing systems available as
commercial off-the-shelf systems (or component systems) or as freeware
available on the world wide web.  Part of speech tagging, for example,
has at least four good trainable systems available attacking it.
These systems are normally results of independent development groups
and independent corporate entities.  We expect that the independence
of these research groups leads them to produce models that specialize
in different ways.  For example, one tagger could more precisely
annotate adjectives than another that more precisely annotates verbs.
Having all of these systems that address a common task is beneficial
for the field because it allows a new kind of experimentation to be
performed: combining the independent hypotheses.

\section{Natural Byproducts of Technological Development}

The situation is not unique to the field of natural language
processing.  Within computer science, one can see the hardware
evolution of the computer leave a trail of processors and platforms
which are succeeded by ever faster and more appropriate machines.
Automobiles become progressively more reliable and more efficient.
Insulated waterproof clothing is losing its bulk and requiring less
maintenance.  These three technological progressions all leave their
useless (or less valuable) forebears to break down and wear out, never
to be directly compared with systems (or products) that result from
later developmental cycles.

Natural language processing, however, produces systems that are not
physically manifest.  As conceptual artifacts, they will not wear out.
Like all algorithmic entities, they can be revived at will and remain
viable even if they are not dominant.

A long history of systems that address a common task can be found for
any task that has been ``solved''.  By solved, we mean that the task
can be performed with high enough accuracy by a machine that no more
resources are being allocated to produce better performance.  There
are still many natural language processing tasks that remain unsolved.
We expect that most if not all of them will have a wide variety of
systems attack them before they are solved.

\section{The Ends to Justify The Means}

There are reasons to attack the task of combination other than the
fact that we can.  First, we can expect to find new lower bounds on
the possible performance that can be achieved on a problem.  Second,
we can build ensemble systems that perform better than any of their
members.

\subsection{New Achievable Bounds}

Corpus-based tasks are inherently open-ended.  It is difficult to
determine how much performance gain can still be achieved on a task at
any given time.  Part of that uncertainty is what makes it a
\emph{research} task, but some of it comes from not knowing the
quality of the data.

Computing inter-annotator agreement is often cited as a good way of
determining how difficult a problem is.  There are three drawbacks to
this approach.  The first two question the dominance claim of
inter-annotator agreement.

First, there is the question of annotator competence.  When one
annotator (or a subset of the annotators) is much better at performing
the task consistently than another simply because the other one is
less capable the inter-annotator agreement will reflect the
performance of the worse annotator (or set of annotators).  Secondly,
in suggesting that human performance is an upper bound on how well a
machine can perform on a task implies that the machine can never
perform better than the human.  The reasons for promoting this belief
are homo-centric (or perhaps bio-centric).  We know that there are
many tasks at which machines can outperform humans.  There is no
reason that learning cannot be one of them.  These two reasons both
suggest that inter-annotator agreement is less reasonable as an
\emph{upper} bound than we thought.

Finally, the drawback that is most applicable to natural language
processing is that human annotators sometimes use information
unavailable in the data to perform their annotations.  This is really
a question of comparing apples and oranges.  Much of that data is not
available to computers simply because it has never been entered or
cannot be indexed well enough.  This suggests that inter-annotator
agreement is too stringent as an upper bound on machine performance.

We see that inter-annotator agreement is both too strong and too weak
to serve as a performance bound.  These are not new arguments, and
they are more or less obvious.  The only reason that the measure is
used as a bound, then, is that it is the only point that is readily
available and computable when a new task is defined and its data is
collected.  Inter-annotator agreement remains a useful upper bound on
how high an accuracy we can \emph{measure}.

Once a few systems have been built that address a task, however, there
are other more reasonable candidates for upper bounds on performance
that can be computed in order to encourage work on a task, estimate
progress versus potential, and determine if a problem has been
``solved''.  The available systems can typically be combined into a
composite system using democratic or other simple principles as
guides.  The performance of this composite system then becomes a
bound on the performance that individual independently-produced
systems can achieve.  One of the goals of this thesis is to propose
such a bound for parsing.

The other advantage of using combination techniques to produce a bound
is that as the individual systems become better the bound can be
re-evaluated.  If the individual systems are truly independently
constructed and highly accurate, then their improvements will make the
upper bound a more accurate bound.  Note that we do not mean it will
make the bound get higher, although that could happen.  We mean that
the bound will become closer to the true bound which is limited by
noise in the data and the knowledge of the task available to the
machine.

\subsection{Better Systems}

In circumstances where the individual systems are not fully utilizing
the resources available for allocation to the pursuit of the learning
task, the bulky composite system can itself be considered a practical
approach for the task.  This is the case in many initial development
domains, where there is little data available for a task.

The other way of looking at this is that when there are more resources
available for a task than is currently required, utilization of
combination methods is a fruitful way to allocate those resources.

Computing power is an example of an underutilized resource when it is
measured globally.  With the rapid growth of wide area networks, it is
plausible to attempt to exploit the wasted computing power that is
currently on many desks to pursue classifier combination techniques.

Voting and other combination methods are powerful techniques for
reducing error, as we show later in this thesis.  There is plenty of
theoretical work to support this claim as well, some of which we
describe in Section \ref{section:comboback:recentcombo}.

\section{The Price of Progress}

There is a cost to all of this that we have alluded to.  Combination
methods require the aggregate computational expense of the ensemble
members plus the cost associated with performing the combination.  If
the individual members of the ensemble were designed to run on modern
computers, then they may already be stretching the resource
utilization to the limit.  

At this point in time, however, the rapid increases in computing
hardware mean that programs that ran on hardware that was current only
3 years ago are barely using half of the resources of the hardware
that is currently available for a similar purchase price and
maintenance cost.  The achievements in increasing computing speed and
space per dollar is a major enabling factor for this work.

Alternatively, if computing speed was not getting faster, the network
is a major facilitator on its own.  For various reasons, including a
general lack of knowledge, most programs run on only one computer.
Combination methods can typically take advantage of parallelism to run
the ensemble members simultaneously by distributing work across
several machines.

The quantity of available computing resources is an issue for
combining hypotheses, but the current conditions in computing hardware
are favorable.  Moreover, there is little reason to believe that rate
of growth of hardware specialization and fast networking will not
continue into the future.

\section{Recent Work on Classifier Combination}
\label{section:comboback:recentcombo}

Wolpert's work on \emph{stacking} was one of the first machine
learning attempts to combine classifiers \cite{wolpert:stacking}.  He
was interested in neural network classifiers, but realized the
technique he developed was general-purpose, not specific to neural net
classifiers.  

Stacking is a hierarchical approach to classification.  At the bottom
level of the stack are $k$ individual classifiers, each trained on a
different partition of the training data.  The data is disjointly
partitioned into $k$ subsets, and the training data for each of the
initial classifiers is the entire training set except for set $k$.

The output of those first level classifiers when run on held-out data
is then fed into the next level of the stack which attempts to predict
based on those outputs alone.


The first level classifiers are then run on the entire dataset in
order to produce a new pseudo dataset consisting of the output of the
classifiers as the values of features.  This resulting dataset is used
to train another classifier on the second level of the stack.  The
goal is to get the second level classifier to learn to correct the
first level classifiers.  Many combining heuristics could be plugged
into the architecture, such as majority voting, but Wolpert was the
first to suggest that position should be occupied by another inductive
learner.
 
The process can be adjusted in order to extend up multiple levels, but
there is no empirical evidence for or against doing so.

Heath et al. experimented with combining decision trees
\cite{heath96:committee}.  They used standard decision tree induction
for producing the ensemble members, and majority voting for combining
hypotheses.  Their work was the first to consider the question of how
to automate the process of making independent, diverse learners.
Their approach was to randomize the learning process.  Their simulated
annealing decision tree induction system, SADT, utilized randomness
during its construction of the tree.  They resampled this process to
create an ensemble.

They gave a theoretical treatment of the error reduction that can be
realized in ideal cases.  Simply put, they showed that classification
errors decrease exponentially in the number of ensemble members, given
that individual members of the ensemble consistently perform better
than random chance at the classification task.

Opitz and Shavlik built ensembles of neural networks
\cite{opitz96:diverse}.  Their main contribution (aside from a good
performance result on several tasks) was to introduce a formalization
of the notion of diversity in their work.  They explicitly maximized a
linear combination of accuracy and diversity in producing their
ensemble.  They generate their ensemble by using a genetic algorithm
that attempts to maximize this metric.  The population for the
algorithm is a set of neural networks, and they are mutated and
crossed-over using topology-modifying operators.  At the conclusion of
the optimization, the resulting population (of fixed size, specified
as input) operates as an ensemble for classification of new data.

\section{Combination in Natural Language}

Independent system combination has recently started appearing in
natural language processing work.  This is in part because of recent
work done in the machine learning community, but also because the
field has grown to the point where there are so many diverse
individual systems available for combination.

The machine learning community and the computational learning
theorists have developed many ensemble theories and architectures for
traditional vector space classification problems.  Natural language is
different from traditional classification problems in that it is
typically sequence-based and often the predictions can be very
structured.  Parsing, for example, is hard to simply reduce to a
binary classification problem.

\subsection*{Part of Speech Tagging}

Part of speech tagging is not a typical machine learning vector space
classification problem.  It involves classifying words and contexts
into part of speech tags, but the individual classifications are not
independent.

Van Halteren et al. \cite{halteren98:combine} provide some methods for
combining state of the art part of speech taggers by treating the task
as a classification problem and applying stacking.  They acquired four
of the best part of speech tagging programs and trained them on the
same data.  Then, a held-out tuning dataset was used to estimate the
accuracy of the taggers and collect statistics on where they
individually make errors.  The experimenters then take two separate
approaches.  In some experiments they generate tagging heuristics for
using the statistics they collect.  The best of these experiments
collects statistics on what the correct tag is given particular
pairwise disagreements between taggers.  When the disagreement was not
seen in training data, it backs off to the best individual tagger.
Note that in this case they are not learning which tagger to trust in
which situations, but rather what the correct tag is given the
situation.

In other experiments they directly train separate classifiers using
the outputs of the individual taggers as input feature values.  This
second experiment is more reminiscent of Wolpert's stacking method.
They use both a memory-based learner and a decision tree learner in a
straightforward manner.  In contrast to the stacking architecture,
they also pass information from features involved in the underlying
text that the component taggers operated on.  In particular, they add
words and other tags to the feature space in some experiments.
  
Surprisingly, the heuristic we mention performs best on this task, and
it is significantly better than all the other algorithms they try,
including the classifier induction techniques.  It achieves a 19\%
tagging error rate reduction.

Brill and Wu studied combining part of speech tagging independent of
van Halteren \cite{brillwu:combine}.  They similarly worked strictly
on the outputs of four taggers, although their set was not the same as
van Halteren et al.  There are two main contributions of their work
that are separate from the other study.  They developed a feasibility
technique for deciding if combination is a worthwhile endeavor.  They
detect if one tagger makes a strict subset of the errors that another
tagger makes.  The other contribution was learning a switch between
taggers instead of just predicting a new tag.  It is counterintuitive,
but this model gave them the lowest error rate.  It is probably a data
scarcity issue.  Instead of choosing best tag from among approximately
30 different tags, the combiner must only choose which of the four
taggers it trusts the most.  Since the prediction set is smaller,
there is less noise to learn from the data, and more samples for each
predicted class.

\subsection*{Named Entity Extraction}

Borthwick et al. have used the maximum entropy principle to combine
outputs of named entity recognition \cite{borthwick98:named}.  They
combine four systems (including their own) that competed in the
Seventh Message Understanding Conference (MUC-7) using a maximum
entropy technique.  Their system was originally based on a maximum
entropy model, so they could simply add the output generated by the
other three systems as features in their system.  The resulting
performance they attain is a dominant result for the task.  

\subsection*{Speech Recognition}

Fiscus combined five speech recognizers that participated in the 1999
LVCSR evaluation to get a statistically significant reduction in word
error rate \cite{rover}.  Speech recognition is not a classification
task, and the output of different systems need not even have the same
length.  He aligned transcriptions given by different recognizers,
then produced the final hypothesis by voting over the columns of the
alignment.  The technique he developed was successful and practical
enough to be incorporated into several speech recognition systems.

\subsection*{Translation}

Machine translation has been an object of combination techniques as
well.  Frederking and Nirenburg combined three translation systems
using a dynamic programming algorithm \cite{mt:threeheads}.  The three
systems they used were all developed in-house: a knowledge-based
system, and example-based system, and a lexical-transfer system.  Each
of these systems produces hypotheses that are recorded in a chart.
Each chart entry points to a start and end position of the input
string, offers a potential translation for that substring, and gives a
score representing the goodness of that translation.  The scores for
the chart elements are normalized to allow comparison between systems,
then a finally hypothesis is created by selecting a set of chart
elements that cover the sentence and have the highest score.  This is
done with a straightforward divide and conquer approach implemented as
an $O(n^3)$ dynamic programming algorithm.




\chapter{Combining Parsers}
\label{chapter:combining}

%
%
%





Progress in corpus-based parser development has been sought after with
incremental results during the majority of this decade.  Many
independent efforts have been made toward replicating the bracketing
style of the Penn Treebank project.  There has been a great deal of
competition among automatically trained parsers, each parser trying to
perform the best on previously unseen data.  This competition has
resulted in a number of parsers that controlled experiments show have
comparable (and good) performance.

\section{Task Description}

In this chapter we explore techniques for combining multiple parsers.
Our goal is to achieve better overall performance.  We explore
supervised methods, in which we allow the machines to learn a few
parameters or rules by inspecting training data to help it decide in
which situations it should trust which parser.  Also, we explore
unsupervised methods, such as democratic voting, in which all parsers
are treated equally and the machine blindly combines without first
explicitly determining which parser to trust in which situations.

We are working with three statistical parsers that have been objects
of independent development efforts.  The three parsers are Michael
Collins's generative parser \cite{collins:parsing97} configured as it
was used in the 1998 Johns Hopkins University Center for Language and
Speech Processing Workshop \cite{hajic:ws98}, a parser created by
Eugene Charniak \cite{charniak:parsing}, and Adwait Ratnaparkhi's
maximum entropy parser, {\tt MXPARSE} \cite{adwait:parsing}.  In some
experiments where we measure the robustness of our combination
techniques, we use a simple PCFG parser developed in our laboratory.

Ratnaparkhi's parser, {\tt MXPARSE}, is a machine that iteratively
builds a parse from the bottom up by chunking noun groups, then
progressively constructing constituents on top of those previously
created.  After each construction phase the work is inspected and some
constituents are deleted by a separate pruning phase.  The decisions
of the phases are made by a separate maximum entropy model.  During
parsing, the potential operations of these machines are searched to
find a most probable sequence of operations given the input sentence.
This in turn uniquely defines a parse tree.

Collins's parser relies on a generative, lexicalized parsing model.
Like {\tt MXPARSE}, Collins's parser is lexicalized.  Each constituent
is parametrized with the lexical head of the phrase it represents.  It
assigns probabilities to sequences of actions that produce parse trees
from the top down.  To do this it treats each labeled constituent as a
separate hidden Markov model producing the sequence of children nodes
for the constituent.  When a sentence is presented to be parsed, the
parser searches top-down for the sequence of productions that produce
the sentence (labelled with part of speech tags) with the highest
probability. Collins's parser relies on Ratnaparkhi's tagger
\cite{adwait96:tagging} to do the preprocessing and assign an initial
set of part of speech tags to the sentence.


Charniak's parser is similar to Collins's except that it does not
compute the probability of a constituent in the same way.  Each
constituent is conditioned on the lexical head of its phrase, but it
is also conditioned on its parent's label and some class information
about the lexical head.  Charniak computes the joint probability of
the tree and the sentence using dynamic programming instead of the
beam search that both Ratnaparkhi and Collins use.


All of the parsers were trained on the same sections of the Penn
Treebank version 2 (02-21), and tuned on various sections which we
leave out of our experimentation (sections 00, 01, 24).  Sections
02-21 contain approximately 40000 sentences.  Every performance
statistic we present concerning the parsers was derived from testing
the parsers on data that they was not part of their training set
(sections 22 and 23).  In our supervised combination experiments we
train the combiners using section 23, which contains 2416 sentences.
Previously reported performance results on these parsers were derived
from this section.

\subsection{Performance Measures}
\label{section:combining:measures}

Parsing performance is measured in a number of ways.  All of them
start with counting the three observable situations that can occur in
a prospective parse.  These situations are illustrated in Table
\ref{table:combining:situations}.  First, we break the reference and
guess parses into a set of constituents.\footnote{Some constituents
  are removed from these sets.  See Section
  \ref{section:combining:evalb} for a more detailed description.} Each
constituent consists of a label and a span.  Then we can observe three
situations: (a) The suggested constituent is in the suggested parse
and in the correct parse.  It is a correctly predicted constituent.
(b) The constituent suggested by our parser is not in the correct
parse.  It is a precision error.  (c) The constituent in the correct
parse is not in our parse.  We missed it: it is a recall error.  Note
that case (d) is not observable in the world of parsing because we
never see a constituent that is not in the suggested parse or in the
correct parse.  However the number of times case (d) occurs in a
particular parse is computable: we can count how many possible
constituents are possible for a particular sentence.

\begin{table}[htbp]
  \begin{center}
    \begin{tabular}{|lr|c|c|}
      \cline{3-4}
      \multicolumn{2}{c}{} & \multicolumn{2}{|c|}{In Reference?} \\
      \multicolumn{2}{c|}{} & yes & no \\
      \hline
      In Our & yes & a  & b \\
      \cline{2-4}
      Guess? & no & c & d \\
      \hline
    \end{tabular}
    \caption{Possible Parsing Constituent Situations}
    \label{table:combining:situations}
  \end{center}
\end{table}

The metrics for parser performance are as follows:

\begin{itemize}
\item
  Precision ($P$) is the fraction of the constituents that the parser
  produces that are correct: $a/(a+b)$.
\item
  Recall ($R$) is the fraction of the correct constituents that the
  parser produces: $a/(a+c)$.
\item
  F-measure is the harmonic mean of precision and recall.  Its
  geometric interpretation is interesting. It is the ratio of the area
  of the rectangle with corners $(0,0)$ and $(P,R)$ to its perimeter,
  normalized such that the maximum value is 1.0.  To calculate:
  $2PR/(P+R)$ or $2a/(2a+b+c)$.  Qualitatively speaking, F-measure is
  the strictest single measure.
\item 
  The other measure we use to evaluate parsers is the arithmetic mean
  of precision and recall: $(P+R)/2$ or $a(2a+b+c)/2(a+b)(a+c)$.
\item 
  In some cases, when parsers are performing very well, we will
  report the percent of sentences that were parsed exactly correctly.
  That is, the number of sentences for which $b=c=0$ divided by the
  total number of sentences.
\end{itemize}

\subsection{Baselines And Oracles}
\label{section:combining:baselines}

Before we begin, it will serve us well to determine what bounds exist
on how well we can perform this task.  There are several baselines and
oracles we can study to get a feel for the difficulty of parser
combination.  Baselines are the lower bounds that we should expect to
surpass with any reasonable system, and oracles are the upper bounds
that we know we cannot surpass with our best systems.

For baselines, we have

\begin{itemize}
\item
  The Winner Takes All combination strategy.  This is the accuracy of
  the best individual parser.  A similar baseline was used by Samuel et
  al. investigating efficacy of committee combination
  \cite{samuel98:committeetbl}.\footnote{Instead of using the best
  individual, however, they compared to the first member added to the
  committee.}  
\item The average performance of the member parsers.  This is the same
  baseline used in Halteren's study of part of speech tagger
  combination \cite{halteren98:combine}.  It is also the constituent
  accuracy we would expect to achieve if we combined the three parsers
  by picking constituents at random from among the three.
\end{itemize}

\begin{table}[htbp]
  \begin{center}
    \begin{tabular}{|l|rr|rr|r|}
      \hline
      & \multicolumn{1}{c}{P}
      & \multicolumn{1}{c|}{R} 
      & \multicolumn{1}{c}{(P+R)/2} 
      & \multicolumn{1}{c|}{F}
      & \multicolumn{1}{c|}{Exact} \\
      \hline
      Parser1 & 85.81 & 85.63 & 85.72 & 85.72 & 28.1 \\
      Parser2 & 86.87 & 86.55 & 86.71 & 86.71 & 29.3 \\
      Parser3 & 88.73 & 88.54 & 88.63 & 88.63 & 34.9 \\
      \hline
      Average & 87.14 & 86.91 & 87.02 & 87.02 & 30.8 \\
      \hline
    \end{tabular}
    \caption{Baseline Parsing Performance}
    \label{table:combining:baseline}
  \end{center}
\end{table}

The performance of the baseline parser combination techniques is
presented in Table \ref{table:combining:baseline}.  We determine the
performance of the average parser by first summing the error
distribution tables for the three parsers as in Table
\ref{table:combining:situations}, then calculating the various metrics
on the resulting table.  The exact sentence accuracy is the average of
exact sentence accuracies of the three parsers. The Winner Takes All
strategy corresponds to the Parser3 row in the table.  The precision
and recall differences between Parser3 and the other parsers are
significant based on a binomial hypothesis test with $\alpha=0.01$.
The set on which these numbers were generated had 44177 constituents
in 2416 sentences.


For oracles, we have

\begin{itemize}
\item
  The parser combiner that picks the best parser for each sentence.
  We call this the Parser Switch Oracle.
\item
  The parser that picks exactly those constituents suggested by the
  member parsers that are found in the correct parse.  This parser
  always gets 100\% precision, and we call it the Maximum
  Precision Oracle.
\end{itemize}

\begin{table}[htbp]
  \begin{center}
    \begin{tabular}{|l|rr|rr|r|}
      \hline
      & \multicolumn{1}{c}{P} 
      & \multicolumn{1}{c|}{R} 
      & \multicolumn{1}{c}{(P+R)/2} 
      & \multicolumn{1}{c|}{F} 
      & \multicolumn{1}{c|}{Exact} 
      \\
      \hline
      Maximum Precision Oracle & 100.00 & 95.41 & 97.70 & 97.65 & 64.5\\
      Parser Switch Oracle     & 93.12  & 92.84 & 92.98 & 92.98 & 46.8 \\
      \hline
    \end{tabular}
    \caption{Oracle Parsing Performance}
    \label{table:combining:oracle}
  \end{center}
\end{table}

\begin{figure}[htbp]
  \begin{center}
    \epsfig{file=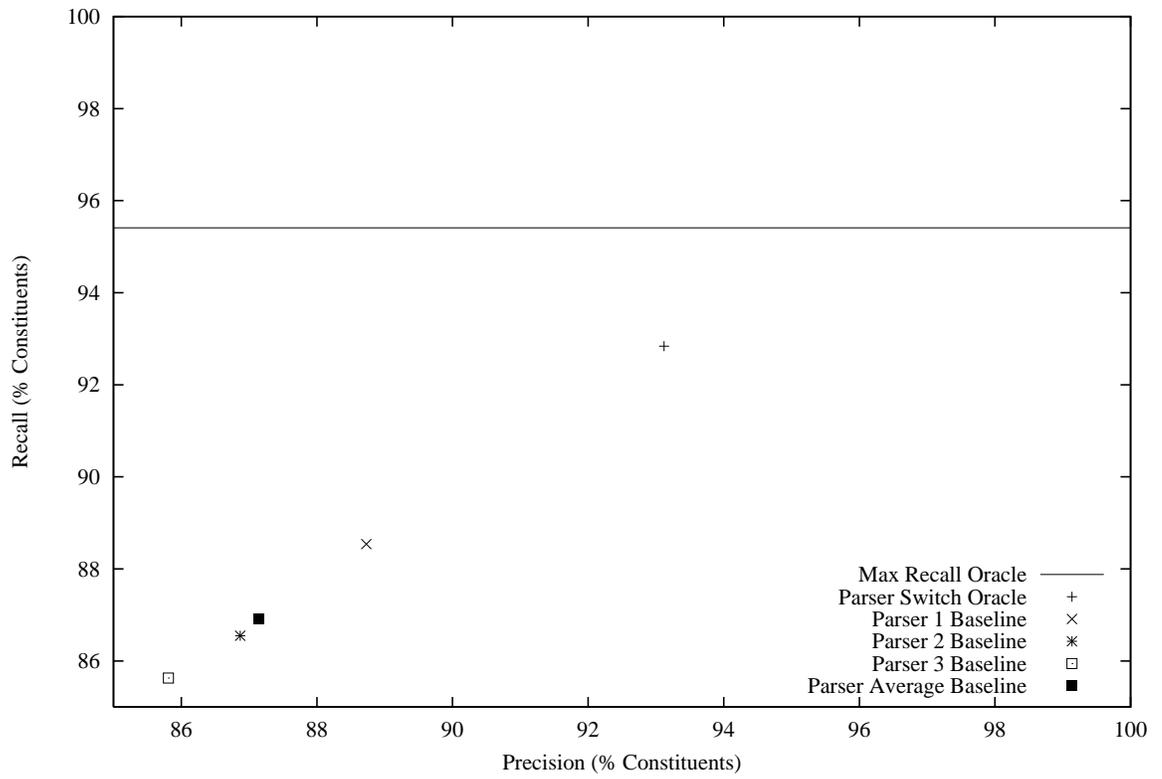, width=1.0 \textwidth}
  \end{center}
  \caption{Bounds on Combination Performance}
  \label{fig:combining:bounds}
\end{figure}

The performance of the oracle parser combination techniques is
presented in table \ref{table:combining:oracle}.  All of the bounds
discussed in this section are presented pictorially in Figure
\ref{fig:combining:bounds}.  It is a precision versus recall plot in
which each parser is represented by a single point.  Notice that if we
could pick exactly the correct constituents from those hypothesized by
the three parsers we could get 95.41\% recall.  We are missing less
than 5\% of the constituents from the set.  
Furthermore, if we could just pick the best parser
for each sentence, but still keep the bad predictions the parser makes
in that sentence we would move to near 93\% precision and recall.
These bounds are well over the state of the art and they encourage us
that we have a lot of room for growth.  However, people in the parsing
community typically feel there is a ceiling of 95-97\% precision and
recall using this dataset \cite{charniak97:aimag,marcus:parsingbound}.

In Table \ref{table:combining:missedconstituents} we show the
distribution of constituent labels in a test set, as well as the
distribution of constituent labels from the subset of that set that
none of the three parsers correctly predicted.  This is the
distribution of recall errors for the maximum precision oracle.  From
this we see that the constituents labelled {\tt S}, {\tt NP} and {\tt
  VP} are covered by the parsers disproportionately with respect to
constituents with the other labels.  Alternatively, this could be an
artifact of noun phrases and verb phrases being more consistently
annotated in the corpus than the other types of constituents.

\begin{table}[htbp]
  \begin{center}
    \begin{tabular}{|c|rr|rr|}
      \hline 
      & \multicolumn{2}{|c|}{Training Set}
      & \multicolumn{2}{|c|}{Recall Errors}
      \\
      \cline{2-5}
      Label & Percent & Count &  Percent & Count \\
      \hline
      ADJP & 2.02 & 891 & 7.40 & 150 \\
      ADVP & 2.75 & 1213 & 4.79 & 97 \\
      CONJP & 0.05 & 21 & 0.15 & 3 \\
      FRAG & 0.11 & 49 & 1.53 & 31 \\
      INTJ & 0.02 & 11 & 0.05 & 1 \\
      NAC & 0.07 & 30 & 0.49 & 10 \\
      NP & 41.96 & 18536 & 26.21 & 531 \\
      NX & 0.27 & 121 & 5.43 & 110 \\
      PP & 12.43 & 5492 & 13.62 & 276 \\
      PRN & 0.32 & 142 & 1.14 & 23 \\
      PRT & 0.36 & 159 & 0.69 & 14 \\
      QP & 1.11 & 490 & 1.14 & 23 \\
      S & 12.81 & 5660 & 11.65 & 236 \\
      SBAR & 4.07 & 1797 & 6.76 & 137 \\
      SBARQ & 0.02 & 10 & 0.25 & 5 \\
      SINV & 0.36 & 157 & 0.64 & 13 \\
      SQ & 0.04 & 18 & 0.15 & 3 \\
      UCP & 0.07 & 32 & 1.04 & 21 \\
      VP & 19.79 & 8743 & 15.79 & 320 \\
      WHADJP & 0.01 & 4 & 0.10 & 2 \\
      WHADVP & 0.31 & 136 & 0.59 & 12 \\
      WHNP & 0.97 & 429 & 0.25 & 5 \\
      X & 0.02 & 7 & 0.15 & 3 \\
      \hline
    \end{tabular}
    \caption{Recall Error Distribution for Maximum Precision Oracle}
    \label{table:combining:missedconstituents}
  \end{center}
\end{table}

\subsection{Measuring Parser Diversity}

While the baselines and oracles place bounds on our hopes, they do 
little to suggest that we should have any hope at all of gaining 
performance by combining a specific set of parsers.  Luckily, there 
is a clue that suggests that individual parsers differ enough to be 
combined.

First, let us establish a metric for measuring the difference between 
two parsers.  Since we have structured our investigation as the 
combination of black-box parsers, we cannot look at their internals 
for describing the differences.  We can only look at how their 
differences affect their function.  In this case that means we will 
look at how the parsers bracket their output differently.

We first must describe what difference we are interested in.  In this
case we are in luck.  We are interested in how many constituents one
parser produces that a second parser misses.  More formally, let
$S_{A}$ be the set of constituents produced by parser A and $S_{B}$ be
likewise for parser B.  Our measure is given in Formula
\ref{eqn:combining:nonrecall}.

\begin{equation}
\bar{R}(A,B)=|S_{A}-S_{B}|/|S_{A}|
\label{eqn:combining:nonrecall}
\end{equation}

We call it $\bar{R}$ because when $S_{A}$ is the set of correct parse
constituents $\bar{R}$ equals $1-recall$ when recall is computed as
described in Section \ref{section:combining:measures} using $A$ as the
reference set.  In this way we can also consider a distance to the
hidden ``correct'' parser which produces the parses given in the
corpus.  This is an asymmetric metric, and its asymmetry is useful.
Each of the following three cases of interest can be detected by this
metric:

\begin{enumerate}
\item  
  Suppose parsers A and B are actually identical.  While we cannot
  determine that there does not exist some input that they will parse
  differently, we can determine the extent to which they are identical
  by $\bar{R}(A,B)$ and $\bar{R}(B,A)$.  The closer these two measures
  are to zero, the more similar the parsers.
  
\item  
  \label{item:combining:subseterrors}
  Suppose parser A always makes more mistakes than parser B, and
  moreover, parser A always makes a subset of the mistakes that parser
  B makes.  In this case we would never trust parser A over parser B,
  and it is pointless to consider combining the two.  We can detect
  this, because the following situations will hold: $\bar{R}(reference,B)
  < \bar{R}(reference,A)$, $\bar{R}(A,B) = 0$, and $\bar{R}(B,A) > 0$.  In
  short, when parser A performs better than parser B and $\bar{R}$ is
  skewed such that the value when B is the first argument is much
  greater than when when A is the first argument, then we should tend
  to believe parser A {\em in every case}.

\item  
  \label{item:combining:indeperrors}
  Suppose parser A and parser B make independent predictions.  Then
  $\bar{R}(A,B) > 0$ and $\bar{R}(B,A) > 0$ as both parsers will
  predict constituents that the other one does not.  Furthermore, if
  parser A and parser B tend to make independent mistakes,
  $\bar{R}(reference,A)$ and $\bar{R}(reference,B)$ will both be near
  the same value.  In fact, if $\bar{R}(reference,A) < \bar{R}(B,A)$
  and $\bar{R}(reference,B) < \bar{R}(A,B)$ then we can say that the
  pair of parsers are closer to the reference than they are to each
  other.
        
\end{enumerate}

\begin{table}[htbp]
  \begin{center}
    \begin{tabular}{|c|ccc|c|}
      \hline 
      \( S_{A}\backslash S_{B} \)
      & Parser1 & Parser2 & Parser3 & reference\\
      \hline 
      Parser1&0&16.87&14.91&14.18\\
      Parser2&16.73&0&13.63&13.12\\
      Parser3&14.89&13.77&0&11.26\\
      \hline 
      reference&14.36&13.44&11.45&0\\
      \hline 
    \end{tabular}
    \caption{A Directed Distance Between Parsers}
    \label{table:combining:parserdistance}
  \end{center}
\end{table}

We can see in Table \ref{table:combining:parserdistance} the values of
$\bar{R}$ for each of our parser pairs as well as the reference.  Notice
that each of the parsers differ from each other more than they differ
from the reference.  This is exactly the situation we describe in case
\ref{item:combining:indeperrors}, and it is a clue that the parsers in
question have independent errors.  Furthermore, since $(\forall A,B)
\bar{R}(A,B) \ne 0$ we can see that no parser makes a strict subset of
the predictions of the others.  This is contrary to case
\ref{item:combining:indeperrors}, and allows us to see that there is
potential for constructive combination between all pairs of these
parsers.


\section{EVALB Transformation}
\label{section:combining:evalb}


Magerman \cite{magerman95:parsing} reports results of an experimental
evaluation of a parser trained on the Penn Treebank.  He used an
evaluation system developed by Black et al.
\cite{black91:parsemetric} for comparing hand-coded parsing systems.
The statistical parsing community has followed this design in
performing evaluations.  The community has focused on the labelled
bracketing method of scoring parsers.  The algorithm has some
important ramifications for developing parser combination techniques.

Let $\pi_T$ be the correct parse, and $\pi_G$ be the hypothesized
parse.  Algorithm \ref{algorithm:combining:evalb} is the algorithm for
comparing two parsers that is in standard use in the Treebank parsing
community.\footnote{Satoshi Sekine and Michael Collins wrote a program
  for parser evaluation called {\tt EVALB} (short for EVALuating
  Brackets) which evaluates parsers using the algorithm we describe
  above.  I use this program as a reference implementation.  At the
  time of this writing, it could be found at {\tt
    http://cs.nyu.edu/cs/projects/proteus/evalb/}.}

\begin{algorithm}
{EVALB Transformation}
{}

\label{algorithm:combining:evalb}
\item Strip all epsilon productions from $\pi_T$, as most parsers do
  not generate epsilon productions.\footnotemark
  \label{algorithm:evalb:transformationbegin}
\item
  Remove all terminal nodes which are POS-tagged with some kinds of
  punctuation from both $\pi_T$ and $\pi_G$.  The punctuation we
  remove is from the ``or''-delimited set \{, or  : or `` or '' or
  .\}.
\item
  Repeatedly remove all constituents from the tree that no longer span
  any tokens from the original sentence due to the pruning we just
  performed.
  \label{algorithm:evalb:transformationend}
\item 
  Create $S_T$ from the reference parse.  This is the set of tuples
  $(s,e,l)$ where $s$ is the number of terminal nodes to the left of
  the left side of the constituent's span, $e$ is the sum of $s$ and
  the number of terminal nodes dominated by the constituent, and $l$
  is the label on the constituent.\footnotemark
  Similarly create $S_G$ from
  the hypothesized (Guess) parse.
\item 
  Remove any constituent that dominates all the other nodes in
  $S_T$.  Do the same in $S_G$.  Every sentence has a topmost
  constituent spanning it, so we need not count it.  It is taken as
  given that all parsers  produce it.
\item
  Now produce the error distribution table as in Table
  \ref{table:combining:situations} using $S_T$ and $S_G$.
\item
  We have already shown how to compute the measures of interest using
  this table.
\end{algorithm}
\addtocounter{footnote}{-1}
\footnotetext{
  Epsilon productions
  appear in the corpus to encode traces describing special
  linguistic phenomena (e.g. wh-movement).  They yield leaf nodes
  that do not correspond to observed tokens.
  }
\stepcounter{footnote}
\footnotetext{
  Some evaluations treat
  this set ($S_T$) as a multi-set because there can be chains of
  unary productions of the same label
}

There are several ramifications of this algorithm that should be
observed.
First,
 the parser may use punctuation to help perform the parse, but
 how the parser brackets punctuation has no effect on the final
 score. For example, it makes no difference where the final period
 attaches, or whether the quotes around a quotation are included in
 the constituent dominating it.  Punctuation is ignored for purely
 historical reasons.  Some of the earliest parsers represented
 punctuation as it is typed -- most often as part of an adjacent
 word, whereas others treated punctuation as separate tokens.
Second,
 the set of productions used in parsing the sentence is not restricted
 to the set found in the correct parse.  Each constituent is
 identified only by its label and span.  Its correctness does not
 depend on the labels on its children.  The parse has been simplified
 at this point to a set of triangles with labels on them.
Third,
 this algorithm has meaning for parses that are not necessarily trees.
 It works with any acyclic graph with the appropriate terminal
 nodes.
 
 Notice that steps \ref{algorithm:evalb:transformationbegin} through
 \ref{algorithm:evalb:transformationend} of the algorithm produce a
 simple graph transformation or rewrite.  We can call it the {\em
   EVALB transformation} which we write $EV(parse)$.  We can say that
 two parses are identical if their images under the EVALB
 transformation are the same.  In light of this observation, we are
 performing all of our parser combination techniques after the EVALB
 transformation takes place.  Essentially, we are inserting the
 combination techniques after step
 \ref{algorithm:evalb:transformationend} of the evaluation algorithm.

\begin{figure}[htbp]
  \begin{center}
    \epsfig{file=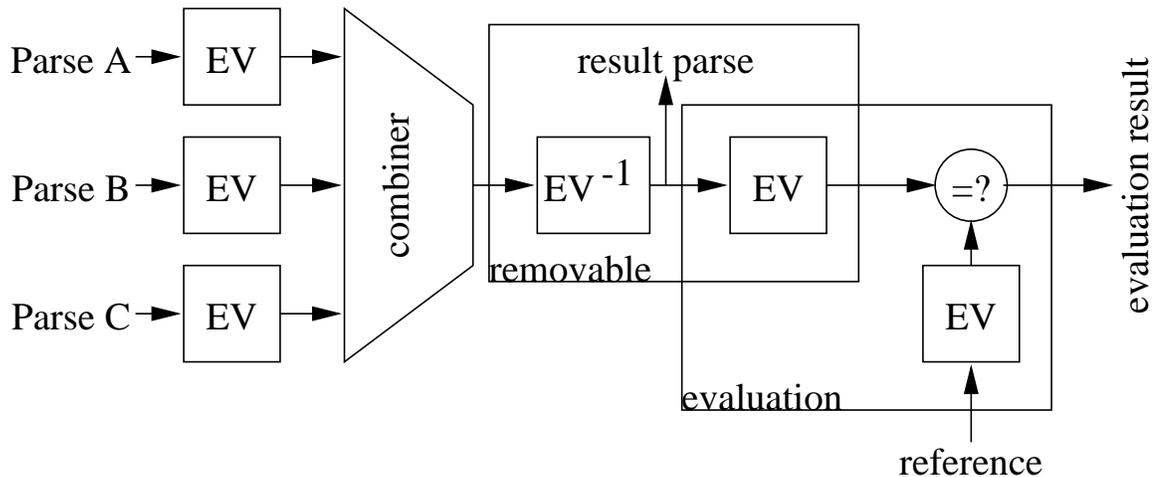, width=1.0\textwidth}
  \end{center}
  \caption{EVALB Transformation in the Combining Framework.}
  \label{fig:combining:evalb}
\end{figure}

Performing the parser combination at this point is not ``cheating''
because although the EVALB transformation is many-to-one, we can pick
an inverse transformation that inserts the punctuation back into the
result of our parser combination.  There always exists such an inverse
transformation, because we can always insert the punctuation into all
constituents that its left non-punctuation neighbor is in, or right
non-punctuation neighbor if there is no appropriate left neighbor.
Furthermore, evaluating the parse which is
$EV^{-1}(combine(EV(A),EV(B),EV(C)),p)$ (where $p$ is the punctuation
we need to replace) gives us the same results as evaluating
$combine(EV(A),EV(B),EV(C)))$ itself.  This is obvious as application
of the EVALB transformation is the first step in the parser evaluation
algorithm, but it is a technical point that is worth mentioning.  While
for the purposes of creating a parse tree for use outside our
evaluation we would use result of
$EV^{-1}(combine(EV(A),EV(B),EV(C)),p)$, for a simpler experimental
framework we use the shorter form.  This point is illustrated in
Figure \ref{fig:combining:evalb}.

The versatility of the EVALB transformation also lets us apply it to
tree-like structures with overlapping brackets and disconnected
forests in addition to typical parse trees.  As discussed in the
previous section, there are some natural language processing tasks
that can be performed with non-tree structures.  The only limitation
that the EVALB transformation puts on what structures we will allow
our combining technique to produce is that the structures must all be
valid inputs to some inverse EVALB transformation.  The result of
applying the inverse EVALB transformation must be a tree with properly
nested constituents.  This restriction was not problematic for any of
the combining strategies we explored.


\section{Non-parametric Approaches}

As mentioned earlier, the parsers we acquired were trained on the
majority of the Penn Treebank.  Only two sections remain (4116
sentences) on which we can tune and test our combining techniques for
these parsers.  This is precious little data, so we held out the
section with 1700 sentences for the final evaluation. 

Every probabilistic model is subject to two types of error: modeling
error and estimation error.  Modeling error comes from the
inadequacies of the model.  In linguistic processes the model is
hidden from us to a large extent and we have to guess at what the real
model is.  Often we knowingly make our models weak or inaccurate
because we know we do not have enough data to accurately estimate the
parameters of a better model.  Estimation error comes from our lack of
access to the true probabilities or parameters which flesh out our
model.  At worst we estimate these parameters by hand, and at best we
estimate them from counting many observed outcomes and relying on the
law of large numbers.  Herein lies a vicious dependency.  We cannot
utilize complex models without accurate probability estimates and we
can only produce accurate estimates for small parameter spaces given
our limited data.

One method of exploring the space of probabilistic models is to first
pick some reasonable non-parametric models and then add parameters to
them to make them more accurate.  In this section we explore some
non-parametric approaches.  The advantage of these approaches is that
their implementation requires no extra training data.  This is good
for our situation, as our remaining data is in short supply.

\subsection{Constituent Voting}
\label{section:combining:constvote}

We start our investigation by treating our parsers as
independently-minded democratic voters.  We require them each to vote
on whether or not each individual constituent belongs in the
hypothesized parse.  The set of candidate constituents they vote on
is the set of constituents in the union of their resulting sets.

\begin{table}[htbp]
  \begin{center}
    \begin{tabular}{|r|rr|rr|r|}
      \hline
      \multicolumn{1}{|c|}{System} 
      & \multicolumn{1}{c}{P} 
      & \multicolumn{1}{c|}{R} 
      & \multicolumn{1}{c}{(P+R)/2} 
      & \multicolumn{1}{c|}{F} 
      & \multicolumn{1}{c|}{Exact} 
      \\
      \hline
      1 Vote Required     & 77.05 & 95.41 &  86.23 & 85.25 & 18.9\\
      2 Votes Required     & 92.09 & 89.18 &  90.64 & 90.61 & 37.0\\
      3 Votes Required     & 96.93 & 76.13 &  86.53 & 85.28 & 21.3\\
      \hline
      Best Individual & 88.73 & 88.54 & 88.63 & 88.63 & 34.9 \\
      \hline
    \end{tabular}
    \caption{Democratic Voting Results}
    \label{table:combining:rawvoting}
  \end{center}
\end{table}

In Table \ref{table:combining:rawvoting} we see the results.  The row
index corresponds to the threshold we set for inclusion in the
hypothesized parse.  For example, the first row of the table is the
result we get when each constituent is required to receive at least
one vote to remain in the hypothesis.  This is the same as the union
of the three parse sets.  From this line we see that less than 5\% of
the bracketings in the Penn Treebank are not captured by one of these
three parsers.

Note that the result described by the first row does not necessarily
consist of parse trees.  It could contain crossing brackets.  While
there are still some tasks for which this output is useful, this would
cause many algorithms that take parse trees as input to require some
careful reworking.  The output can be seen as corresponding to
multiple possible parse trees when the bracketings cross.  Still, it
is an unfortunate situation which bears more investigation later in
this chapter.

The result described by the second row of the table corresponds to
well-formed parse trees as we prove in Lemma
\ref{lemma:combining:treeguarantee}, below.  Furthermore, the quality
of the combination parse requiring the simple majority vote in this
case is competitive with the results we present later in this chapter.
This result is a significant improvement over the individual parsers,
and all other parsers of this data known to date.

The third row in the table represents the parser which requires
unanimous votes for inclusion in the hypothesis.  This is the most
precise of the three parsers, and less than 4\% of the bracketings it
suggests are incorrect.

To summarize the important result of this section: we can achieve an
absolute 3.36\% gain in precision and an absolute 0.64\% gain in
recall by combining three independent parsers using a simple
non-parametric technique.  This corresponds to a relative 30\%
reduction in precision errors and a relative 6\% reduction in recall
errors.  Furthermore the technique is simple.  It does not require any
knowledge of the internal workings of these parsers, nor does it
explicitly enforce any global constraints concerning dependencies
between parse constituents.  The robustness of this technique is
explored further in Section \ref{section:combining:robustness}.

\subsubsection{Strictly More Than 50\% Vote Guarantees The Result Is A
  Tree} 

Whenever all constituents in the hypothesized parse are given strictly
more than 1/2 of the votes (e.g. 3 of 5 or 4 of 6), we are guaranteed
that the parse is a tree.  By this we mean it will have no crossing
brackets.  This is not obvious, but it is simple to prove.  Each
individual parser produces a tree and hence has no crossing brackets.
Once a constituent acquires more than 1/2 of the votes, there are more
than 1/2 of the parsers which contain that constituent.  None of those
parsers contain a crossing bracket, so no crossing bracket can have
more than 1/2 of the votes.  There are simply not enough votes
remaining to allow any crossing bracket to receive more than 1/2 of
the votes.

\begin{lemma}[Tree Guarantee]
\label{lemma:combining:treeguarantee}
If the number of votes
required by constituent voting is (strictly) greater than half of the
parsers under consideration, the resulting structure has no crossing
constituents.
\end{lemma}

\begin{proof}
 Assume a pair of crossing constituents appears
in the output of the constituent voting technique.  Each of the
constituents must have received at least $\lceil\frac{k+1}{2}\rceil$
votes from the $k$ parsers.  Let $s$ be the sum of the votes for the
assumed constituents.  $s \leq k$ because none of the parsers contains
crossing brackets so none of them vote for both of the assumed
constituents.  But by addition $s = 2\lceil\frac{k+1}{2}\rceil > k$, a
contradiction.
\end{proof}

This principle guarantees that the set of constituents that receive
any threshold number of votes where the threshold is set at 1/2 of the
parsers corresponds to a valid parse tree.  A simple non-parametric
version of this creates a hypothesis parse from all constituents
receiving a vote of more than 1/2.

\subsection{Parser Switching}
\label{section:combining:nonparametricparserswitching}

Unlike the original parsers as seen in Table
\ref{table:combining:baseline}, the result in the second row of Table
\ref{table:combining:rawvoting} does not have balanced precision and
recall.  The raw counts suggest that this combined parser
under-generates constituents when compared with the individual parsers.
The Parser Switch Oracle of Section \ref{section:combining:baselines}
has balanced precision and recall, and its performance is still well
above the raw voting.  If we could use an algorithm that utilized our
knowledge of how well raw voting works in building a parser switch,
perhaps the result would generate more constituents without
sacrificing overall performance.

We experimented with a few algorithms to produce parser switches.
There was a strikingly large performance difference between the
distance-based and similarity-based switching methods.  The
similarity-based parser switching algorithm is shown below.

\begin{algorithm}
{Similarity-based Unsupervised Parser  Switching}
{}
\item
  From each candidate parse, $\pi_i$, for a sentence create the
  constituent set $S_i$ in the usual fashion.  
\item 
  Compute the similarity score for $\pi_i$ and $\pi_j$, the
  number of constituents that match in the two parses.
\begin{equation}
    m(\pi_i,\pi_j) = |S_j \cap S_i|
\end{equation}
\item
  Switch to (use) the parser with the highest similarity to the other parses.
  Ties are broken arbitrarily. 
\begin{equation}
  \pi^*=\argmax{\pi_i}\sum\limits_{j\ne i}m(\pi_i,\pi_j)
\end{equation}
\end{algorithm}

Instead of considering the similarity between parses, we can imagine
that there exists some {\em true parse} that was modified to make all
the parses we observe from our parsers.  The process of turning that
true, hidden parse into the parses we observe is akin to Shannon's
noisy channel model \cite{shannon49}.  That true parse is modified
using simple edit operations by the removal of its structure and the
attempted recovery of that same structure by the ``noisy'' parsers. We
observe the result of this noisy channel in the hypotheses generated
by the individual parsers.  To recover the true parse we would want to
explore the space of possible parses, picking the one that minimizes
the number of editing operations required to produce all of the
observed parses.  
It is the most likely candidate to be the true
parse because it presents us with the simplest process for producing
the observed parses.
One should note however, that the space of
possible parses for a given sentence is too large to make a
straightforward exploration tractable. The number of ways to bracket a
sentence of length $n$ is the Catalan number $C(n-1)$ if we restrict
ourselves to binary branching. Since we are allowing n-ary branching
in our parses, the Catalan number is just a lower bound.  Furthermore
for each bracketing containing $n$ brackets there are $k^n$ ways to
label those brackets with nonterminal labels, where $k$ is the size of
the set of nonterminal labels.  Writing the closed-form expression or
even just the recurrence for the number of parse trees on $n$ words
with $k$ different bracketing labels is a non-trivial exercise.

The distance between a pair of parses in that space would be the cost
of editing one parse into another.  We will call that the {\em edit
 distance} or just {\em distance between parses} in the discussion
below.  The goal of our next switching algorithm is to pick from the
candidate parses the parse that is closest to the true parse by
choosing the parse that minimizes the edit distance to all of the
others.

\begin{algorithm}
{Distance-based Unsupervised Parser Switching}
{}
\item
  From each candidate parse, $\pi_i$, for a sentence create the
  constituent set $S_i$ in the usual fashion.  
\item
  The distance between $\pi_i$ and $\pi_j$ is the number of mismatched
  constituents in the two parses.
\begin{equation}
  d(\pi_i,\pi_j)   = |(S_j \cup S_i) - (S_j \cap S_i) |
\end{equation}
\item
  Switch to (use) the parse with the lowest distance to the other
  parses.  Ties are broken arbitrarily. 
\begin{equation}
  \pi^*=\argmin{\pi_i}\sum\limits_{j}d(\pi_i,\pi_j)
\end{equation}
\end{algorithm}

The relationship between the similarity and distance measures for
individual parses comes from the definitions given above.  It is shown
in Equation \ref{eqn:combining:distancesim}, where $c(\pi)$ is the
count of the number of constituents in parse $\pi$.

\begin{equation}
  \label{eqn:combining:distancesim}
  d(\pi_i,\pi_j)  =  c(\pi_i)+c(\pi_j)-2m(\pi_i,\pi_j)
\end{equation}

This leads us to a straightforward interpretation of the difference
between the similarity-based algorithm and the distance-based
algorithm.  We see that the distance-based algorithm is the same as
the similarity-based algorithm with an extra term inside the
maximization.  That term is a weight on the number of constituents in
the particular parse ($\pi_i$) that we are considering.  In essence,
it linearly penalizes the parses with more constituents.

\begin{eqnarray}
  \argmin{\pi_i}
  \sum\limits_{j}d(\pi_i,\pi_j)
  &=& 
  \argmin{\pi_i}
  \left(
    \sum\limits_{j}c(\pi_i)
    +  \sum\limits_{j}c(\pi_j)
    -2  \sum\limits_{j}m(\pi_i,\pi_j)
  \right)
\nonumber
  \\
  &=& 
  \argmin{\pi_i}
  \left(
    nc(\pi_i)
    -2  \sum\limits_{j}m(\pi_i,\pi_j)
  \right)
\nonumber
  \\
  &=&
  \argmax{\pi_i}
  \left(
    2  \sum\limits_{j\ne i}m(\pi_i,\pi_j)
    + 2m(\pi_i,\pi_i)
    -    nc(\pi_i)
  \right)
\nonumber
  \\
  &=&
  \argmax{\pi_i}
  \left(
    \sum\limits_{j\ne i}m(\pi_i,\pi_j)
    -    \frac{(n-2)c(\pi_i)}{2}
  \right)
\end{eqnarray}

Another interesting property of the distance-based algorithm is that
it can be described in terms of a bound on the optimality of the
choice we make.

\begin{lemma}[Centroid Approximation Bound]
  The parse chosen by the distance-based unsupervised parser switching
  algorithm requires no more than 2 times the number of edits that the
  optimal choice in parse space needs to be transformed into all of
  the observed candidates.
  \label{lemma:combining:centroidapprox}
\end{lemma}

\begin{proof}
  
  The technique for this proof comes from Gusfield's work on multiple
  sequence alignment, although his goal was
  to show that a particular biological sequence alignment technique
  was good under a given goodness measure \cite{gusfield-alignment}.
  
  The edit distance in question must be symmetric.  That is, it must
  take the same number of edits to transform parse A into parse B as
  it does to transform parse B into parse A.  This is reasonable,
  given that the concept of an edit includes the ability to ``undo''
  it.
  
  Also, the edit distance should submit to the {\em triangle
    inequality}.  It should be at least as easy to edit parse A into
  parse B as it is to edit parse A into parse C and then edit parse C
  into parse B.  This is also obviously reasonable.
  
  The first observation is that the centroid we've chosen is minimal
  among the choices we could make.  That is, the number of edits
  incurred by transforming it into each of the other parses is at
  least as small as the total number of edits required using each of
  the other candidate points as the centroid.  That comes from the
  decision rule we used to pick it.  Next we will relate the cost of
  editing this chosen parse into all of the other parses to the cost
  of editing the optimal parse into all of the other parses.
  Remember, the optimal parse is some parse hidden in the parse space
  that is too large to simply search.  We define $K$ to be the total
  cost of editing all parses into all other candidate parses, and we
  give a quick bound on how much work we will do using this centroid.

  \begin{figure}[htbp]
    \begin{center}
      \epsfig{file=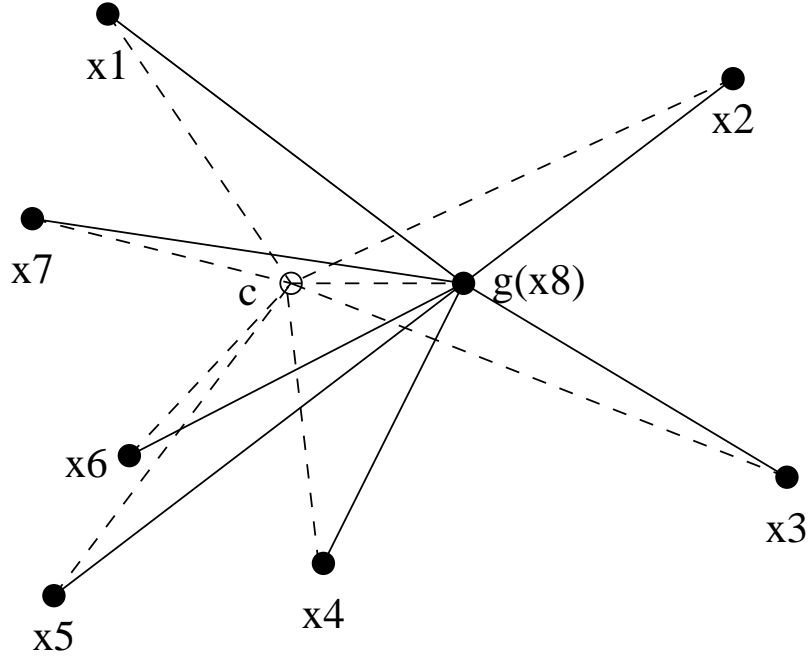, width= 0.7 \textwidth}
    \end{center}
    \caption{Edit Distances in Parse Space}
    \label{fig:combining:parsespace}
  \end{figure}
  
  A diagrammatic view of what we intend to accomplish is presented in
  Figure \ref{fig:combining:parsespace}.  The filled points are the
  parses given as input.  The point marked $c$ is the true parse,
  hidden from us unless we are willing to explore the entire space.
  The dotted lines represent the minimum possible edit distance.
  Those lengths are the cost of editing the true parse into the
  observed parses.  Point $x_8$ is also marked $g$ because it is the
  centroid chosen by minimizing the sum of pairwise distances (using
  the algorithm given).  The cost we incur by using it is represented
  by the solid lines.  We are claiming that the edit distance using
  $g$ is less than twice the edit distance using $c$.

\begin{eqnarray}
  n\sum_{i}d(\pi_i,g)
  & \leq & 
  \sum_{i}\sum_{j}d(\pi_i,\pi_j)
  \nonumber
  \\
  K 
  & \doteq &
  \sum_{i}\sum_{j}d(\pi_i,\pi_j)
  \nonumber
  \\
  \sum_{i}d(\pi_i,g)
  & \leq & 
  \frac{K}{n}
  \label{eqn:combining:centroidupper}
\end{eqnarray}

The next observation of interest is that even the optimal choice for a
centroid must obey the triangle inequality.  The true parse, the best
parse in parse space, is denoted here by $c$.

\begin{eqnarray}
  \sum_{i}\sum_{j}d(\pi_i,\pi_j)
  & \leq & 
  \sum_{i}\sum_{j}\left(d(\pi_i,c)+d(c,\pi_j)\right)
\nonumber
  \\
  &  = &
  2(n-1)\sum_{i}d(\pi_i,c)
\nonumber
  \\
  \sum_{i}d(\pi_i,c) 
  & \geq &
  \frac{K}{2(n-1)}
  \label{eqn:combining:centroidlower}
\end{eqnarray}

Now we have bounded our hypothesis, $g$, from above with respect to
$K$, and the optimal parse, $c$, from below with respect to $K$.
This gives us a way to bound the extra cost we incur by using this
suboptimal choice using simple substitution from Equations
\ref{eqn:combining:centroidupper} and
\ref{eqn:combining:centroidlower}.

\begin{equation}
  \label{eqn:combining:centroidbound}
  \frac{\sum_{i}d(\pi_i,g)}{\sum_{i}d(\pi_i,c) }
  \leq
  \frac{2(n-1)}{n}
  < 
  2
\end{equation}

In Equation \ref{eqn:combining:centroidbound} we see that the number
of edits required to change our hypothesis, $g$, into each of the
other parses is less than twice the number of edits required to change
the optimal centroid hypothesis, $c$, (from the space of all parses)
into the observed parses.  We take this to be a reassuring bound on
this approximation, as it was unlikely we could explore the space of
parses to find $c$ in the first place. 
\end{proof}

The bound that we have just derived is interesting theoretically, but
we cannot measure its behavior empirically because we are not able to
find the optimal centroid hypothesis for comparison with the candidate
that is picked.  We did, however perform an experiment to address the
effect of this heuristic.  Consider picking the worst candidate for
the centroid approximation instead of the best.  The result for using
that method is given under the entry {\tt bad distance} in Table
\ref{table:combining:rawswitch}.  Picking a centroid at random is the
same as picking a parser at random, so that result would be
approximately the same as the average individual parser accuracy.  In
short, we see that picking according to the heuristic with the
provable bound gives significantly better results than these other
(admittedly weak) techniques.

\begin{table}[htbp]
  \begin{center}
    \begin{tabular}{|r|rr|rr|r|}
      \hline
      \multicolumn{1}{|c|}{Technique}
      & \multicolumn{1}{c}{P} 
      & \multicolumn{1}{c|}{R} 
      & \multicolumn{1}{c}{(P+R)/2} 
      & \multicolumn{1}{c|}{F} 
      & \multicolumn{1}{c|}{Exact} 
      \\
      \hline
      Best Individual & 88.73 & 88.54 & 88.63 & 88.63 & 34.9 \\
      \hline
      similarity          & 89.50 & 89.88 & 89.69 & 89.69 & 35.3 \\
      distance            & 90.24 & 89.58 & 89.91 & 89.91 & 38.0 \\
      \hline
      bad distance        & 82.70 & 82.81 & 82.75 & 82.75 & 20.9 \\
      average performance & 87.14 & 86.91 & 87.02 & 87.02 & 30.8 \\
      \hline 
    \end{tabular}
    \caption{Non-parametric Parser Switching}
    \label{table:combining:rawswitch}
  \end{center}
\end{table}

Combining with these algorithms produces the results in Table
\ref{table:combining:rawswitch}.  The similarity switching parser is
a better parser than any of the individual parsers and it gets higher
recall than combining the parsers with constituent voting.  However,
the loss of precision makes the overall performance suffer.  The
distance switching parser is significantly better at precision and
exact sentence accuracy than the similarity switching parser.  The
loss it incurs in recall is significant, but it is more than offset by
the gain in precision, as we can see by the significantly different
F-measure.  We can see that the penalty the distance measure places on
sentences with more constituents is appropriate in this case, as it
correctly penalizes the parses that over-generate.

One of the main advantages of the parser switching framework is that
the final predictions are as useful as the input because they maintain
all the constraints that the input parses maintain.  There are no
crossing brackets, and as long as the switching algorithm is
reasonably unbiased the trees are as dense as the input trees.  If
there are limits on the productions available for the parsers and the
input parsers obey this limit, then we can guarantee our output will
have the same guarantee.  This can be important, for example, if we
are dealing with a translation grammar that is specified as operations
on productions in the grammar, or if we have partial database queries
or other semantic information associated with the nodes in the parse
tree.  Maintaining an entire tree intact allows us to guarantee that
we do not invalidate the translation or the database query in the
process of producing a better hypothesis.

The secondary advantage we will see later is that it performs better
at getting sentences exactly correct than the hybridization methods of
constituent voting and na\"ive Bayes constituent combination.

\subsection{Parse Tree Alignment}

We have observed that parsing using a simple edit distance between
parses proportional to the number of mismatched constituents gives us
good results.  There is no reason to believe that this particular
choice of edit distance is the best one, though.  In this section we
explore other edit distances, and provide a general technique for
editing complete parses using arbitrary (but constrained) costs of
editing constituents.

The  edit distance based on mismatched constituents is very
coarse-grained.  It allows no partially-matched constituents, which we
might desire.  Consider the sentence:

\begin{center}
 He mowed the grass down.
\end{center}

There are at least two acceptable parses for this sentence based on
different interpretations of the word {\tt down}.  In Parse
\ref{parse:combining:align1}, the man is mowing down the grass,
probably with a lawn mower, but perhaps with an automatic rifle.  In
Parse \ref{parse:combining:align2}, the man is mowing something that
is a cross between grass and soft fine feathers.  If we keep only the
matching constituents from those two parses, we get the structure in
Parse \ref{parse:combining:align:matchresult}.  It gives no hint that
the verb is transitive and there is very likely a noun phrase included
inside the verb phrase.  We have lost some information from these
hypotheses that we would like to preserve.

\begin{parse}
  \Tree
  [.S [.NP He ]
  [.VP mowed 
  \qroof {the grass}.NP 
  down ] ]
  \label{parse:combining:align1}
\end{parse}

\begin{parse}
  \Tree
  [.S [.NP He ]
  [.VP mowed  
  \qroof{the grass down}.NP ] ]
  \label{parse:combining:align2}
\end{parse}

\begin{parse}
  \Tree
  [.S [.NP He ] 
  \qroof{mowed the grass down}.VP ]
  \label{parse:combining:align:matchresult}
\end{parse}

If a third parser produced Parse \ref{parse:combining:align3}, we
would feel very confident that $grass$ is part of a noun phrase inside
the verb phrase, {\em even if we had no other knowledge of English}.
Keeping only the matching constituents from any pair, or all of the
parses (\ref{parse:combining:align1}, \ref{parse:combining:align2},
and \ref{parse:combining:align3}), we still arrive at Parse
\ref{parse:combining:align:matchresult}.  This is precisely because
the matched constituent edit distance does not differentiate in any
way among the differences between these parses.  The distance between
any pair under this metric is exactly two edits: one constituent must
be removed, and one inserted.

\begin{parse}
  \Tree
  [.S [.NP He ]
  [.VP mowed the \qroof{grass down}.NP ] ]
  \label{parse:combining:align3}
\end{parse}

The only way we should prefer Parse \ref{parse:combining:align2}
(which we do), is if it is cheaper to edit it into both Parses
\ref{parse:combining:align1} and \ref{parse:combining:align3} than it
is to edit them into each other.

We have found a set of constituents that should have been edited in a
way that yields an intuitive cost structure that does not match the
reality of the distance measure we are using.  It seems that it should
be easy to work out a distance that is compatible with our intuition
on a constituent-by-constituent basis.  To this end we will describe a
novel method for utilizing a given constituent-by-constituent editing
cost function for computing an edit distance (and alignment) between
complete parses.

Consider the relationship between alignment and editing.  By alignment
we mean a relation between the sets of constituents in two parses.  In
practical terms, an alignment describes a mapping between constituents
in one parse and constituents in another parse, where any particular
constituent needs not be mapped.

Each alignment corresponds to editing one set of constituents into
another.  Constituents that are not mapped (in the relation) are said
to be {\em insertions} or {\em deletions} depending on which way the
editing operation is being viewed.  All of the rest of the nodes are
{\em substitutions}, one (or many) for the other.  In this way we can
view the elements of the relation together with the constituents
missing from the relation as editing operations.  Several facts
quickly become clear:

\begin{itemize}

\item 

  For each alignment there is a unique editing cost.  That is the
  sum of the cost of substituting the constituents in the relation
  together with the cost of inserting the constituents not involved in
  the relation.

\item 

  Depending on the editing cost function, there may be many
  alignments that produce the same editing cost between sets of
  constituents.  We need only give an example to prove this.  Consider
  the distance function we gave earlier, mismatched constituents.  If
  we have two constituents on the left hand side that match a single
  constituent on the right hand side, it will be cheapest to align a
  pair of them, and the remainder remains unaligned.  The choice we
  make in picking which constituent from the pair yields our proof.

\item 

  The minimal edit distance between sets of constituents can be
  proven by showing an alignment for the set whose cost is the edit
  distance.  The alignment is a certificate for the edit distance.

\item

  Verifying an alignment associated with an edit distance is a
  polynomial undertaking because verifying the alignment itself is
  polynomial (given that the edit cost function is polynomial).
  Unless there is some algebraic shortcut, verifying a minimal edit
  distance will require us to find an alignment.  For this reason it
  is typically considered more prudent (and possible) to set out to
  find the minimal alignment first, instead of looking for shortcuts
  to computing a minimal edit distance.

\end{itemize}




Below we will give a polynomial algorithm for finding minimum-cost
alignments with a few constraints on the edit cost function via a
reduction to finding a minimum weight edge cover of a bipartite graph.

Both Oflazer \cite{oflazer96:treematch} and Calder
\cite{calder97:aligntrees} have previously presented techniques for
aligning linguistic trees.  Oflazer's technique first converts the
tree representation into a list of paths from the root to the leaves
of the tree.  It then compares those path lists using standard dynamic
programming approaches to computing edit distance.  The motivation for
his approach is computing approximate match between trees to
facilitate database search.  It is not clear that the induced
alignment between the path lists represents simple edit operations on
trees.

Calder's technique for aligning trees is a bottom-up exact match
strategy.  A correspondence between the yields of two trees is made,
and once grounded on that map between yields, the constituents can be
compared by comparing their yields.  This technique allows no
partial constituent matches, and is well suited to producing
alignments with the goal of comparing parses to a reference corpus.

Our work is significantly different from Calder's in that we are not
requiring aligned constituents to be strictly nested one inside the
other. We are implicitly ignoring the global structure in picking
aligning constituents, and we explore many distance measures between
constituents.  Furthermore, our algorithm is arrived at from a
different set of constraints than Calder's.  The work is different
from Oflazer's tree-matching algorithm in that this work is not
performing an approximate match.  We are directly minimizing the
metrics we show.  Our representation is different from Oflazer's, as
well.  We use a bag of constituents, and he uses a vertex list
sequence.

\subsubsection{Constraints and Formalities}

We assume we are given a well-defined distance (edit cost) between
constituents. By well-defined we mean:

\begin{itemize}
\item The edit distance be strictly positive, $d(X,Y)\geq 0$.
  Negative costs for edits are meaningless.
\item The distance must be conservative, $d(X,X)=0$.  There is no
  editing cost required to leave a constituent unedited.
\item The distance must be symmetric, $d(X,Y)=d(Y,X)$.
  Editing is naturally a symmetric operation, as an insertion into on
  parse is equivalent to a deletion from the other.  Also,
  substitutions should cost the same amount regardless of their
  directionality.
\item The distance must handle insertions and deletions by recognizing
  the $NULL$ constituent, $d(X,NULL)$.  The cost of deleting a
  constituent $X$ is $d(X,NULL)$.  To preserve symmetry we must
  likewise constrain the cost of inserting a new constituent to
  $d(X,NULL)$.
\item We do not want to constrain the distance to prevent constituents
  from moving large distances, or even outside of parenting
  constituents. 
\end{itemize}

We will give the parse editing (alternatively alignment) process some
liberty, especially in light of the constraints imposed by the edit
distance criteria.  Our requirements for the parse alignment are:

\begin{itemize}
\item Each left side constituent must map to zero, one, or more right
  side constituents.  The mapping to zero constituents will be
  indicated by an alignment with $NULL$.
\item We want to recover the cheapest alignment, corresponding to the
  cheapest sequence of constituent edits producing the right side
  parse from the left side parse.
\item The edit distance associated with the alignment must be
  symmetric and obey the triangle inequality.  This is so we can use
  it in conjunction with the distance-based parser switching algorithm
  and still enjoy the good performance bounds from 
  Lemma \ref{lemma:combining:centroidapprox}.
\end{itemize}

\subsubsection{Edge Covering Weighted Bipartite Graphs}

\begin{algorithm}
{Aligning Parses by Aligning Constituents}
{}
\item
  From the two parses, $\pi_i$ and $\pi_j$, for a sentence create the
  constituent sets $S_i$ and $S_j$ in the usual fashion.  
\item
  Add the distinguished element, $NULL$, to each of the constituent
  sets.
\item 
  Create a bipartite graph, $G=(V,E)$, with bipartition $(S_i,S_j)$
  such that $E\subseteq (S_i\times S_j)$.
\item
  Let each edge be weighted by the cost of editing between its
  endpoints into each other:  $w(v_1,v_2)=d(v_1,v_2)$.
\item
  Convert to linear program.
  We want to find $a_{ij}$ to minimize
\begin{equation}
  \sum_{(v_i,v_j)\in E} a_{ij} w(v_i,v_j)
\end{equation}
  subject to the constraints that the vertices must be covered by at
  least one incoming or outgoing edge:
\begin{eqnarray}
  (\forall v_i) \sum_{v_j\in S_j} a_{ij} \geq 1 
  \\
  (\forall v_j) \sum_{v_i\in S_i} a_{ij} \geq 1
\end{eqnarray}
\item 
  Those edges for which the corresponding $a_{ij}=1$ are the ones
  included in the final alignment.
\end{algorithm} 

It is well known that solving these types of linear programs results
in integer (binary in this case) weights on the edges
\cite{groetschel:combinatorics}.  In these cases we are solving what
looks like an NP-complete integer programming problem using available
polynomial algorithms for linear programming.\footnote{We used the
  freely available simplex-based linear programming package written by
  Michel Berkelaar, LP\_SOLVE, to solve these problems.  It is
  available from {\tt ftp://ftp.es.ele.tue.nl/pub/lp\_solve}.  While
  there exist cases for the simplex method that make it worst-case
  non-polynomial, we had no difficulties in using it in our
  experiments.}

Recall that linear programming is a technique for maximizing or
minimizing a linear function subject to a convex set of constraints.
The simplex algorithm is a worst-case exponential time algorithm for
solving linear programming instances, but the bad cases are rare.  The
simplicity of the implementation of the simplex algorithm makes it the
algorithm of choice for most linear programming applications, even
though there exist theoretically better (polynomial) algorithms for
finding solutions.  Furthermore, the bad cases for the simplex
algorithm are rare.  In our experiments, use of the simplex algorithm
was not a bottleneck.  It is much faster than the individual parsers
we combined.

The alignment produced by the algorithm yields an edit distance
between the two parses equal to the value of the resulting cover,
$\sum\limits_{a_{ij} = 1} a_{ij} d(v_i,v_j)$.

\begin{figure}[htbp]
    \begin{center}
      \epsfig{file=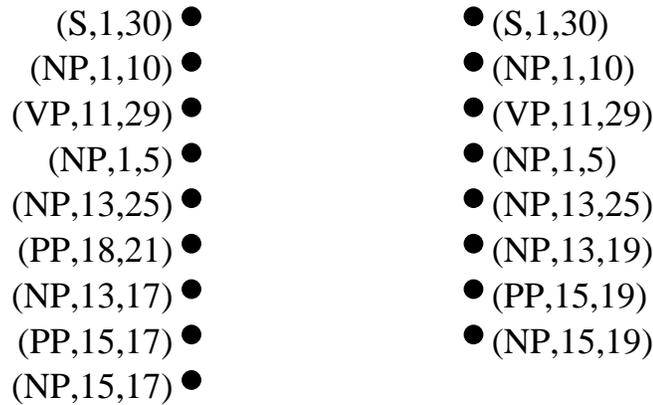, height= 0.25 \textheight}
    \end{center}
    \caption{Alignment -- Two Parses As Bipartite Graph}
    \label{fig:combining:align:1}
  \end{figure}

\begin{figure}[htbp]
    \begin{center}
      \epsfig{file=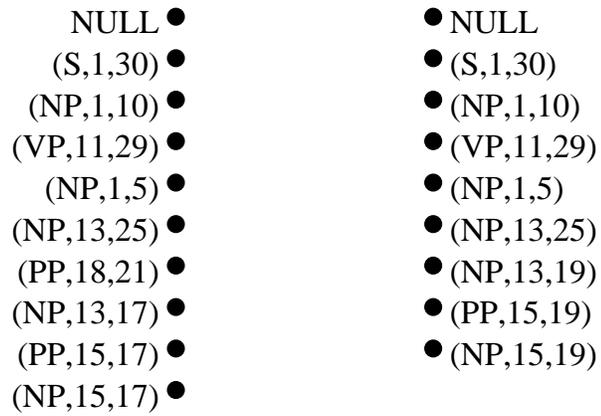, height= 0.25 \textheight}
    \end{center}
    \caption{Alignment -- Adding Null Nodes}
    \label{fig:combining:align:2}
  \end{figure}

\begin{figure}[htbp]
    \begin{center}
      \epsfig{file=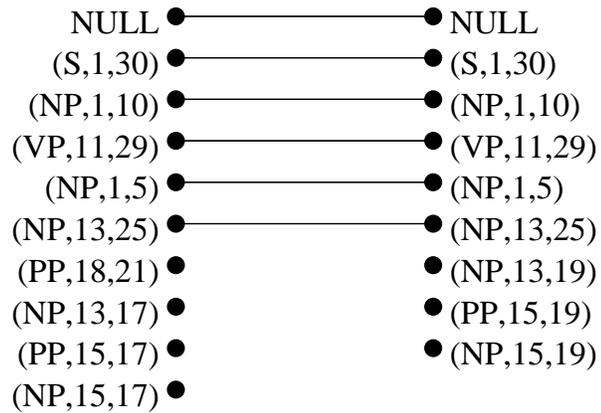, height= 0.25 \textheight}
    \end{center}
    \caption{Alignment -- Exact Matches Aligned}
    \label{fig:combining:align:3}
  \end{figure}

\begin{figure}[htbp]
    \begin{center}
      \epsfig{file=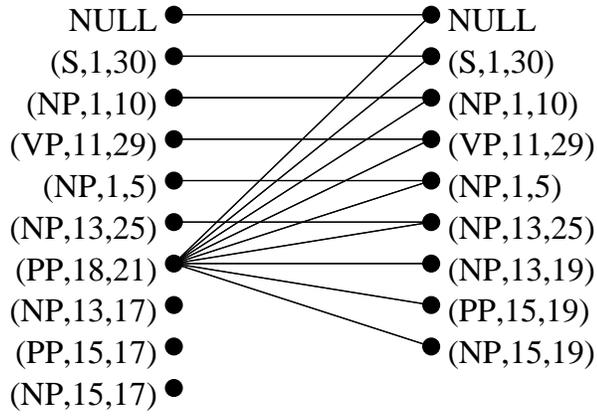, height= 0.25 \textheight}
    \end{center}
    \caption{Alignment -- Unaligned Nodes Are Fully Connected}
    \label{fig:combining:align:4}
  \end{figure}

\begin{figure}[htbp]
    \begin{center}
      \epsfig{file=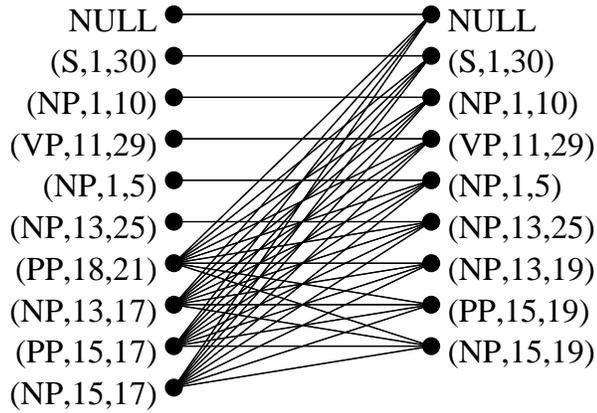, height= 0.25 \textheight}
    \end{center}
    \caption{Alignment -- Remaining Forward Edges}
    \label{fig:combining:align:5}
  \end{figure}

\begin{figure}[htbp]
    \begin{center}
      \epsfig{file=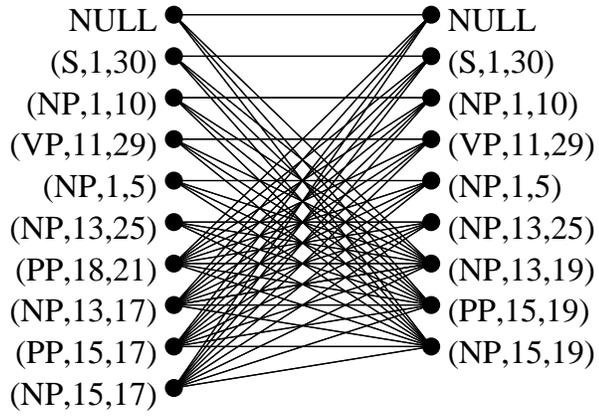, height= 0.25 \textheight}
    \end{center}
    \caption{Alignment -- Remaining Reverse Edges}
    \label{fig:combining:align:6}
  \end{figure}

\begin{figure}[htbp]
    \begin{center}
      \epsfig{file=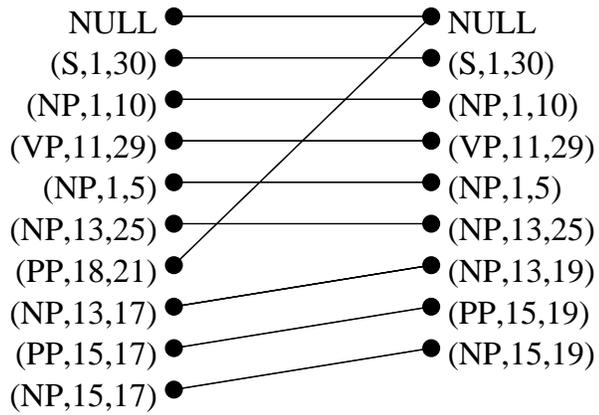, height= 0.25 \textheight}
    \end{center}
    \caption{Alignment -- Result of Linear Program}
    \label{fig:combining:align:7}
  \end{figure}

Figures \ref{fig:combining:align:1} through
\ref{fig:combining:align:7} depict the steps of the algorithm as it
would be run on an artificial but realistic example.  The edge
weights are omitted for aesthetic purposes.

\subsubsection{Constituent Edit Distances}

To explore the utility of the alignment algorithm, we require a
test bed of constituent alignment distances.  We experimented with many
different distances, and creating or choosing one remains an art.  We
may have implicitly over-fit our model to the development test data
through our experimentation.  This is one of the main shortcomings of
non-parametric methods, and it is very hard to avoid if one wants to
do any exploration of non-parametric methods in an empirical research
setting.  The results on the separate test set are presented in
section \ref{section:combining:testset}, and that evaluation provides
a sanity check on the methods.

Here we will list a sampling of the distances we used, together with a
brief description of each one.  There are some notational issues to
discuss first, though.  If $X$ is a constituent, then we denote its
label by $X_l$.  Its left index is $X_i$ and its right index is
$X_j$.  If the constituent $X$ matches constituent $Y$ in all three of
these features, we say $X=Y$.  Individual predicates are conjunctively
joined with ``,'' and disjunctively joined with ``or''.

Finally, we must comment on our use of $\infty$ in the distance
measures.  Linear programs and linear programming packages typically
require finite, real-valued weights.  In order to accommodate this in
a practical manner, we replaced the $\infty$ value in these distances
with a number larger than the weight of any possible alignment
excluding an $\infty$.  We could bound the value by simply summing the
weights on all the finite-weighted edges and doubling it.  That value
substituted for $\infty$ was large enough that we would notice them as
spurious output when the program was run.  As expected, no
``infinite''-valued edge was ever chosen as an edge for an alignment.

\begin{equation}
  d_{Kronecker}(X,Y)=\left\{
      \begin{array}{rl}
        0 & X=Y
        \\
        1 &  X\neq Y \mbox{, }X=NULL \mbox{ or } Y=NULL
        \\
        \infty & X\neq Y \mbox{, } X\neq NULL \mbox {, } Y\neq NULL
      \end{array}
      \right.
    \label{eqn:combining:align:kronecker}
\end{equation}

The first distance, named Kronecker after the Kronecker delta
function, is given in Equation \ref{eqn:combining:align:kronecker}.
The value for this alignment is the number of mismatched constituents,
as they will each be aligned to $NULL$ with a cost of 1.

\begin{equation}
  d_{piecewise}(X,Y)=\left\{
      \begin{array}{rl}
        0 & X=Y
        \\
        2 & X\neq Y\mbox{, }X=NULL \mbox{ or } Y=NULL
        \\
        3& \mbox{only one of}
        \left\{
          \begin{array}{l}
            X_i \neq Y_i
            \\
            X_j\neq Y_j
            \\
            X_l\neq Y_l
          \end{array}
        \right.
        \\
        \infty & \mbox{otherwise}
      \end{array}
      \right.
    \label{eqn:combining:align:piecewise}
\end{equation}

In Equation \ref{eqn:combining:align:piecewise} we have a distance
that is similar to the Kronecker distance, except that it allows a
pair of constituents to be aligned to each other if they differ in
exactly one feature: label, left index, or right index.  The cost of
such a matching is 3, versus a cost of 4 to align each of the
constituents to the corresponding $NULL$ (a cost of 2 for each of the
mismatched pair).

\begin{equation}
  d_{looselabel}(X,Y)=\left\{
      \begin{array}{rl}
        0 & X=Y
        \\
        2 & X\neq Y \mbox{, }X=NULL \mbox{ or } Y=NULL
        \\
        3& X_i = Y_i \mbox{, } X_j = Y_j \mbox{, } X_l\neq Y_l
        \\
        \infty & X_i\neq Y_i \mbox{ or } X_j \neq Y_j
      \end{array}
      \right.
    \label{eqn:combining:align:looselabel}
\end{equation}

The looselabel distance given in Equation
\ref{eqn:combining:align:looselabel} is similar to the piecewise
distance, except only the label on the constituent is allowed to
mismatch.

\begin{equation}
  d_{linear}(X,Y)=\left\{
      \begin{array}{rl}
        0 & X=Y
        \\
        X_j-X_i & Y=NULL
        \\
        Y_j-Y_i & X=NULL
        \\
        \infty & X_l \neq Y_l
        \\
        \infty & X_i \neq Y_i, X_j\neq Y_j
        \\
        |X_j-Y_j|+|X_i-Y_i| &\mbox{otherwise}
      \end{array}
      \right.
    \label{eqn:combining:align:linear}
\end{equation}

The linear edit distance given in Equation
\ref{eqn:combining:align:linear} is an attempt to penalize editing
constituents that have wildly different spans into each other.  It
does so by assigning a cost to editing constituents proportional to
the difference in spans between the constituents.  It also requires
that the labels on the constituents match, as well as at least one
edge.  This is not the first distance we tried that introduced linear
penalties for editing constituents, but it was one of the better ones.

\begin{equation}
  d_{stringent}(X,Y)=\left\{
      \begin{array}{rl}
        0 & X=Y
        \\
        2 & Y=NULL \mbox{ or } X=NULL
        \\
        \infty & X_i \neq Y_i\mbox{, }X_j\neq Y_j
        \\
        \infty & X_l \neq Y_l\mbox{, }(X_i \neq Y_i \mbox{ or }X_j\neq Y_j)
        \\
        3 & X_l\neq Y_l , X_i =Y_i , X_j = Y_j
        \\
        3(|X_j-Y_j|+|X_i-Y_i|) &\mbox{otherwise}
      \end{array}
      \right.
    \label{eqn:combining:align:stringent}
\end{equation}

The stringent distance allows constituent labels to mismatch if the
spans are the same, and it allows one edge of the span to mismatch if
the label and the other edge of the span is the same.  Still, it
does not allow the span to mismatch by more than one token.

\begin{table}[htbp]
  \begin{center}
    \begin{tabular}{|r|rr|rr|r|}
      \hline 
      \multicolumn{1}{|c|}{Distance}
      & \multicolumn{1}{c}{P} 
      & \multicolumn{1}{c|}{R} 
      & \multicolumn{1}{c}{(P+R)/2} 
      & \multicolumn{1}{c|}{F} 
      & \multicolumn{1}{c|}{Exact} 
      \\
      \hline
      linear     & 90.04 & 89.39 & 89.71 & 89.71 & 38.0\\
      piecewise  & 90.17 & 89.55 & 89.86 & 89.86 & 38.0\\
      Kronecker  & 90.22 & 89.55 & 89.88 & 89.88 & 37.9\\
      loose label& 90.26 & 89.63 & 89.95 & 89.95 & 38.3\\
      stringent  & 90.27 & 89.63 & 89.95 & 89.95 & 38.3\\
      \hline
      \hline
      Best Individual & 88.73 & 88.54 & 88.63 & 88.63 & 34.9 \\
      \hline
    \end{tabular}
    \caption{Parser Switching Using Centroid Approximation}
    \label{table:combining:aligncentroid}
  \end{center}
\end{table}

In Table \ref{table:combining:aligncentroid} we see the result of
performing distance-based parser switching using the alignment cost
produced us the various constituent edit distances.  The leftmost
column indicates the constituent edit distance that was used in
conjunction with the alignment and distance algorithms.  The other
columns are the same as in the other performance tables.  The
difference between the piecewise and Kronecker models is not
significant, and neither is the difference between the loose label and
stringent systems.

It appears that the loose label distance is the best one to use for
aligning Treebank parses.  This could be because there are constituent
labels in the Treebank that behave similarly enough that interchanging
them does not make a big difference on resolving syntactic ambiguity.
Another reason could be that the weaker parsers might be good at
finding the spans for constituents but not as good at labelling them.

The performance difference between the Kronecker system and the
previously discussed distance-based parser switching algorithm using
mismatched constituents is a result of the two programs breaking ties
in a different (arbitrary) way.  The two algorithms are equivalent in
analysis otherwise.  We can see from this how little the arbitrary
tie-breaking affects performance.  Ties were broken in the same manner
for all of the systems in Table \ref{table:combining:aligncentroid}.

\subsubsection{The Consensus Parse}

This method for approximating centroids in the parse space can also be
used as a first step for building a new kind of consensus parse,
similar to constituent voting.

Given these alignments between pairs of parses and a threshold $t$, we
can build an ad hoc hybrid parse in the following way:

\begin{algorithm}
{Consensus Parse from Pairwise Alignments}
{Input: Bipartite alignment graphs and cost threshold $t$ for deciding
  when to stop hypothesizing constituents.}  
\item Initialize $C$ to the empty set
  and $G$ as the obvious union of the bipartite alignment graphs.
\item
  Merge all $NULL$ nodes in $G$.
\item 

  For each constituent $c$ in each parse, compute the cost $f(c)$ to
  edit that constituent into each of its neighbors $N(c)$ given by the
  alignments.
\begin{equation}
  f(c) = \sum_{c'\in N(c)} d(c,c')
\end{equation}
\item Let $c^* =\argmin{c\in V(G),c\neq NULL} f(c)$
\item If $(f(c^*) > t)$ then output the current hybrid, $C$ and
 quit.
\item $C \leftarrow C \cup \{c^*\}$.
\item Remove $c^*$ and all $c'\in N(c)$ from $G$.
\item If the graphs are empty (aside from $NULL$), output $C$ and
  quit. 
\end{algorithm}


This is an ad hoc greedy algorithm, attempting to maximize the
confidence on the constituents that are being put into the hybrid.
Typically $t$ is chosen to match the constituent editing function and
in the same manner, {\em not} by estimating it on data.

\begin{table}[htbp]
  \begin{center}
    \begin{tabular}{|r|rr|rr|}
      \hline 
      \multicolumn{1}{|c|}{Distance}
      & \multicolumn{1}{c}{P} 
      & \multicolumn{1}{c|}{R} 
      & \multicolumn{1}{c}{(P+R)/2} 
      & \multicolumn{1}{c|}{F} 
      \\
      \hline
      linear     & 87.89 &  89.27 &  88.58 &  88.58\\
      piecewise  & 92.24 &  88.83 &  90.54 &  90.50 \\
      stringent  & 92.11 &  89.13 &  90.62 &  90.60\\
      loose label& 92.10 &  89.15 &  90.63 &  90.60\\
      Kronecker  & 92.09 &  89.18 &  90.64 &  90.61 \\
      \hline
      \hline
      Best Individual & 88.73 & 88.54 & 88.63 & 88.63 \\
      \hline
    \end{tabular}
    \caption{Parser Switching Using Consensus Approximation}
    \label{table:combining:alignconsensus}
  \end{center}
\end{table}

\subsubsection{Limitations}

We build the consensus in this ad hoc, greedy fashion because we must
work with pairwise alignments.  Multiple alignments of this sort are
intractable as we add parsers, and it is not clear what goodness
measure we would want to maximize in producing a multiple alignment in
the first place.  In short, edge covering $k$-partite graphs is
exponential in $k$.  This algorithm is a greedy approximation to it.


\section{Adding Parameters}

Non-parametric methods help us develop initial results and get a sense
for the feasibility of our method.  In this section we develop
parametric versions of combining by constituent voting and parser
switching.  We use few parameters in this process because we have very
little training data.  Estimating too many parameters will undoubtedly
yield a model with estimates based on insufficient statistics.

\subsection{Independent Constituents}

As in the non-parametric case, each member hands the combiner a set
of tuples of the form $(s,e,l)$ for each sentence, where $s$ is the
start index for the constituent, $e$ is the ending index, and $l$ is
the label. 

We then formulate the combination of voters as a binary classification
problem.  First we make the constituency independence assumption:
assume each constituent is independently selectable.  This is
inconsistent with the notion of a parse {\em tree}, but it can still
produce a useful structure.

For each constituent $c$ we are interested in $P(\pi(c) | M_1 \ldots
M_k)$ where $M_i$ is the random variable which takes a value from
$\{true,false\}$ depending on whether parser $i$ contains that
constituent in its final parse.  

We can use a na\"ive Bayes model \cite{dudahart} to produce an
estimate of this probability.  Na\"ive Bayes makes the assumption that
all of the random variables we condition on are independent.  This
assumption exactly matches the assumption we are making in endeavoring
to combine these parsers in the first place.  In this way the na\"ive
Bayes modeling technique is well matched to our problem.

In more detail we first uses Bayes's law to make the transformation:

\begin{eqnarray}
  P(\pi(c) | M_1\ldots M_k) 
  & = &
  \frac{P(M_1\ldots M_k |\pi(c))P(\pi(c))}
  {P(M_1\ldots M_k)}
\end{eqnarray}

Then we assume the $M_i$ variables are pairwise independent.

\begin{eqnarray}
  P(\pi(c))\frac{P(M_1\ldots M_k|\pi(c))}{P(M_1\ldots M_k)}
  & = &
  P(\pi(c))\prod_{i=1}^k{\frac{P(M_i|\pi(c))}{P(M_i)}}
  \label{eqn:combining:bayesform}
\end{eqnarray}

We can throw away the denominator because we are actually only
interested in the value of $\pi(c)$ that is larger.  We can then
transform the expression into terms we can collect from a corpus.

\begin{eqnarray}
  P(\pi(c))\prod_{i=1}^k{\frac{P(M_i|\pi(c))}{P(M_i)}}
  & = &
  P(\pi(c))\prod_{i=1}^k{P(M_i|\pi(c))}
\end{eqnarray}

\begin{eqnarray}  
  \lefteqn{  P(\pi(c)=true)\prod_{i=1}^k{P(M_i|\pi(c)=true)} = }
  \nonumber
  \\
& &  \frac{C(\pi(c)=true)}{\sum_X{C(\pi(c)=X)}}\prod_{i=1}^k{\frac{C(M_i,\pi(c)=true)}{C(\pi(c)=true)}}
  \label{eqn:combining:bayescounts}
\end{eqnarray}

The $C(\bullet)$ family of functions return the count instances of
co-occurrences of their arguments in a training set.

We use Laplacian (sometimes Lidstone's) smoothing while estimating to
avoid assigning zero probability to novel events
\cite{laplaceprob,lidstonelaw}.  Laplacian smoothing, sometimes known
as ``add-one'' smoothing, is equivalent to adding one to the number of
times each possible event was seen in the corpus before estimating
probabilities.  Lidstone's smoothing is similar, except an unspecified
parameter, $\lambda$ is added to the number of times each possible
event was seen.  Both smoothing schemes are linear combinations of the
observed frequencies with the uniform distribution.

This is the simplest form of a na\"ive Bayes classifier for this
problem.  It uses one parameter per parser.  On our training set it
performs identically to the second row of Table
\ref{table:combining:rawvoting}.  This is not surprising since we are
using only three parsers and they differ very little in accuracy.  The
robustness of this model when adding a poor parser is described in
Section \ref{section:combining:robustness}.

\subsubsection{Context}

There are a number of candidate contexts that may indicate how we
should distribute our trust across the ensemble members:

\begin{itemize}

\item Constituent Label ($l$)

\item Constituent length ($e-s$)

\item Parent label (ancestor label)

\item Sentence length
  
\end{itemize}

Polling pattern (i.e. for candidate constituent $x$,
$\pi_1(x)=1\wedge\pi_2(x)=0\wedge\pi_3(x)=1$) is not a reasonable
candidate for a context.  There are too many polling patterns to
choose from (the set grows exponentially in the ensemble size).  The
parameter space is simply too large to yield any reliable probability
estimates on our small datasets.


In the following formulation, $\pi(c)$ is the binary random variable we are
estimating.  Its value indicates whether we feel this constituent
should be in the parse.  $T$ is the random variable indicating the
label (e.g. NP, VP) on the constituent.  $M_i$ is the binary prediction
parser $i$ provides for the particular labelled constituent in
question.  Alternatively, for some $i$, $M_i$ can describe the value
of contextual features around the constituent in question.  In fact,
one can view the votes of the member parsers as merely more features
to throw into the classifier.  We call this model the Coprediction
Model.  It requires many parameters, specifically $O(kn)$ where $k$ is the
number of values the copredicting feature can take on and $n$ is the
number of parameters in a model without context.

\begin{eqnarray}
  P(\pi(c),T=t | M_1\ldots M_k) 
  & = &
  \frac{P(M_1\ldots M_k |\pi(c),T)P(\pi(c),T)}
  {P(M_1\ldots M_k)}
  \\
  & = &
  P(\pi(c),T)P(M_1\ldots M_k|\pi(c),T)
  \\
  & = &
  P(\pi(c),T)\prod_{i=1}^k{P(M_i|\pi(c),T)}
  \\
  P(\pi(c)|T,M_1\ldots M_k)
  &=&
  \sum_{t}P(\pi(c),t)P(M_1\ldots M_k|\pi(c),t)
\end{eqnarray}

This derivation is exactly the same as in Equation
\ref{eqn:combining:bayesform} except that we are predicting
both membership in the parse and the context of the parse.  Since we
are sure of the context of the parse, the result we use is
$P(\pi(c),T=t)$ where $t$ is the particular observed context.  The
probabilities are estimated similarly to those in Equation
\ref{eqn:combining:bayescounts}.

Another way to add context is shown below.  We call this the
Independent Context Model.  Here the context serves only to change the
threshold at which we use our estimate of $P(\pi(c)|M_1\ldots M_k)$.
We adjust the threshold by $P(T|\pi(c))$ to account for the particular
context we observe.  This process can easily be repeated by inserting
an adjustment factor for each of the contexts desired.  This
formulation uses as few parameters as possible among formulations
including contexts.  It needs only $O(k+n)$ where $k$ and $n$ are as
we mentioned before.  Since so few parameters are needed, it is much
easier to gather sufficient statistics for each of them.  It is
crucial that the contexts be independent in this case, as well as
independent of the predictors, as that is the assumption we use to
move from Equation \ref{eqn:combining:bayescontext} to Equation
\ref{eqn:combining:bayescontextindep}.

\begin{eqnarray}
  P(\pi(c) | T, M_1\ldots M_k) 
  & = &
  \frac{P(M_1\ldots M_k |\pi(c),T)P(\pi(c))}
  {P(T,M_1\ldots M_k)}
  \\
  & = &
  P(\pi(c))P(T,M_1\ldots M_k|\pi(c))
  \label{eqn:combining:bayescontext}
  \\
  & = &
  P(\pi(c))P(T|\pi(c))\prod_{i=1}^k{P(M_i|\pi(c),T)}
  \label{eqn:combining:bayescontextindep}
\end{eqnarray}

In the Coprediction Model, we are predicting the label on the
constituent and its membership in the hypothesis parse simultaneously
given the predictions of the parsers.  In the Independent Context
Model we first predict the label given the predictions of the parsers,
and then membership in the hypothesis parse based on the label and the
predictions of the parsers.

These models are each somewhat arbitrary ways to introduce context
without requiring a tabular estimation of the entire joint
distribution $P(\pi(c), M_1\ldots M_k)$.  The reason we avoid it is
that we expect the size of the ensemble to eventually grow to include
many parsers.  As $k$ gets large, filling the table to estimate that
distribution directly from relative frequencies requires a large
corpus that could potentially be better used to train the member
parsers.

We tested each of these models of context on our parser outputs using
the contexts described above.  The results can be seen in Table
\ref{table:combining:bayescontextresults}.  The model types are {\bf
  indep} or {\bf copredict} to describe whether the particular model
was using the coprediction or independent context techniques.  Among
the contexts, {\bf tag} represents the tag on the constituent (e.g.
NP, VP), {\bf parenttag} represents similarly the tag on the parent
constituent, {\bf clength} is a continuous feature representing the
span of the constituent, and {\bf slength} is the length of the
sentence.  The singular appearance of {\bf tag\&parenttag} represents
a feature whose values are pairs consisting of the tag of the
constituent and the tag of the parent of the constituent.

Fewer and smaller contexts are used with the coprediction model
because of the way it blows up the parameter space.  These results are
from the training set, the same set used for estimating the
probabilities.  None of the context added to the model gave large
improvements to the F-measure.

\begin{table}[htbp]
  \begin{center}
    \begin{tabular}{|cc|cc|cc|}
      \hline 
      Model  & Context & P     & R     &(P+R)/2 & F \\
      \hline
      indep & tag,clength,parenttag & 91.81 & 89.10 & 90.46 & 90.43 \\
      copredict & tag & 91.26 & 89.71 & 90.49 & 90.48 \\
      indep & clength & 91.63 & 89.43 & 90.53 & 90.52 \\
      indep & tag\&parenttag & 91.96 & 89.17 & 90.56 & 90.54 \\
      indep & tag & 92.05 & 89.22 & 90.63 & 90.61 \\
      indep & slength & 92.06 & 89.20 & 90.63 & 90.61 \\
      copredict & clength & 92.14 & 89.15 & 90.64 & 90.62 \\
      copredict & slength & 91.95 & 89.44 & 90.69 & 90.68 \\
      \hline
      \hline
      \multicolumn{2}{|c|}{Best Individual} & 88.73 & 88.54 & 88.63 & 88.63 \\
      \multicolumn{2}{|c|}{Na\"ive Bayes}   & 92.09 & 89.18 & 90.64 & 90.61 \\
      \hline
    \end{tabular}
    \caption{Results of Bayes with Context (Training Set)}
    \label{table:combining:bayescontextresults}
  \end{center}
\end{table}

\subsubsection{Negative Results}

The results in Table \ref{table:combining:bayescontextresults} are
discouraging.  None of the contexts added much to the predictive power
of our models.  Furthermore, the gain seen in the last row of that
table versus combining by non-parametric democratic voting or by
context-less na\"ive Bayes is not enough to show that the training set
precision and recall of the hypotheses are significantly different in
their predictions on the training set at a 90\% confidence level.  In
short, nothing helped.  The estimation error induced by using these
models and adding parameters overshadowed any reduction we achieved by
utilizing more descriptive models.

We cannot prove that there is not some set of contexts that will give
us a gain in accuracy.  However, we can analyze our data using these
particular contexts to get a feel for why context does not provide a
gain.  In particular, we are interested in instances where it is
desirable to trust one parser more than the consensus of the other two
parsers.  If there is no context in which a single parser performs
better than the other two, then there is no way we can use context
information to perform better than majority vote or simple na\"ive
Bayes.

A statistic we are interested in is the precision of a parser on those
samples for which it disagrees with the majority opinion.  In our
scenario this can only happen when the other two parsers agree and the
parser in question disagrees with their hypothesis.  The formula for
the precision is given in Equation \ref{eqn:combining:isolated} where
$\majority$ is the operator that produces a set consisting of elements
appearing in a majority of the given sets, and $S_{P_i}$ is the set of
constituents produced by Parser $i$.  We call the measure {\em
  isolated precision} because it is the precision the parser can
achieve on constituents that only it believes should be in the parse.
When the isolated precision is less than 50\%, adding the constituents
in question to the set will result in adding more errors than correct
predictions.  When it is greater than 50\%, adding those predictions
will result in a gain over the majority predictor.  We can get some
idea of whether partitioning the prediction space using a particular
context will be helpful by looking for places in that partitioning
where the isolated precision is greater than 50\%.

\begin{eqnarray}
P_{isolated}(P_i) 
&=&
\frac{|(S_{P_i}-\majority\limits_{j\not= i}S_{P_i}) \cap S_{true}|}
{|S_{P_i}-\majority\limits_{j\not= i}S_{P_i}|}
\label{eqn:combining:isolated}
\end{eqnarray}

\begin{table}[htbp]
  \begin{center}
    \begin{tabular}{|l|rr|rr|rr|}
      \hline 
      \multicolumn{1}{|c|}{Constituent}
      & \multicolumn{2}{c|}{Parser1}
      & \multicolumn{2}{c|}{Parser2}
      & \multicolumn{2}{c|}{Parser3} 
      \\ 
      \multicolumn{1}{|c|}{Label}
      & \multicolumn{1}{c}{count} & \multicolumn{1}{c|}{P}
      & \multicolumn{1}{c}{count} & \multicolumn{1}{c|}{P}
      & \multicolumn{1}{c}{count} & \multicolumn{1}{c|}{P}
      \\
      \hline 
      ADJP    & 132   & 28.78 & 215   & 21.86  & 173   & 34.10 \\
      ADVP    & 150   & 25.33 & 129   & 21.70  & 102   & 31.37 \\
      CONJP   & 2     & 50.00 & 8     & 37.50  & 3     & 0.00 \\
      FRAG    & 51    & 3.92  & 29    & 27.58  & 11    & 9.09 \\
      INTJ    & 3     & 66.66 & 1     &100.00  & 2     & 50.00 \\
      LST     & 0     & NA    & 0     & NA     & 0     & NA \\
      NAC     & 0     & NA    & 13    & 53.84  & 7     & 14.28 \\
      NP      & 1489  & 21.08 & 1550  & 18.38  & 1178  & 27.33 \\
      NX      & 7     & 85.71 & 9     & 22.22  & 3     & 0.00 \\
      PP      & 732   & 23.63 & 643   & 20.06  & 503   & 27.83 \\
      PRN     & 20    & 55.00 & 33    & 54.54  & 38    & 15.78 \\
      PRT     & 12    & 16.66 & 20    & 40.00  & 16    & 37.50 \\
      QP      & 21    & 38.09 & 34    & 44.11  & 76    & 14.47 \\
      RRC     & 1     & 0.00  & 1     & 0.00   & 2     & 0.00 \\
      S       & 757   & 13.73 & 482   & 23.65  & 434   & 38.94 \\
      SBAR    & 331   & 11.78 & 196   & 23.97  & 178   & 34.83 \\
      SBARQ   & 0     & NA    & 6     & 16.66  & 3     & 0.00 \\
      SINV    & 3     & 66.66 & 11    & 81.81  & 13    & 30.76 \\
      SQ      & 2     & 0     & 11    & 18.18  & 3     & 33.33 \\
      UCP     & 6     & 16.66 & 12    & 8.33   & 8     & 12.50 \\
      VP      & 868   & 13.36 & 630   & 24.12  & 477   & 35.42 \\
      WHADJP  & 0     & NA    & 0     & NA     & 1     & 0.00 \\
      WHADVP  & 2     &100.00   & 5   & 40.00  & 1     & 100.00 \\
      WHNP    & 33    & 33.33 & 8     & 25.00  & 17    & 58.82 \\
      WHPP    & 0     & NA    & 0     & NA     & 2     & 100.00 \\
      X       & 0     & NA    & 2     & 100.00 & 1     & 0.00 \\
      \hline 
    \end{tabular}
    \caption{Isolated Constituent Precision By Context}
    \label{table:combining:isolated:tag}
  \end{center}
\end{table}

\begin{table}[htbp]
  \begin{center}
    \begin{tabular}{|l|rr|rr|rr|}
      \hline 
      \multicolumn{1}{|c|}{Parent}
      & \multicolumn{2}{c|}{Parser1}
      & \multicolumn{2}{c|}{Parser2}
      & \multicolumn{2}{c|}{Parser3} 
      \\ 
      \multicolumn{1}{|c|}{Label}
      & \multicolumn{1}{c}{count} & \multicolumn{1}{c|}{P}
      & \multicolumn{1}{c}{count} & \multicolumn{1}{c|}{P}
      & \multicolumn{1}{c}{count} & \multicolumn{1}{c|}{P}
      \\
      \hline 
      ADJP    & 46    & 30.43& 87    & 20.68& 58    & 24.13 \\
      ADVP    & 21    & 19.04& 31    & 22.58& 26    & 34.61 \\
      FRAG    & 37    & 21.62& 10    & 0.00& 9     & 33.33 \\
      NAC     & 0     & NA& 3     & 66.66& 2     & 100.00 \\
      NP      & 1081  & 22.57& 1320  & 19.01& 1034  & 25.82 \\
      NULL    & 194   & 0& 0     & NA& 0     & NA \\
      NX      & 4     & 100.00& 0     & NA& 0     & NA \\
      PP      & 445   & 26.06& 447   & 21.47& 360   & 27.77 \\
      PRN     & 27    & 59.25& 22    & 40.90& 28    & 25.00 \\
      RRC     & 1     & 0.00& 1     & 0.00& 1     & 0.00 \\
      S       & 1111  & 10.53& 672   & 22.47& 543   & 33.70 \\
      SBAR    & 240   & 19.58& 184   & 24.45& 177   & 35.02 \\
      SBARQ   & 0     & NA& 8     & 50.00& 4     & 25.00 \\
      SINV    & 15    & 60.00& 33    & 27.27& 16    & 31.25 \\
      SQ      & 6     & 33.33& 11    & 27.27& 9     & 33.33 \\
      TOP     & 8     & 100.00& 59    & 30.50& 24    & 37.50 \\
      UCP     & 4     & 25.00& 9     & 44.44& 2     & 100.00 \\
      VP      & 1378  & 20.10& 1146  & 22.94& 952   & 34.24 \\
      WHADJP  & 1     & 100.00& 0     & NA& 0     & NA \\
      WHADVP  & 0     & NA& 3     & 66.66& 0     & NA \\
      WHNP    & 1     & 0.00& 2     & 50.00& 7     & 71.42 \\
      WHPP    & 0     & NA& 0     & NA& 0     & NA \\
      X       & 2     & 100.00& 0     & NA& 0     & NA \\
      \hline 
    \end{tabular}
    \caption{Isolated Constituent Precision by Parent Label}
    \label{table:combining:isolated:parent}
  \end{center}
\end{table}

Tables \ref{table:combining:isolated:tag} and
\ref{table:combining:isolated:parent} give values for
$P_{isolated}(P_i)$ under restriction to constituent label and
parent's constituent label contexts respectively.  Notice that in most
of the situations in which the precisions are greater than 50\% the
number of times those contexts appear is insignificant.  In the
training set from which these numbers were calculated a 0.1 percent
improvement in precision requires approximately 40 more correct
predictions (or about 40 fewer incorrect predictions).

\begin{figure}[htbp]
  \begin{center}
    \epsfig{file=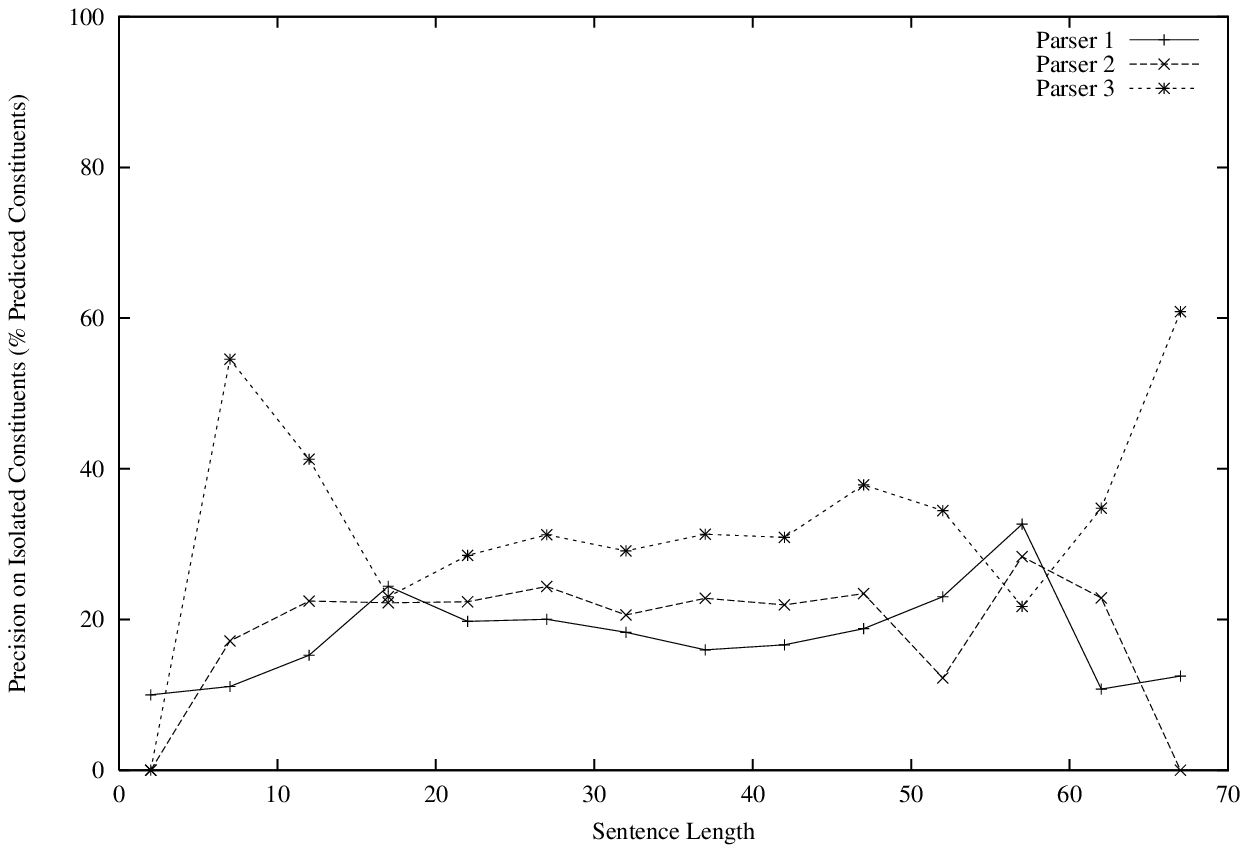, height=0.45\textheight}
    \epsfig{file=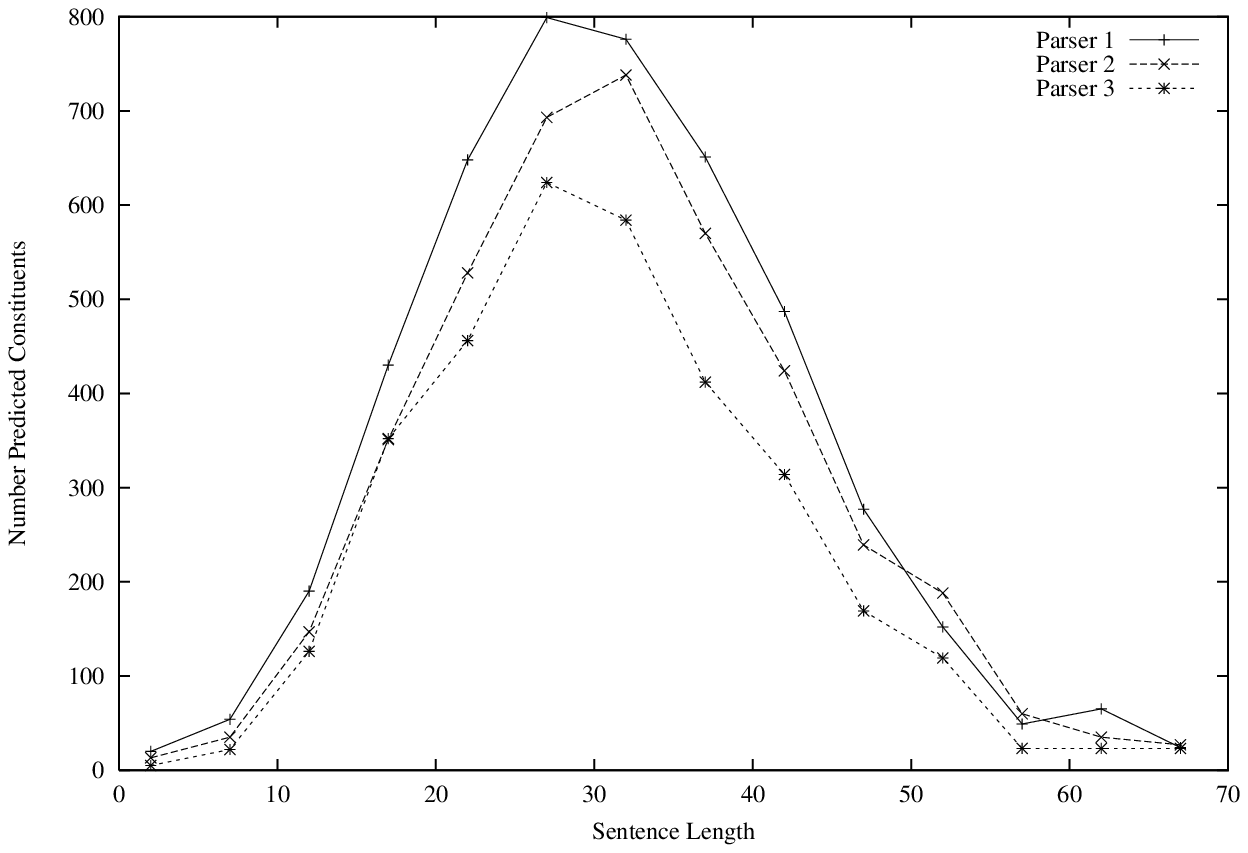, height=0.45\textheight}
  \end{center}
  \caption{Isolated Constituent Parser Precision and Sentence Length}
  \label{fig:combining:isolated:slength}
\end{figure}

\begin{figure}[htbp]
  \begin{center}
    \epsfig{file=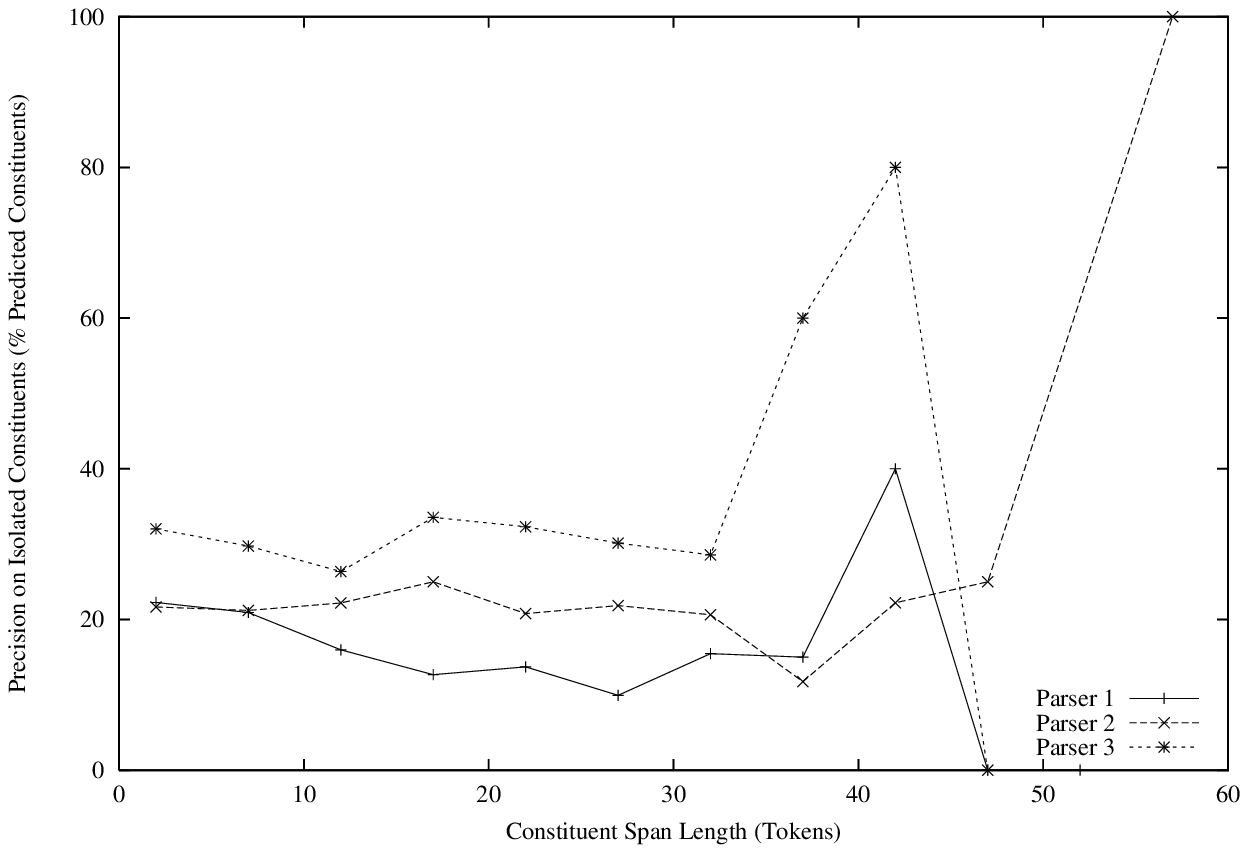, height=0.45\textheight}
    \epsfig{file=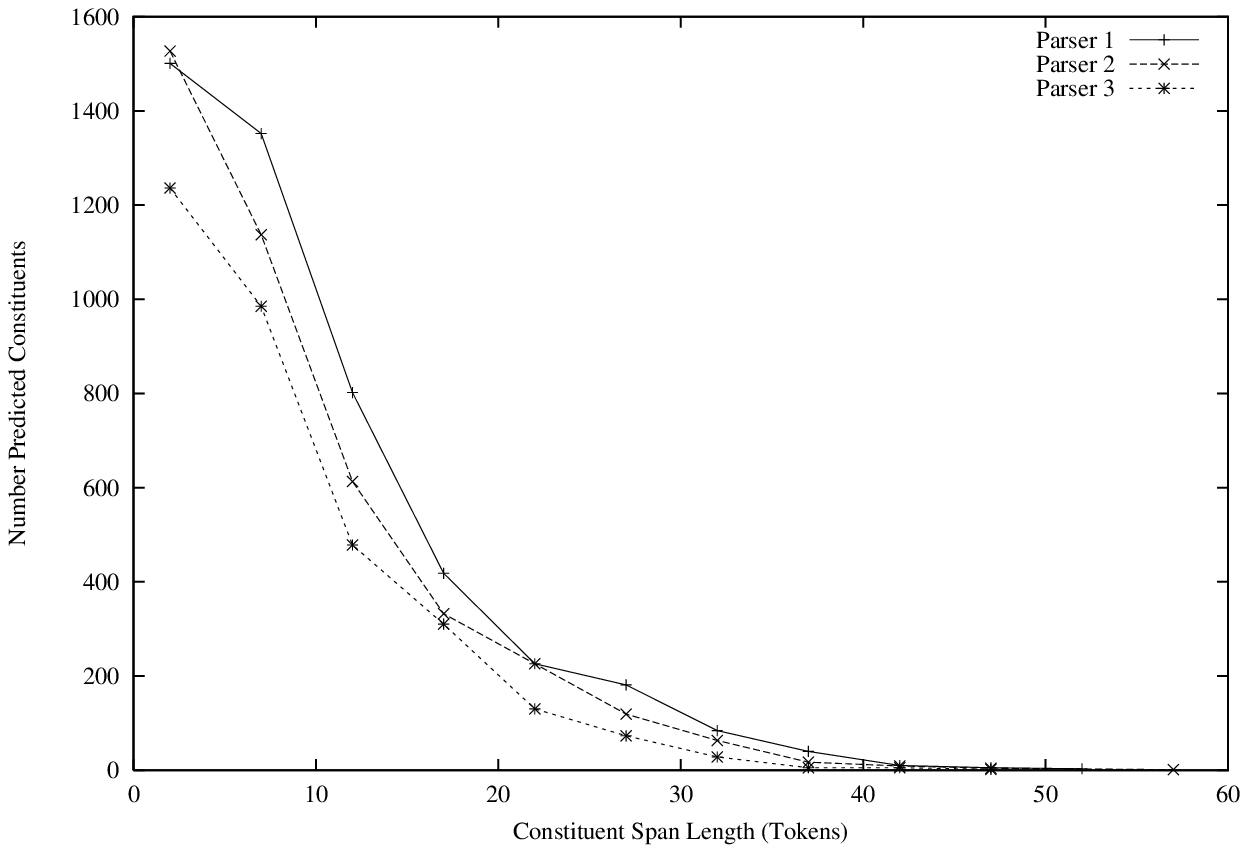, height=0.45\textheight}
  \end{center}
  \caption{Isolated Constituent Parser Precision and Span Length}
  \label{fig:combining:isolated:clength}
\end{figure}

The graphs in Figures \ref{fig:combining:isolated:slength} and
\ref{fig:combining:isolated:clength} show how the same value
varies by sentence length and constituent length respectively.  We see
the same effect in these graphs.  When the graph of isolated precision
is over 50\%, the number of occurrences of the particular context is
so small that the possible gain is small and often insignificant.

In Tables \ref{table:combining:precision:tag} and
\ref{table:combining:precision:parent} we break down the (standard)
precision that each parser is able to attain by the constituent label
and the label on the parent of the constituent.  This gives us an idea
of how the parsers perform in isolation and how little they differ in
assorting their accuracy across the constituent types.  

From Table \ref{table:combining:precision:tag} we can calculate the
precision and recall we can achieve by assembling a parser that
decides to trust the most precise parser for each constituent.  For
example, we would trust Parser3 for constituents with label NP because
it has the highest precision for that label.  If we built this parser
we would get a precision of 88.97\% and a recall of 88.07\%.  This
yields an F-measure of 88.52\%, substantially worse than the best
individual parser.  The F-measure is worse because precision goes up
but recall would decrease in this case when compared to the best
individual parser.

Our goal is to maximize F-measure but we picked the parser with the
highest precision for each of the partitions induced by the
constituent label context.  This is not because choosing in this
manner maximizes F-measure.  F-measure is a global measure, and as
such is very hard to maximize \cite{goodman96:parsingalgorithms}.
This is because an imbalance favoring precision for one partition and
an imbalance favoring recall for another partition can combine to
yield a higher F-measure than when the partitions are individually set
to optimize F-measure.

\begin{table}[htbp]
  \begin{center}
    \begin{tabular}{|l|rr|rr|rr|}
      \hline 
      \multicolumn{1}{|c|}{Constituent}
      & \multicolumn{2}{c|}{Parser1}
      & \multicolumn{2}{c|}{Parser2}
      & \multicolumn{2}{c|}{Parser3} 
      \\ 
      \multicolumn{1}{|c|}{Label}
      & \multicolumn{1}{c}{count} & \multicolumn{1}{c|}{P}
      & \multicolumn{1}{c}{count} & \multicolumn{1}{c|}{P}
      & \multicolumn{1}{c}{count} & \multicolumn{1}{c|}{P}
      \\
      \hline 
      ADJP    & 688   & 75.58& 832   & 68.75& 810   & 73.95 \\
      ADVP    & 1227  & 82.23& 1182  & 82.74& 1195  & 85.02 \\
      CONJP   & 19    & 73.68& 25    & 64.00& 15    & 60.00 \\
      FRAG    & 65    & 13.84& 43    & 32.55& 14    & 14.28 \\
      INTJ    & 8     & 87.50& 6     & 83.33& 9     & 77.77 \\
      LST     & 4     & 100.00& 3     & 100.00& 2     & 100.00 \\
      NAC     & 5     & 100.00& 24    & 75.00& 19    & 68.42 \\
      NP      & 18747 & 88.92& 18884 & 88.39& 18718 & 90.25 \\
      NX      & 10    & 90.00& 12    & 41.66& 3     & 0.00 \\
      PP      & 5620  & 82.36& 5544  & 81.78& 5530  & 84.61 \\
      PRN     & 102   & 85.29& 114   & 81.57& 116   & 68.96 \\
      PRT     & 142   & 77.46& 177   & 76.27& 170   & 77.05 \\
      QP      & 472   & 88.77& 503   & 85.88& 539   & 78.47 \\
      RRC     & 1     & 0.00 &2     & 0.00 & 3     & 0.00 \\
      S       & 5753  & 83.45& 5584  & 88.07& 5671  & 89.70 \\
      SBAR    & 1776  & 78.15& 1703  & 85.49& 1742  & 87.25 \\
      SBARQ   & 4     & 75.00& 10    & 40.00& 7     & 57.14 \\
      SINV    & 28    & 78.57& 141   & 92.19& 149   & 88.59 \\
      SQ      & 10    & 80.00& 22    & 59.09& 15    & 86.66 \\
      UCP     & 11    & 45.45& 22    & 40.90& 17    & 52.94 \\
      VP      & 8811  & 85.81& 8733  & 89.12& 8758  & 90.64 \\
      WHADJP  & 3     & 66.66& 1     & 0.00& 4     & 50.00 \\
      WHADVP  & 122   & 96.72& 130   & 93.07& 124   & 95.96 \\
      WHNP    & 438   & 93.60& 293   & 96.92& 423   & 96.69 \\
      WHPP    & 21    & 100.00& 17    & 100.00& 25    & 100.00 \\
      X       & 1     & 100.00& 4     & 100.00& 3     & 66.66 \\
      \hline
    \end{tabular}
    \caption{Constituent Precision By Context}
    \label{table:combining:precision:tag}
  \end{center}
\end{table}

\begin{table}[htbp]
  \begin{center}
    \begin{tabular}{|l|rr|rr|rr|}
      \hline 
      \multicolumn{1}{|c|}{Parent}
      & \multicolumn{2}{c|}{Parser1}
      & \multicolumn{2}{c|}{Parser2}
      & \multicolumn{2}{c|}{Parser3} 
      \\ 
      \multicolumn{1}{|c|}{Label}
      & \multicolumn{1}{c}{count} & \multicolumn{1}{c|}{P}
      & \multicolumn{1}{c}{count} & \multicolumn{1}{c|}{P}
      & \multicolumn{1}{c}{count} & \multicolumn{1}{c|}{P}
      \\
      \hline 
      ADJP    & 362   & 80.38& 410   & 72.43& 390   & 77.17 \\
      ADVP    & 216   & 83.33& 221   & 81.90& 230   & 83.47 \\
      FRAG    & 72    & 55.55& 49    & 71.42& 48    & 85.41 \\
      NAC     & 2     & 100.00& 6     & 83.33& 6     & 100.00 \\
      NP      & 10670 & 85.41& 10831 & 83.63& 10661 & 85.90 \\
      NULL    & 194   & 0.00&   0    & NA&   0    & NA \\
      NX      & 8     & 100.00& 4     & 100.00& 3     & 100.00 \\
      PP      & 5621  & 88.82& 5667  & 87.87& 5636  & 89.51 \\
      PRN     & 149   & 81.87& 151   & 74.17& 154   & 70.12 \\
      RRC     & 1     & 0.00& 1     & 0.00& 1     & 0.00 \\
      S       & 11232 & 87.38& 10975 & 91.37& 11019 & 92.59 \\
      SBAR    & 2315  & 86.86& 2198  & 88.85& 2369  & 90.20 \\
      SBARQ   & 10    & 100.00& 21   & 80.95& 17    & 82.35 \\
      SINV    & 339   & 94.98& 442   & 92.30& 445   & 95.05 \\
      SQ      & 30    & 80.00& 36    & 69.44& 34    & 76.47 \\
      TOP     & 2294  & 97.25& 2431  & 96.25& 2409  & 97.50 \\
      UCP     & 29    & 89.65& 36    & 86.11& 27    & 100.00 \\
      VP      & 10501 & 81.92& 10489 & 83.14& 10583 & 85.39 \\
      WHADJP  & 1     & 100.00&  0   & NA      & 0      & NA \\
      WHADVP  & 0      & NA& 3       & 66.66   &  0     & NA \\
      WHNP    & 14    & 78.57& 12    & 91.66   & 20    & 80.00 \\
      WHPP    & 24    & 100.00& 25   & 100.00 & 25    & 100.00 \\
      X       & 4     & 100.00& 3    & 100.00 & 4     & 100.00 \\
      \hline
    \end{tabular}
    \caption{Constituent Precision by Parent Label}
    \label{table:combining:precision:parent}
  \end{center}
\end{table}

We should not be surprised that the contexts we investigate make
little difference in our decision-making capability for combining
these parsers.  The parsers were all trained with these contexts (and
more) in mind.  Their creators have done a good job of taking
advantage of the tendencies of certain structures to be found only in
certain contexts.

\subsubsection{Experiment: Adding a base noun phrase chunker}

In order to again further test the efficacy of our context-dependent
combination techniques, we added the Ramshaw and Marcus base noun
phrase chunker \cite{ramshaw99:basenp} to the ensemble.  The chunker
attempts to predict which phrases appear as non-recursive noun phrases
in a parse, the noun phrases that are lowest in the tree.  It is
designed for applications that do not require full parsing, and
generally runs much faster than a full-blown parser.  Ramshaw and
Marcus used transformation-based learning to induce a set of
transformation rules that can be quickly run over a large corpus to
produce base NP brackets.

The expectation for adding the chunker to the ensemble was that the
Bayes technique would learn to trust the chunker, or at least utilize
the chunker's decisions when dealing with {\tt NP} tags.

\begin{table}[htbp]
  \begin{center}
    \begin{tabular}{|cc|cc|cc|}
      \hline 
      Model  & Context & P     & R     &(P+R)/2 & F \\
      \hline
      indep     & tag & 91.62  & 89.44 & 90.53 & 90.52 \\
      copredict & tag & 91.44  & 89.14 & 90.29 & 90.28 \\
      \hline
      \hline
      \multicolumn{2}{|c|}{Best Individual} & 88.73 & 88.54 & 88.63 & 88.63 \\
      \multicolumn{2}{|c|}{Na\"ive Bayes (3)} & 92.09 & 89.18 & 90.64 & 90.61 \\
      \multicolumn{2}{|c|}{Na\"ive Bayes (4)} & 91.60 & 89.57 & 90.59 & 90.57\\
      \hline
    \end{tabular}
    \caption{Results of Including a Noun Phrase Chunker}
    \label{table:combining:bayesnpcontextresults}
  \end{center}
\end{table}

In Table \ref{table:combining:bayesnpcontextresults} we can see the
result of using the independent constituents and coprediction models
for combining the four systems.  Both of these results are worse than
the na\"ive Bayes model for the three parsers, as well as the na\"ive
Bayes result when the chunker is added to the group.  Also, both of
these models predict with significantly lower precision than the
na\"ive Bayes(3) model.

\begin{table}[htbp]
  \begin{center}
    \begin{tabular}{|c|cc|cc|}
      \hline 
      Model  &  P     & R     &(P+R)/2 & F \\
      \hline
      Parser1 & 88.92 &  89.93 &  89.43 &  89.42 \\
      Parser2 & 88.39 &  90.05 &  89.22 &  89.21 \\
      Parser3 & 90.25 &  91.14 &  90.69 &  90.69 \\
      \hline
      Majority(1-3)& 93.30 &  92.17 &  92.73 &  92.73 \\
      \hline
      NP chunker& 93.59  & 68.87 &  81.23  & 79.34 \\
      \hline
    \end{tabular}
    \caption{Parser Performance on Noun Phrases}
    \label{table:combining:parsingNP}
  \end{center}
\end{table}

Further study enables us to decide why we gained nothing from
introducing such an intuitively appealing source of noun phrase
annotations.  In Table \ref{table:combining:parsingNP} we show the
performance of the parsers and chunker on only those constituents that
are labelled with {\tt NP}.  While the precision for the chunker is
significantly higher than the majority vote of the three parsers, note
that the recall for the chunker is much lower than the other systems.
This happens because the chunker is predicting only non-recursive noun
phrases.  All noun phrases that contain a nested noun phrase are by
definition not targets for the chunker to predict.  That is why the
combined system takes a performance hit when the chunker is added:
when the chunker says a constituent should be a noun phrase it is a
little more accurate than the majority vote, but when it says the
constituent is not a noun phrase it is wrong on all of the non-base
noun phrases.

To summarize the claims, we estimated the portion of the noun phrase
constituent inclusion decisions that were correctly predicted by the
chunker, and on which the majority of the three parsers disagreed.
This value was only 48\%, indicating that the chunker was performing
worse than chance on predictions it was asked to make about
constituents that it had a chance to help predict.  This is probably
not a coincidence.  The chunker was only designed to predict the
non-recursive noun phrases.  Incorporating it into the combined model
would require a specialized model that included a notion of
``base-ness'' of a noun phrase.  While it could prove useful to pursue
combination techniques in which such a feature can easily be
specified, none of our models can take it directly into account.

\subsection{Pruning into Trees}


The parametric models we have developed so far have not enforced a
tree constraint.  There is nothing stopping these
independently-predicted constituents from inducing crossing structures
in the parse tree.  In this case, pruning of the crossing constituents
could be explored.  Our negative results from this section did not
suggest there would be any value in pursuing this line of research at
the present time, though.

However, when we are using the simplest na\"ive Bayes configuration
with no context, requiring estimated constituent probabilities
strictly larger than 0.5 to include them in the parse, the result from
Lemma \ref{lemma:combining:treeguarantee} enables us to say that no
crossing brackets will appear in the final hypothesis.

\subsection{Parser Switching}

Just as in Section
\ref{section:combining:nonparametricparserswitching}, we can try to
beat the Parser Switch Oracle using our models.  As shown above,
context gives us very little if any gain, so we will not incorporate
it into our switching model.  It could be the case that small gains in
the na\"ive Bayes probability model can make larger gains in the
switching algorithm, but that is not the purpose of this
investigation.  We first reformulate the problem as shown below.  We
are interested choosing the parse among the input parses that
maximizes the probability of correctness for each of its constituents
(and predictions on missing constituents).  We treat those predictions
as independent.

\begin{eqnarray*}
  \argmax{\pi_i} P(\pi_i|M_1\ldots M_k)
  & = &
  \argmax{\pi_i} \prod_{c}{P(\pi_i(c)|M_1\ldots M_k)}
  \\
  & = &
  \argmax{\pi_i} \prod_{c}\frac{P(M_1\ldots M_k|\pi_i(c))P(\pi_i(c))}
  {P(M_1\ldots M_k)}
  \\
  & = &
  \argmax{\pi_i} \prod_{c}{P(\pi_i(c))\prod_{j=1}^{k}{\frac{P(M_j|\pi_i(c))}
  {P(M_j)}}}
  \\
  & = &
  \argmax{\pi_i} \prod_{c}{P(\pi_i(c))\prod_{j=1}^{k}{P(M_j|\pi_i(c))}}
\end{eqnarray*}

\begin{table}[tbp]
  \begin{center}
    \begin{tabular}{|rr|rr|r|}
      \hline
      \multicolumn{1}{|c}{P} 
      & \multicolumn{1}{c|}{R} 
      & \multicolumn{1}{c}{(P+R)/2} 
      & \multicolumn{1}{c|}{F} 
      & \multicolumn{1}{c|}{Exact} 
      \\
      \hline
      90.13  & 89.65 &  89.89  & 89.89 & 38.4 \\
      \hline 
    \end{tabular}
    \caption{Na\"ive Bayes Parser Switching}
    \label{table:combining:probswitch}
  \end{center}
\end{table}

The results (as shown in Table \ref{table:combining:probswitch}) are
better than those achieved in the unsupervised, non-parametric parser
switching experiment (from Table \ref{table:combining:rawswitch}).
Intuitively, this is because we have more faith in the predictions of
the better parsers.  The candidate parse that agrees more with the
better parsers is preferred to those that agree more with the
parsers that perform worse.


\section{Final Evaluation}
\label{section:combining:robustness}
\label{section:combining:testset}

After developing all of our models we evaluated them on the 1700
sentences in the test corpus.  This section gives a full account of
those results.

\subsection{Test Set}

We have made some performance claims on our training data.  The claims
are summarized below and in Table \ref{table:combining:trainingset}.

\begin{enumerate}
\item
  A significant precision and recall boost can be attained using
  simple non-parametric democratic voting on constituents.
\item
  A significant precision and recall boost can be attained using
  non-parametric parser switching.  This is useful when we want to
  preserve constraints on the productions in the parses.
\item
  A significant precision and recall boost can be attained using
  parametric parser switching, and the gain is larger than the
  non-parametric version.
\item
  Parser switching by approximating the centroid using parse edit
  distance suggests we can more precisely pick parses than by using
  Bayes parser switching.  This is odd because this is an unsupervised
  method that surpasses the comparable supervised method.  The
  difference in F-measure is merely suggestive, though.
\end{enumerate}

\begin{table}[htbp]
  \begin{center}
    \begin{tabular}{|l|cc|cc|c|}
      \hline 
      Reference / System 
      &\multicolumn{1}{c}{P} 
      & \multicolumn{1}{c|}{R} 
      & \multicolumn{1}{c}{(P+R)/2} 
      & \multicolumn{1}{c|}{F} 
      & \multicolumn{1}{c|}{Exact} \\
      \hline 
      Average Individual Parser & 87.14 & 86.91 & 87.02 & 87.02 & 30.8\\
      Best Individual Parser    & 88.73 & 88.54 & 88.63 & 88.63 & 35.0\\
      \hline
      Parser Switch Oracle   & 93.12 & 92.84 & 92.98 & 92.98 & 46.8\\
      Maximum Precision Oracle       & 100.00& 95.41 & 97.70 & 97.65 & 64.5\\
      \hline
      Similarity Switching    & 89.50 & 89.88 & 89.69 & 89.69 & 35.3\\
      Distance Switching      & 90.24 & 89.58 & 89.91 & 89.91 & 38.0\\
      Alignment Switching     & 90.26 & 89.63 & 89.95 & 89.95 & 38.3\\
      Bayes Switching         & 90.13 & 89.65 & 89.89 & 89.89 & 38.4\\
      \hline
      Constituent Voting      & 92.09 & 89.18 & 90.64 & 90.61 & 37.0\\
      Alignment and Consensus & 92.10 & 89.15 & 90.63 & 90.60 & 37.0\\
      Na\"ive Bayes           & 92.09 & 89.18 & 90.64 & 90.61 & 37.0\\
      \hline 
    \end{tabular}
    \caption{Summary of Training Set Performance}
    \label{table:combining:trainingset}
  \end{center}
\end{table}

In this section we evaluate the models that produced those claims on
the 1700 sentences in the test set.

\begin{table}[htbp]
  \begin{center}
    \begin{tabular}{|l|rr|rr|r|}
      \hline 
      Reference / System 
      &\multicolumn{1}{c}{P} 
      & \multicolumn{1}{c|}{R} 
      & \multicolumn{1}{c}{(P+R)/2} 
      & \multicolumn{1}{c|}{F}  
      & \multicolumn{1}{c|}{Exact} \\
      \hline 
      Average Individual Parser & 87.61 & 87.83 & 87.72 & 87.72 & 31.6\\
      Best Individual Parser    & 89.61 & 89.73 & 89.67 & 89.67 & 35.4\\
      \hline
      Parser Switch Oracle             & 93.78 & 93.87 & 93.82 & 93.82 & 48.3\\
      Maximum Precision Oracle         & 100.00& 95.91 & 97.95 & 97.91 & 65.6\\
      \hline
      Similarity Switching      & 90.04 & 90.81 & 90.43 & 90.43 & 36.6\\
      Distance Switching        & 90.72 & 90.47 & 90.60 & 90.60 & 38.4\\
      Alignment Switching       & 90.70 & 90.47 & 90.59 & 90.59 & 38.4\\
      Bayes Switching           & 90.78 & 90.70 & 90.74 & 90.74 & 38.8\\
      \hline 
      Constituent Voting        & 92.42 & 90.10 & 91.26 & 91.25 & 37.9\\
      Alignment and Consensus   & 92.43 & 90.08 & 91.26 & 91.24 & 37.9\\
      Na\"ive Bayes             & 92.42 & 90.10 & 91.26 & 91.25 & 37.9\\
      \hline
    \end{tabular}
    \caption{Summary of Test Set Performance}
    \label{table:combining:testset}
  \end{center}
\end{table}

Table \ref{table:combining:testset} shows the results. All of the
parsers performed as well on this set as they did on their original
test set.  The best parser performed significantly better on this than
on the original test set.  One possible explanation for this is that
this set contains many sentences that are systematically easier to
parse.  This is likely because these parses are not randomly
partitioned.  They were partitioned based on the order in which they
were published.  If, for example, the complexity of the news varies
(and consequently contains simpler sentences) with respect to time,
then we would expect to observe this behavior.  Verifying this
hypothesis is beyond the scope of this thesis, however.

We can push aside the question of baseline performance differences
between the different testing corpus sections (22 and 23) because we
are more interested in the improvement we can make via combination
than the raw accuracy numbers.  In some sense, if this dataset
(section 22) is easier to parse, it should be harder to get a gain
from combining parsers.

The combining techniques all perform substantially better on this set,
probably because the developers (and reviewers of their published
works) never investigated it.  That is, the combining techniques
reduce the error rate more on this set than on the previous set.
There was no bias or incentive for performing well on this set, and
implicitly training on the test set (either individually or through
peer review) was not investigated.  Implicit training on the test set
tends to make the systems similar, because there exists a strong
competitive drive to tune the systems to do at least as well as the
other system on whatever metrics are used, rather than to simply
perform well at modeling the phenomena in the corpora.\footnote{Or,
  stated another way, {\em excessive attention to the media defeats
    independent thinking.}}

The big surprise from this set is that the Alignment Switching method
(that rivaled Bayes Switching on the training set) performed very
poorly on this set.  This is probably due to excessive
experimentation.  We investigated many edit distance functions in
building the algorithm and may have implicitly over-fit the training set
by making our choice.  This is a well-known shortcoming of the
winner-takes-all approach to choosing between multiple algorithms
during training \cite{ng97:overfitting}.  The best algorithm for the
training data is not the best algorithm for the test data, and the
challenge is to find an algorithm that is accurate on the training
data without capturing unique or rare phenomena present in it.
Repeated experimentation to find a good algorithm for the training
data tends to find algorithms that model the noise in the training
data as well as the underlying phenomena of interest.

Note that on this test set, that had a higher initial accuracy than
the training set we have still managed to reduce the error rates by
approximately the same amount using our best methods (Distance
Switching, Bayes Switching, Constituent Voting and Na\"ive Bayes
Hybridization).

\begin{table}[htbp]
  \begin{center}
    \begin{tabular}{|l|rr|rr|r|}
      \hline 
      Reference / System 
      &\multicolumn{1}{c}{P} 
      & \multicolumn{1}{c|}{R} 
      & \multicolumn{1}{c}{(P+R)/2} 
      & \multicolumn{1}{c|}{F}  
      & \multicolumn{1}{c|}{Exact} \\
      \hline 
      Best Individual Parser    & 89.61 & 89.73 & 89.67 & 89.67 &
      35.4\\
      B.I.P. plus Section 23 & 89.60 &  89.76  & 89.68 &
      89.68 &  35.7 \\ 
      \hline
    \end{tabular}
    \caption{Best Individual Test Set Differences}
    \label{table:combining:testdiff}
  \end{center}
\end{table}

As we have mentioned a number of times, the parametric parsers are
using more data than the non-parametric ones.  They use the data
from section 23 of the Treebank to estimate their parameters.  In
order to be fair, we would like to let the individual parsers train on
this section when they are being combined in a non-parametric manner.
Unfortunately, we were only provided with training code for one of the
parsers, namely the Best Individual Parser.  In Table
\ref{table:combining:testdiff} we show how well the best individual
parser performs on section 22 when it is given the original corpus
versus how it performs when it is additionally given the section 23
data for training.  Notice that the performance changes very little.
We take this result to suggest that we are not missing very much by
not being able to perform the experiment we just described.  The
non-parametric parser is not losing out for lack of training data
for the individual parsers.

\begin{table}[htbp]
  \begin{center}
    \begin{tabular}{|l|r|r|}
      \hline
      \multicolumn{1}{|c|}{Parser}
      &
      \multicolumn{1}{c|}{Sentences}
      &
      \multicolumn{1}{c|}{\%}
      \\
      \hline
      Parser 1 & 279  & 16 \\
      Parser 2 & 216  & 13 \\
      Parser 3 & 1204 & 71 \\
      \hline
    \end{tabular}
    \caption{Bayes Switching Parser Usage}
    \label{table:combining:switchingusage}
  \end{center}
\end{table}

Table \ref{table:combining:switchingusage} shows how much the Bayes
switching algorithm uses each of the parsers on the test set.  Parser
3, the most accurate parser, was chosen 71\% of the time, and Parser
1, the least accurate parser was chosen 16\% of the time.
Furthermore, the reliance on Parser 3 is not a result of arbitrary
tie-breaking.  There is very little chance of a tie ever occurring
because algorithm uses a very fine grained model.  Many probabilities
are involved in setting the switch for each sentence.

\subsection{Robustness}

In the course of investigating the combination of these three parsers,
we were not able to quantify the impact of their relative accuracies.
These three parsers are all trained to high accuracies, and the
precision/recall tradeoff is in balance for each of them.  There exist
parsers that perform with very high precision at the expense of
recall.  Also, there may be parsers that perform very well on the
constituents that these three parsers get incorrect, but which are not
very good elsewhere.  

We have access to a PCFG-based parser, which performs rather poorly.  In
this section we will explore the sensitivity of our combining methods
to this parser by re-evaluating the methods using an ensemble of four
instead of just three.

\begin{table}[htbp]
  \begin{center}
    \begin{tabular}{|l|rr|rr|r|}
      \hline 
      Reference / System 
      &\multicolumn{1}{c}{P} 
      & \multicolumn{1}{c|}{R} 
      & \multicolumn{1}{c}{(P+R)/2} 
      & \multicolumn{1}{c|}{F}  
      & \multicolumn{1}{c|}{Exact} \\
      \hline 
      Average Individual Parser & 84.55 & 80.91 & 82.73 & 82.69 & 24.6\\
      Best Individual Parser    & 89.61 & 89.73 & 89.67 & 89.67 & 35.4\\
      \hline
      Parser Switch Oracle             & 93.92 & 93.88 & 93.90 & 93.90 & 48.4\\
      Maximum Precision Oracle         & 100.00& 96.66 & 98.33 & 98.30 & 69.4\\
      \hline
      Similarity Switching      & 89.90 & 90.89 & 90.40 & 90.39 & 36.7\\
      Distance Switching        & 90.92 & 90.16 & 90.54 & 90.54 & 37.8\\
      Alignment Switching       & 90.94 & 90.21 & 90.57 & 90.57 & 38.0\\
      Bayes Switching           & 90.94 & 90.70 & 90.82 & 90.82 & 39.1\\
      \hline 
      Constituent Voting        & 89.78 & 91.80 & 90.79 & 90.78 & 33.5\\
      Na\"ive Bayes             & 92.42 & 90.10 & 91.26 & 91.25 & 37.9\\
      Alignment and Consensus   & 95.70 & 82.82 & 89.26 & 88.80 & 25.7\\
      \hline
    \end{tabular}
    \caption{Summary of Robust Test Set Performance}
    \label{table:combining:robusttestset}
  \end{center}
\end{table}

Table \ref{table:combining:robusttestset} contains the results of
running these algorithms after adding the poor parser to the set.
Observe that the Average Individual Parser baseline has been lowered
significantly by the addition of this parser.  The oracles have been
affected a little, and the Parser Switch Oracle shows that in at least
two cases the hypothesis produced by the poor parser best matched the
reference sentence.

The precision for all of the switching algorithms except similarity
switching has gone up significantly, but generally those gains are
offset by a loss in recall. The exception is Bayes switching which
gains in precision and holds steady in recall, managing an overall
gain which is not significant, but which is indicative that the Bayes
model is more robust.  This is not surprising, considering that the
Bayes model is the only one that uses parameters indicating how much
to believe each of the parsers.  Overall, though, none of the
differences in F measure for the switching algorithms between this
result and the result for three parsers are significant.

While the voting results for the committee of 3 parsers were a wash,
the robustness results are where the different models show their
colors.  We see that the Alignment and Consensus technique is
extremely fragile to its belief that the parsers all perform the same,
and the Constituent Voting technique loses a considerable amount of
precision.  An even more dramatic loss is seen if we look at the Exact
match measure.  It shows that these models are statistically
significantly different (with confidence level $>.99$), and the two
non-parametric are no longer performing even as well as the Best
Individual Parser baseline.

The results of the Bayes methods, as well as constituent voting and
similarity parser switching were published by the author in another
source \cite{henderson:parsercombo}.

\section{Conclusions}

Parser diversity can be exploited to produce more accurate parsers in
many different ways.  Our oracle experiments suggested there was much
gain to be had on this task, and it is likely that there is still
more.

We gave non-parametric algorithms that perform well at this task, and
proved that under certain combining scenarios the issue of crossing
brackets does not need to be addressed.  We also proved that the
centroid-approximating switching algorithms that are based on edit
distance gave a bounded approximation of the edit distance of the true
centroid, the cheapest possible centroid parse in the space of all
parses.

The non-parametric switching algorithms were almost as resistant to
noise as the parametric algorithms.  Non-parametric algorithms are
useful in this case because parameters need to be learned from held
out data.  We would rather not hold out any data during the training
of our individual parsers.

The parametric algorithm we gave for parse hybridization dominates the
other methods in robustness, but since it is a very coarse-grained
model it performs exactly the same as the non-parametric algorithms
when it is utilized for combining parsers that all have the same base
accuracy.  It gave us the largest overall reduction in precision and
recall errors over the best individual parser, a precision error rate
reduction of 30\% and 6\% for recall.  The Bayes parser switching
algorithm was likewise the best for the algorithm for maximizing the
exact sentence accuracy metric, yielding an absolute gain of 3.7\%
with the combination of four parsers.  These gains are each as
significant as the gains that were made by each of these parsers over
their previous competition.

The parsers we have created are not practical in all situations.
Running three parsers will take three times as much computer power as
one.  However, in cases where accuracy is much more important than
speed, or where computing resources are underutilized by current
parsing technology these methods can be employed.  CPU speeds are
getting faster, as well, and machine memory is getting larger.  Many
tasks that seemed computationally ridiculous a decade ago are now
common practice on desktop PCs.

Throughout this chapter we were utilizing the fact that these parsers
were created independently and would therefore tend to have
independently distributed errors (to the extent that the corpus is
noise-free).  One should notice that there exists an even better way
to combine parsers: use parsers with orthogonal, or complementary
error distributions.  There is reason to believe that if one trains a
$(k+1)$-th to add to an ensemble of $k$ parsers with some knowledge of
the errors those $k$ parsers tend to make, one will get the best
performance gain from the ensemble not by training the $(k+1)$-th
parser to minimize raw error on the training set.  This will be
explored further in Chapter \ref{chapter:varying}.


\chapter{Varying Parsers}
\label{chapter:varying}

In Chapter \ref{chapter:combining} we showed that parsers that were
results of independent human research efforts could be combined for a
boost in accuracy and a new bound on the achievable accuracy for
parsing the Penn Wall Street Journal corpus.  It would be much better
for us to find an ensemble of parsers which complement each other.
The parsers would have to be the result of a unified research effort,
though, in which the errors made by one parser were made a priority
target for the developer of another parser.

We would willingly accept five parsers that each achieved only 40\%
exact sentence accuracy as long as they made those errors in such a
way that at least two of the five were correct on any given sentence
(and the others abstained or were wrong in different ways).  We could
achieve 100\% sentence accuracy simply by selecting the parse that was
suggested by two of the parsers.

In this chapter we will separate the issue of creating complementary
parsers from the task of creating a parser, with the goal of finding a
good method for automating the task of building complementary parsers.
Our goal is to find a method to achieve a parser performance gain by
creating an ensemble of parsers all of which are produced by the same
parser induction algorithm.

\section{Task Description}

We will start out with some definitions in order to specify our
algorithms as completely as possible.  First, let $s =
(w_1,w_2,\ldots,w_n)$ be a sentence containing $n$ words.  We will
represent a parse tree referring to that sentence as $t = \{(i,j,l) :
i,j \in \{0\ldots n\}, j\geq i, l\in \sigma_{NT}$.  Here we mean that
$(i,j)$ denotes the span of the constituent by giving the indices of
the start and end points in the sentence.  Index 0 is the position
prior to the first word and index $n$ represents the position after
the final word.  The label for the constituent is $l$, and it comes
from the set of possible nonterminal labels for the constituents,
$\sigma_{NT}$.  The constituents must be properly nested, as well.  By
that we mean 
\begin{displaymath}
(\forall  (i_a,j_a,l_a),(i_b,j_b,l_b) \in t)
(i_a \leq i_b \wedge j_a \geq j_b)
\vee
(i_a \geq i_b \wedge j_a \leq j_b)
\end{displaymath}
There is traditionally a formal dominance specified when $i_a=i_b$
and $j_a=j_b$, but we are not including that in our model.
Typically it just follows simple global rules on the constituent
labels, such as constituents marked as sentences dominating
constituents marked as verb phrases in these cases.

Let $f:S\rightarrow T$ (a function specified by an algorithm) be a
parser that produces a tree $t\in T$ given the observed sentence $s\in
S$.  A bracketed corpus $corp_\phi\in Corp_\phi$, a bag of examples,
can be seen as a function from the set of possible examples $\phi$
into the whole numbers, $corp_\phi: \phi\rightarrow N, corp_\phi$,
where the number associated with a particular example denotes the
number of times the example appears in the bag.  We will typically use
the notation for a collection instead, though, as it is more
straightforward in a number of cases: $corp =
\langle(s_1,t_1),(s_2,t_2),\ldots,(s_m,t_m) \rangle$.  Here there are
$m$ samples in the corpus, and some of the $(s,t)$ entries may be
repeated.  When not specified, we will be talking about $corp_{S\times
  T}$, a bracketed parse tree corpus, where $S$ and $T$ will be
understood.

An unbracketed corpus can be seen as a projection of a bracketed
corpus in such a way that the trees are removed:
\begin{eqnarray*}
  uncorp_S:S\rightarrow N
  \\
  uncorp(corp) = \sum_{t\in T} corp(s,t)
\end{eqnarray*}
Alternatively, it is the collection of $s_i$ from the $(s_i,t_i)$
pairs of a bracketed corpus.
The
corpus resulting from applying a parser to an unbracketed corpus
$uncorp_S$ is the function that can be tabularly specified as
\begin{displaymath}
  corp_{S\times T}=\{((s,f(s)),uncorp_S(s))\}
\end{displaymath}
We will alternatively call this construction $f\bullet uncorp_S$.
It has a straightforward equivalent in the collection notation.
Note that at this point we are allowing only deterministic parsers to
be considered.  A nondeterministic parser might not produce the same
tree each time it encounters a particular sentence.

A parser induction algorithm creates a parser $f \in F$ from a corpus
of sentences and their associated trees, $corp_{S\times T}$.
\begin{displaymath}
  g: Corp_{S\times T}\rightarrow F
\end{displaymath}

The formal statement of our goal is that we want to find a general
method for using only a single parser induction algorithm $g$ and a
single given corpus $c$ to produce a parser that performs better
(makes fewer errors under some metrics) than the parser that is the
result of $g(c)$ when possible. 

Given $Err: Corp_\phi\times Corp_\phi \rightarrow\Re$, the ultimate
goal of corpus-based parsing is to find $g^{**}$:
\begin{displaymath}
g^{**} = \argmin{g} E_D[Err(corp_{test},g(corp_{train})\bullet
uncorp(corp_{test}))] 
\end{displaymath}
where $corp_{test}$ and $corp_{train}$ are both drawn from the same unknown
underlying distribution, $D$.  For practical reasons, however, parser
induction algorithms typically attempt to minimize
$Err(corp_{train},g(corp_{train})\bullet uncorp(corp_{train})$ because
it is fully observable.  However, the designers keep in mind that they
want to be well-defined and useful over the distribution of possible
sentences, and straightforward memorization of the training corpus is
not enough.

Our goal is to try to build an ensemble of parsers $F_{ensemble}
\subset F$, each one created using induction algorithm $g$ together
with a function for combining their outputs such that the composite
parser, $f^\prime$, has the following property:

\begin{displaymath}
Err(corp,f^\prime\bullet uncorp(corp)) 
\leq
Err(corp,g(corp)\bullet uncorp(corp))
\end{displaymath}

Moreover, we would like

\begin{displaymath}
\lim\limits_{|F_{ensemble}|\rightarrow \infty}
Err(corp,f^\prime\bullet uncorp(corp))
=
\min_{f^{\prime\prime} \in F}
Err(corp,f^{\prime\prime}\bullet uncorp(corp))
\end{displaymath}
That is, in the limit our techniques should do as good as any possible
parser at parsing the training corpus.  This is reasonable precisely
because of our uncertainty about $D$.

We will describe two methods for this which give different results and
different restrictions on the $g$ for which they are successful.  We
will also discuss the computational issues involved and some useful
side effects.

\section{Creating A Diverse Ensemble}

We have already seen in Chapter \ref{chapter:combining} that a set of
independently-created parsers tend to make independent decisions that
can be combined to reduce errors.  It can be argued that the parsers
we used were not really independently created, however, for a number
of reasons:

\begin{itemize}
\item The parsers were all trained on the same training data.
\item The authors consulted much of the same linguistic theories in
  the course of their research.
\item The parsers were selected because they were published and
  publicly available.  Any bias in reviewers that makes them tend to
  accept some papers and not others are included here.  For example,
  the authors are all excellent writers.  It could be the case,
  however unlikely, that the best people at designing parser induction
  systems are barely literate or just academically shy.
\item The parsers were all designed by humans.  We really have no way
  to determine what bias humans bring to the world of designing parser
  induction systems.  Presumably it is a positive bias, but we cannot
  determine this experimentally without a comparison.
\end{itemize}

There is a recently-developed statistical technique for removing
biases and reducing variance known to the machine learning community:
bagging.  We will show that bagging does perform a diverse ensemble
using only a single parser induction algorithm and a single dataset.
Furthermore the ensemble can be combined for a parsing performance
gain.

\subsection{Background: Bagging}

Efron and Tibshirani developed methods for estimating statistics
describing a dataset using a machine-intensive technique called
bootstrap estimation.  In short, they found that they could reduce the
systematic biases introduced by many estimation techniques by
aggregating estimates that they made on randomly drawn representative
resamplings of those datasets.\footnote{The representative resamplings
  were designed to be the same size as the original datasets, and each
  sample was chosen uniformly at random \emph{with replacement}.}
That seminal work \cite{efron93:bootstrap} led to Breiman's refinement
and application of their techniques for machine learning
\cite{breiman96:bagging}.  His technique is called \emph{bagging},
short for ``bootstrap aggregating''.

Bagging attempts to find a set of classifiers which are consistent
with the training data, different from each other, and smoothly
distributed such that the most likely classifier to be added to the
ensemble is the classifier created based on the training data.

\begin{algorithm}
{Bagging Predictors (Breiman, 1996)}
{ Given: training set $\mathcal L = \{(y_i,x_i), i \in \{1\ldots
  m\}\}$ where $y_i$ is the label for example $x_i$, classification
  induction algorithm $\Psi:Y\times X \rightarrow \Phi$ with
  classification algorithm $\phi \in \Phi$ and $\phi:X\rightarrow Y$.
}
  
\item Create $k$ bootstrap replicates of $\mathcal L$ by sampling $m$
  items from $\mathcal L$ \emph{with replacement}.  Call them
  $\mathcal L_1 \ldots \mathcal L_k$.
  
\item For each $j\in \{1\ldots k\}$, Let $\phi_j = \Psi(\mathcal L_j)$
  be the classifier induced using $\mathcal L_j$ as the training set.
  
\item If $Y$ is a discrete set, then for each $x_i$ observed in the
  test set, $y_i = \mathrm{mode} \langle \phi_j(x_i)\ldots
  \phi_j(x_i)\rangle$.  We are taking $y_i$ to be the value predicted
  by the most predictors, the majority vote.
\footnotemark
\end{algorithm}
\footnotetext{
  When $|Y|=|\Re|$, the regression form of bagging is $y_i =
  \sum\limits_j 
  \phi_j(x_i)$.
} 

There are two interesting qualitative properties of bagging.  First,
bagging relies on the chosen classifier induction algorithm's lack of
\emph{stability}.  By this we mean the chosen algorithm should be
easily perturbed.  A small change in the training set should produce a
significant change in the resulting classifier.  Neural networks and
decision trees are examples of unstable classifier systems, whereas
k-nearest neighbor is a stable classifier.  Secondly, bagging is
theoretically resistant to noise in the data and bias in the learning
algorithm.  Unfortunately it is resistant to bias in the learning
algorithm even when that bias is favorable.  In some cases classifier
induction algorithms that perform well in isolation can perform poorly
in ensemble for this reason.  Empirical results have verified both of
these claims \cite{quinlan96:bagging_boosting,
  maclin97:baggingvboosting, bauer99:bagboost}.

\subsection{Bagging A Parser By Sentences}

In Algorithm \ref{algorithm:varying:parsebagging}, we give an
algorithm that applies the technique of bagging to parsing.  Here we
are leveraging our previous work on combining independent parsers to
produce the combined parser.  The rest of the algorithm is a
straightforward transformation of bagging for classifiers.  Some
exploratory work in this vein was described in \cite{hajic:ws98}.  Our
work validates their result, and explores alternative formulations.

\begin{algorithm}
{Bagging A Parser}
{Given corpus $corp$ with size $m=|corp|=\sum_{s,t}corp(s,t)$ and
  parser induction algorithm $g$.}
\label{algorithm:varying:parsebagging}
\item Draw $k$ bootstrap replicates $corp^1\ldots corp^k$ of $corp$
  each containing $m$ samples of $(s,t)$ pairs randomly picked from
  the domain of $corp$ according to the distribution $D(s,t) =
  corp(s,t)/|corp|$.  Each bootstrap replicate is a bag of samples,
  where each sample in a bag is drawn randomly with replacement from
  the bag corresponding to $corp$.
\item Create parser $f^i = g(corp^i)$ for each $i$. $F_{ensemble} =
  \bigcup_i \{f^i\}$.
\item Given a novel sentence $s_{test}\in corp_{test}$, combine the
  collection of hypotheses $ t_i =f^i(s_{test})$ using the unweighted
  constituent voting scheme of Section
  \ref{section:combining:constvote}.
\end{algorithm}

\subsubsection{Uniform Distribution over Sentences}

The first set of experiments we carried out investigated Algorithm
\ref{algorithm:varying:parsebagging} as it is specified.  Later we
will present results of experiments using modified versions.

\begin{sidewaysfigure}[htbp]
  \centering
\fbox{
\begin{tabular}{rl}
    \epsfig{file=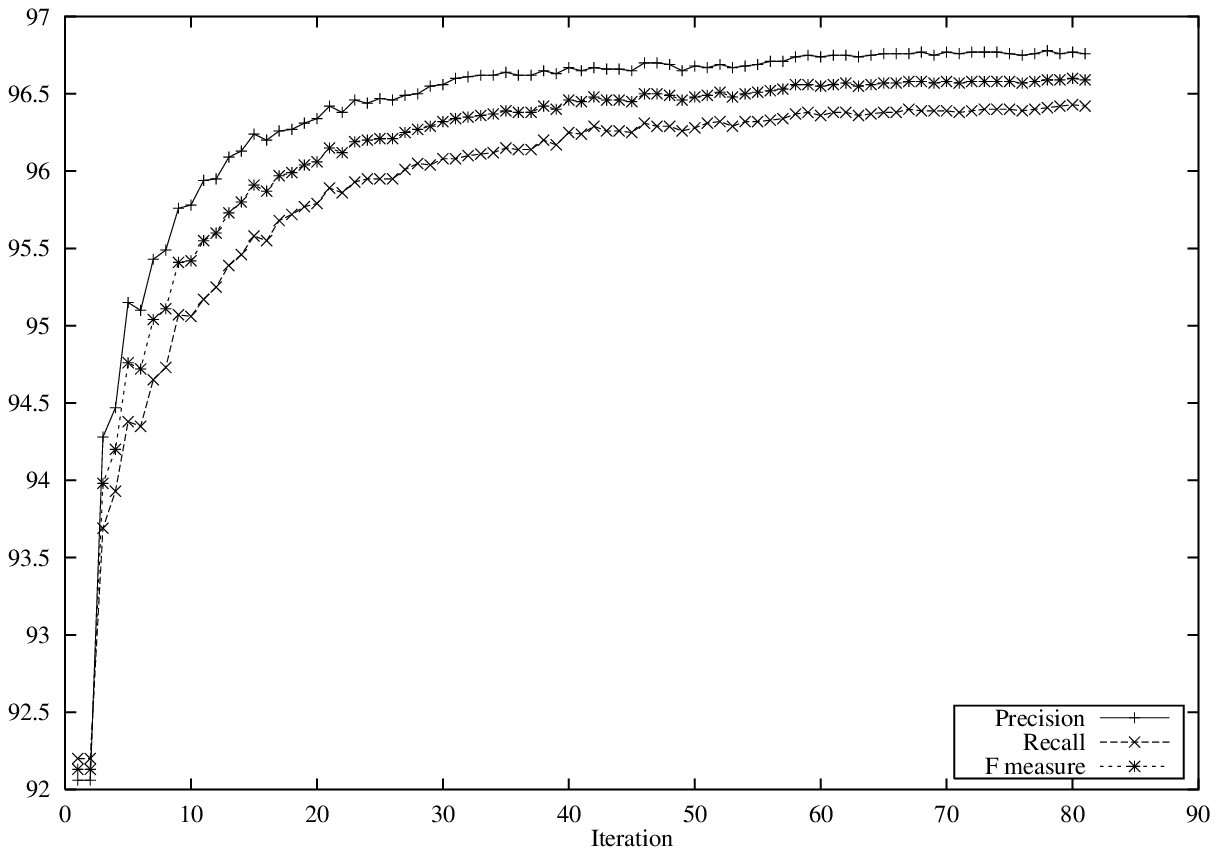, width= 0.45\textwidth}
&
    \epsfig{file=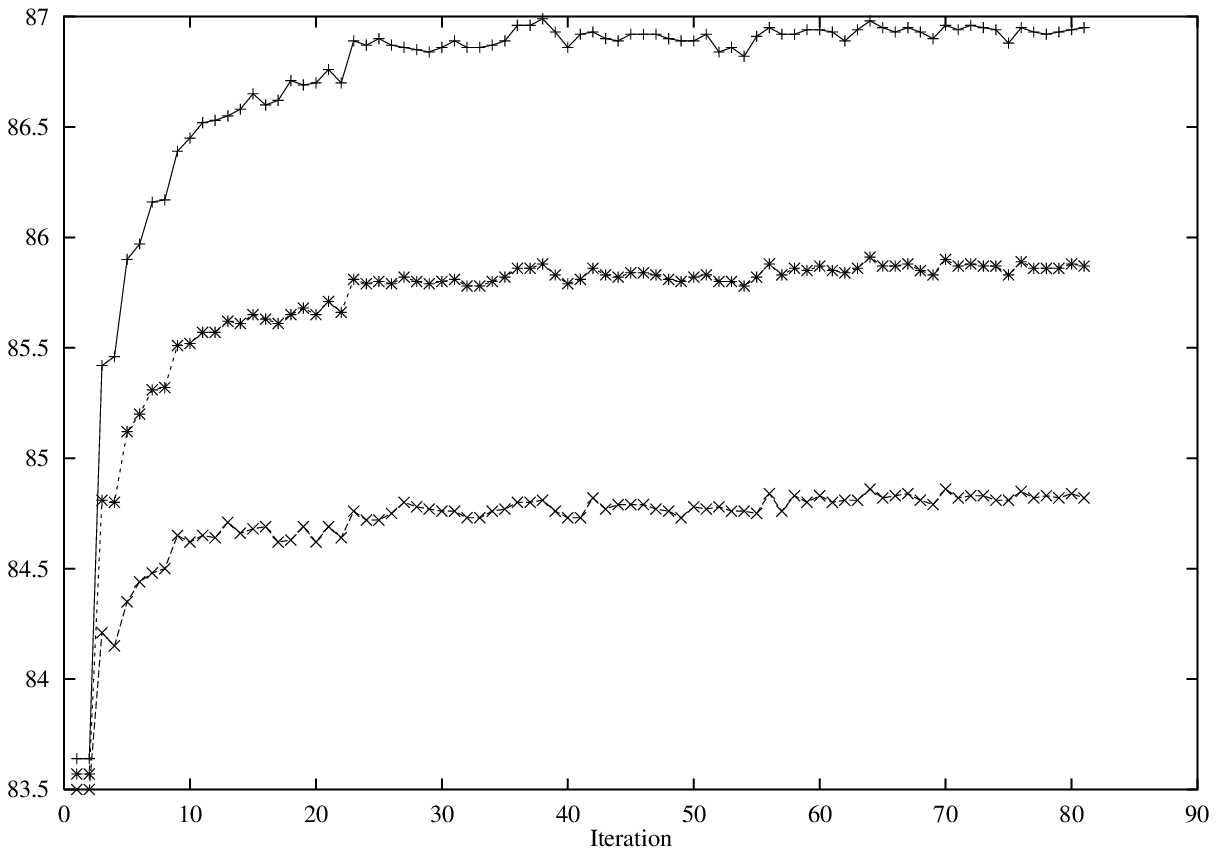, width= 0.45\textwidth}
\\
    \epsfig{file=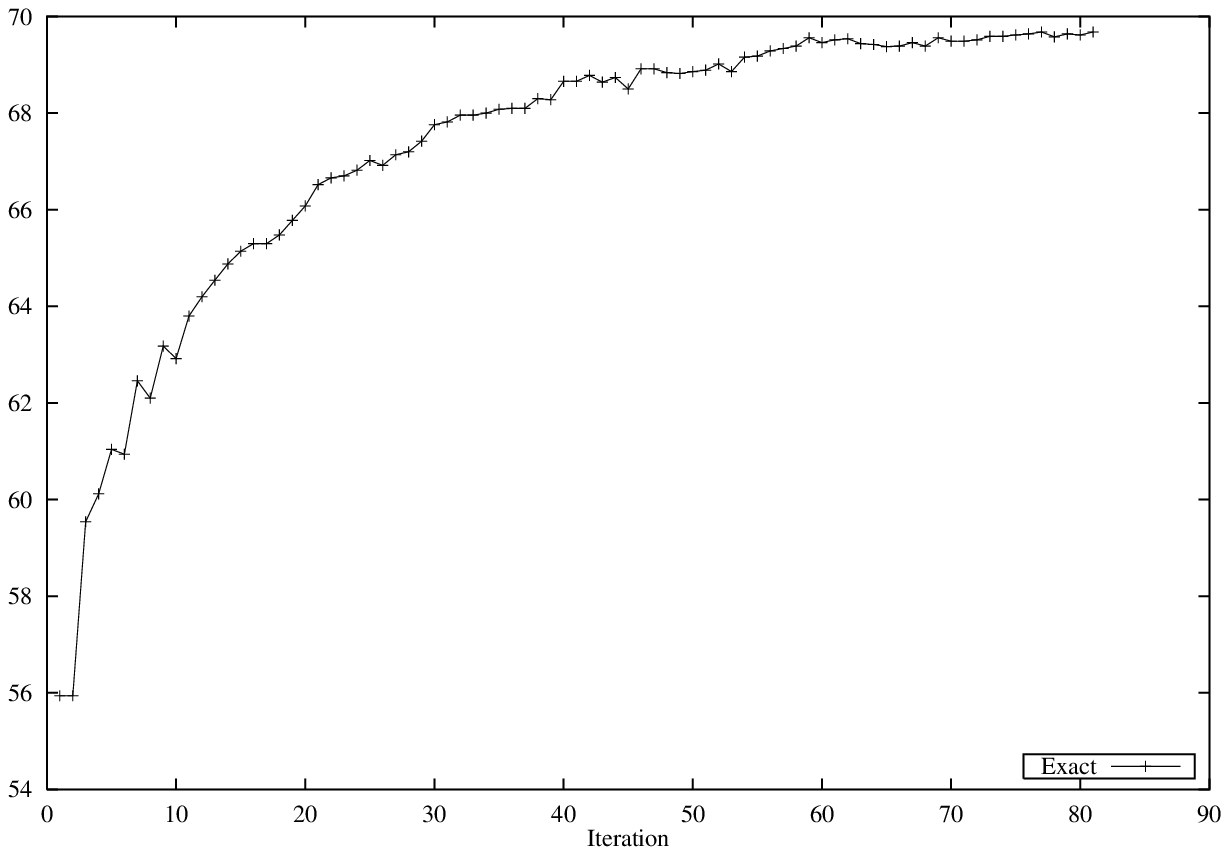, width= 0.45\textwidth}
&
    \epsfig{file=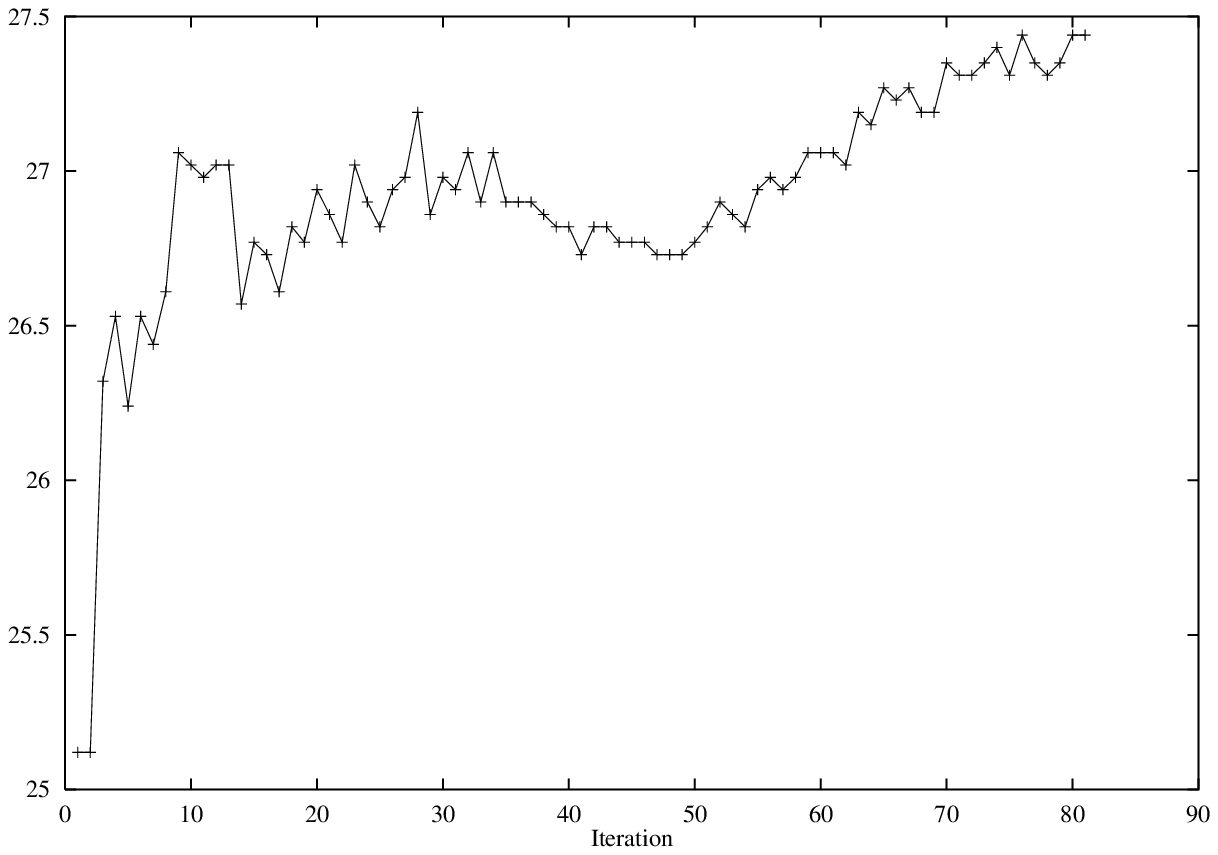, width= 0.45\textwidth}
\end{tabular}
}


  \caption{Bagging a Small Training Set} 
  \label{fig:varying:bag.small.sent.uni}
\end{sidewaysfigure}

\begin{table}[htbp]
  \centering
\begin{tabular}{|c|l|rr|rr|rr|}
       \hline
      Set
      &\multicolumn{1}{c|}{Instance} 
      &\multicolumn{1}{c}{P} 
      &\multicolumn{1}{c|}{R} 
      &\multicolumn{1}{c}{F}  
      &\multicolumn{1}{c|}{Gain}  
      &\multicolumn{1}{c}{Exact}
      &\multicolumn{1}{c|}{Gain} \\
      \hline 
Training&Original Parser  & 97.07 &  97.30 &  97.18 &NA&    73.3 & NA 
\\
&       Initial & 92.06 & 92.20 & 92.13 &  0.00 & 55.9  &  0.0
\\
&       BestF(80)       & 96.77 & 96.43 & 96.60 &  4.47 & 69.6  & 13.7
\\
&       Final(81)       & 96.76 & 96.42 & 96.59 &  4.46 & 69.7  & 13.7
\\
      \hline 
Test & Original Parser  & 86.03 & 85.43 & 85.73  & NA & 28.6 & NA
\\
&   Initial & 83.64 & 83.50 & 83.57 &  0.00 & 25.1  &  0.0
\\
&       TrainBestF(80)  & 86.94 & 84.84 & 85.88 &  2.31 & 27.4  &  2.3
\\
&       TestBestF(64)   & 86.98 & 84.86 & 85.91 &  2.34 & 27.1  &  2.0
\\
&       Final(81)       & 86.95 & 84.82 & 85.87 &  2.30 & 27.4  &  2.3
\\
\hline
\end{tabular}


  \caption{Bagging a Small Training Set} 
  \label{table:varying:bag.small.sent.uni}
\end{table}

In Figure \ref{fig:varying:bag.small.sent.uni} we see the result of
running a bagging experiment using 5000 sentences in the training set.
These were the first 5000 sentences of the Penn Treebank sections
01-21.  The parser induction algorithm we used in all of these
experiments of this chapter was Collins's model 2 parser
\cite{collins:parsing97}.  It produced the best parser we had access
to for the experiments in Chapter \ref{chapter:combining}, and we were
given access to the code for training the parser.  The ensemble that
was produced for this experiment contained 81 parsers upon
completion.\footnote{It is intuitively more reasonable to have an odd
  number of parsers when possible to eliminate the issue of breaking
  ties during construction of the final hypothesis.}  The graphs on
the left are from the training set, and the ones on the right are from
the test set.  The curves in the upper graphs are precision, recall,
and F-measure.  In the lower graph the curves represent exact sentence
accuracy.  The independent variable for all of these graphs is the
number of bags that are being combined.

Table \ref{table:varying:bag.small.sent.uni} gives some details from
the curves.  It gives the values of the various metrics for the first
bag, combining the first $n$ generated bags that give the best
F-measure, and combining all of the bags, all computed on the training
set.  The lower entry gives the same values, along with the choice an
omniscient observer would make for $n$ if it could look at the test
set.

In the figure we see that on the training set all of our measures
increase, and that precision increases only slightly more than recall.
On the test set, however, we notice that recall does not get nearly as
large a gain as precision.  Also, the exact sentence accuracy gain,
while significantly better than the initial state at every point, does
not increase monotonically.  These curves also suggest an asymptotic
effect: there is not much more gain to be had by increasing the
ensemble size.

\subsubsection{Uniform Distribution over Constituents}

In the previous experiment each sentence is treated equally important
in the training set by giving it equal weight during resampling.  Each
sentence from the training corpus is not equally informative to the
parser induction algorithm, however.  Longer sentences contain more
constituents and lexical items, presenting more potential information
for learning algorithms.

\begin{sidewaysfigure}[htbp]
\centering
\fbox{
\begin{tabular}{rl}
    \epsfig{file=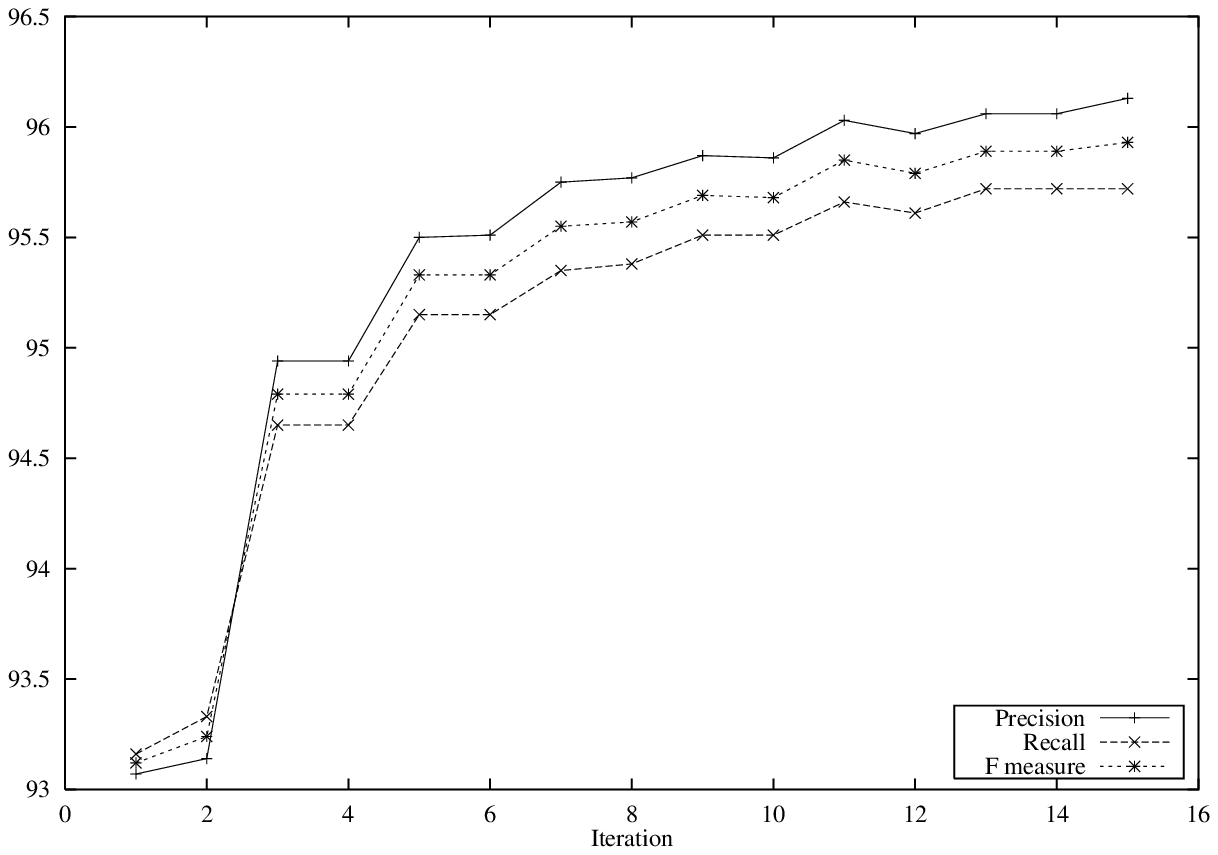, width= 0.45\textwidth}
&
    \epsfig{file=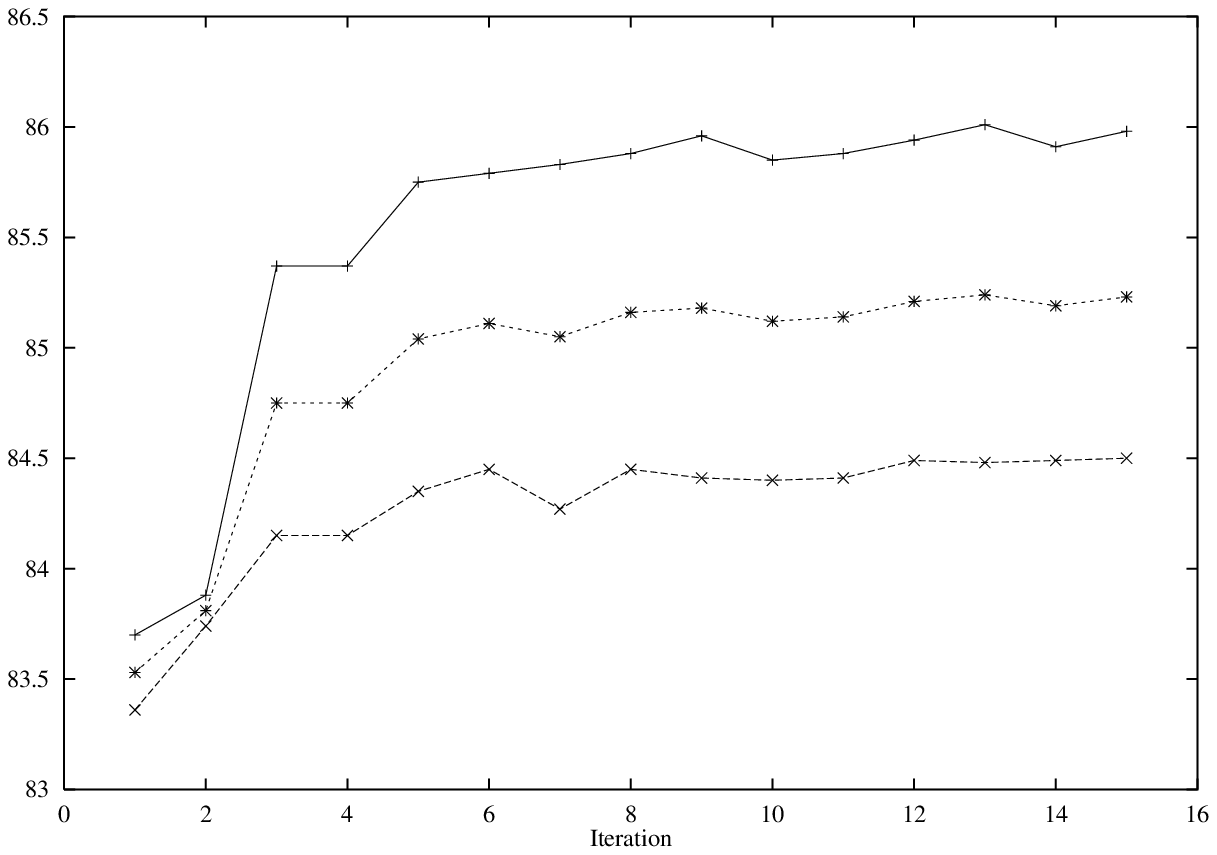, width= 0.45\textwidth}
\\
    \epsfig{file=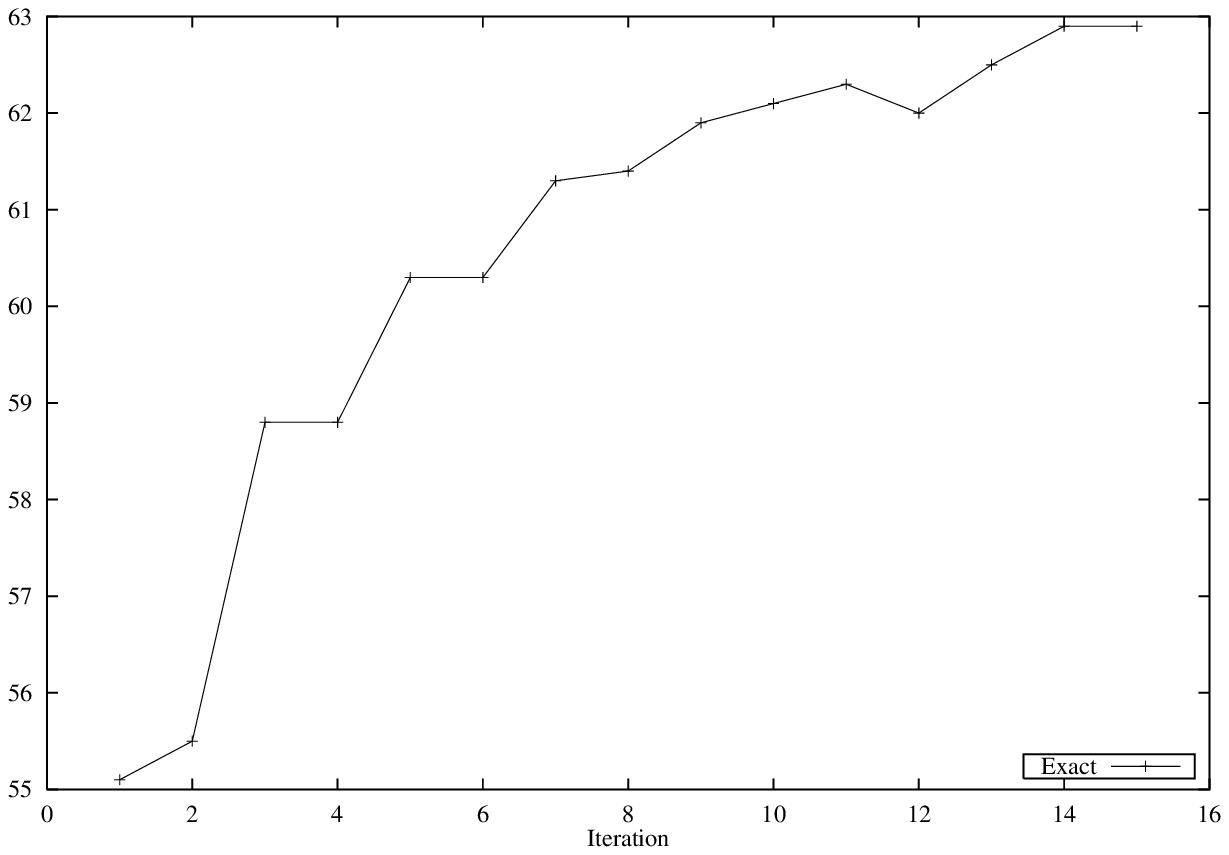, width= 0.45\textwidth}
&
    \epsfig{file=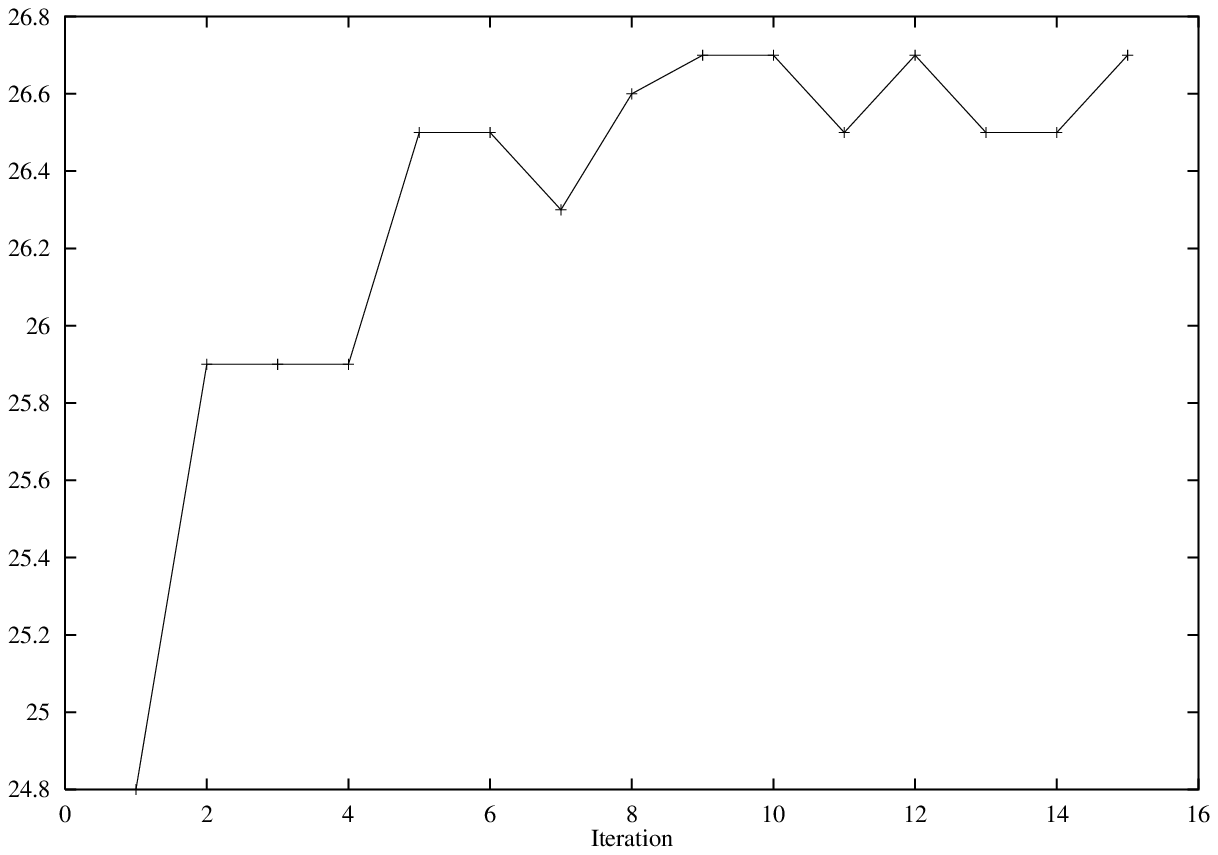, width= 0.45\textwidth}
\end{tabular}
}


  \caption{Bagging with a Uniform Distribution over Constituents} 
  \label{fig:varying:bag.const.uni}
\end{sidewaysfigure}

\begin{table}[htbp]
  \centering
\begin{tabular}{|c|l|rr|rr|rr|}
       \hline
      Set
      &\multicolumn{1}{c|}{Instance} 
      &\multicolumn{1}{c}{P} 
      &\multicolumn{1}{c|}{R} 
      &\multicolumn{1}{c}{F}  
      &\multicolumn{1}{c|}{Gain}  
      &\multicolumn{1}{c}{Exact}
      &\multicolumn{1}{c|}{Gain} \\
      \hline 
Training
&Original Parser  & 97.07 &  97.30 &  97.18 &NA&    73.3 & NA 
\\
&       Initial & 93.07 & 93.16 & 93.12 &  0.00 & 55.1  &  0.0
\\
&       BestF(15)       & 96.13 & 95.72 & 95.93 &  2.81 & 62.9  &  7.8
\\
&       Final(15)       & 96.13 & 95.72 & 95.93 &  2.81 & 62.9  &  7.8
\\
      \hline 
Test & Original Parser  & 86.03 & 85.43 & 85.73  & NA & 28.6 & NA
\\
&   Initial & 83.70 & 83.36 & 83.53 &  0.00 & 24.8  &  0.0
\\
&       TrainBestF(15)  & 85.98 & 84.50 & 85.23 &  1.70 & 26.7  &  1.9
\\
&       TestBestF(13)   & 86.01 & 84.48 & 85.24 &  1.71 & 26.5  &  1.7
\\
&       Final(15)       & 85.98 & 84.50 & 85.23 &  1.70 & 26.7  &  1.9
\\
\hline
\end{tabular}


  \caption{Bagging with a Uniform Distribution over Constituents} 
  \label{table:varying:bag.const.uni}
\end{table}

A bagging experiment was performed in which the distribution over
sentences was calculated in proportion to their length ($s_{len}$):
\begin{displaymath}
  D(s,t) = \frac{s_{len}corp(s,t)}{\sum\limits_{s^\prime,t^\prime}s^\prime_{len}corp(s^\prime,t^\prime)}
\end{displaymath}
In this way we were approximating a distribution that was weighted
based on the number of constituents in a sentence.  The number of
constituents in a parse tree is loosely proportional to the number of
tokens in the sentence to the extent that the valency of a constituent
(number of children) is constant.

The results from the experiment is given in Figure
\ref{fig:varying:bag.const.uni} and Table
\ref{table:varying:bag.const.uni}.  When comparing just the first 15
parsers from the previous experiment we see that this modification
performs almost exactly the same.  The only significant difference is
that the parsers from the previous experiment have a higher exact
sentence accuracy on the training set.  Since we picked shorter
sentences less often for inclusion in the training sets of the
bootstrap replicates, they were memorized by parsers less often.  This
explanation looks plausible because the observation does not hold on
the test set to the same extent.

One interesting (an unexpected) difference is that the individual
parsers generated in this way have a higher average training set
precision and recall than those of the previous experiment.

\subsubsection{Uniform Distribution over Constituent Possibilities}

A tree for an entire sentence is not the smallest-scale measurable
decision that a parser must make.  Each parser can be viewed as acting
as a constrained binary classifier acting on potential labelled
constituents in the parse.  The constraints come from the fact that
the set of constituents for a particular sentence must form a nested
bracketing, a tree.  For the purposes of this chapter, however, we
will be ignoring the tree constraints.  We will be using the result
from Lemma \ref{lemma:combining:treeguarantee} and its consequences to
allow us to do this.

The number of possible constituents for a sentence, disregarding
structure, is $\frac{|\sigma_{NT}|s_{len}(s_{len}+1)}{2}$ where
$\sigma_{NT}$ is the set of nonterminals available for annotating
constituents.  The other factor is the number of possible places for a
constituent to begin and end: $\frac{s_{len}(s_{len}+1)}{2} =
{s_{len}+1\choose 2}$.  Therefore, to set sentence weights based on
the number of possible constituents:
\begin{displaymath}
  D(s,t) =
  \frac{s_{len}(s_{len}+1)corp(s,t)}
  {\sum\limits_{s^\prime,t^\prime}s^\prime_{len}(s^\prime_{len}+1)corp(s^\prime,t^\prime)} 
\end{displaymath}

\begin{sidewaysfigure}[htbp]
\centering
\fbox{
\begin{tabular}{rl}
    \epsfig{file=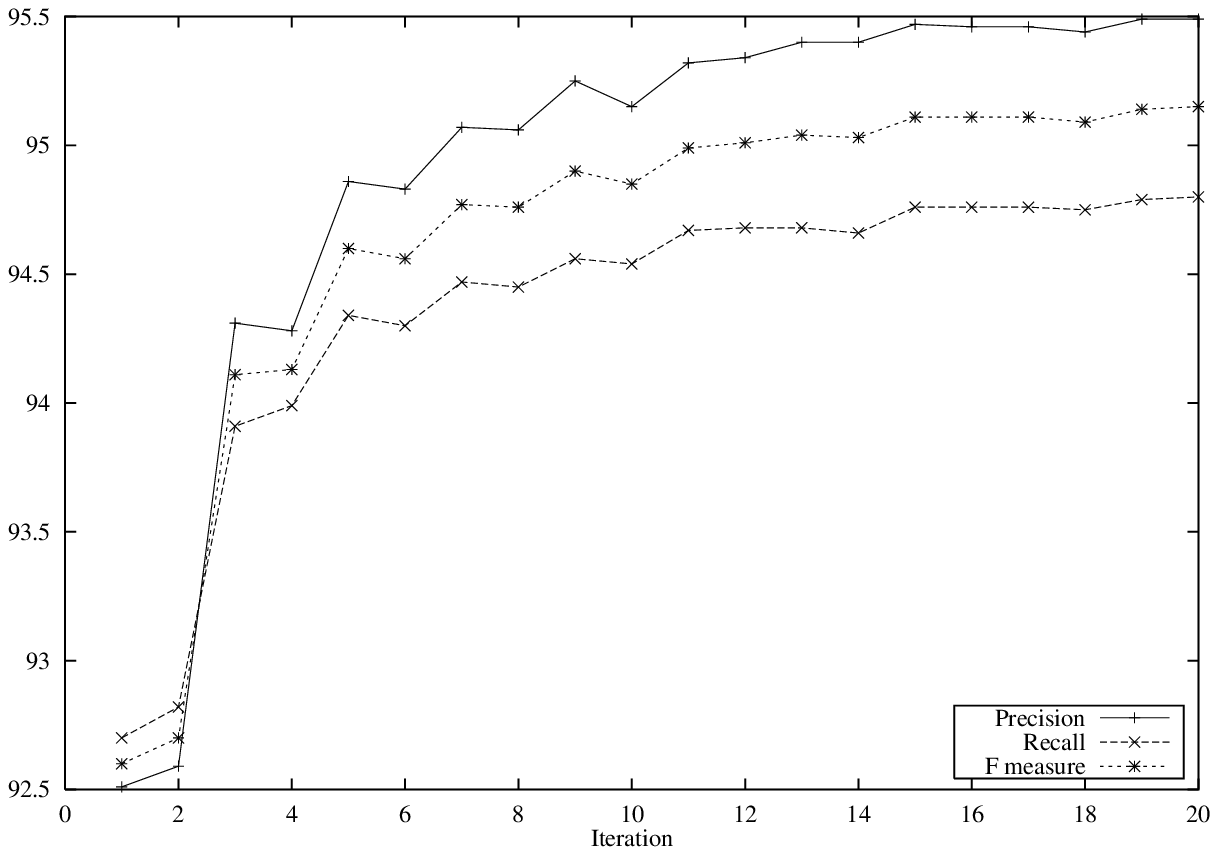, width= 0.45\textwidth}
&
    \epsfig{file=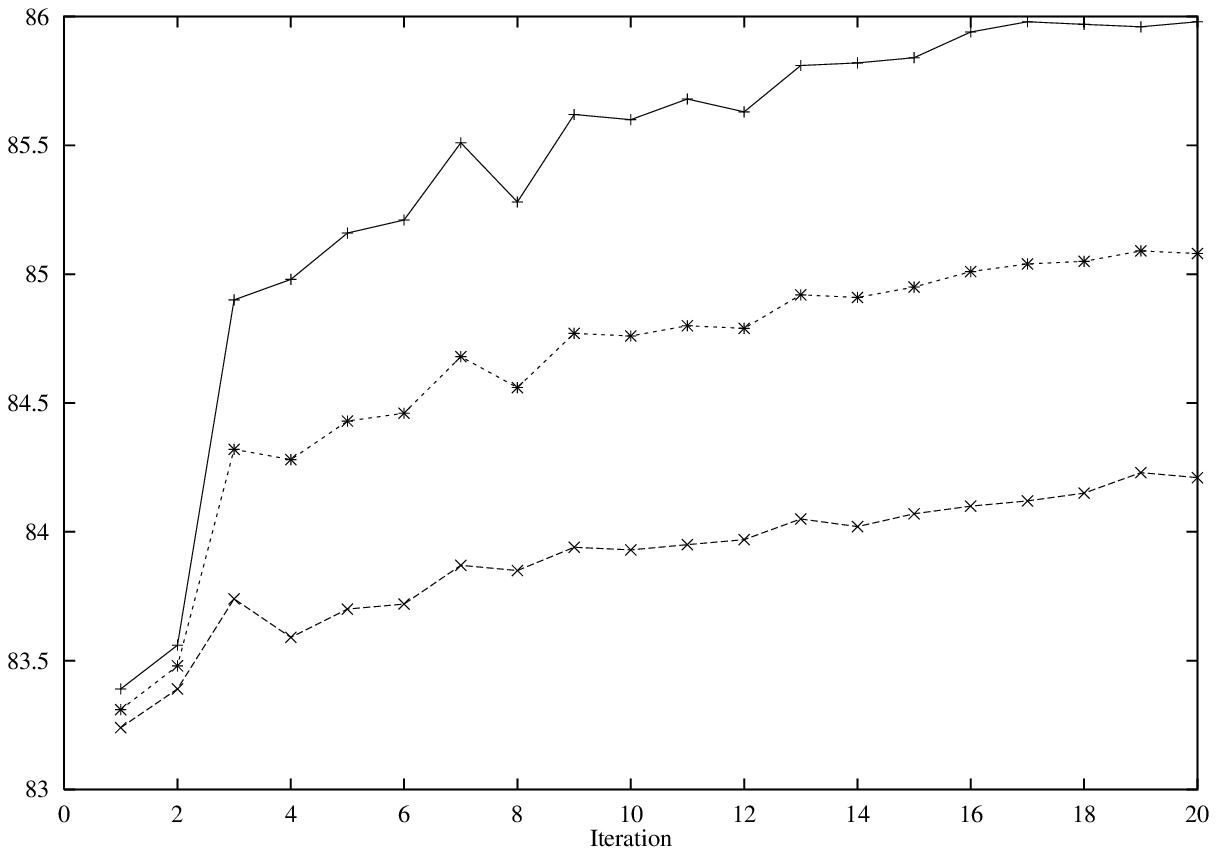, width= 0.45\textwidth}
\\
    \epsfig{file=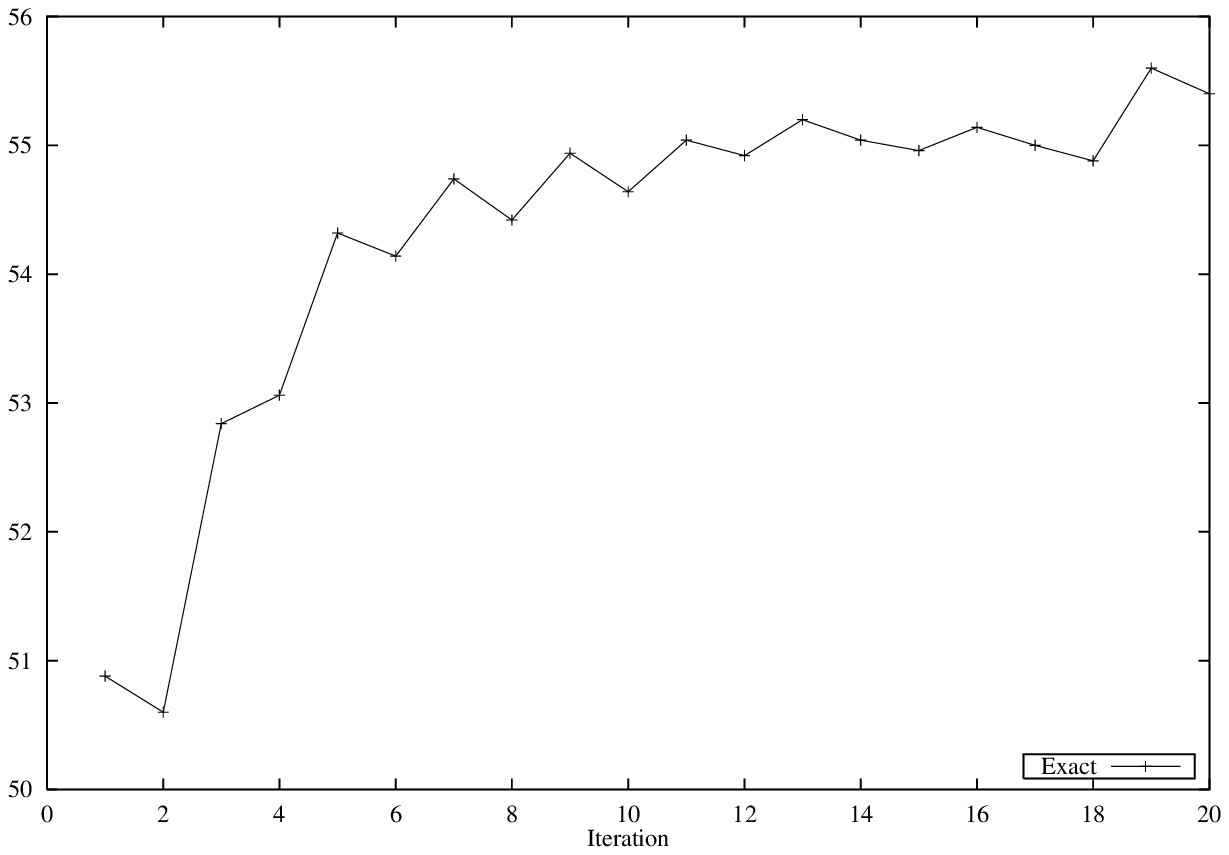, width= 0.45\textwidth}
&
    \epsfig{file=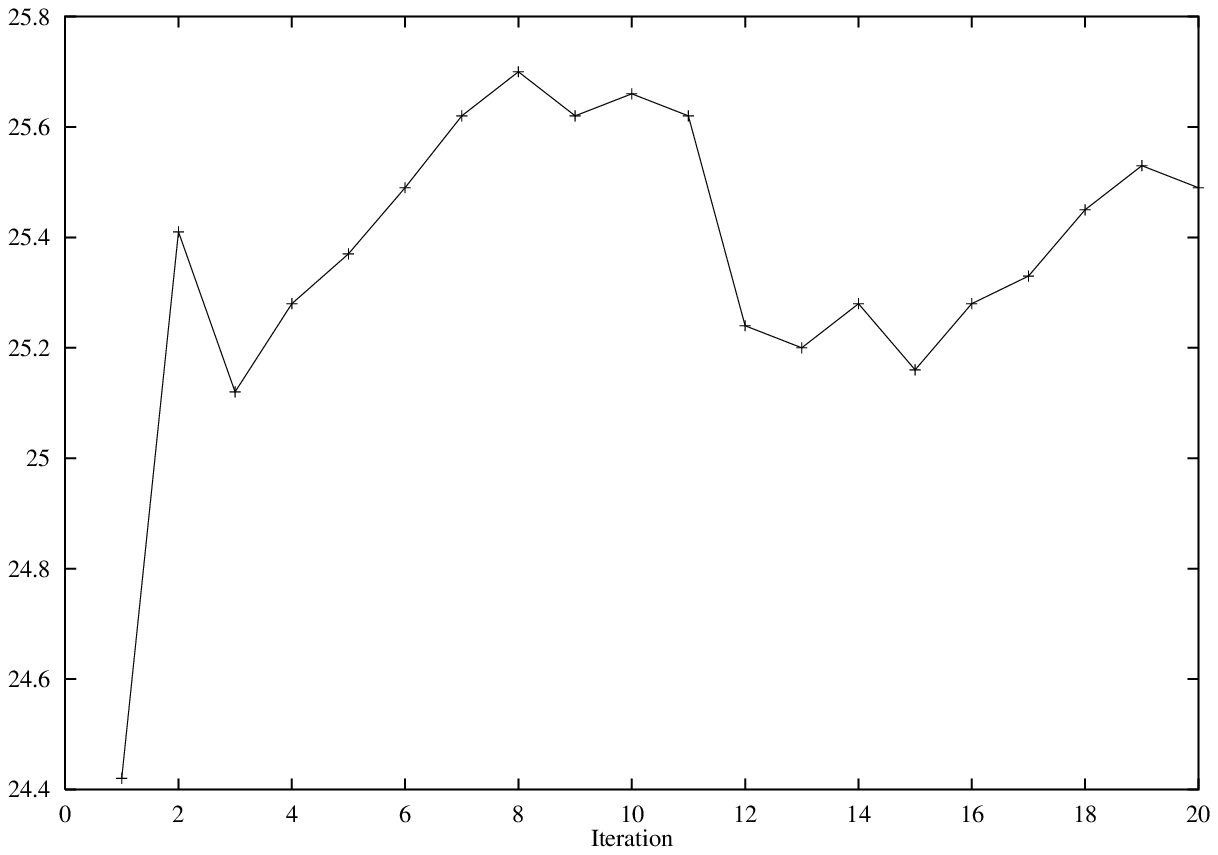, width= 0.45\textwidth}
\end{tabular}
}


  \caption{Bagging with a Uniform Distribution over Constituent
    Possibilities} 
  \label{fig:varying:bag.const.dec}
\end{sidewaysfigure}

\begin{table}[htbp]
  \centering
\begin{tabular}{|c|l|rr|rr|rr|}
       \hline
      Set
      &\multicolumn{1}{c|}{Instance} 
      &\multicolumn{1}{c}{P} 
      &\multicolumn{1}{c|}{R} 
      &\multicolumn{1}{c}{F}  
      &\multicolumn{1}{c|}{Gain}  
      &\multicolumn{1}{c}{Exact}
      &\multicolumn{1}{c|}{Gain} \\
      \hline 
Training
&Original Parser  & 97.07 &  97.30 &  97.18 &NA&    73.3 & NA 
\\
&       Initial & 92.51 & 92.70 & 92.60 &  0.00 & 50.9  &  0.0
\\
&       BestF(20)       & 95.49 & 94.80 & 95.15 &  2.55 & 55.4  &  4.5
\\
&       Final(20)       & 95.49 & 94.80 & 95.15 &  2.55 & 55.4  &  4.5
\\
      \hline 
\hline
Test & Original Parser  & 86.03 & 85.43 & 85.73  & NA & 28.6 & NA
\\
&   Initial & 83.39 & 83.24 & 83.31 &  0.00 & 24.4  &  0.0
\\
&       TrainBestF(20)  & 85.98 & 84.21 & 85.08 &  1.77 & 25.5  &  1.1
\\
&       TestBestF(19)   & 85.96 & 84.23 & 85.09 &  1.78 & 25.5  &  1.1
\\
&       Final(20)       & 85.98 & 84.21 & 85.08 &  1.77 & 25.5  &  1.1
\\
\hline
\end{tabular}


  \caption{Bagging with a Uniform Distribution over Constituent
    Possibilities} 
  \label{table:varying:bag.const.dec}
\end{table}

Figure \ref{fig:varying:bag.const.dec} and Table
\ref{table:varying:bag.const.dec} show the results of this experiment.
We first notice that this set of parsers has lower average
performance than those of the previous two sections.  This can be seen
in the initial classifier results.  While this set of parsers gets a
gain that is close to the other two, the final precision and recall on
the training set is significantly lower, and the final recall on the
test set is low as well.  Overall this experiment failed to produce a
better method for combining bagged parsers.

\subsubsection{Experiment: Preferring Shorter Sentences}

It has been observed that children learn language by being exposed to
simple sentences first \cite{carroll92:parsing}.  Also, we have seen
that both attempts we have made to weigh sentences more heavily based
on length has failed to produce better composite parsers.  These two
facts led us to another experiment, motivated by completely empirical
evidence, in which we weight the sentences such that shorter sentences
are preferred:
\begin{displaymath}
  D(s,t) =
  \frac{corp(s,t)/s_{len}}
  {\sum\limits_{s^\prime,t^\prime}corp(s^\prime,t^\prime)/s^\prime_{len}} 
\end{displaymath}

\begin{sidewaysfigure}[htbp]
\centering
\fbox{
\begin{tabular}{rl}
    \epsfig{file=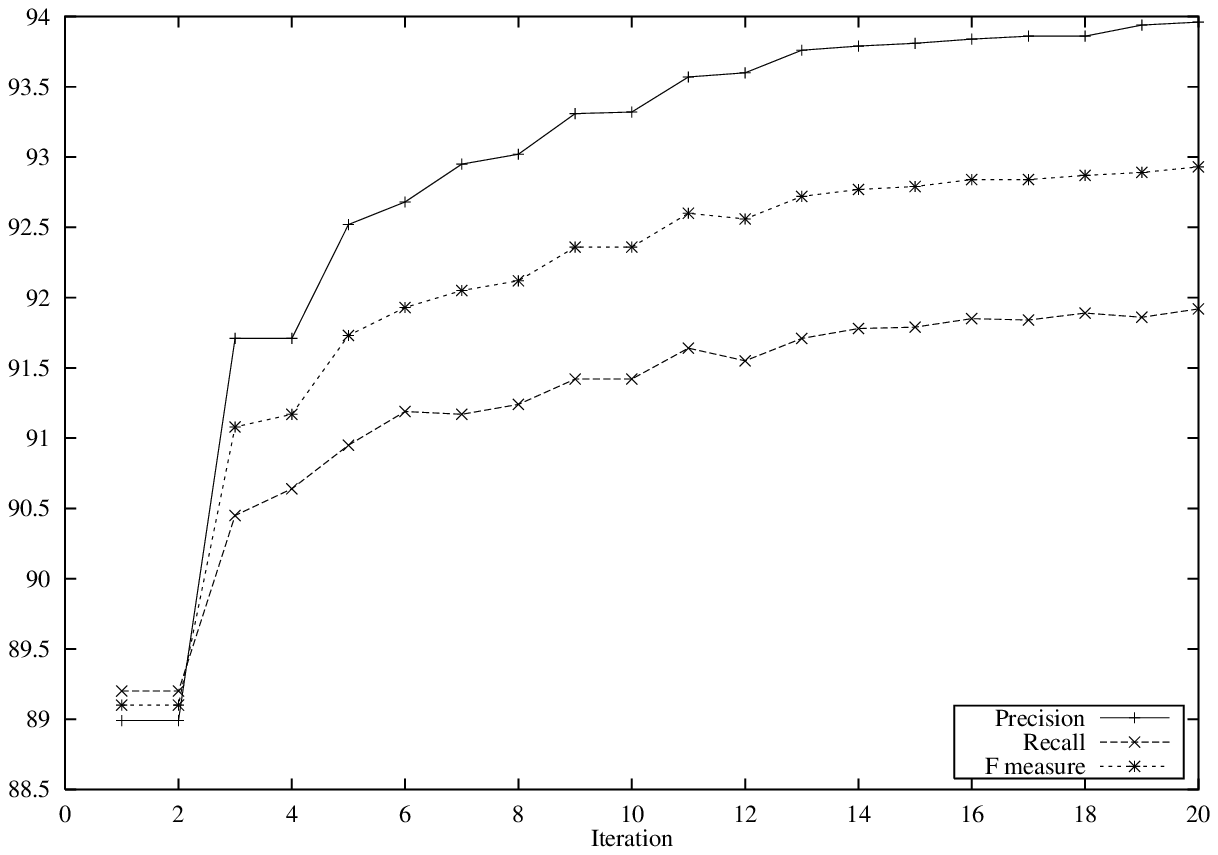, width= 0.45\textwidth}
&
    \epsfig{file=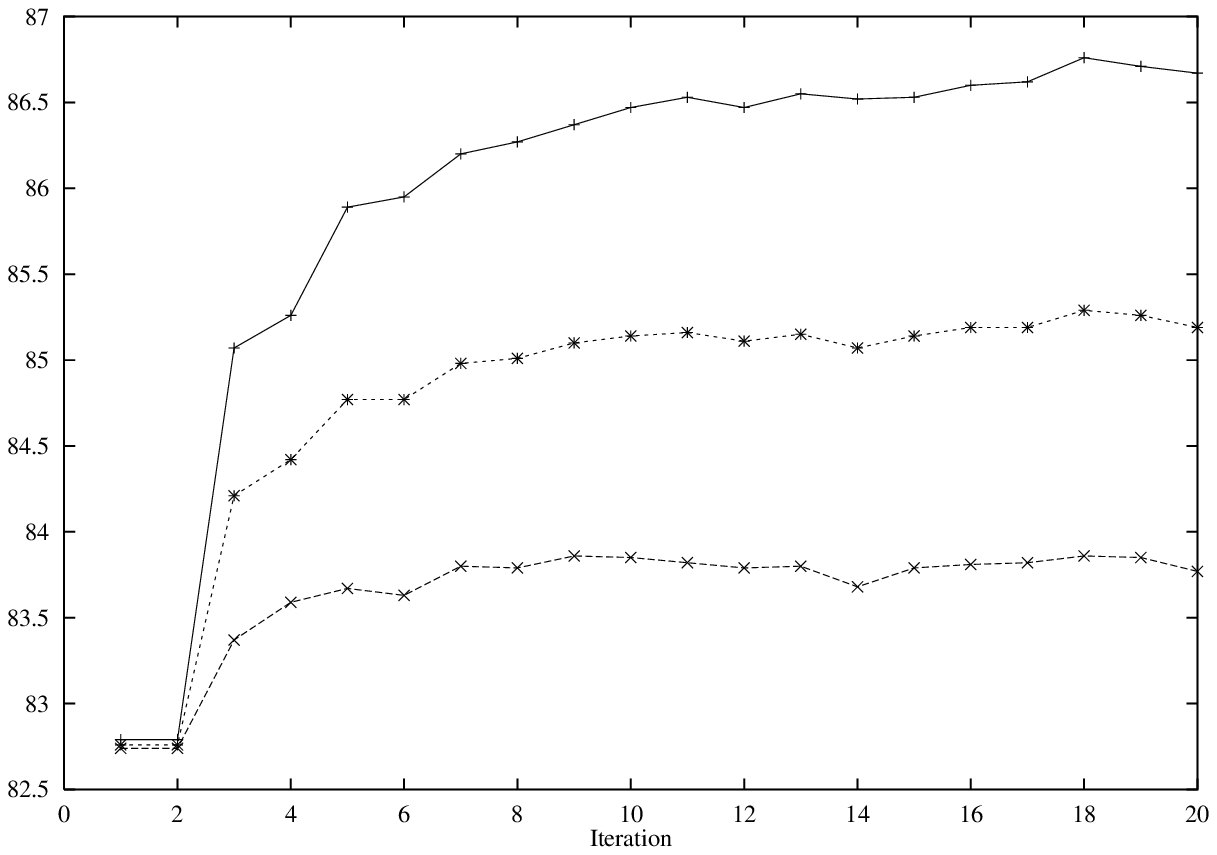, width= 0.45\textwidth}
\\
    \epsfig{file=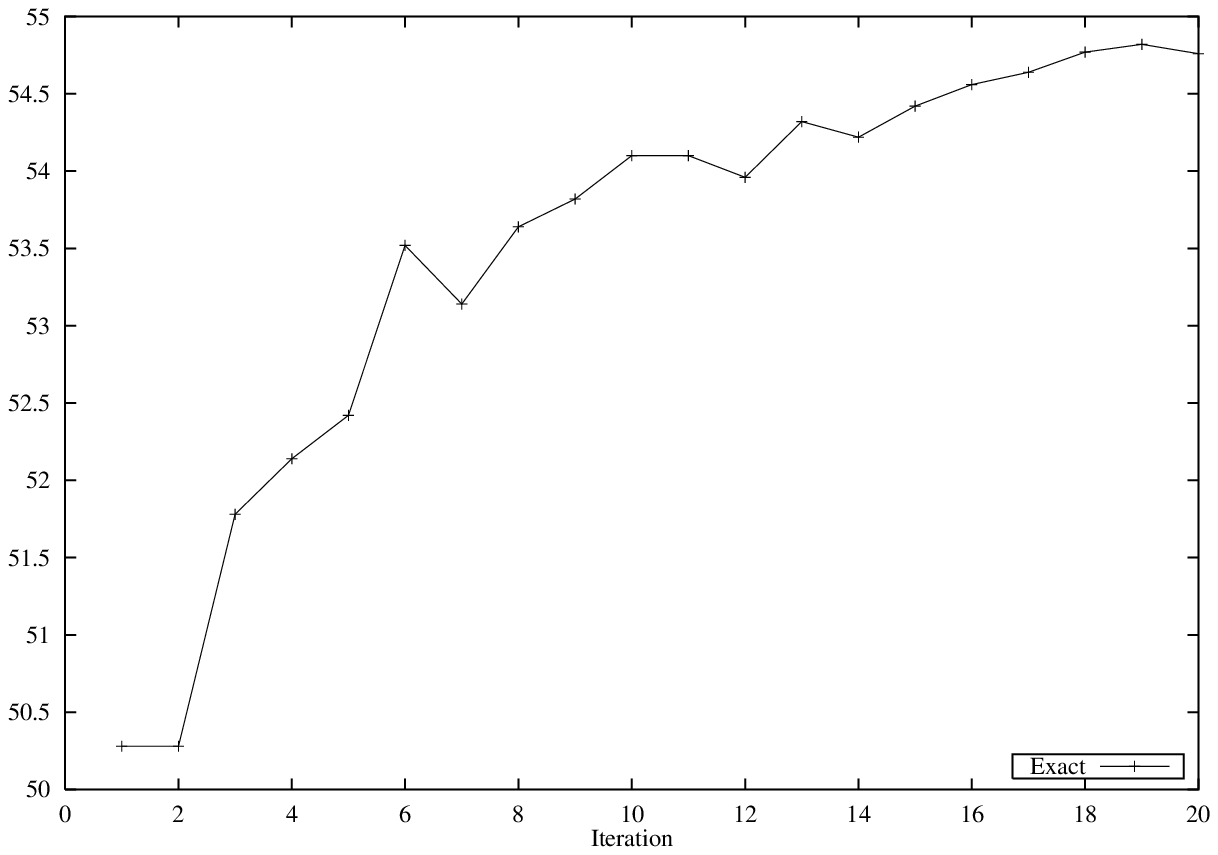, width= 0.45\textwidth}
&
    \epsfig{file=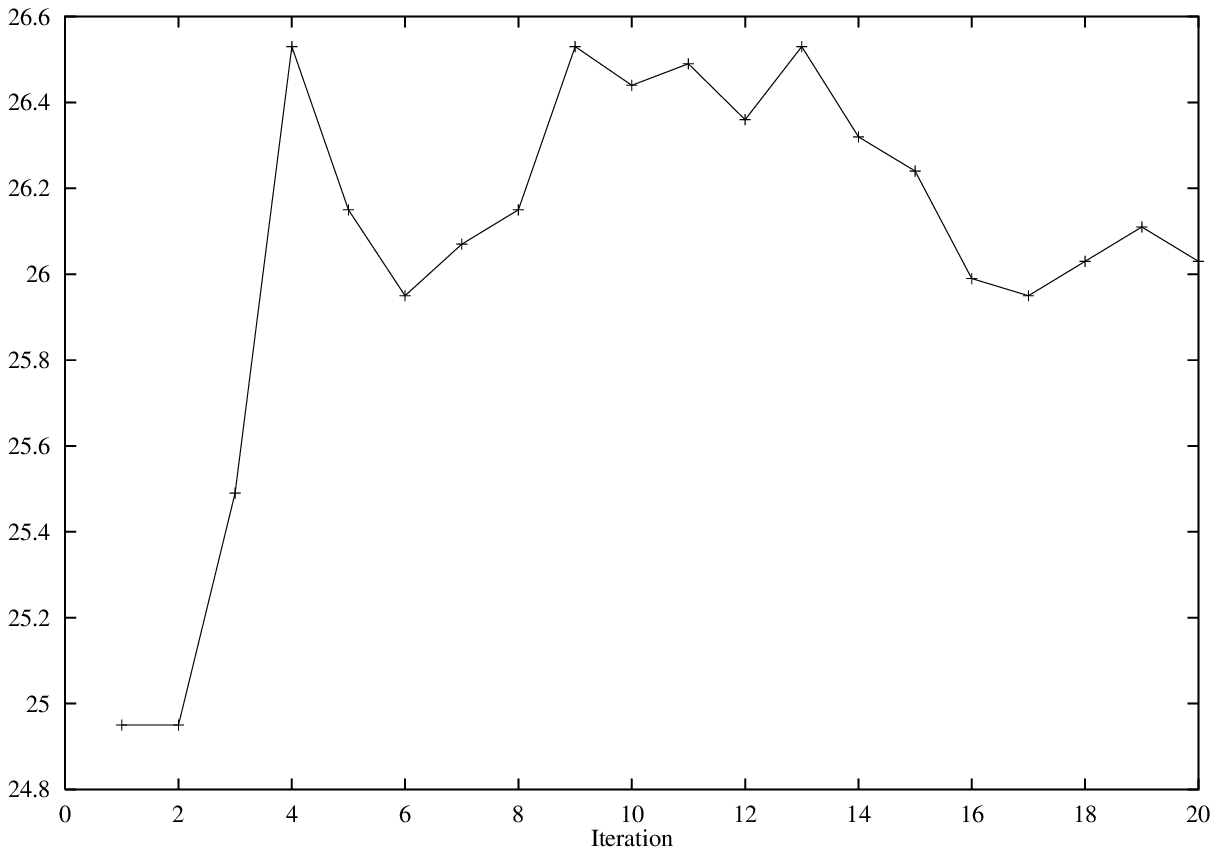, width= 0.45\textwidth}
\end{tabular}
}


  \caption{Bagging with a Preference for Shorter Sentences} 
  \label{fig:varying:bag.sent.inv}
\end{sidewaysfigure}

\begin{table}[htbp]
  \centering
\begin{tabular}{|c|l|rr|rr|rr|}
       \hline
      Set
      &\multicolumn{1}{c|}{Instance} 
      &\multicolumn{1}{c}{P} 
      &\multicolumn{1}{c|}{R} 
      &\multicolumn{1}{c}{F}  
      &\multicolumn{1}{c|}{Gain}  
      &\multicolumn{1}{c}{Exact}
      &\multicolumn{1}{c|}{Gain} \\
      \hline 
Training&Original Parser  & 97.07 &  97.30 &  97.18 &NA&    73.3 & NA 
\\
&       Initial & 88.99 & 89.20 & 89.10 &  0.00 & 50.3  &  0.0
\\
&       BestF(20)       & 93.96 & 91.92 & 92.93 &  3.83 & 54.8  &  4.5
\\
&       Final(20)       & 93.96 & 91.92 & 92.93 &  3.83 & 54.8  &  4.5
\\
      \hline 
Test & Original Parser  & 86.03 & 85.43 & 85.73  & NA & 28.6 & NA
\\
&   Initial & 82.79 & 82.74 & 82.76 &  0.00 & 24.9  &  0.0
\\
&       TrainBestF(20)  & 86.67 & 83.77 & 85.19 &  2.43 & 26.0  &  1.1
\\
&       TestBestF(18)   & 86.76 & 83.86 & 85.29 &  2.53 & 26.0  &  1.1
\\
&       Final(20)       & 86.67 & 83.77 & 85.19 &  2.43 & 26.0  &  1.1
\\
\hline
\end{tabular}


  \caption{Bagging with a Preference for Shorter Sentences} 
  \label{table:varying:bag.sent.inv}
\end{table}

The results of this experiment are shown in Figure
\ref{fig:varying:bag.sent.inv} and Table
\ref{table:varying:bag.sent.inv}.  While bagging these parsers gets a
larger gain in both precision and recall than the prior three
experiments, the base accuracy of these parsers is significantly lower
than before.  Put another way: bagging is \emph{more successful} at
raising the accuracy of this poorly biased parser than it was at
raising the accuracy of the prior parsers that we showed.  In general
this is observed in all of the experiments: bagging can make an
ensemble of poorly performing parsers perform well, as long as they
can be perturbed by small changes in the training corpus.  

This is the conclusion of our bagging experiments.  We present the
results of our best method in Section \ref{section:varying:evaluation},
where the training set is the entire Treebank.

\section{Adding a Complementary Parser}

Bagging parsers has proved itself to be a successful technique for
automatically creating a diverse ensemble, but the design of an
ensemble in which the parsers are designed to make complementary (not
just independent) errors remains to been explored.  As before, the
only freedom that remains in creating the ensemble is the distribution
of the training data.  For experimental purposes the parser induction
system $g$ will once again be fixed to a single strategy.

There are two basic ways we can consider building this ensemble of
parsers.  First, we could divide up the data in some fixed strategy,
building classifiers out of possibly-overlapping subsets.  If we did
this without any randomization this would look very much like the
cross-validation method of Wolpert \cite{wolpert:stacking}.  If we
performed this in a purely randomized way it would look a lot like
bagging, discussed above.  There is really no other option without
exploring some other intrinsic knowledge of the data.  In this section
we assume we have no such knowledge. 

The alternative is to sequentially build classifiers, one at a time,
adjusting the sub-corpus we use to produce the next classifier based
on the errors that are made by the ensemble that has been already
created.  This is the approach that we take.  We add a $k+1$-th
classifier to an ensemble of $k$ classifiers by noticing where those
$k$ classifiers make mistakes.  This is the general class of
algorithms of which AdaBoost is an example.

In this section we will investigate the application of the principles
of boosting and AdaBoost in particular to the job of creating parsers
with complementary errors.

\subsection{Background: Boosting}

The AdaBoost algorithm was presented by Freund and Schapire in 1996
\cite{freund96:adaboost_experiments, freund97:adaboost}.  Both authors
had performed prior theoretical work on boosting that lacked practical
appeal because it required knowledge that was not generally available
for popular learning algorithms \cite{schapire90,freund95:boostmaj}.
The algorithms relied on knowledge of the inductive bias of the
underlying learning algorithm, or required a known achievable accuracy
be specified.

The AdaBoost algorithm, on the other hand, requires only one thing of
its underlying learner.  It is allowed to abstain from making
predictions about some labels, but it must consistently be able to get
more than 50\% accuracy on the samples that it commits to a decision
on.  That accuracy is measured over the distribution describing the
importance of samples that it is given.  So, if each sample is
weighted by its importance, the weak learner must be able to get more
correct samples than incorrect samples \emph{by mass of importance} on
those that it labels.  This particular statement of the restriction
comes from Schapire and Singer's study \cite{ss:boost98:colt}.

\begin{algorithm}
{AdaBoost (Freund and Schapire, 1997)}
{
Given: Training set $\mathcal L = \{(y_i,x_i), i \in \{1\ldots m\}\}$
    where $y_i \in \{-1,1\}$ is the label for example $x_i$,
    classification    induction algorithm $\Psi:Y\times X \rightarrow
    \Phi$ with classification algorithm (weak learner) $\phi \in \Phi$
    and $\phi:X\rightarrow Y$.    Initial uniform distribution $D_1(i)
    = 1/m$.  Number of iterations, $T$. Counter $t=1$.
}
\label{algorithm:varying:adaboost}

\item Create $L_t$ by randomly choosing with replacement $m$ samples
 from $L$ using distribution $D_t$.
  \label{algorithm:varying:adaboost:loopstart}

\item $\phi_t \leftarrow \Psi(L_t)$

\item Choose $\alpha_t \in \Re$.

\item Adjust and normalize the distribution.  $Z_t$ is a normalization
  coefficient. 
\begin{displaymath}
  D_{t+1}(i) = \frac{D_t(i)\exp(-\alpha_t y_i \phi_t(x_i))}{Z_t}
\end{displaymath}

\item Increment $t$.  Quit if $t > T$.

\item Repeat from step \ref{algorithm:varying:adaboost:loopstart}.
\item The final hypothesis is 
\begin{displaymath}
  \phi_{boost}(x) = \mathrm{sign} \sum\limits_t  \alpha_t\phi_t(x)
\end{displaymath}
\end{algorithm}

Schapire and Singer extended AdaBoost to describe how to choose the
hypothesis mixing coefficients in certain circumstances, how to
incorporate a general notion of confidence scores, and also provided a
better formulation of theoretical performance \cite{ss:boost98:colt}.
In Algorithm \ref{algorithm:varying:adaboost} we show the version of
AdaBoost used in their work, as it is the most recent and mature
description.  We show a variant based on resampling, as that is what
we use in our work.

The value of $\alpha_t$ should generally be chosen to minimize
\begin{displaymath}
  \sum_i D_t(i)\exp(-\alpha_t y_i \phi_t(x_i))
\end{displaymath}
in order to minimize the expected per-sample training error of the
ensemble, which Schapire and Singer show can be concisely expressed
by $\prod\limits_t Z_t$.  Schapire and Singer give several examples
for how to pick an appropriate $\alpha$, and the moral is that it
depends on the possible outputs of the underlying weak learner.

A few studies have been done comparing bagging and boosting
\cite{maclin97:baggingvboosting, bauer99:bagboost,
  quinlan96:bagging_boosting}.  Their conclusions have generally been
similar:
\begin{itemize}
\item Bagging works in every environment.  It rarely produces ensembles
  worse than isolated classifiers.
\item When boosting works it typically has a much greater effect than
  bagging.
\item Boosting is extremely sensitive to noise.  Any noise or
  inconsistencies in the corpus get magnified and the later
  classifiers ``obsess'' over them, focusing the distribution's
  mass on them.
\item The serial nature of boosting makes it a much slower process
  during training than bagging because bagging can exploit
  the parallelism of modern ubiquitous computing.
\end{itemize}


Margineantu and Dietterich used AdaBoost to reduce the size of a
nearest neighbor classifier, and also provided a method for weeding
weak (or redundant) ensemble members from the ensemble
\cite{dietterich97:prune_boost}.  Two separate experimenters
investigated the use of AdaBoost using decision tree
induction\footnote{Decision trees are hierarchical rule-based
  classifiers, e.g. a taxonomy, and beyond the scope of this thesis.
  See \cite{cart,ind} for more information.}  as weak learners
\cite{drucker96:boosting_trees, quinlan96:bagging_boosting}.

Breiman's \emph{Arcing} (adaptive resampling for classification),
technique is a competitor of AdaBoost \cite{Breiman96:arcing}.  He
uses the same general algorithm, but an altered re-weighting formula.
Controlled empirical work comparing the two techniques finds incomplete
dominance, with a slight advantage to AdaBoost if there is any, and
AdaBoost's theoretical properties and reputation give a reason for us
to use modifications of it rather than Arcing in our experiments.

There are a few results suggesting that AdaBoost has weaknesses, or
at least that it is not as well understood as theories suggest.
Maclin's study of the resampling version of AdaBoost points out that
it suffers from using only one weight per classifier
\cite{maclin98:boostregion}.  The later classifiers that are generated
are given very little weight even though they perform exceptionally
well on the samples they focus on.

Grove and Schuurmans show that the concept of maximizing the minimum
margin does not explain the efficacy of boosting
\cite{grove98:boostlimit}.  The creators of AdaBoost had previously
provided this theory to explain its efficacy \cite{sfbl:marginboost}.
They find new coefficients for combining the classifiers created by
AdaBoost in an optimal way, using linear programming, such that the
minimum margin is maximized on the training data.  The result of their
experiment is a system whose training accuracy is superior to
AdaBoost's, but which surprisingly has worse generalization ability on
test data.  In this way they refute the argument that AdaBoost
performs well because it maximizes the minimum margin on the training
set.  They also dispel the rumor of AdaBoost's resistance to
over-fitting training data in this work.  In short, they empirically
refute two standing theoretical arguments for the efficacy of AdaBoost.

Boosting has been used in a few NLP systems, with positive results.
First, Haruno et al. \cite{boostingparsing98} used boosting to produce
more accurate classifiers which were embedded as a control mechanism
in a parser for Japanese.  They develop a dependency parser in which a
probabilistic classifier is used to give a probability of one bunsetsu
modifying another (a dependency link).  Then, as all Japanese
dependency links point to the left, they use an $O(n^2)$ dynamic
programming algorithm to produce a parse using dynamic programming.
Initially they used a decision tree (similar to Magerman
\cite{magerman95:parsing}) as the probabilistic classifier embedded in
this parser, but found they could get better results by boosting that
classifier using AdaBoost in its original form.

The creators of AdaBoost used it to perform
text classification \cite{boosttext99}.  Abney et al.
\cite{abney99:boosttagpp} performed part-of-speech tagging and
prepositional phrase attachment using AdaBoost as a core component.
They found they could achieve accuracies on both tasks that were
competitive with the state of the art.  There were two interesting
side effects of this study: they found that embedding the predictions
of boosted classifiers in a Viterbi-like \cite{viterbi}
dynamic-programming search algorithm severely degraded performance.
Also, they found that inspecting the samples that were consistently
given the most weight during boosting revealed some faulty annotations
in the corpus.  In all of these systems, AdaBoost has been used as a
traditional classification system.

\subsection{Empirical Boosting for Precision}

The first parse boosting algorithm we present is empirically
motivated.  Precision is a difficult measure to maximize for parsing
as pointed out by Goodman \cite{goodman98:phd}, so we present this ad
hoc algorithm.

\begin{algorithm}
{Boosting A Parser}
{Given corpus $corp$ with size $m=|corp|=\sum_{s,\tau}corp(s,t)$ and
  parser induction algorithm $g$.    Initial uniform distribution $D_1(i)
    = 1/m$.  Number of iterations, $T$. Counter $t=1$.}
\label{algorithm:varying:parseboosting}
\item Create $corp_t$ by randomly choosing with replacement $m$
  samples from $corp$ using distribution $D_t$.
  \label{algorithm:varying:boostparse:loopstart}
\item Create parser $f_t \leftarrow g(corp_t)$.
\item Choose $\alpha_t \in \Re$.
  \label{algorithm:varying:boostparse:choosealpha}
  
\item Adjust and normalize the distribution.  $Z_t$ is a normalization
  coefficient.  For all $i$, let parse tree $\tau^\prime_i \leftarrow
  f_t(s_i)$.  Let $\delta(\tau,c)$ be a function indicating that $c$
  is in parse tree $\tau$, and $|\tau|$ is the number of constituents
  in tree $\tau$.  $T(s)$ is the set of constituents that are
  found in the reference or hypothesized annotation for $s$.
\label{algorithm:varying:parseboosting:tricky}
\begin{displaymath}
  D_{t+1}(i) = 
  \frac{D_t(i)  \sum_{c\in T(s_i)}{\left(\alpha + (1-\alpha)|\delta(\tau^\prime_i,c)-\delta(\tau_i,c)|\right)} }
    {Z_t}
\end{displaymath}
\item Increment $t$.  Quit if $t > T$.

\item Repeat from step \ref{algorithm:varying:boostparse:loopstart}.

\item The final hypothesis is arrived at by combining the individual
  constituents.  Each parser $\phi_t$ in the ensemble gets vote
  $\alpha_t$ for the constituents they predict.  Any constituents that
  get strictly more than $\frac{1}{2}\sum_t\alpha_t$ weight is put
  into the final hypothesis.

\end{algorithm}

In step \ref{algorithm:varying:parseboosting:tricky} of Algorithm
\ref{algorithm:varying:parseboosting}, we are performing a simple
AdaBoost on the constituents in $\tau^\prime_i$ and giving the
distribution value for the sentence the sum of the distribution values
that would be realized for the constituents if they were independently
predictable. 

In step \ref{algorithm:varying:boostparse:choosealpha} of the
algorithm, we do not specify how to choose $\alpha_t$.  This is what
will vary for our experiments.  The rest of the structure can remain
the same for boosting, but the weight we give to various errors in
choosing $\alpha$ will specialize the algorithm.

In order to boost precision, we should reduce the weight on those
constituents that are predicted correctly, and leave the weight the
same on those constituents that are predicted by the parser but which
are not in the reference.  This is given in Equation
\ref{eqn:varying:alphaprecision}, and it follows the form set by
Schapire and Singer \cite{ss:boost98:colt} when working with weak
learners that can abstain.  In this case, when the parser does not
predict a constituent should be in the parse, we say it is abstaining.
The numerator is the mass of those constituents that were hypothesized
but not in the reference parse and the denominator is the mass of
those constituents that were predicted correctly.  We give a
step-by-step sample derivation of a tailored $\alpha$ for parsing in
Section \ref{section:varying:boostfmeasure}.

\begin{equation}
\label{eqn:varying:alphaprecision}
\alpha_p = \frac{
  \sum_i \frac{D(i)}{|T(s_i)|}
  \sum_{c\in T(s_i)}(1-\delta(\tau_i,c))\delta(\tau^\prime_i,c)
}
{
  \sum_i \frac{D(i)}{|T(s_i)|}
  \sum_{c\in T(s_i)}\delta(\tau_i,c)\delta(\tau^\prime_i,c)
}
\end{equation}

\begin{sidewaysfigure}[htbp]
\centering
\fbox{
\begin{tabular}{rl}
    \epsfig{file=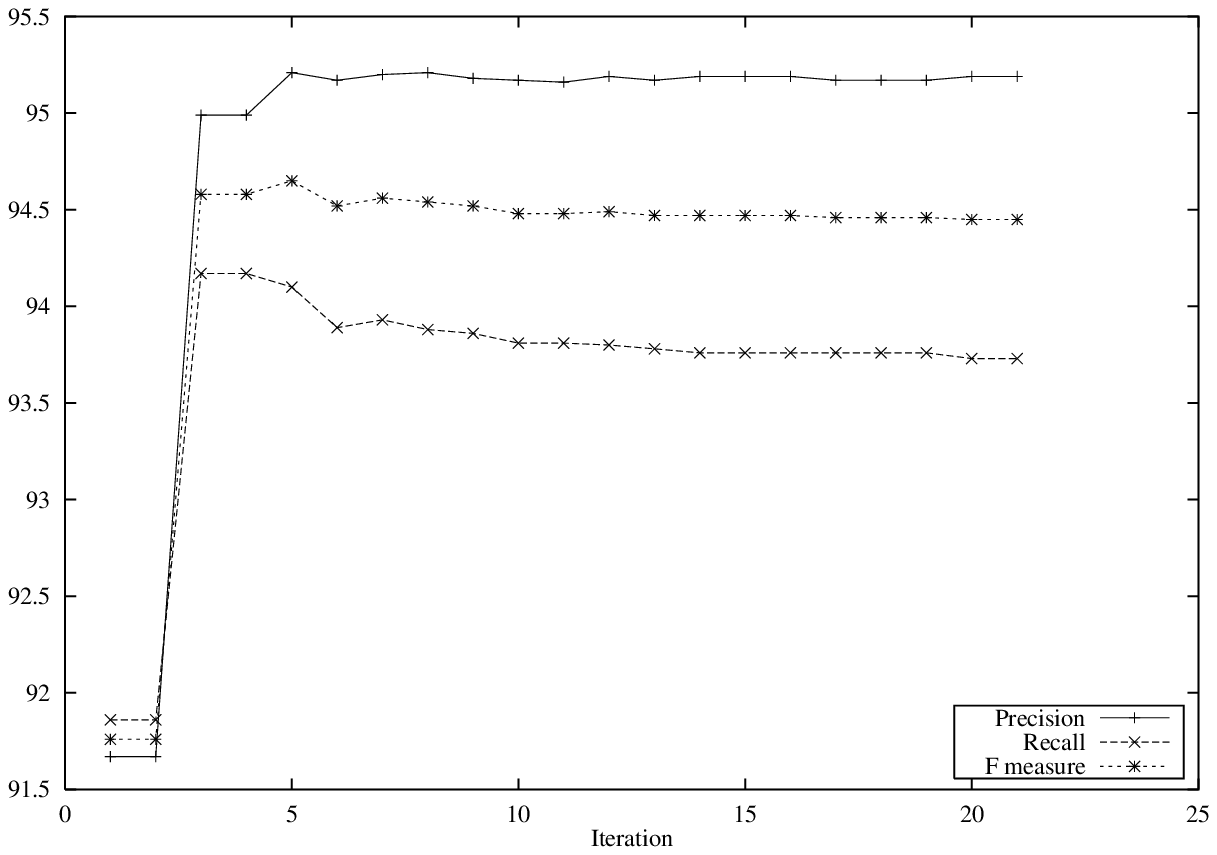, width= 0.45\textwidth}
&
    \epsfig{file=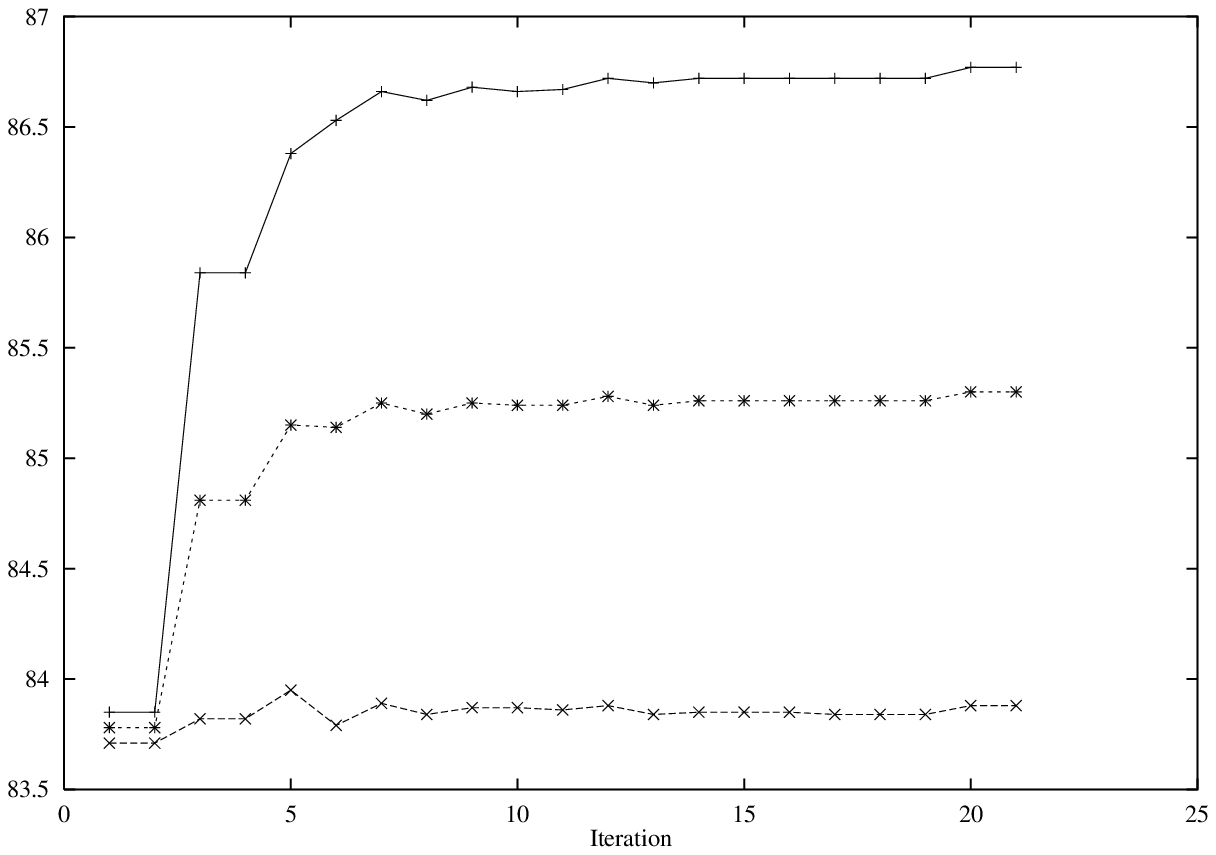, width= 0.45\textwidth}
\\
    \epsfig{file=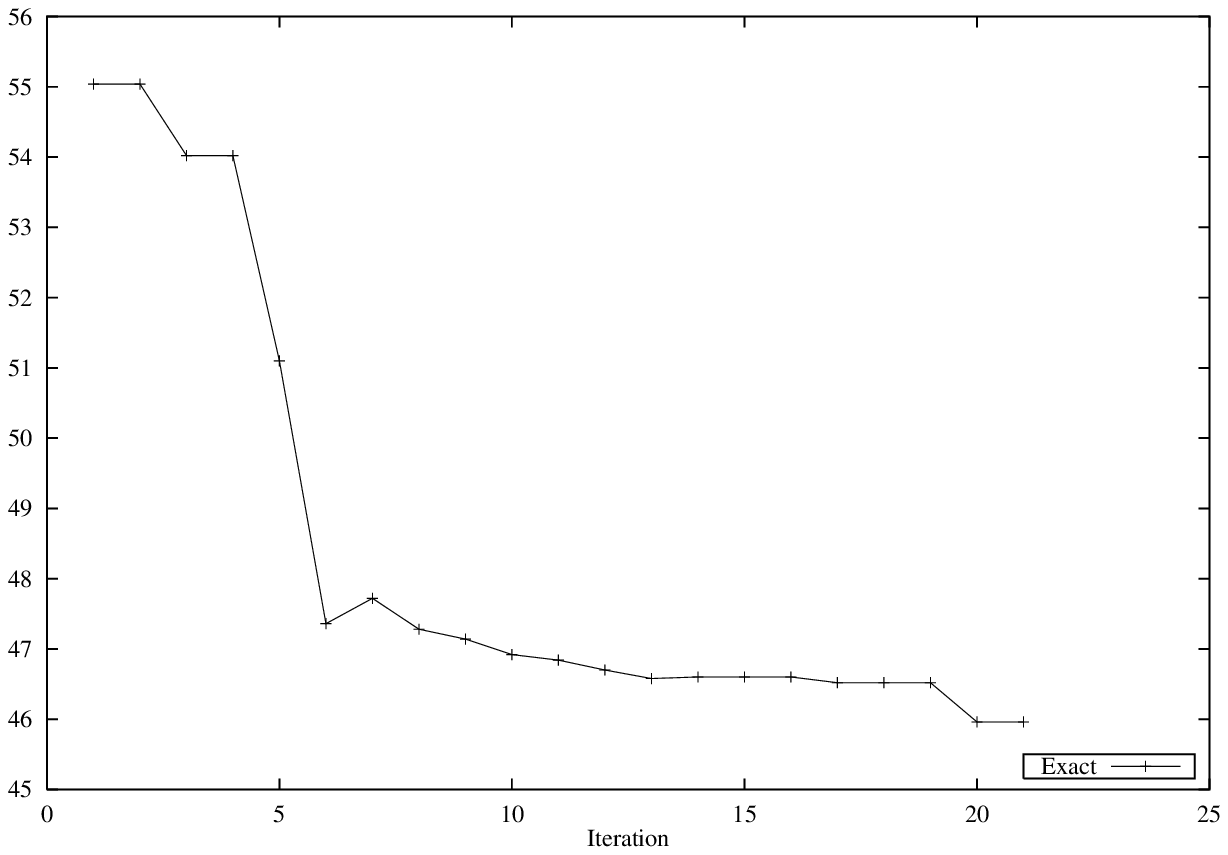, width= 0.45\textwidth}
&
    \epsfig{file=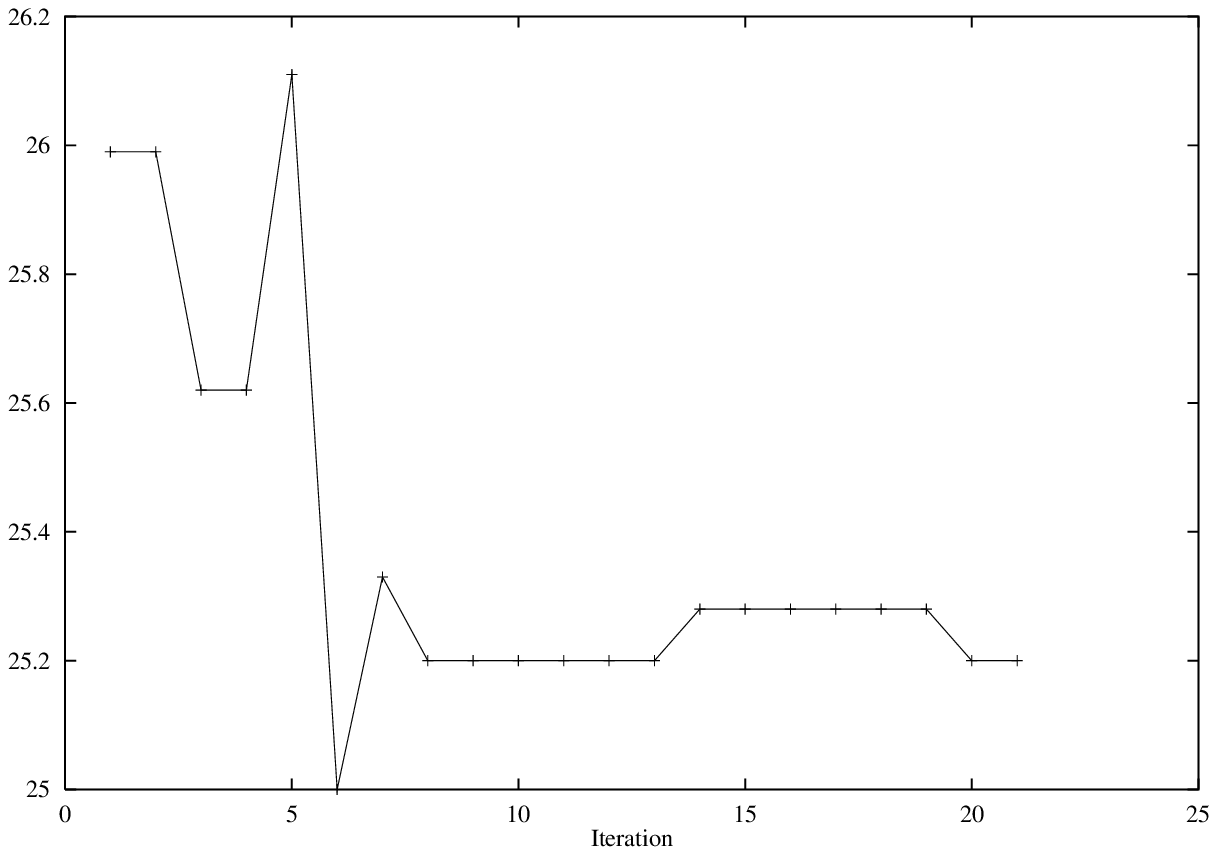, width= 0.45\textwidth}
\end{tabular}
}


  \caption{Boosting Precision} 
  \label{fig:varying:boost.small.prec}
\end{sidewaysfigure}

\begin{table}[htbp]
\centering
\begin{tabular}{|c|l|rr|rr|rr|}
       \hline
      Set
      &\multicolumn{1}{c|}{Instance} 
      &\multicolumn{1}{c}{P} 
      &\multicolumn{1}{c|}{R} 
      &\multicolumn{1}{c}{F}  
      &\multicolumn{1}{c|}{Gain}  
      &\multicolumn{1}{c}{Exact}
      &\multicolumn{1}{c|}{Gain} \\
      \hline 
Training&Original Parser  & 97.07 &  97.30 &  97.18 &NA&    73.3 & NA 
\\
&       Initial & 91.67 & 91.86 & 91.76 &  0.00 & 55.0  &  0.0
\\
&       BestF(5)        & 95.21 & 94.10 & 94.65 &  2.89 & 51.1  & -3.9
\\
&       Final(21)       & 95.19 & 93.73 & 94.45 &  2.69 & 46.0  & -9.1
\\
      \hline 
Test & Original Parser  & 86.03 & 85.43 & 85.73  & NA & 28.6 & NA
\\
&   Initial & 83.85 & 83.71 & 83.78 &  0.00 & 26.0  &  0.0
\\
&       TrainBestF(5)   & 86.38 & 83.95 & 85.15 &  1.37 & 26.1  &  0.1
\\
&       TestBestF(20)   & 86.77 & 83.88 & 85.30 &  1.52 & 25.2  & -0.8
\\
&       Final(21)       & 86.77 & 83.88 & 85.30 &  1.52 & 25.2  & -0.8
\\
\hline
\end{tabular}


  \caption{Boosting Precision} 
  \label{table:varying:boost.small.prec}
\end{table}

In Figure \ref{fig:varying:boost.small.prec} and Table
\ref{table:varying:boost.small.prec} we see the results of using this
algorithm to boost a parser based on a training set of 5000
sentences.  In the figure we see that on the test set the algorithm
achieves significant increases in precision.  However, both recall and
exact sentence accuracy is reduced as a tradeoff.

\subsection{Boosting for Recall}

Boosting the recall of a parser is more theoretically plausible.  It
seems just like a classification problem.  There is a fixed set of
constituents in the reference and the goal is to get as many of them
correct as possible.

\begin{equation}
\label{eqn:varying:alpharecall}
\alpha_r = \frac{
  \sum_i \frac{D(i)}{|T(s_i)|}
  \sum_{c\in T(s_i)}\delta(\tau_i,c)(1-\delta(\tau^\prime_i,c))
}
{
  \sum_i \frac{D(i)}{|T(s_i)|}
  \sum_{c\in T(s_i)}\delta(\tau_i,c)\delta(\tau^\prime_i,c)
}
\end{equation}

In Equation \ref{eqn:varying:alpharecall} we show how to calculate
$\alpha_r$, the weighing parameter to be used in boosting recall.  The
numerator here is the mass on constituents that are found in the
reference transcription but not the hypothesis, and the denominator is
the same as we used for $\alpha_p$.

\begin{sidewaysfigure}[htbp]
\centering
\fbox{
\begin{tabular}{rl}
    \epsfig{file=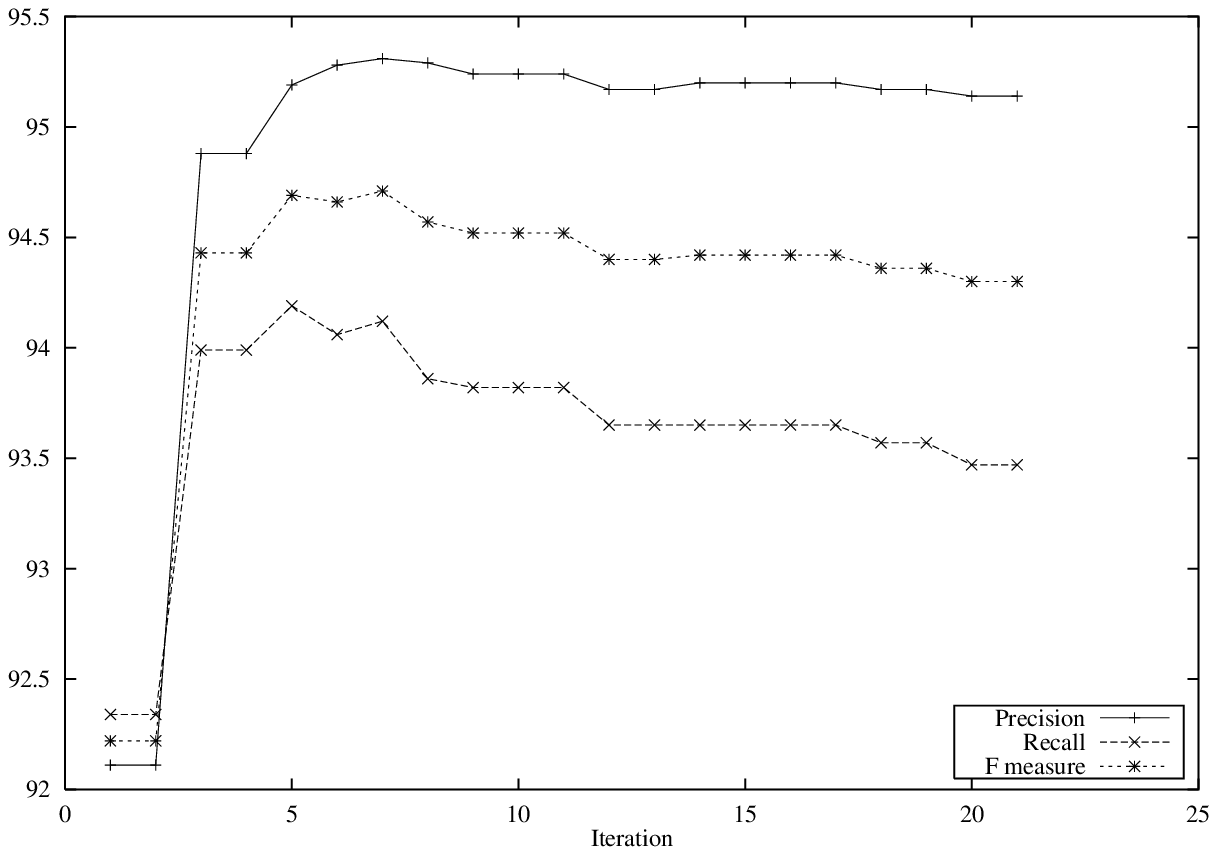, width= 0.45\textwidth}
&
    \epsfig{file=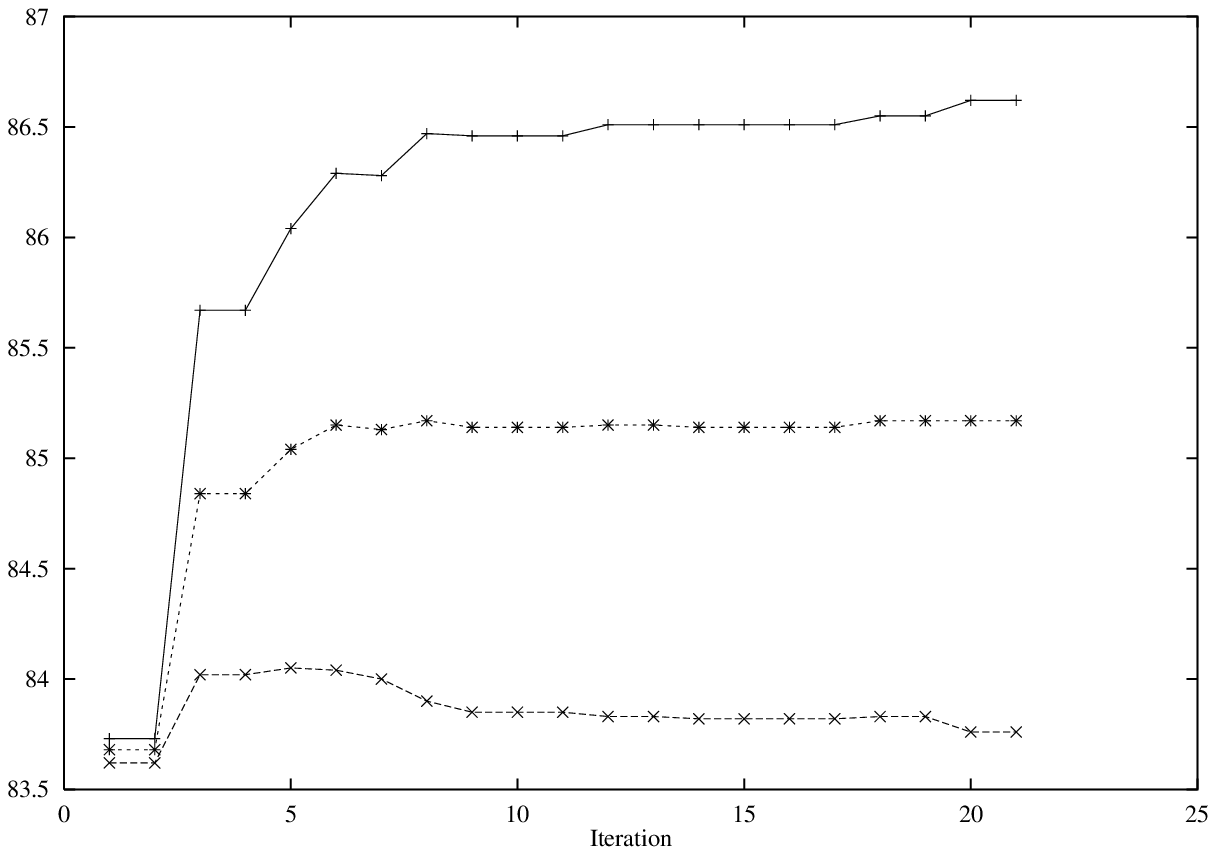, width= 0.45\textwidth}
\\
    \epsfig{file=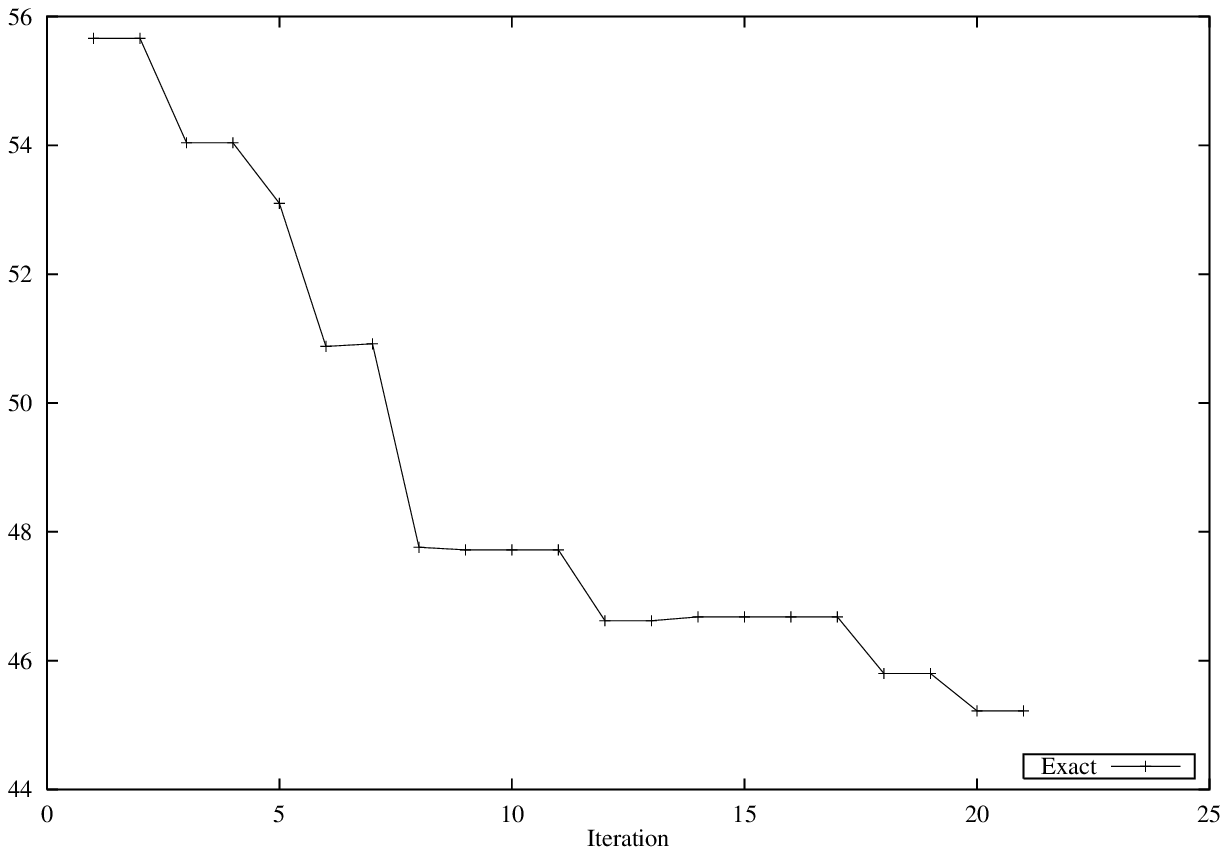, width= 0.45\textwidth}
&
    \epsfig{file=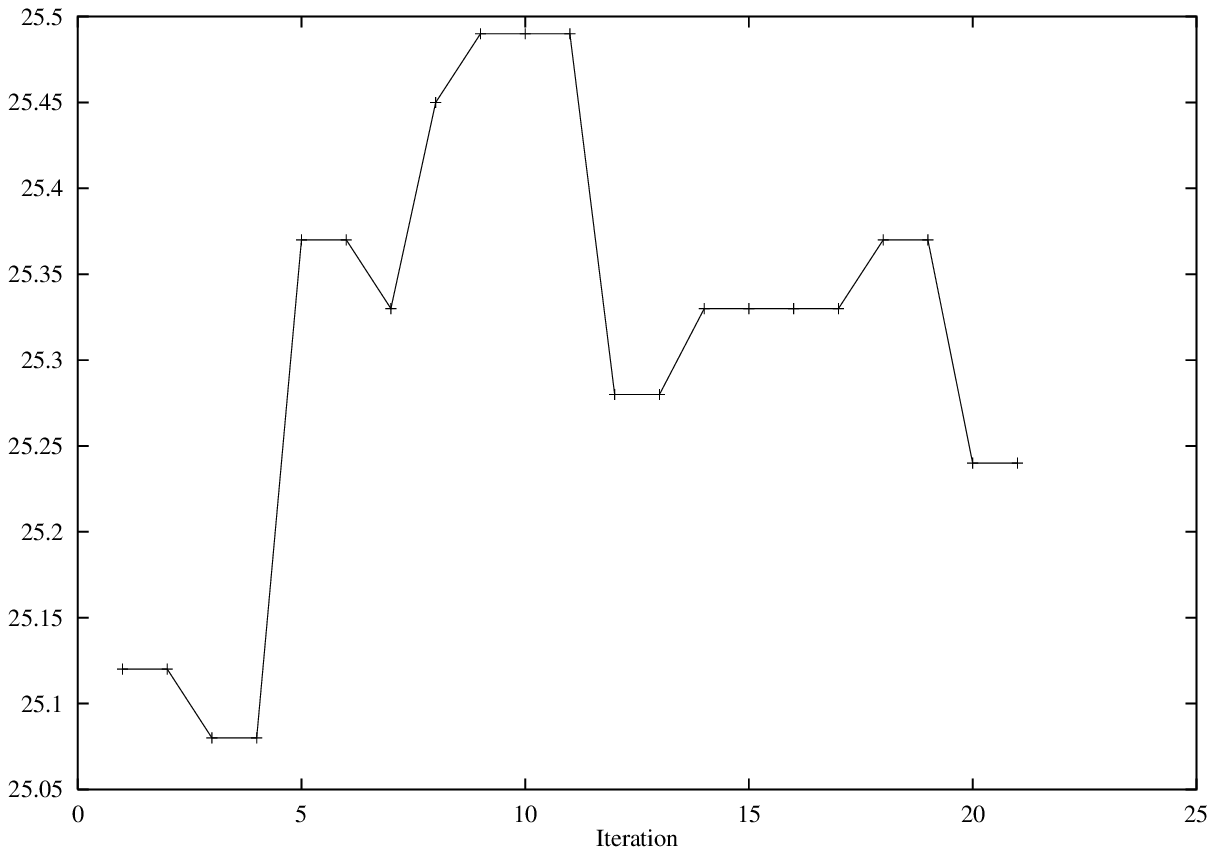, width= 0.45\textwidth}
\end{tabular}
}


  \caption{Boosting Recall}
  \label{fig:varying:boost.small.recall}
\end{sidewaysfigure}

\begin{table}[htbp]
\centering
\begin{tabular}{|c|l|rr|rr|rr|}
       \hline
      Set
      &\multicolumn{1}{c|}{Instance} 
      &\multicolumn{1}{c}{P} 
      &\multicolumn{1}{c|}{R} 
      &\multicolumn{1}{c}{F}  
      &\multicolumn{1}{c|}{Gain}  
      &\multicolumn{1}{c}{Exact}
      &\multicolumn{1}{c|}{Gain} \\
      \hline 
Training&Original Parser  & 97.07 &  97.30 &  97.18 &NA&    73.3 & NA 
\\
&       Initial & 92.11 & 92.34 & 92.22 &  0.00 & 55.7  &  0.0
\\
&       BestF(7)        & 95.31 & 94.12 & 94.71 &  2.49 & 50.9  & -4.7
\\
&       Final(21)       & 95.14 & 93.47 & 94.30 &  2.08 & 45.2  & -10.4
\\
      \hline 
Test & Original Parser  & 86.03 & 85.43 & 85.73  & NA & 28.6 & NA
\\
&   Initial & 83.73 & 83.62 & 83.68 &  0.00 & 25.1  &  0.0
\\
&       TrainBestF(7)   & 86.28 & 84.00 & 85.13 &  1.45 & 25.3  &  0.2
\\
&       TestBestF(8)    & 86.47 & 83.90 & 85.17 &  1.49 & 25.4  &  0.3
\\
&       Final(21)       & 86.62 & 83.76 & 85.17 &  1.49 & 25.2  &  0.1
\\
\hline
\end{tabular}


  \caption{Boosting Recall} 
  \label{table:varying:boost.small.recall}
\end{table}

In Figure \ref{fig:varying:boost.small.recall} and Table
\ref{table:varying:boost.small.recall} we see the result.  It didn't
work.  We get almost identical results to those we got during
precision boosting.  There are two possible explanations for this.
The constraints on the independence of classifications dictated by
parsing tend to make a parser predict few possible constituents
because they must be properly nested with no crossing bracketings.
Since the parser is not allowed to over-generate constituents and
create a structure with crossing brackets, it cannot create parsers
that err on the side of excessive recall.  The second possibility is
similar.  Perhaps the parser induction algorithm will not allow
parsers to be made that produce excessive predicted constituents in
the training set.  The real answer is probably a mixture of these
two possibilities.

\subsection{Empirical Boosting for F-measure}
\label{section:varying:boostfmeasure}

In Chapter \ref{chapter:combining} we motivated the use of F-measure
for evaluating parsers.  It is a measure of accuracy as well as
balance because it lies near the lower value of precision or recall.
As a good overall measure of accuracy it presents a valuable target
measure for minimization during boosting.  Let us develop the
equations once so the technique is illustrated (and can be validated
on the other $\alpha$ computations).

First, let $a$ be the number of constituents that are hypothesized by
the parser and are in the reference.  Likewise $b$ is the number of
constituents hypothesized by the parser but not found in the reference
and $c$ is the number of constituents in the reference that were not
hypothesized by the parser.  By definition and a little algebra, we
see that $F = 2a/(2a+b+c)$.  The little-known E-measure is $(1-F)$, and
hence $E = (b+c)/(2a+b+c)$.  E-measure is what we want to minimize.  

Freund and Schapire suggest that $\alpha=\epsilon/(1-\epsilon)$ is a
useful way to compute $\alpha$, where $\epsilon$ is the error rate of
the classifier.  Substituting E-measure for $\epsilon$, we get the ad
hoc F-measure boosting algorithm.  Hence we want $\alpha=(b+c)/2a$.

There are more details.  First, we want the mass on constituents instead of
their count for $a$, $b$, and $c$.  Also, we only have distribution
values available on a sentence-by-sentence basis, so the mass of the
distribution  will have to be proportioned to the constituents within
a sentence.  Observe that we really want 
\begin{displaymath}
a = \sum_i \frac{D(i)}{|T(s_i)|}
  \sum_{c\in T(s_i)}\delta(\tau_i,c)\delta(\tau^\prime_i,c)
\end{displaymath}
instead of a simple count for $a$.  The multiplied $\delta$s ensure
that the constituent is in both the hypothesis and the reference.  The
distribution value is divided among the $|T(s_i)|$ potential
constituents in the sentence.  The rest follows, below.

\begin{eqnarray}
\alpha_F &= &
\frac{
  \sum_i \frac{D(i)}{|T(s_i)|}
  \sum_{c\in T(s_i)}
    \delta(\tau_i,c)(1-\delta(\tau^\prime_i,c))
    +
    (1-\delta(\tau_i,c))\delta(\tau^\prime_i,c)
}
{
  \sum_i \frac{D(i)}{|T(s_i)|}
  \sum_{c\in T(s_i)}\delta(\tau_i,c)\delta(\tau^\prime_i,c)
}
\nonumber
\\
&=&
\frac{
  \sum_i \frac{D(i)}{|T(s_i)|}
  \sum_{c\in T(s_i)}
    \delta(\tau_i,c)
    +\delta(\tau^\prime_i,c)
    -2\delta(\tau_i,c)\delta(\tau^\prime_i,c)
}
{
  2 \sum_i \frac{D(i)}{|T(s_i)|}
  \sum_{c\in T(s_i)}\delta(\tau_i,c)\delta(\tau^\prime_i,c)
}
\label{eqn:varying:alphaf}
\end{eqnarray}

This is an ad hoc algorithm because there is nothing that formally
justifies that the boosting proofs work when one substitutes $E$ for
$\epsilon$, but the techniques and principles of AdaBoost are present.

\begin{sidewaysfigure}[htbp]
\centering
\fbox{
\begin{tabular}{rl}
    \epsfig{file=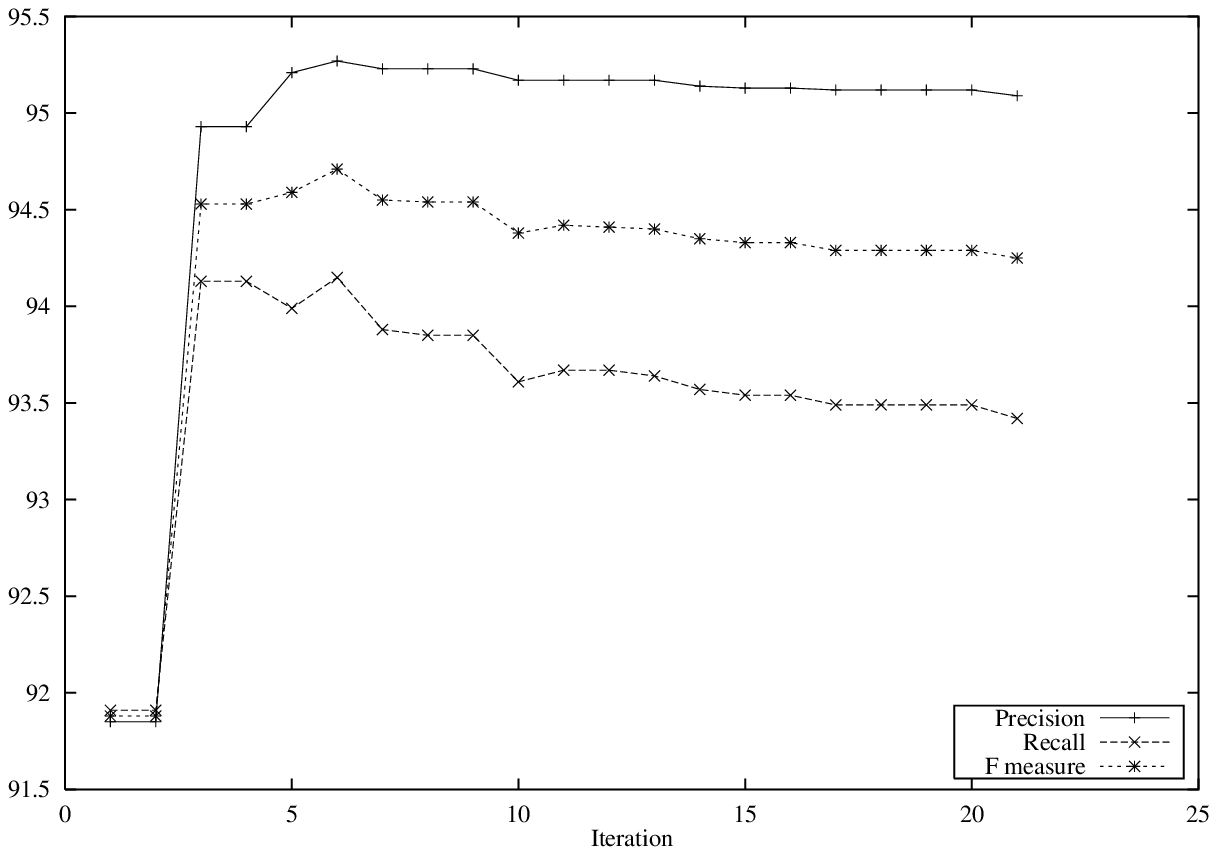, width= 0.45\textwidth}
&
    \epsfig{file=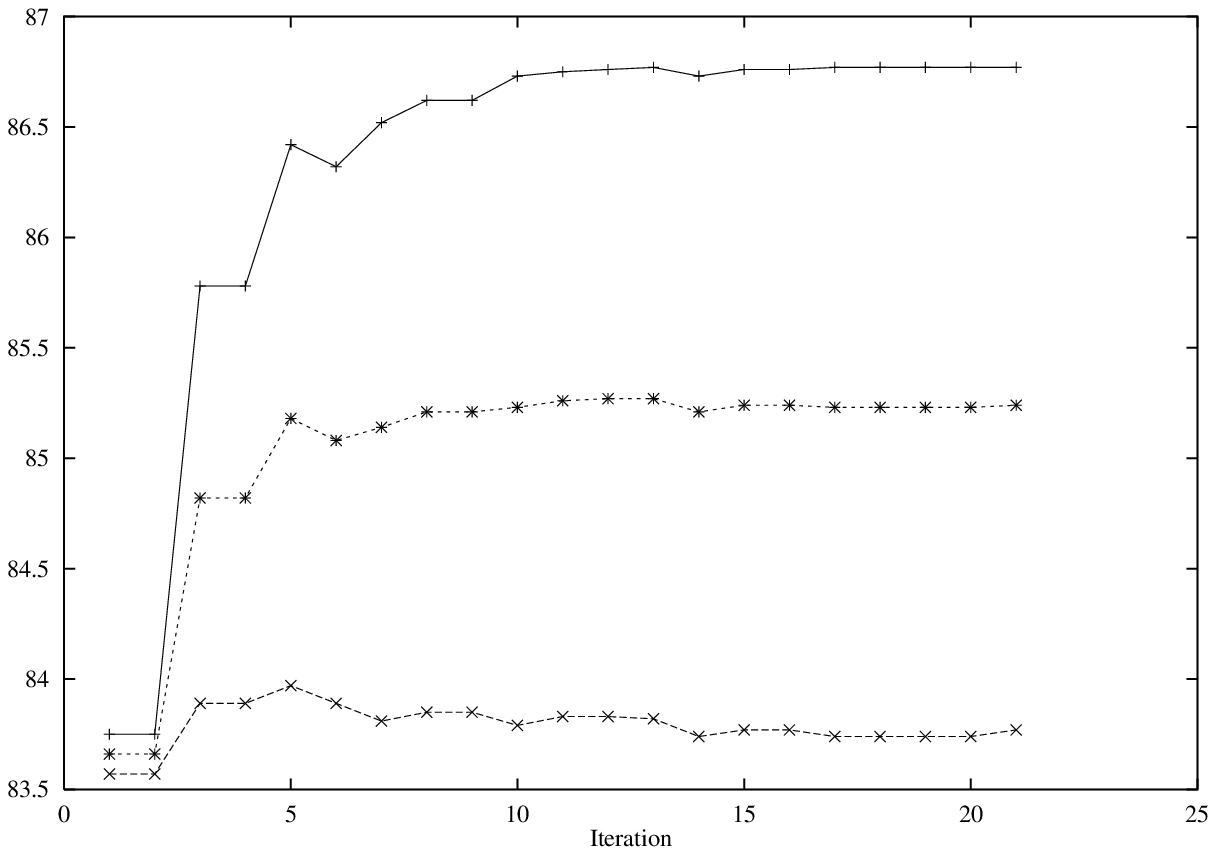, width= 0.45\textwidth}
\\
    \epsfig{file=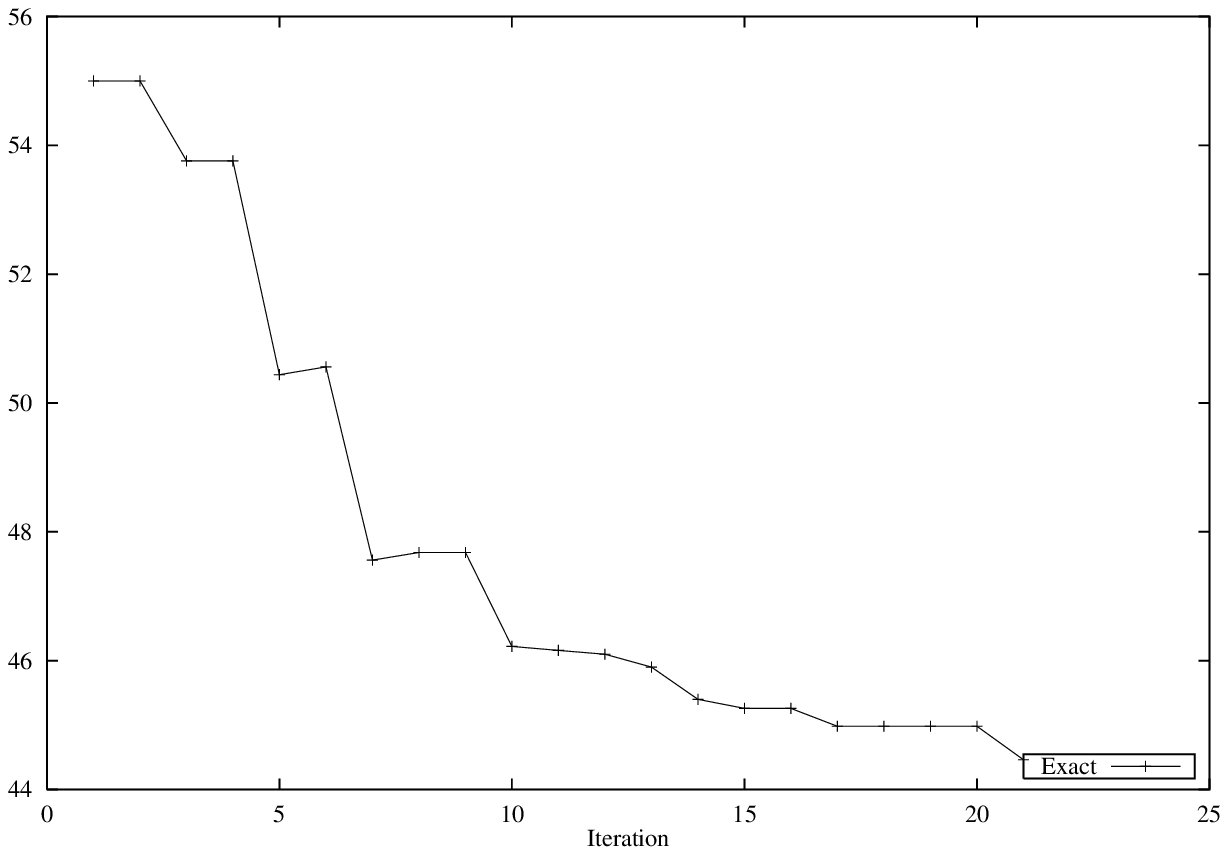, width= 0.45\textwidth}
&
    \epsfig{file=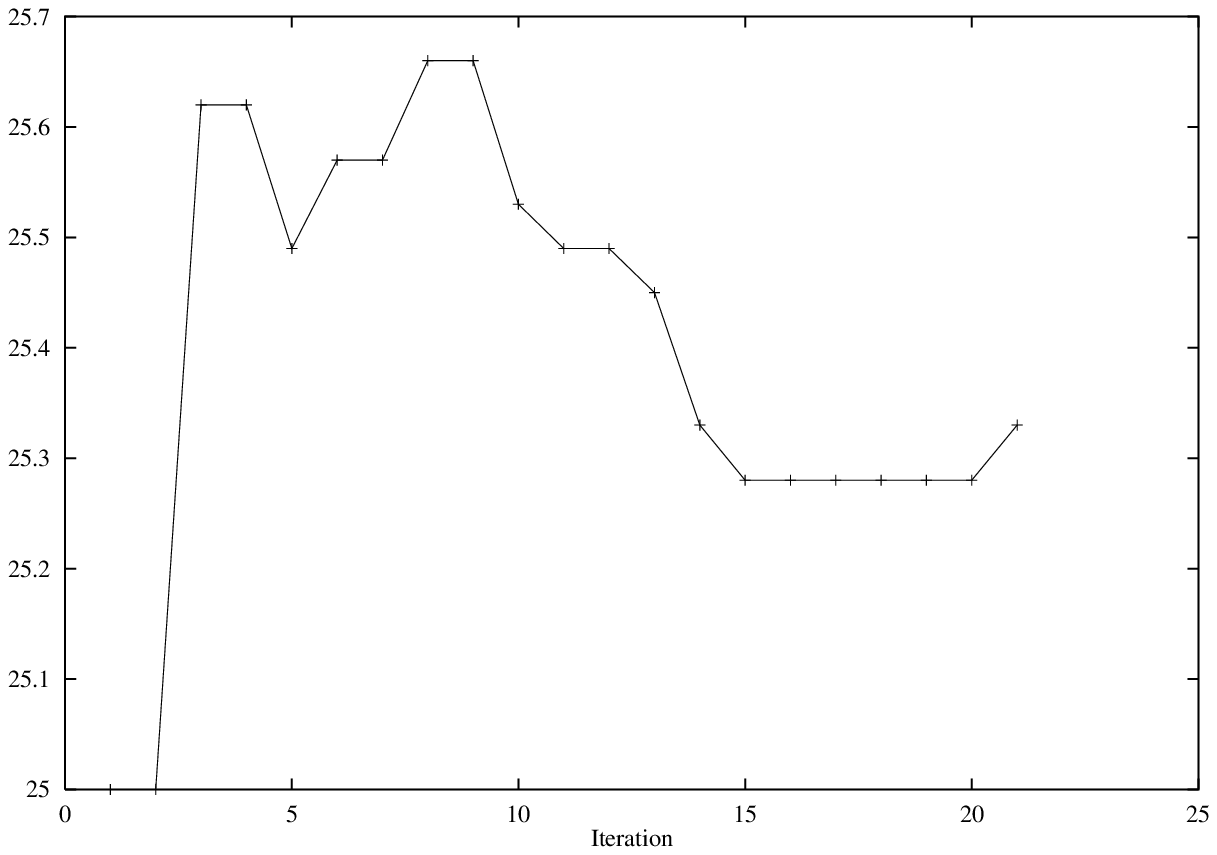, width= 0.45\textwidth}
\end{tabular}
}


  \caption{Boosting F-measure} 
  \label{fig:varying:boost.small.fmeas}
\end{sidewaysfigure}

\begin{table}[htbp]
\centering
\begin{tabular}{|c|l|rr|rr|rr|}
       \hline
      Set
      &\multicolumn{1}{c|}{Instance} 
      &\multicolumn{1}{c}{P} 
      &\multicolumn{1}{c|}{R} 
      &\multicolumn{1}{c}{F}  
      &\multicolumn{1}{c|}{Gain}  
      &\multicolumn{1}{c}{Exact}
      &\multicolumn{1}{c|}{Gain} \\
      \hline 
Training&Original Parser  & 97.07 &  97.30 &  97.18 &NA&    73.3 & NA 
\\
&       Initial & 91.85 & 91.91 & 91.88 &  0.00 & 55.0  &  0.0
\\
&       BestF(6)        & 95.27 & 94.15 & 94.71 &  2.83 & 50.6  & -4.4
\\
&       Final(21)       & 95.09 & 93.42 & 94.25 &  2.37 & 44.5  & -10.5
\\
      \hline 
Test & Original Parser  & 86.03 & 85.43 & 85.73  & NA & 28.6 & NA
\\
&   Initial & 83.75 & 83.57 & 83.66 &  0.00 & 25.0  &  0.0
\\
&       TrainBestF(6)   & 86.32 & 83.89 & 85.08 &  1.42 & 25.6  &  0.6
\\
&       TestBestF(12)   & 86.76 & 83.83 & 85.27 &  1.61 & 25.5  &  0.5
\\
&       Final(21)       & 86.77 & 83.77 & 85.24 &  1.58 & 25.3  &  0.3
\\
\hline
\end{tabular}


  \caption{Boosting F-measure} 
  \label{table:varying:boost.small.fmeas}
\end{table}

In Figure \ref{fig:varying:boost.small.fmeas} and Table
\ref{table:varying:boost.small.fmeas} we see the result of boosting
using this value.  In the first iterations, boosting F-measure is
successful.  Why recall on the training set deteriorates is unclear,
though.  Also, exact sentence accuracy deteriorates quickly, without a
compensating gain in the constituent accuracy metrics.  On both the
training and test set we see a large gain in precision, and the
asymptotic effect shown in the test set curves is comforting that
boosting F-measure is not doing anything systematically incorrect.

\subsection{Boosting for Constituent Accuracy}

Throughout the boosting discussion we have assumed an underlying model
of constituent accuracy.  A potential constituent can be considered
correct if it is predicted in the hypothesis and it exists in the
reference, or it is not predicted and it is not in the reference.
Earlier we made the assertion that potential constituents that do not
appear in the hypothesis or the reference should not make a big
contribution to the accuracy computation.  There are many such
potential constituents, and if we were maximizing a function that
treated getting them incorrect the same as getting a constituent that
appears in the reference correct, we would most likely decide not to
predict any constituents.

Our model of constituent accuracy, then, is simple.  Each prediction
correctly made over $T(s)$ will be given equal weight.  That is,
correctly hypothesizing a constituent in the reference will give us
one point, but a precision or recall error will cause us to miss one
point.  Constituent accuracy is then $a/(a+b+c)$, where $a$ is the
number of constituents correctly hypothesized, $b$ is the number of
precision errors and $c$ is the number of recall errors.

Equation \ref{eqn:varying:alphabca} shows how to compute $\alpha_{ca}$
for the measure we have described.  It is interesting to note that
(comparing to Equation \ref{eqn:varying:alphaf}) $\alpha_F =
2\alpha_{ca}$, even though the motivation used to arrive at the
different formulae was completely different.  The constant factor of 2
makes a difference in the performance of the algorithms, as the
experiment shows.

\begin{eqnarray}
\alpha_{ca} &= &
\frac{
  \sum_i \frac{D(i)}{|T(s_i)|}
  \sum_{c\in T(s_i)}
    \delta(\tau_i,c)
    +\delta(\tau^\prime_i,c)
    -2\delta(\tau_i,c)\delta(\tau^\prime_i,c)
}
{
  \sum_i \frac{D(i)}{|T(s_i)|}
  \sum_{c\in T(s_i)}\delta(\tau_i,c)\delta(\tau^\prime_i,c)
}
\label{eqn:varying:alphabca}
\end{eqnarray}

\begin{sidewaysfigure}[htbp]
  \centering
\fbox{
\begin{tabular}{rl}
    \epsfig{file=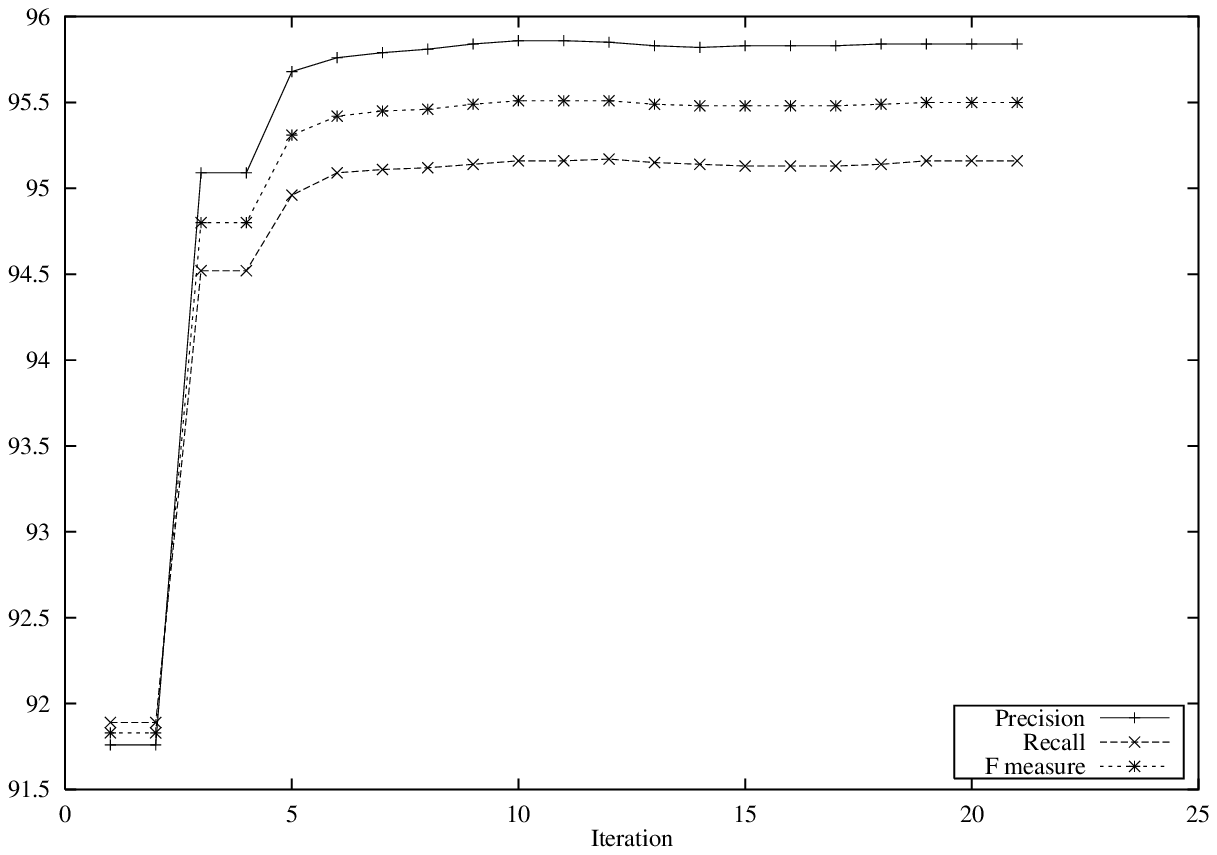, width= 0.45\textwidth}
&
    \epsfig{file=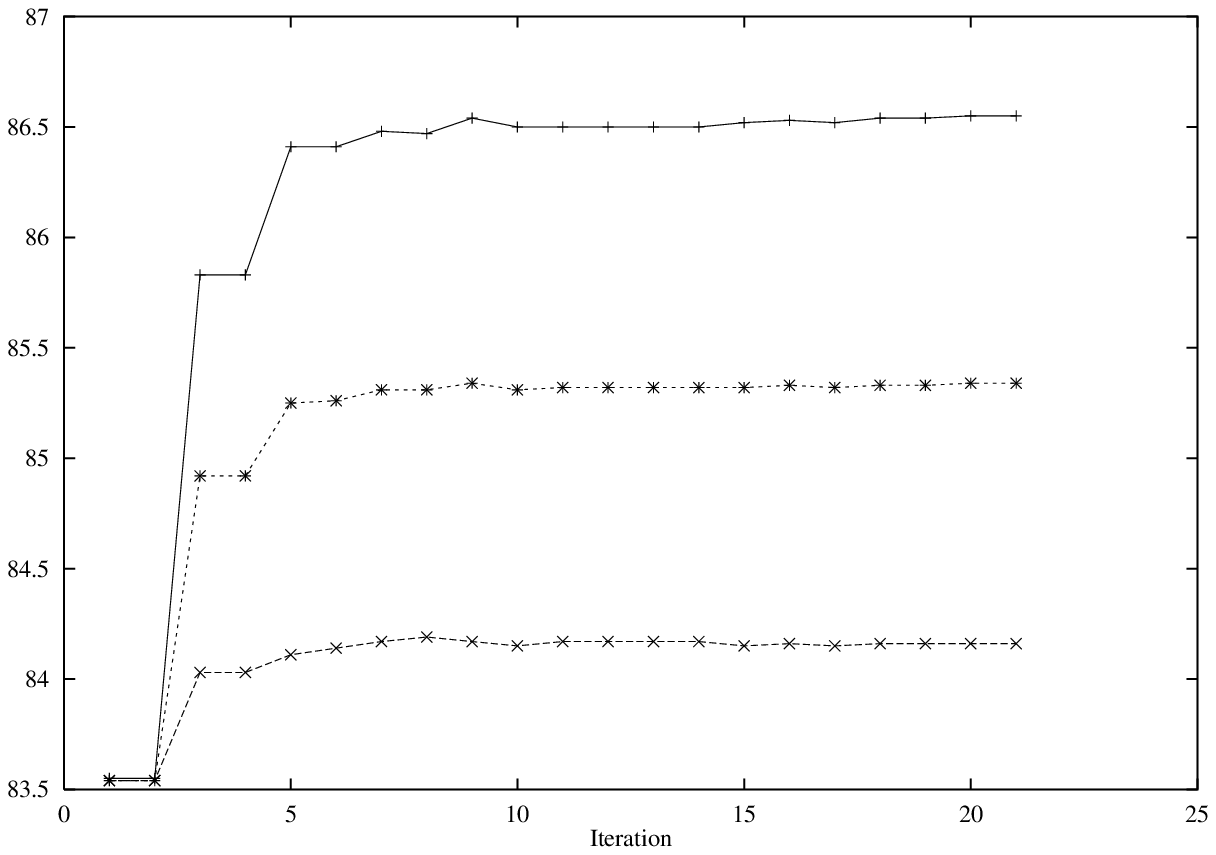, width= 0.45\textwidth}
\\
    \epsfig{file=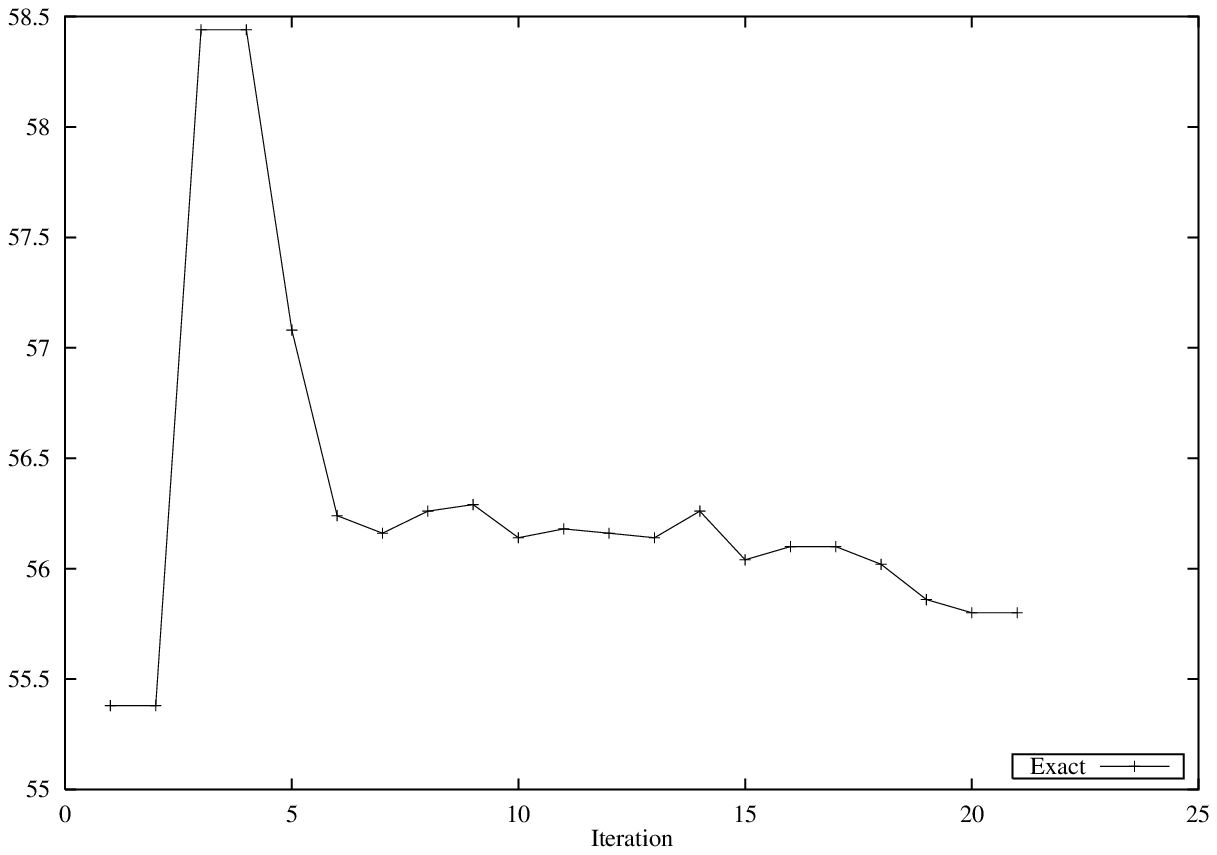, width= 0.45\textwidth}
&
    \epsfig{file=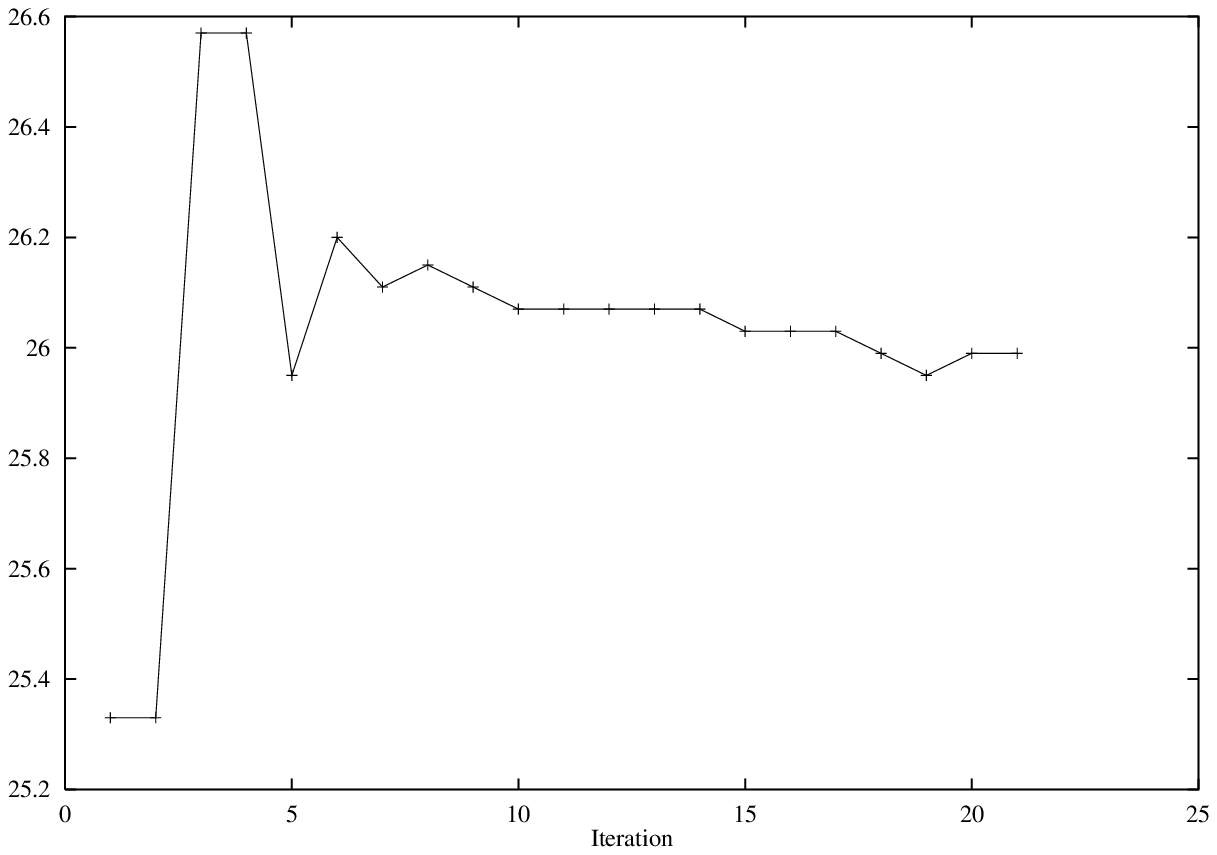, width= 0.45\textwidth}
\end{tabular}
}


  \caption{Boosting Constituent Accuracy}
  \label{fig:varying:boost.small.bca}
\end{sidewaysfigure}

\begin{table}[htbp]
  \centering
\begin{tabular}{|c|l|rr|rr|rr|}
       \hline
      Set
      &\multicolumn{1}{c|}{Instance} 
      &\multicolumn{1}{c}{P} 
      &\multicolumn{1}{c|}{R} 
      &\multicolumn{1}{c}{F}  
      &\multicolumn{1}{c|}{Gain}  
      &\multicolumn{1}{c}{Exact}
      &\multicolumn{1}{c|}{Gain} \\
      \hline 
Training&Original Parser  & 97.07 &  97.30 &  97.18 &NA&    73.3 & NA 
\\
&       Initial & 91.76 & 91.89 & 91.83 &  0.00 & 55.4  &  0.0
\\
&       BestF(10)       & 95.86 & 95.16 & 95.51 &  3.68 & 56.1  &  0.8
\\
&       Final(21)       & 95.84 & 95.16 & 95.50 &  3.67 & 55.8  &  0.4
\\
      \hline 
Test & Original Parser  & 86.03 & 85.43 & 85.73  & NA & 28.6 & NA
\\
&   Initial & 83.55 & 83.54 & 83.54 &  0.00 & 25.3  &  0.0
\\
&       TrainBestF(10)  & 86.50 & 84.15 & 85.31 &  1.77 & 26.1  &  0.7
\\
&       TestBestF(9)    & 86.54 & 84.17 & 85.34 &  1.80 & 26.1  &  0.8
\\
&       Final(21)       & 86.55 & 84.16 & 85.34 &  1.80 & 26.0  &  0.7
\\
\hline
\end{tabular}


  \caption{Boosting Constituent Accuracy}
  \label{table:varying:boost.small.bca}
\end{table}

We see from Figure \ref{fig:varying:boost.small.bca} and Table
\ref{table:varying:boost.small.bca} that boosting constituent accuracy
is the most successful of our boosting attempts.  This is likely the
result of a reasonable decomposition of the problem into a binary
classification.  We are not over-weighing correct constituents, nor
over-weighing our errors.  In the exact sentence accuracy graphs we
once again see that boosting trades off exact sentence accuracy for
small gains in precision and recall.  Furthermore, there is very
little movement in the model after the twelfth iteration.  That is
because like the other boosted versions, the confidence or voting
weights given to the parsers produced in the later iterations are
naturally small.  This problem is discussed below.

As this is our best boosting version, when unspecified data is
analyzed in the following sections, it comes from the result of this
system.

\subsection{Violating The Weak Learning Criterion}

As mentioned earlier in this chapter, AdaBoost has one requirement of
the induction algorithm.  It must focus on the mass.  In the case of
boosting for constituent accuracy, we can detect when the parser
induction algorithm fails to be a weak learner in the same way that
Freund and Schapire detect it for binary classification
\cite{freund97:adaboost}.  When the training error under the distribution
exceeds 0.5 we can say with certainty that the classifier induction
algorithm did not have the weak learning property.  Similarly, when
boosting for constituent accuracy gets fewer constituent inclusion
decisions correct than 0.5 of the mass we can say it has failed.

We detected that the learner exhibited this property after only 10-12
iterations.  Breiman proposed a solution or work-around for this,
however \cite{Breiman96:arcing}. He suggests to dispose of the faulty
classifier and restart with the original distribution when the learner
gets stuck in this situation.

\subsubsection{Backing Off to Bagging}

Since we are performing resampling on our corpus restarting with the
original distribution is the same as creating a new bootstrap
replicate during boosting.  We say that bagging is our back-off
strategy in this case.  In counting iterations, we will include the
iteration that fails to exhibit the weak learning technique, but we
will not include that parser in the ensemble.  We are disposing of the
parser, but including it when we count iterations because we have paid
the computational price of developing a parser in that iteration.

\begin{sidewaysfigure}[htbp]
\centering
\fbox{
\begin{tabular}{rl}
    \epsfig{file=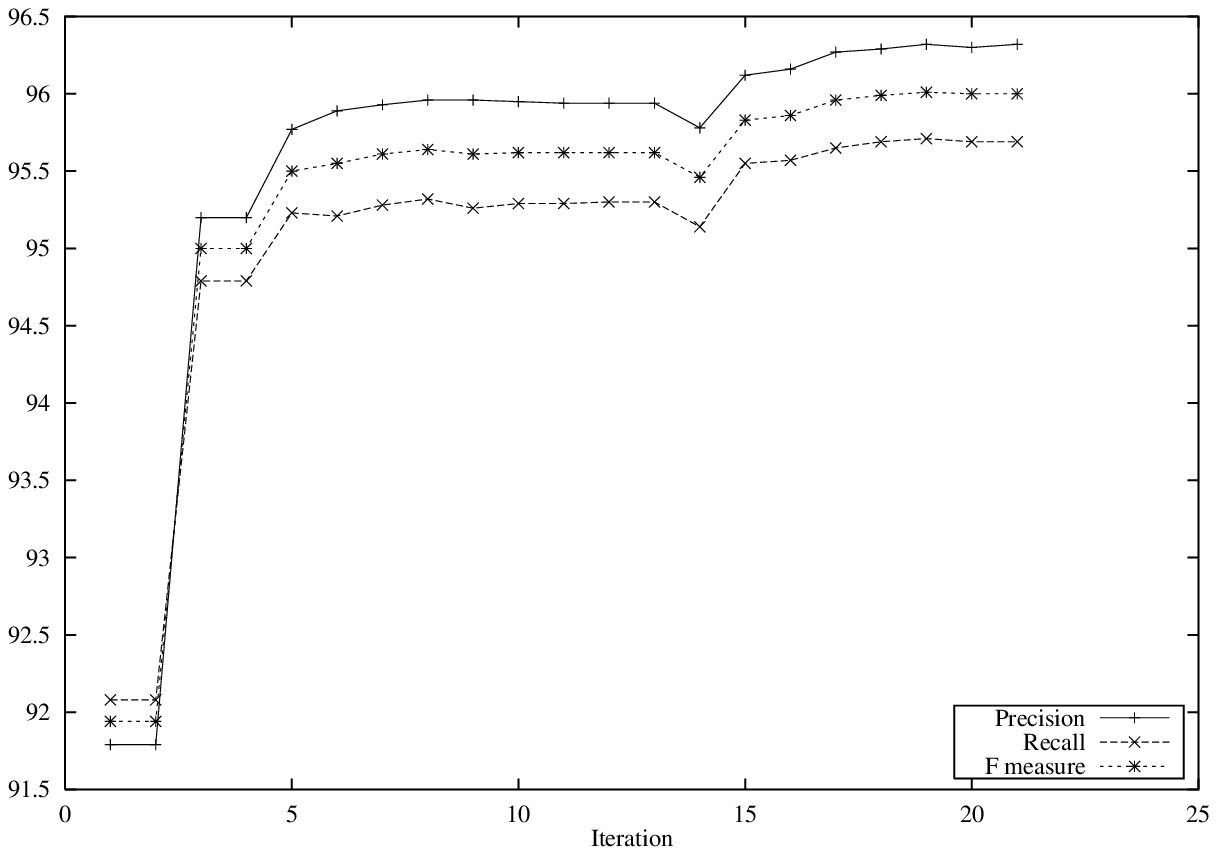, width= 0.45\textwidth}
&
    \epsfig{file=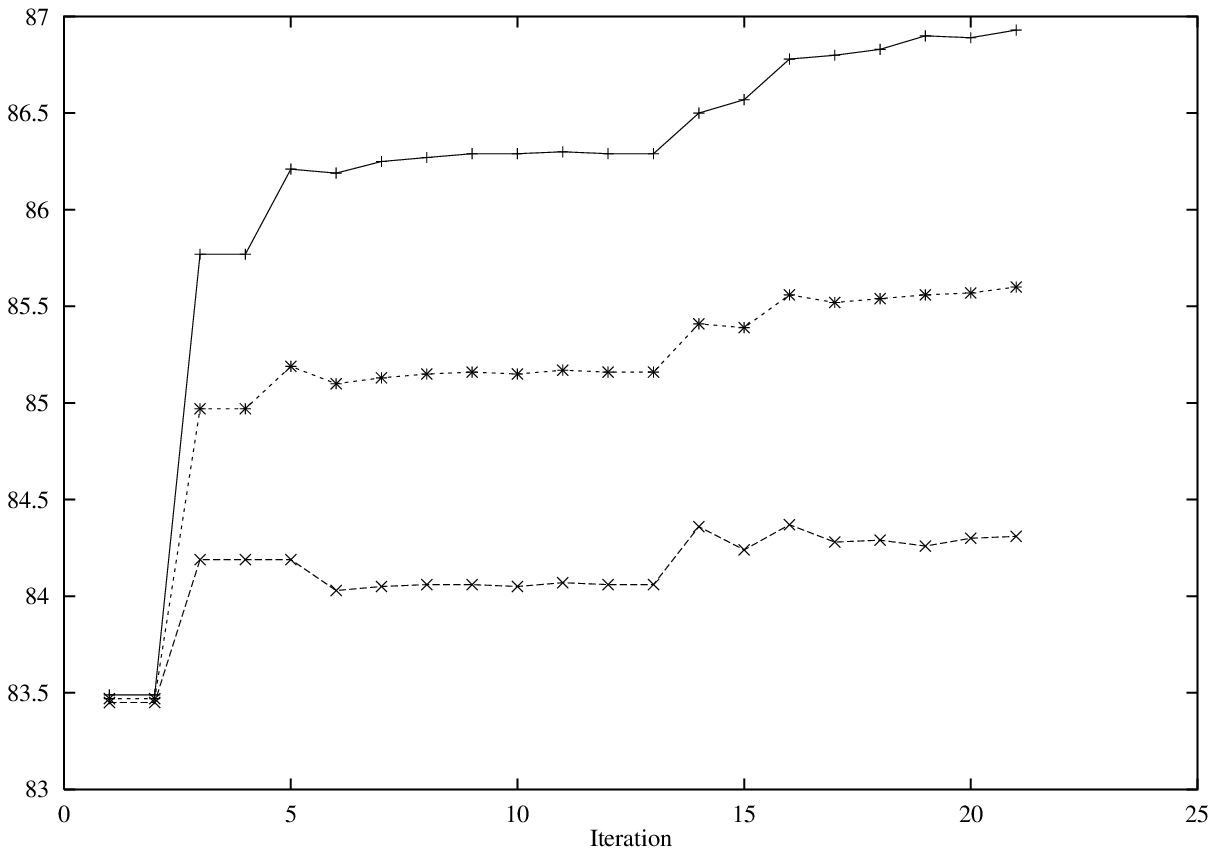, width= 0.45\textwidth}
\\
    \epsfig{file=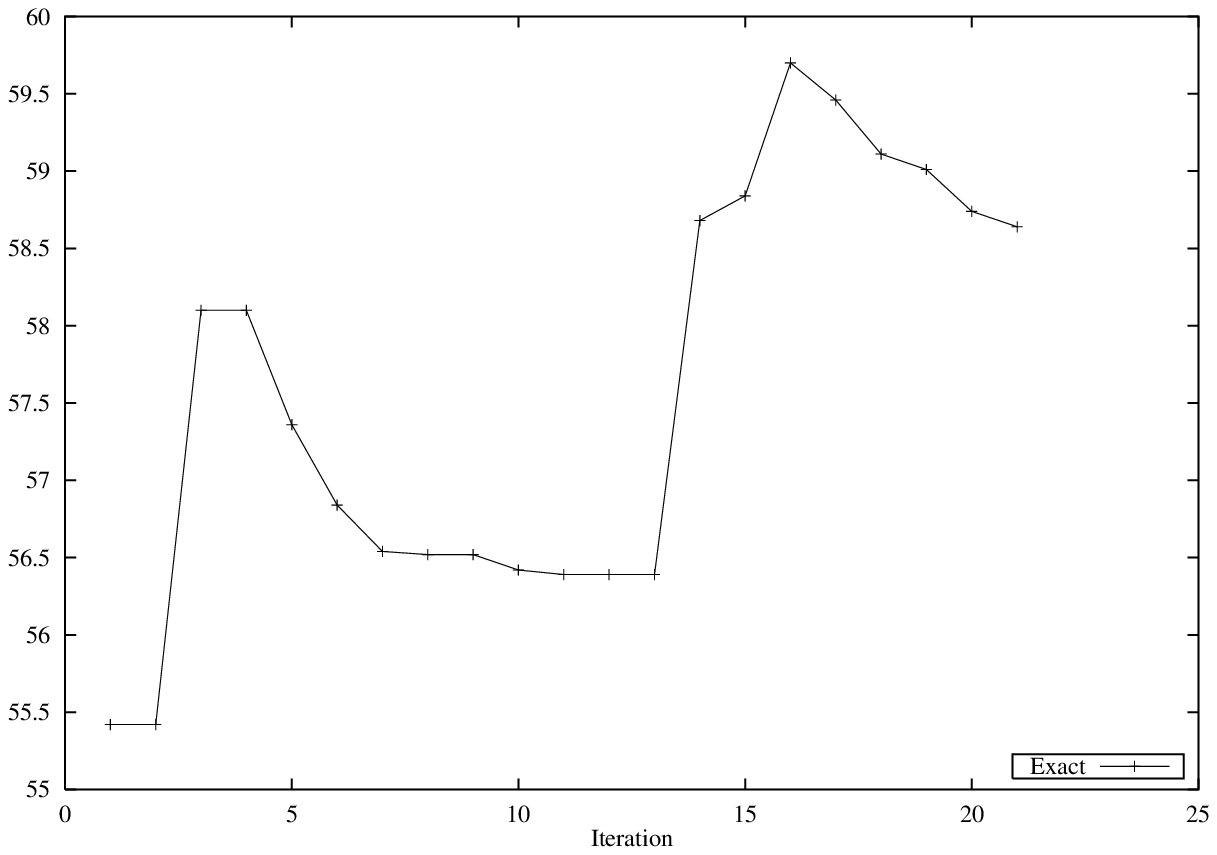, width= 0.45\textwidth}
&
    \epsfig{file=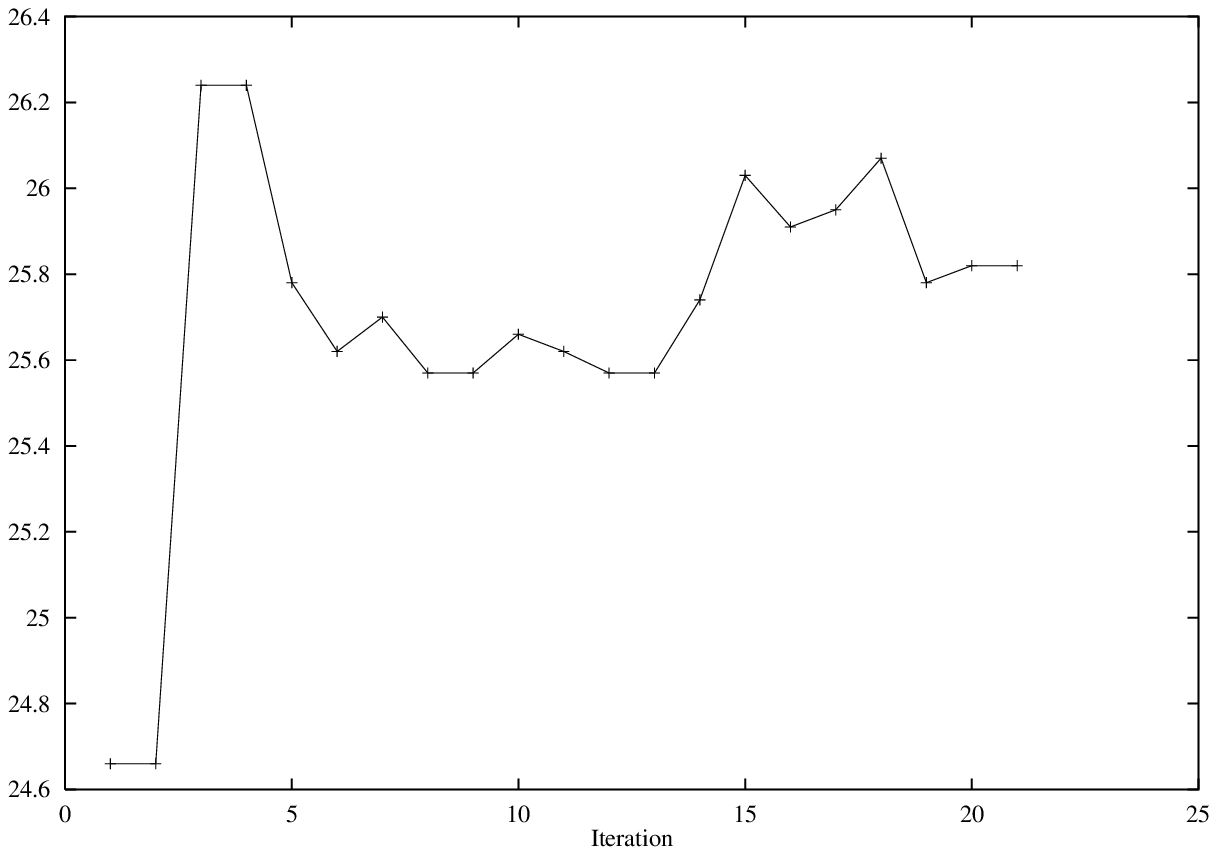, width= 0.45\textwidth}
\end{tabular}
}


  \caption{Backing Off to Bagging (21)} 
  \label{fig:varying:boost.small.bca.rebag.21}
\end{sidewaysfigure}

\begin{table}[htbp]
\centering
\begin{tabular}{|c|l|rr|rr|rr|}
       \hline
      Set
      &\multicolumn{1}{c|}{Instance} 
      &\multicolumn{1}{c}{P} 
      &\multicolumn{1}{c|}{R} 
      &\multicolumn{1}{c}{F}  
      &\multicolumn{1}{c|}{Gain}  
      &\multicolumn{1}{c}{Exact}
      &\multicolumn{1}{c|}{Gain} \\
      \hline 
Training&Original Parser  & 97.07 &  97.30 &  97.18 &NA&    73.3 & NA 
\\
&       Initial & 91.79 & 92.08 & 91.94 &  0.00 & 55.4  &  0.0
\\
&       BestF(19)       & 96.32 & 95.71 & 96.01 &  4.07 & 59.0  &  3.6
\\
&       Final(21)       & 96.32 & 95.69 & 96.00 &  4.06 & 58.6  &  3.2
\\
      \hline 
Test & Original Parser  & 86.03 & 85.43 & 85.73  & NA & 28.6 & NA
\\
&   Initial & 83.49 & 83.45 & 83.47 &  0.00 & 24.7  &  0.0
\\
&       TrainBestF(19)  & 86.90 & 84.26 & 85.56 &  2.09 & 25.8  &  1.1
\\
&       TestBestF(21)   & 86.93 & 84.31 & 85.60 &  2.13 & 25.8  &  1.2
\\
&       Final(21)       & 86.93 & 84.31 & 85.60 &  2.13 & 25.8  &  1.2
\\
\hline
\end{tabular}


  \caption{Backing Off to Bagging (21)} 
  \label{table:varying:boost.small.bca.rebag.21}
\end{table}

In Figure \ref{fig:varying:boost.small.bca.rebag.21} and Table \ref
{table:varying:boost.small.bca.rebag.21} we show the effect of backing
off to bagging on the first 21 iterations of boosting.  In these
iterations, the algorithm backed off to bagging only once, immediately
after iteration 13.  We can see from the top two graphs that this
produces the desired precision and recall results on both the training
and test sets.  The exact sentence accuracy curves, however, display
the fight between the basic tendencies of bagging and boosting.
Bagging tends to make exact sentence accuracy get better, and boosting
tends to make it worse\footnote{Recall that boosting does not attempt
  to maximize exact sentence accuracy except as a side effect of
  maximizing precision or recall.}.

\begin{sidewaysfigure}[htbp]
\centering
\fbox{
\begin{tabular}{rl}
    \epsfig{file=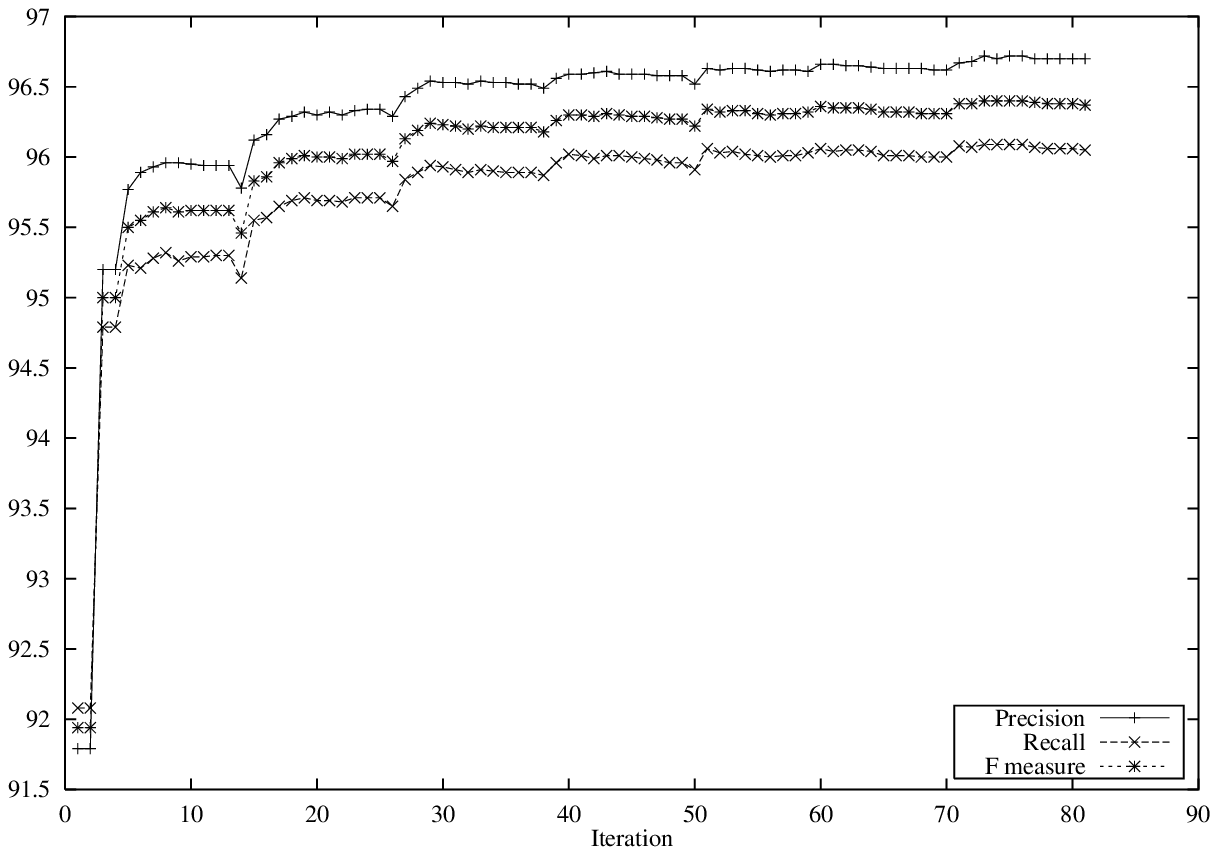, width= 0.45\textwidth}
&
    \epsfig{file=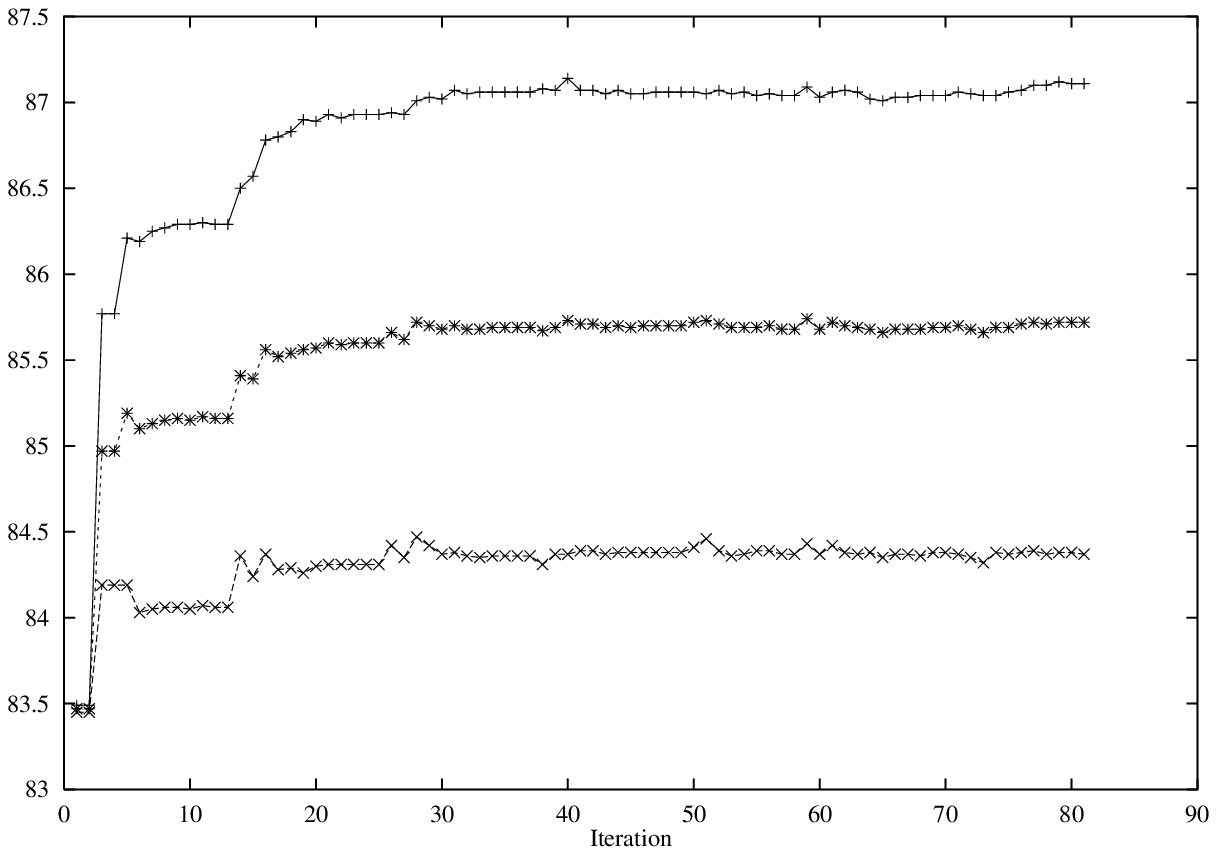, width= 0.45\textwidth}
\\
    \epsfig{file=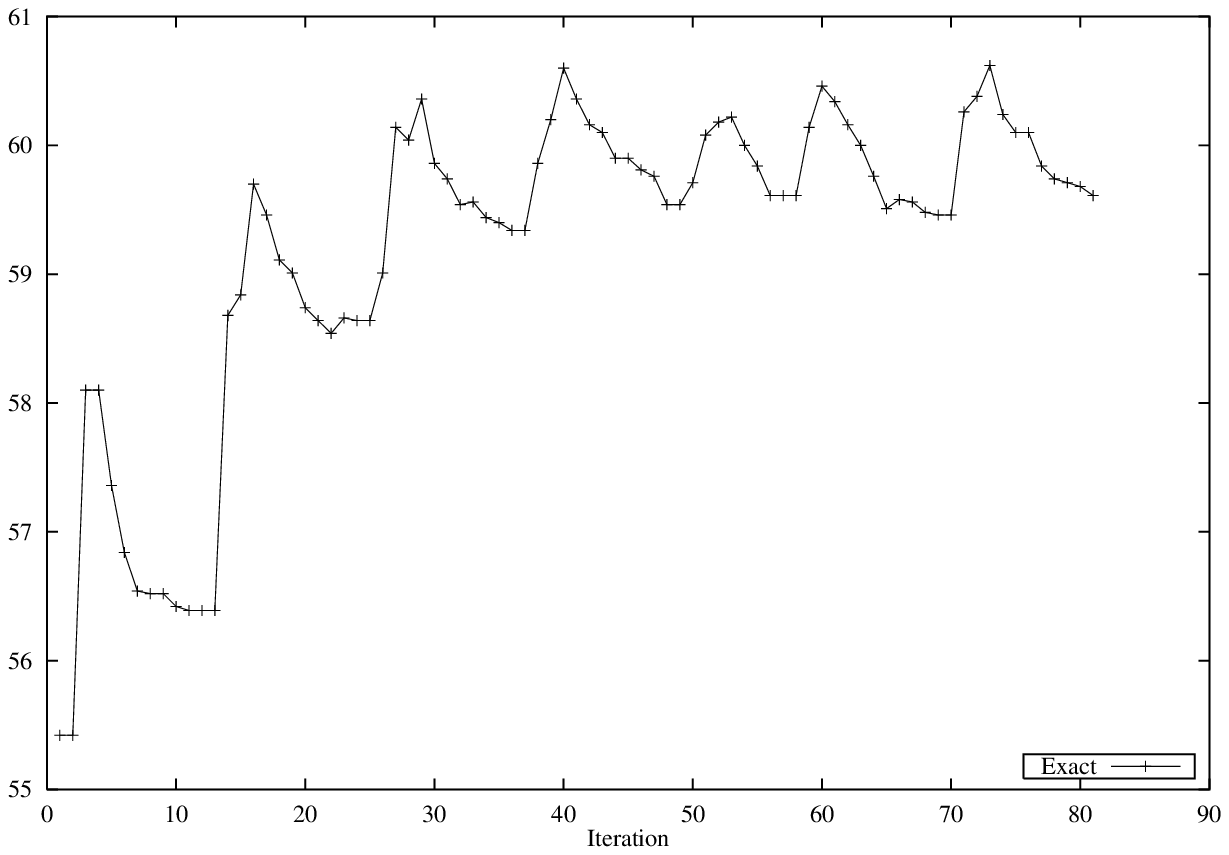, width= 0.45\textwidth}
&
    \epsfig{file=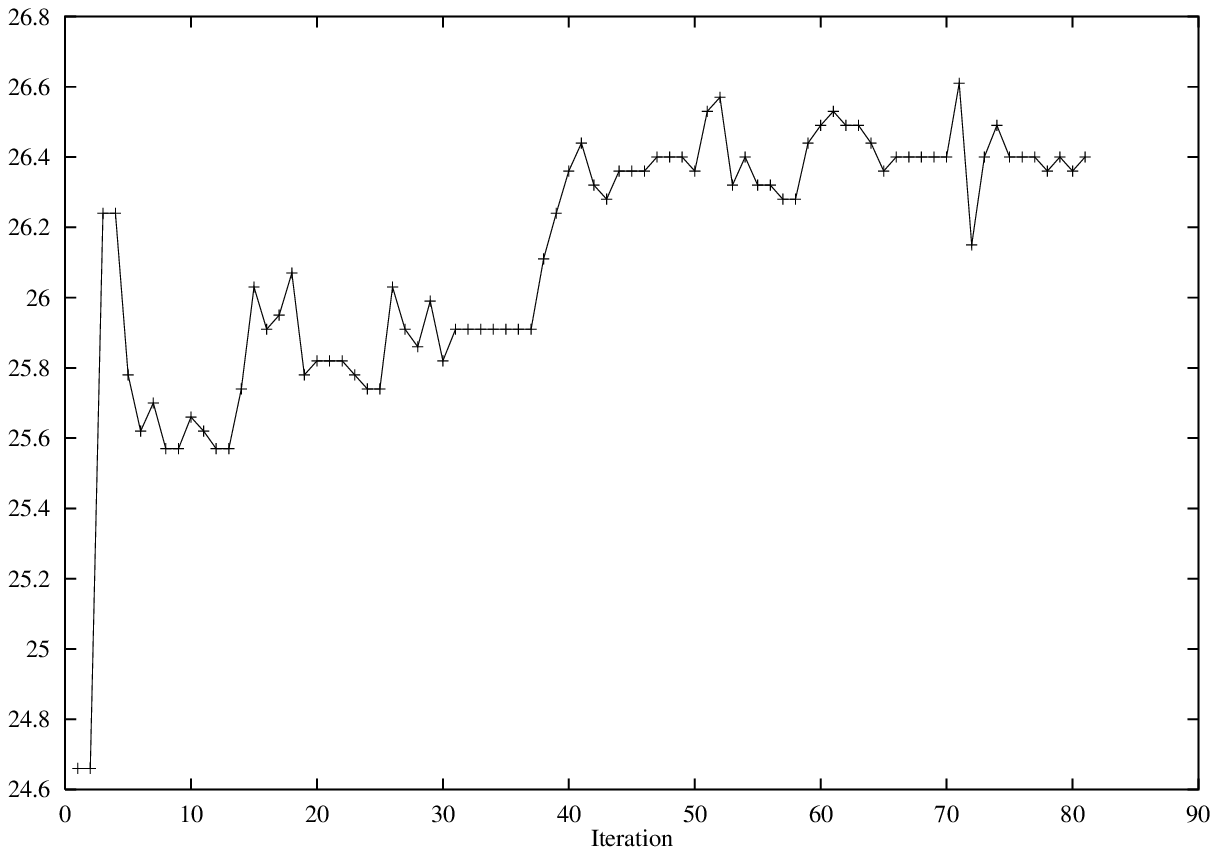, width= 0.45\textwidth}
\end{tabular}
}


  \caption{Backing Off to Bagging} 
  \label{fig:varying:boost.small.bca.rebag}
\end{sidewaysfigure}

\begin{table}[htbp]
\centering
\begin{tabular}{|c|l|rr|rr|rr|}
       \hline
      Set
      &\multicolumn{1}{c|}{Instance} 
      &\multicolumn{1}{c}{P} 
      &\multicolumn{1}{c|}{R} 
      &\multicolumn{1}{c}{F}  
      &\multicolumn{1}{c|}{Gain}  
      &\multicolumn{1}{c}{Exact}
      &\multicolumn{1}{c|}{Gain} \\
      \hline 
Training&Original Parser  & 97.07 &  97.30 &  97.18 &NA&    73.3 & NA 
\\
&       Initial & 91.79 & 92.08 & 91.94 &  0.00 & 55.4  &  0.0
\\
&       BestF(73)       & 96.72 & 96.09 & 96.40 &  4.46 & 60.6  &  5.2
\\
&       Final(81)       & 96.70 & 96.05 & 96.37 &  4.43 & 59.6  &  4.2
\\
      \hline 
Test & Original Parser  & 86.03 & 85.43 & 85.73  & NA & 28.6 & NA
\\
&   Initial & 83.49 & 83.45 & 83.47 &  0.00 & 24.7  &  0.0
\\
&       TrainBestF(73)  & 87.04 & 84.32 & 85.66 &  2.19 & 26.4  &  1.7
\\
&       TestBestF(59)   & 87.09 & 84.43 & 85.74 &  2.27 & 26.4  &  1.8
\\
&       Final(81)       & 87.11 & 84.37 & 85.72 &  2.25 & 26.4  &  1.7
\\
\hline
\end{tabular}


  \caption{Backing Off to Bagging} 
  \label{table:varying:boost.small.bca.rebag}
\end{table}

In Figure \ref{fig:varying:boost.small.bca.rebag} and Table
\ref{table:varying:boost.small.bca.rebag} we show the effect of
backing off to bagging on 81 replicates.  Here we can better see the
staircase effect of bagging combined with the tapering gains of
boosting.

\subsection{Effects of Noisy or Inconsistent Training Data}

\subsubsection{Detecting Violations of the Weak Learner Criterion}

The parser we worked with was not a weak learner.  This was discovered
after most of the boosting experiments were performed.  It was noted
that the distribution became very skewed as boosting continued.
Inspection of the sentences that were getting much mass placed upon
them revealed that their weight was being boosted \emph{in every
  iteration}.  The hypothesis was that the parser was simply unable to
learn them.

In order to test this hypothesis, we built 39,832 parsers, one for each
sentence in our training set.  Each of these parsers was trained on
only a single sentence\footnote{The sentence was replicated 10 times
  to avoid poor probability estimates.} and evaluated on the same
sentence.  In doing this we found that a full 4764 (11.2\%) of these
sentences could not be parsed correctly.  The parser did not have the
weak learner property for this dataset.

\subsubsection{Data Trimming}

\begin{sidewaysfigure}[htbp]
\centering
\fbox{
\begin{tabular}{rl}
    \epsfig{file=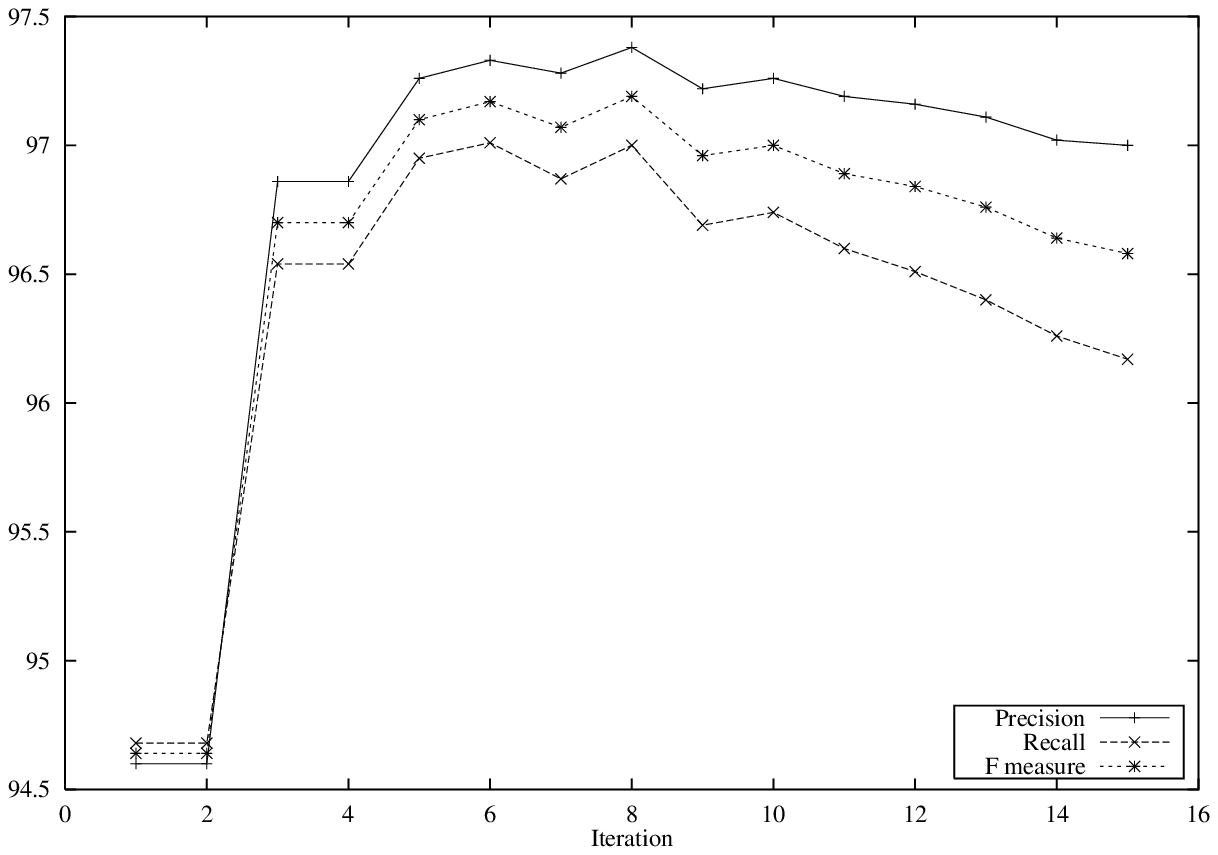, width= 0.45\textwidth}
&
    \epsfig{file=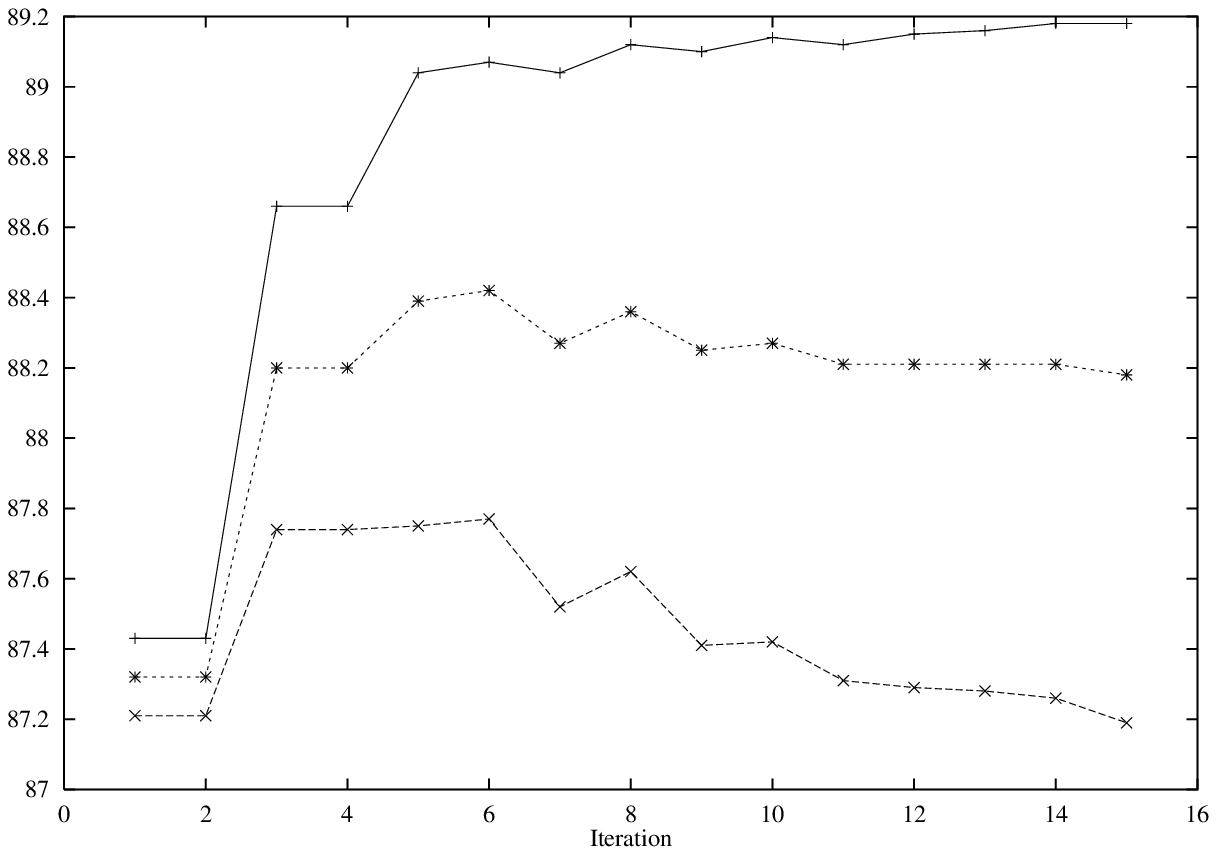, width= 0.45\textwidth}
\\
    \epsfig{file=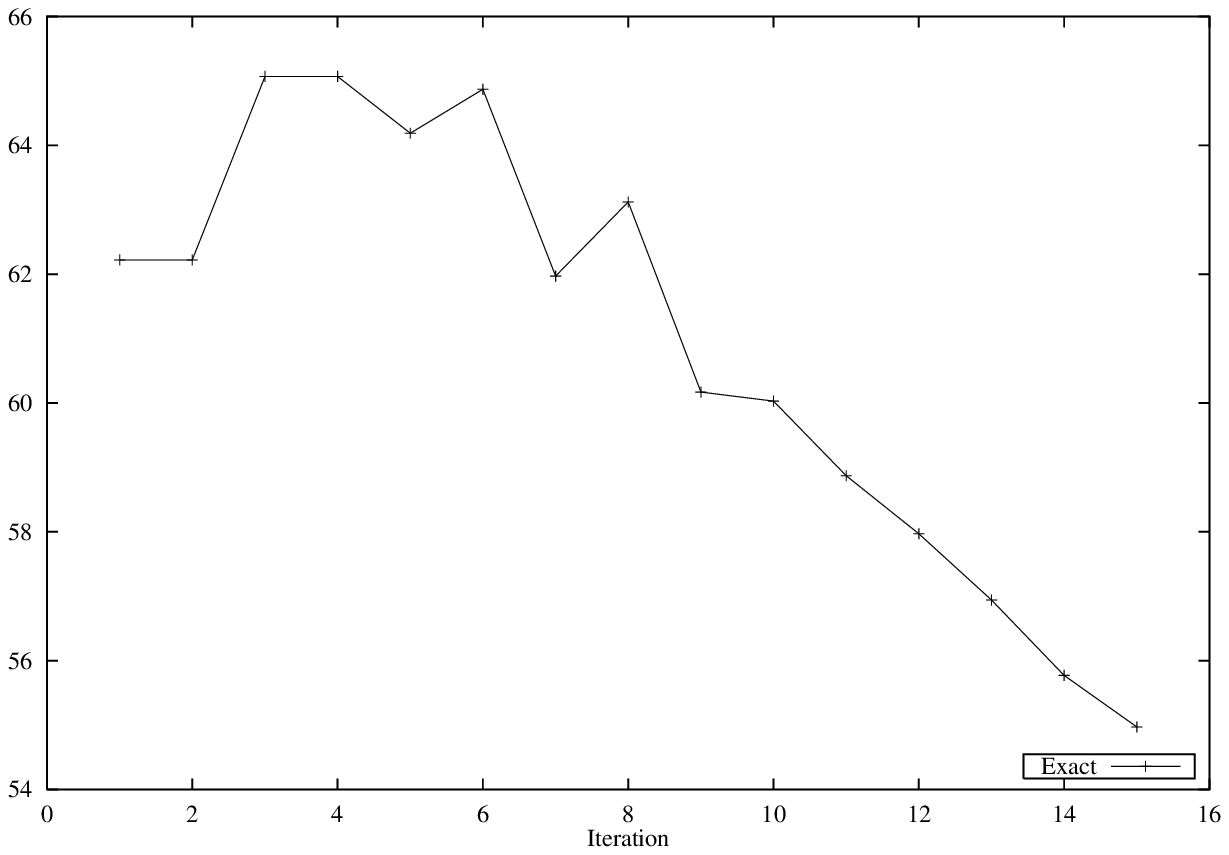, width= 0.45\textwidth}
&
    \epsfig{file=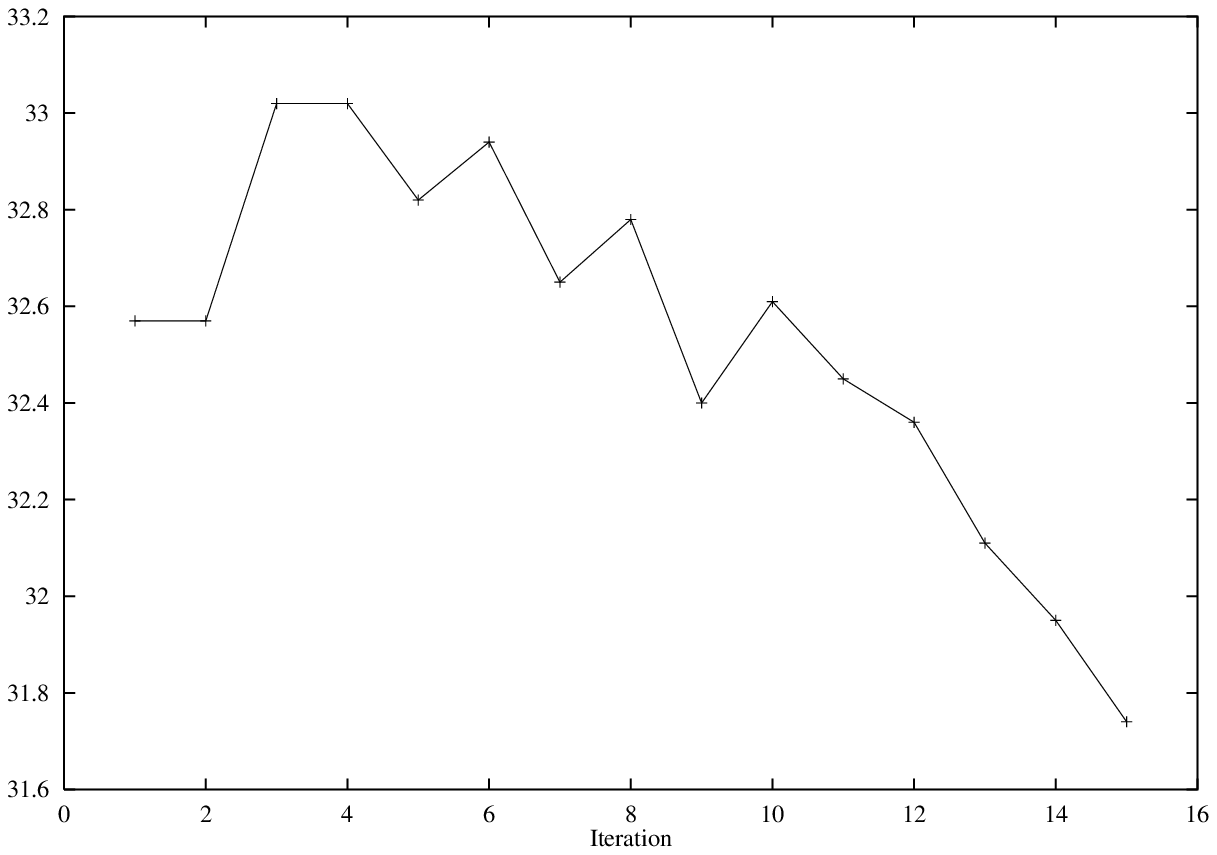, width= 0.45\textwidth}
\end{tabular}
}


  \caption{Boosting the Stable Corpus} 
  \label{fig:varying:boost.stable}
\end{sidewaysfigure}

In order to evaluate how well boosting worked with a weak learner, we
removed those sentences in the corpus that could not be memorized in
isolation by the parser.  We reran the best boosting experiment
(boosting for constituent accuracy) on the entire Treebank minus the
troublesome sentences.  The results are in Table
\ref{table:varying:boost.stable} and Figure
\ref{fig:varying:boost.stable}.  When comparing to the results using
the entire Treebank, we notice that this dataset gets a much larger
gain.  The initial accuracy, however, is much lower.  We conclude that
the boosting algorithm did perform better here, but the parser was
learning useful information in those sentences that it couldn't
memorize that was applied to the test set.

\begin{table}[htbp]
\centering
\begin{tabular}{|c|l|rr|rr|rr|}
       \hline
      Set
      &\multicolumn{1}{c|}{Instance} 
      &\multicolumn{1}{c}{P} 
      &\multicolumn{1}{c|}{R} 
      &\multicolumn{1}{c}{F}  
      &\multicolumn{1}{c|}{Gain}  
      &\multicolumn{1}{c}{Exact}
      &\multicolumn{1}{c|}{Gain} \\
      \hline 
Training 
&       Original Parser & 96.25 & 96.31 & 96.28 & NA    & 64.7 & NA
\\
&       Initial         & 94.60 & 94.68 & 94.64 &  0.00 & 62.2  &  0.0
\\
&       BestF(8)        & 97.38 & 97.00 & 97.19 &  2.55 & 63.1  &  0.9
\\
&       Final(15)       & 97.00 & 96.17 & 96.58 &  1.94 & 55.0  & -7.2
\\
      \hline 
Test
& Original Parser &   88.73 &   88.54 &   88.63  & NA &34.9 &NA
\\
&   Initial & 87.43 & 87.21 & 87.32 &  0.00 & 32.6  &  0.0
\\
&       TrainBestF(8)   & 89.12 & 87.62 & 88.36 &  1.04 & 32.8  &  0.2
\\
&       TestBestF(6)    & 89.07 & 87.77 & 88.42 &  1.10 & 32.9  &  0.4
\\
&       Final(15)       & 89.18 & 87.19 & 88.18 &  0.86 & 31.7  & -0.8
\\
\hline
\end{tabular}


  \caption{Boosting the Stable Corpus} 
  \label{table:varying:boost.stable}
\end{table}

In this manner we managed to clean our dataset to the point that the
parser could learn each sentence in isolation.  We cannot blame the
corpus-makers for the sentences that could not be memorized, however.
The parser's model just would not accommodate them, for better or for
worse.\footnote{Some of the annotation for sentences we threw out
  looked questionable, but we cannot distinguish them in any
  principled manner from those that were simply too complex for the
  parsing model.}

The question of the existence of inconsistent annotation arises.
There may be sentences in the corpus that can be learned by the parser
induction algorithm in isolation but not in concert.  For example,
they could contain conflicting information.  Finding these sentences
would lead us to a better understanding of the quality of our corpus,
and give an idea for where improvements in annotation quality can be
made.

\subsubsection{Informative Simulation (Gedanken Experiment)}

We will first investigate a noisy dataset by simulation to get a feel
for how susceptible boosting is to inconsistency.  Imagine a strange
dataset which consists of only three samples.  Two of the samples have
identical features, but inconsistent labels.  The third sample is
completely different.  We are completely simplifying the concept of a
noisy dataset in this way but we will empirically witness the
effectiveness of boosting a classifier trained on this dataset.  The
first assumption is that the weak learner can always learn the
consistently labelled sample.  Also, it can learn to predict the label
of one of the two inconsistently labelled samples.  It cannot predict
both as that is completely useless, and ties must be broken in some
deterministic way.

\begin{table}[htbp]
\centering
\begin{tabular}{|r|ccc|cc|}
  \hline
  &\multicolumn{3}{c|}{Sample Weight}
  & Weighted &
  \\
  \multicolumn{1}{|c}{Iteration}
  & \multicolumn{1}{|c}{(a,-1)}
  &\multicolumn{1}{c}{(a,1)} 
  & \multicolumn{1}{c|}{(b,1)} 
  & \multicolumn{1}{c}{Error}  
  & \multicolumn{1}{c|}{$\exp(\alpha)$}
  \\
  \hline
  1 &  
  1/3 & 1/3* & 1/3 
  & 1/3 & 1/2
  \\
  2 &  
  1/4* & 1/2 & 1/4
  & 1/4 & 1/3
  \\
  3 &  
  1/2 & 1/3* & 1/6 
  & 1/3 & 1/2
  \\
  4 &  
  3/8* & 1/2 & 1/8 
  & 3/8 & 3/5
  \\
  5 &  
  1/2 & 2/5* & 1/10
  & 2/5 & 2/3
  \\
  6 &  
  5/12* & 1/2 & 1/12
  & 5/12 & 5/7
  \\
  \hline
\end{tabular}
\caption{Simulation: Boosting an Inconsistent Dataset}
\label{table:varying:simulation}
\end{table}

In Table \ref{table:varying:simulation} we see the result of our
simulation.  We are considering how the weights change on three sample
data points from a corpus, \{(a,-1),(a,1),(b,1)\}.  In each iteration,
we mark the sample that is by necessity predicted incorrectly by the
classifier with an asterisk.  The Weighted Error Column shows the overall
error of the corpus, as weighted by the distribution.  The rightmost
column gives the vale of the distribution updating parameter.  When
the value is larger, correctly predicted samples are reduced in weight
more during boosting.

The first thing to note in the table is that the weights on our
inconsistent examples increases and the weight on the easily learned
example decreases.  The effect is so strong that in the limit, all of
the weight would be focused on the inconsistent examples.  Secondly,
the examples didn't present any problem for the weak learner.  In
every case it was able to produce a classifier with an error less than
1/2.  In the limit, though, the error rate will go to 1/2, as all of
the weight is focused on the inconsistent samples.

\subsubsection{Empirical Evidence of Noise}

Thought experiments can give some theoretical insights into phenomena
of interest, but only data analysis can provide real evidence to
ground the insights in the real world.

To acquire experimental evidence of noisy data, we inspected the
distributions that were used during boosting.  We expected to see
the distribution become very skewed if there is noise in the data, or
remain uniform with slight fluctuations if it is doing a good job of
fitting the data.


\begin{figure}
  \centering
  \epsfig{file=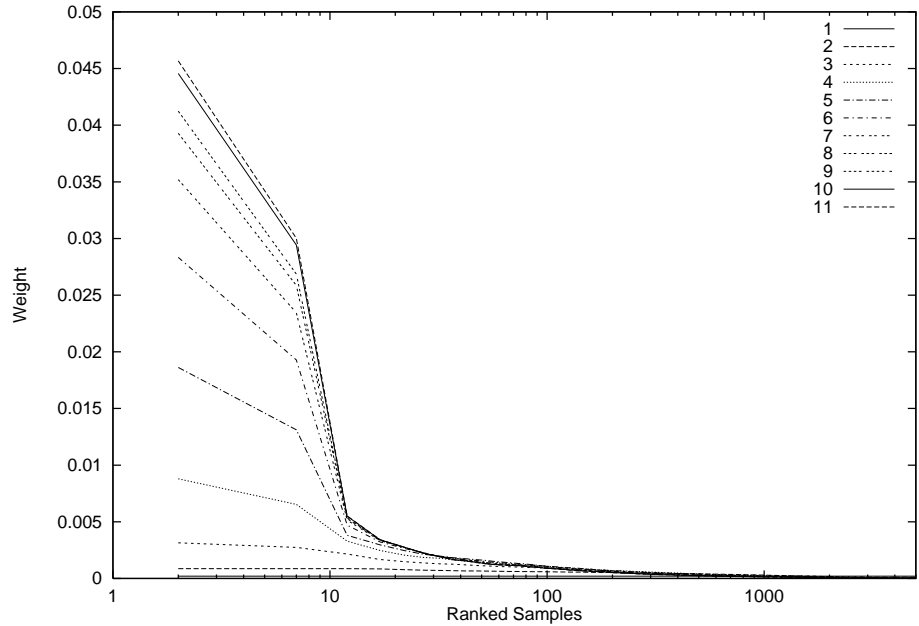,height= .4  \textheight}
  \caption{Weight Change During Boosting}
  \label{figure:varying:freqrank} 
\end{figure}

In Figure \ref{figure:varying:freqrank} we see how the boosting weight
distribution changes.  This was the weight during a training run using
a corpus of 5000 sentences. We rank the sentences by the weight they
are given by the distribution, and sort them in decreasing order by
weight along the x-axis.  The samples were then placed into bins each
containing an equal number of samples, and the average mass of samples
in the bin is reported on the y-axis.  The labels of the curves on
this graph correspond to iterations of a boosting run.  We used 1000
bins for this graph, and a log scale on the x-axis.  Since there were
5000 samples, all samples initially had a y-value of 0.0002.

There are a couple interesting things shown in this graph.  The left
endpoints of the lines move from bottom to top in order of boosting
iteration.  The distribution becomes monotonically more skewed as
boosting progresses.  Secondly we see by the last iteration that most
of the weight is focused on less than 100 samples.  Also the highest
weight appears to be converging at this point, suggesting that there
is some asymptotic effect taking place.  In all, this graph suggests
there is noise in the corpus.



In Section \ref{section:varying:qualitycontrol} we describe the
inconsistencies of the data in more detail.

\section{Evaluation}
\label{section:varying:evaluation}

For practical reasons, many of the experiments we performed earlier in
this chapter were working with a training set of only 5000 sentences
instead of the full Treebank training set which is nearly 8 times
that size.  Boosting in particular is a very computationally expensive
procedure because it requires the parsers to be created in a serial
manner, and a complete re-parsing of the training set in each
iteration.  Bagging on the other hand does not use a feedback loop.
The training set is resampled, and the learning algorithm is simply
run once for each parse.  No re-parsing of the training set is
required.

In this section we will show the results of the best parser
diversification algorithms when they are run using the entire training
portion of the Treebank.  When we refer to bagging, we are using
simple bagging, with a uniform distribution over sentences.  When we
refer to boosting, we are using boosting for constituent accuracy with
backing-off to boosting when the parser displays non-weak learner
behavior.

\begin{sidewaysfigure}[htbp]
  \centering
\fbox{
\begin{tabular}{rl}
    \epsfig{file=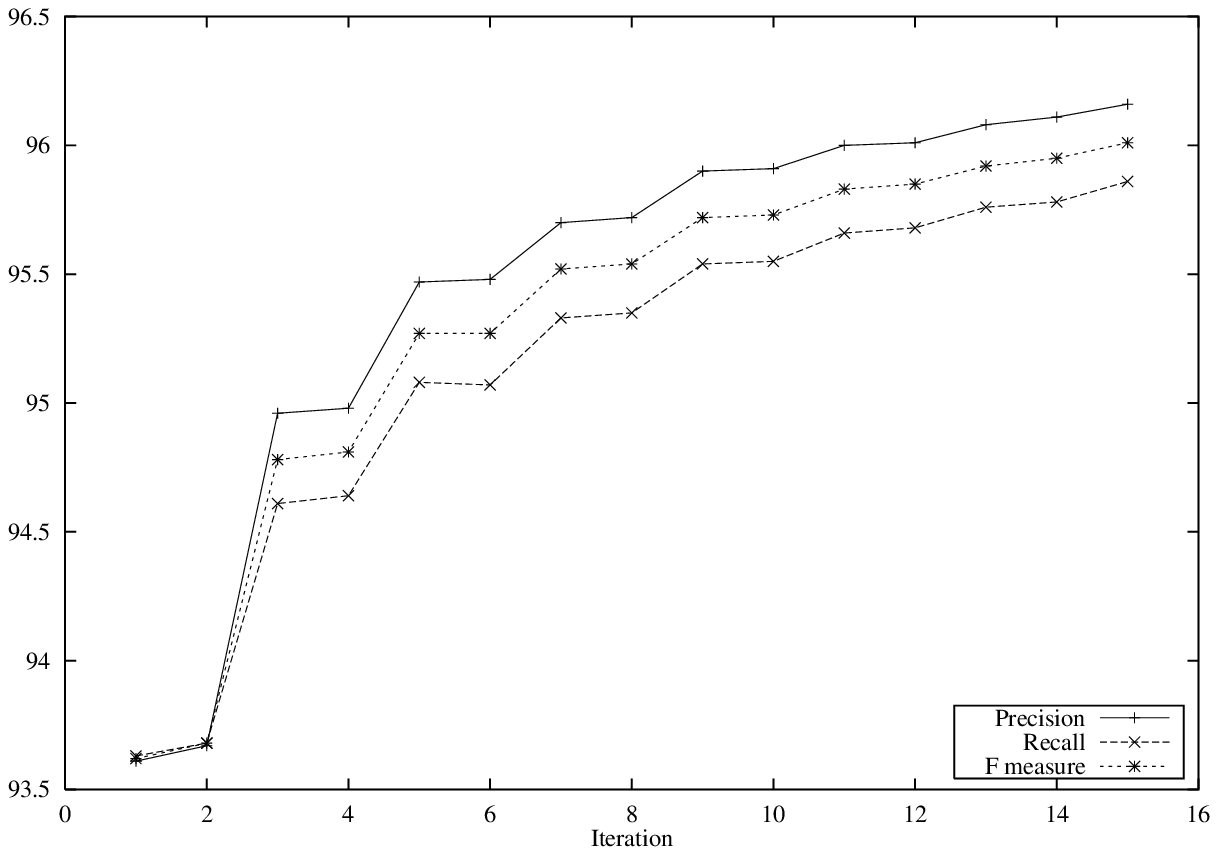, width= 0.45\textwidth}
&
    \epsfig{file=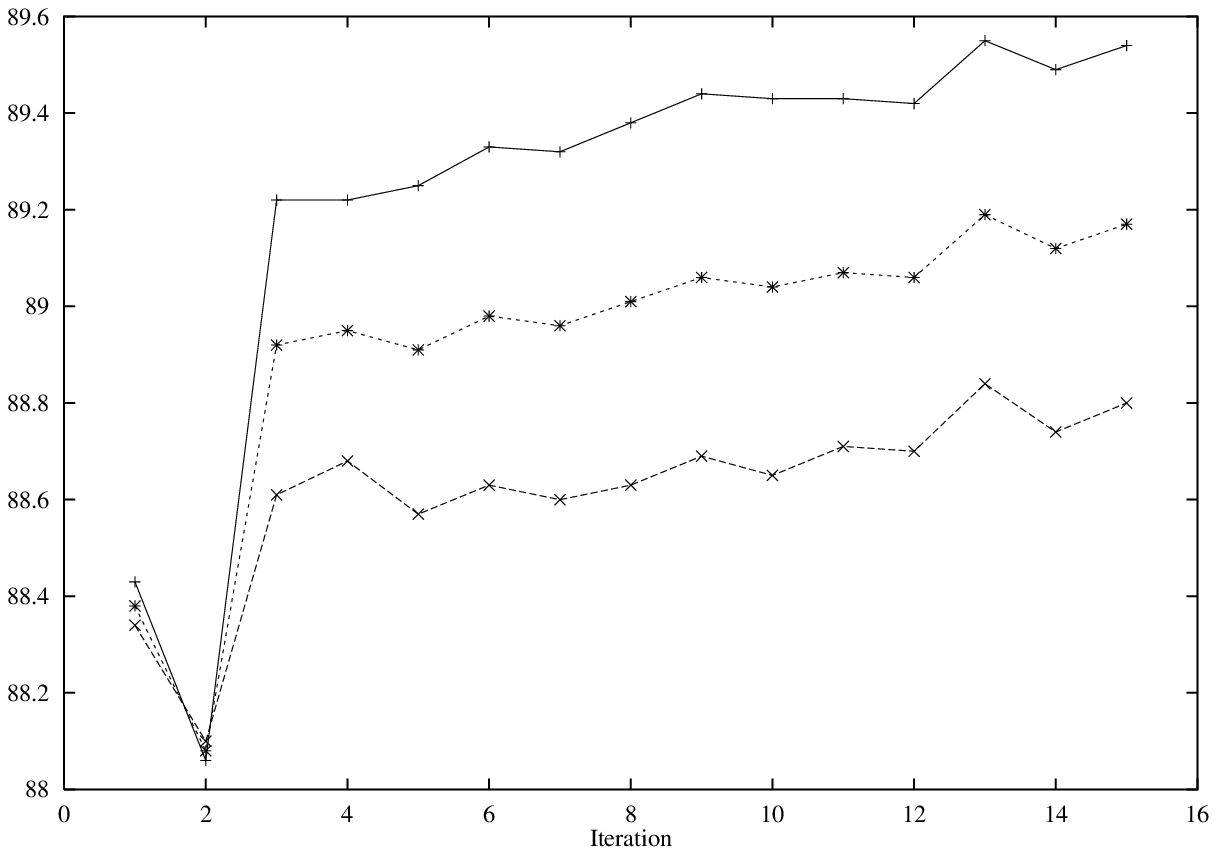, width= 0.45\textwidth}
\\
    \epsfig{file=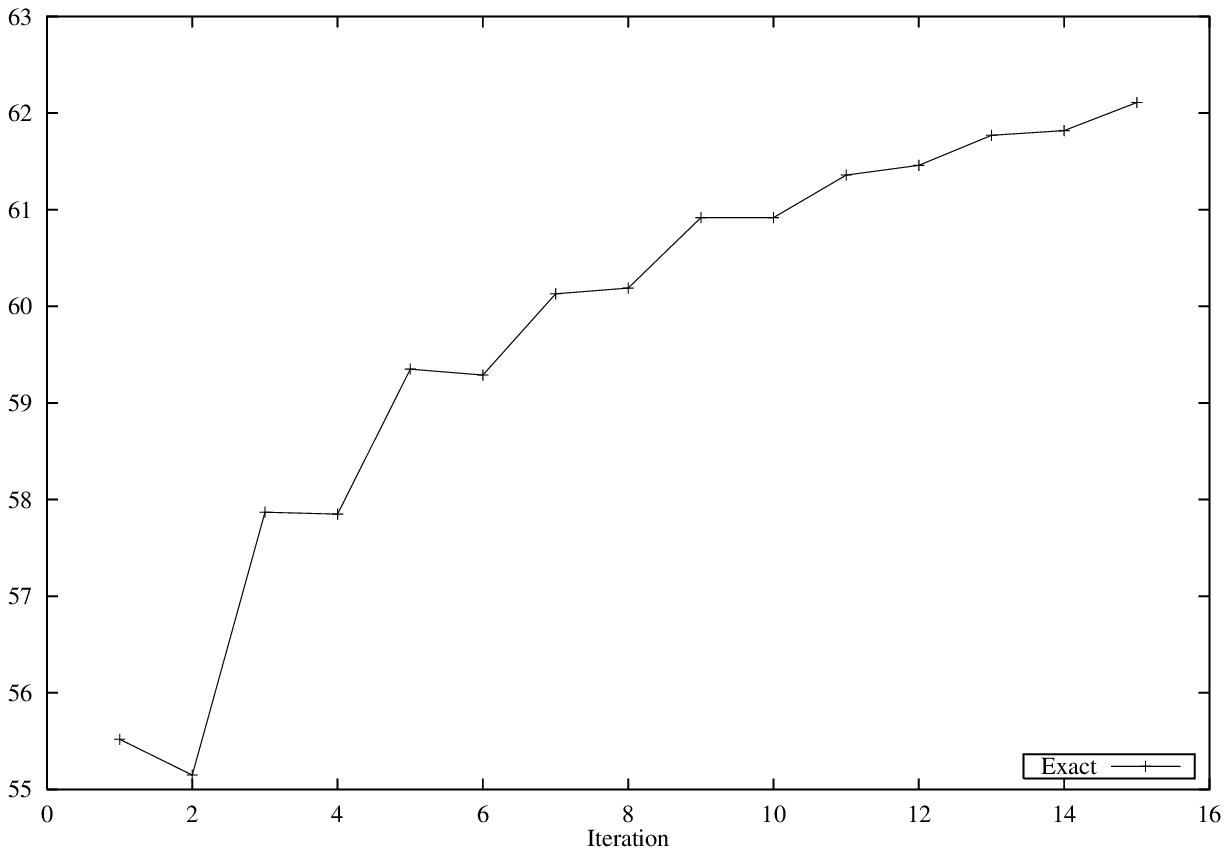, width= 0.45\textwidth}
&
    \epsfig{file=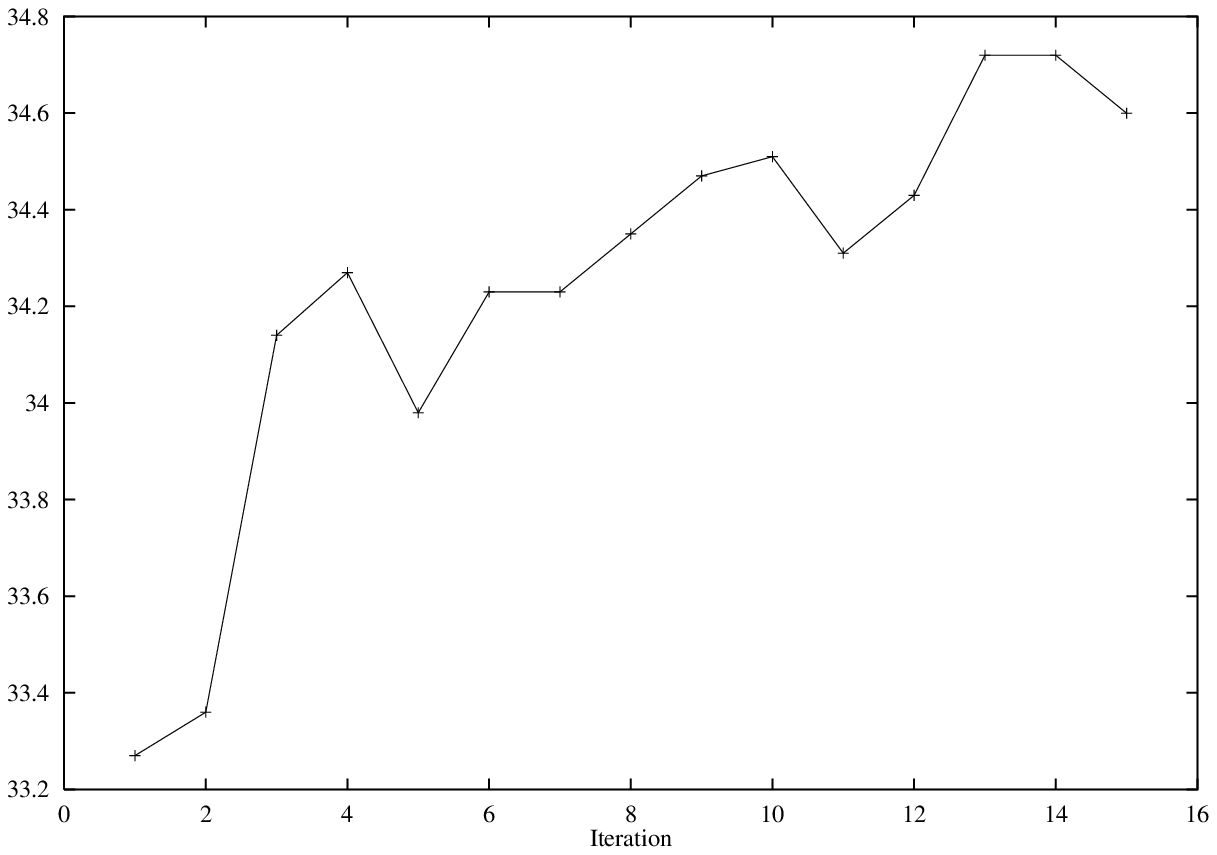, width= 0.45\textwidth}
\end{tabular}
}


  \caption{Bagging the Treebank} 
  \label{fig:varying:bag.sent.uni}
\end{sidewaysfigure}

\begin{table}[htbp]
  \centering
\begin{tabular}{|c|l|rr|rr|rr|}
       \hline
      Set
      &\multicolumn{1}{c|}{Instance} 
      &\multicolumn{1}{c}{P} 
      &\multicolumn{1}{c|}{R} 
      &\multicolumn{1}{c}{F}  
      &\multicolumn{1}{c|}{Gain}  
      &\multicolumn{1}{c}{Exact}
      &\multicolumn{1}{c|}{Gain} \\
      \hline 
Training
&       Original Parser & 96.25 & 96.31 & 96.28 & NA    & 64.7 & NA
\\
&       Initial & 93.61 & 93.63 & 93.62 &  0.00 & 55.5  &  0.0
\\
&       BestF(15)       & 96.16 & 95.86 & 96.01 &  2.39 & 62.1  &  6.6
\\
&       Final(15)       & 96.16 & 95.86 & 96.01 &  2.39 & 62.1  &  6.6
\\
      \hline 
Test
& Original Parser &   88.73 &   88.54 &   88.63  & NA &34.9 &NA
\\
&   Initial & 88.43 & 88.34 & 88.38 &  0.00 & 33.3  &  0.0
\\
&       TrainBestF(15)  & 89.54 & 88.80 & 89.17 &  0.79 & 34.6  &  1.3
\\
&       TestBestF(13)   & 89.55 & 88.84 & 89.19 &  0.81 & 34.7  &  1.4
\\
&       Final(15)       & 89.54 & 88.80 & 89.17 &  0.79 & 34.6  &  1.3
\\
\hline
\end{tabular}


  \caption{Bagging the Treebank} 
  \label{table:varying:bag.sent.uni}
\end{table}

In Figure \ref{fig:varying:bag.sent.uni} and Table
\ref{table:varying:bag.sent.uni} we see the results for bagging.  On
the training set all of the accuracy measures are improved, and on the
test set there is clear improvement in precision and recall.  The
improvement on exact sentence accuracy for the test set is
significant, but only marginally so.  The dip on the test set curves
after iteration number two is due to a tie-breaking issue.  The second
of the training set parsers was the chosen leader for this iteration.
It is suggestive of an unlucky resampling of the data for that
iteration. 

The overall gain achieved on the test set by bagging was 0.8 units of
F-measure, but because the entire corpus is not used in each bag the
initial performance is approximately 0.2 units below the best
previously reported result.  The net gain by this technique is 0.6
units of F-measure, which we show in the next section is close to
the amount that is achieved by doubling the training set size from
20000 to approximately 40000 sentences.  The gain we are reporting is
cumulative above the gain that is achieved with the larger corpus.

Our bagging performance increases have not levelled off by the final
iteration, as well.  Because of constraints on computational resources
and time we did not extend the experiment longer, but we would expect
to see more gains from the process it were.

\begin{sidewaysfigure}[htbp]
  \centering
\fbox{
\begin{tabular}{rl}
    \epsfig{file=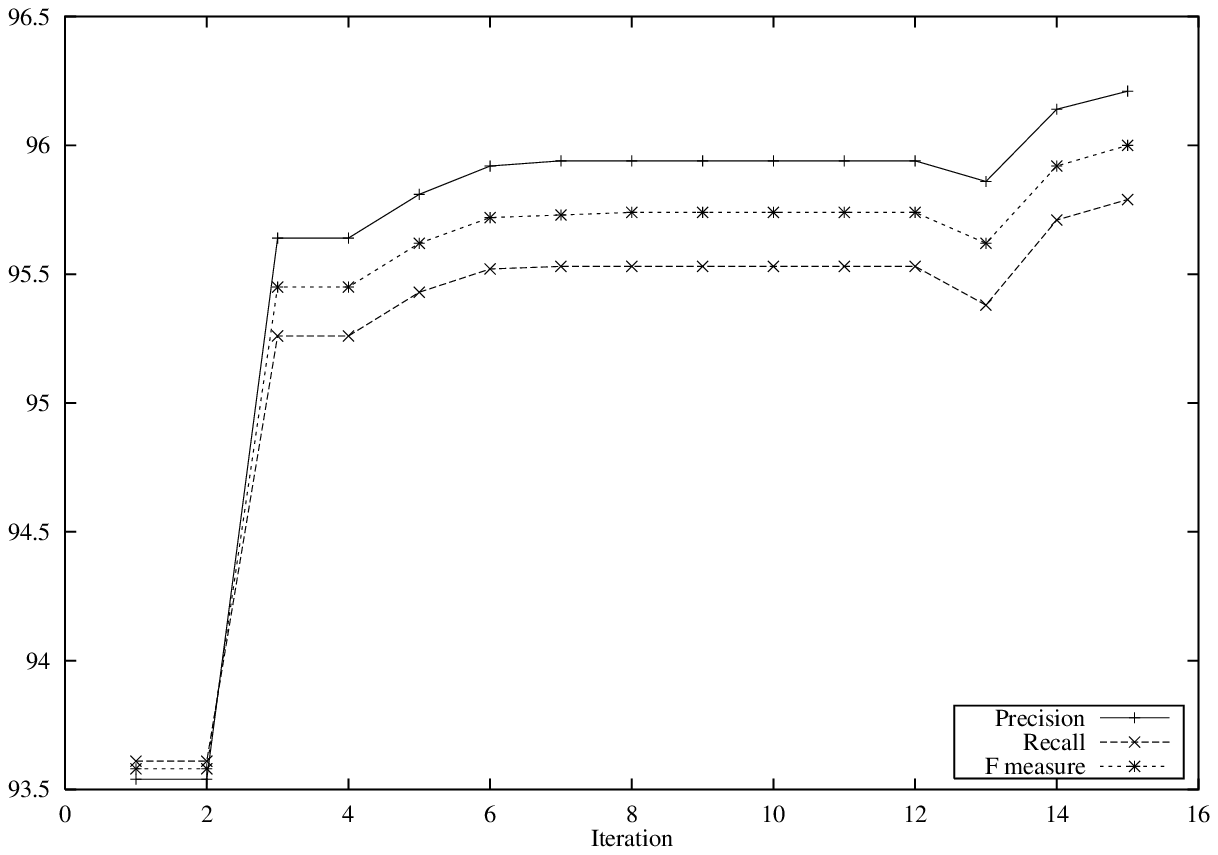, width= 0.45\textwidth}
&
    \epsfig{file=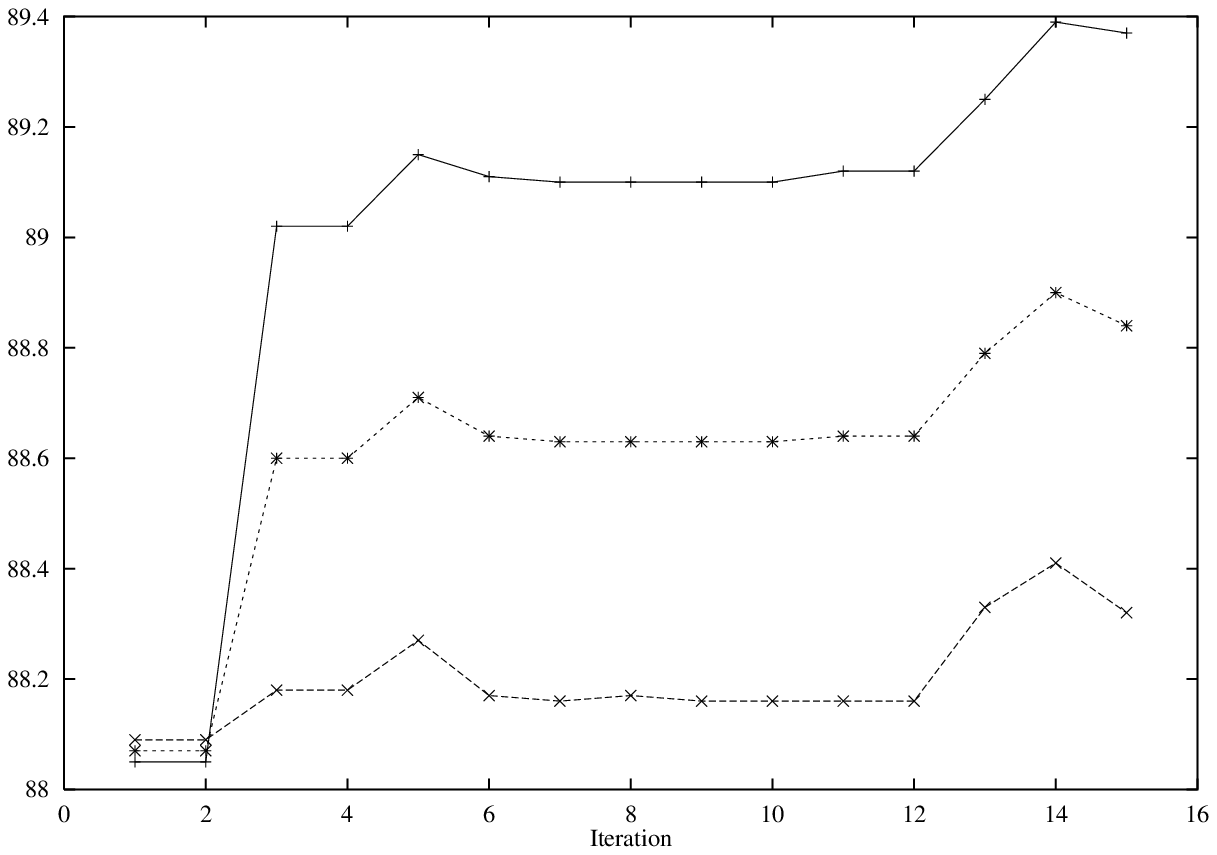, width= 0.45\textwidth}
\\
    \epsfig{file=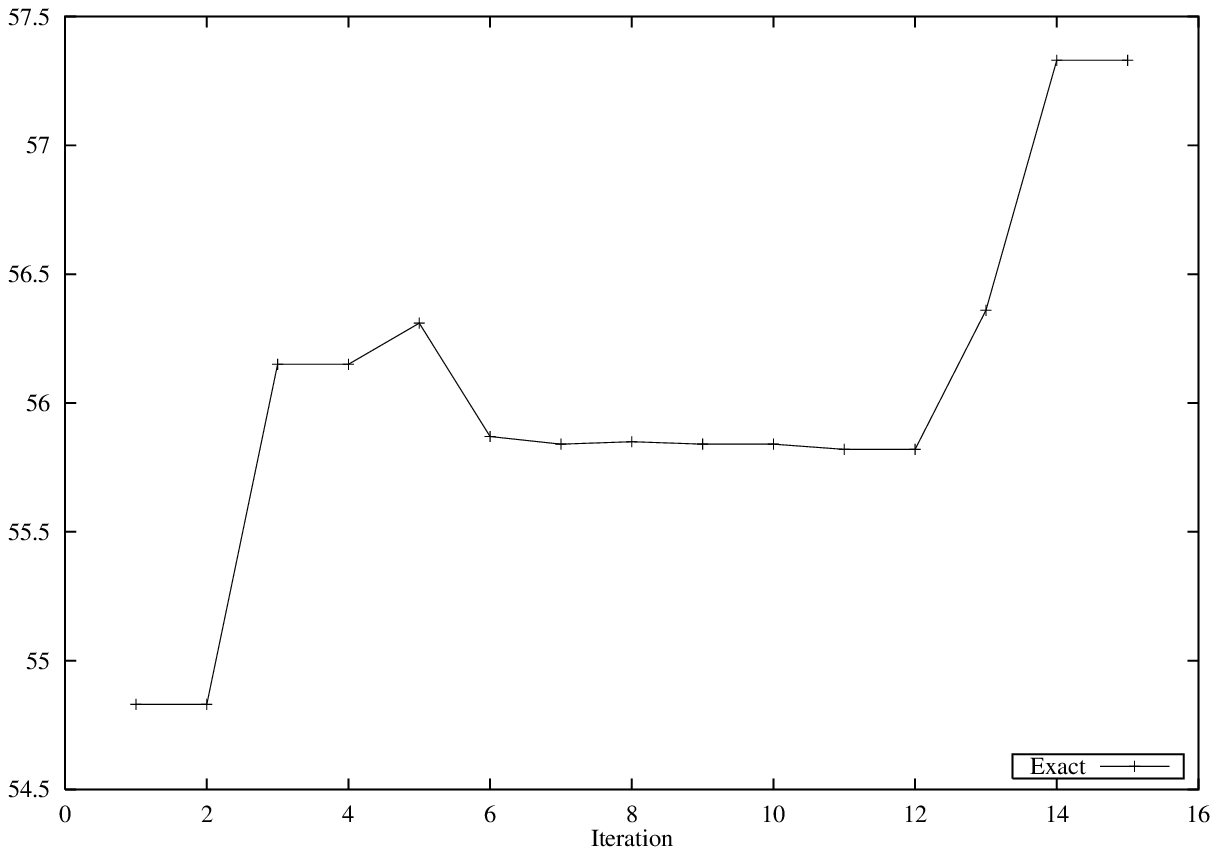, width= 0.45\textwidth}
&
    \epsfig{file=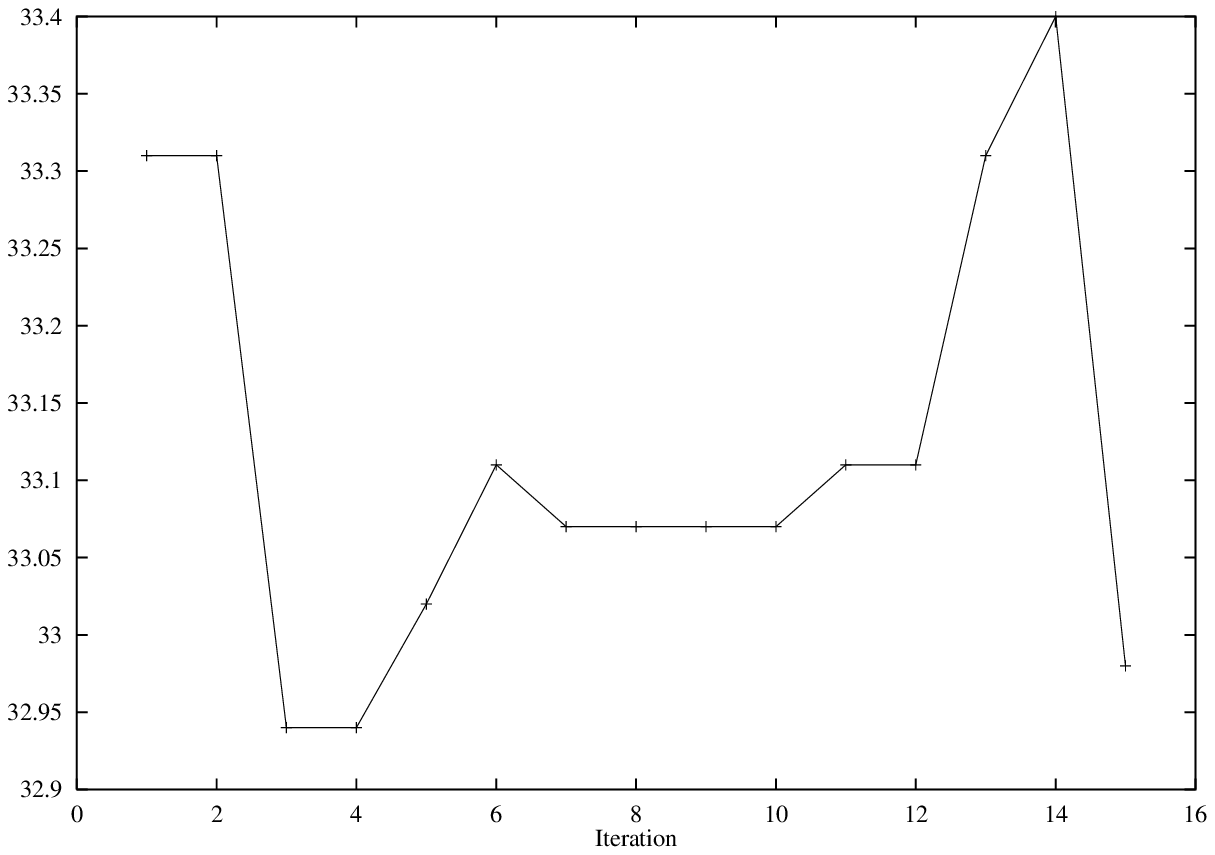, width= 0.45\textwidth}
\end{tabular}
}


  \caption{Boosting the Treebank} 
  \label{fig:varying:boost.sent}
\end{sidewaysfigure}

\begin{table}[htbp]
  \centering
\begin{tabular}{|c|l|rr|rr|rr|}
       \hline
      Set
      &\multicolumn{1}{c|}{Instance} 
      &\multicolumn{1}{c}{P} 
      &\multicolumn{1}{c|}{R} 
      &\multicolumn{1}{c}{F}  
      &\multicolumn{1}{c|}{Gain}  
      &\multicolumn{1}{c}{Exact}
      &\multicolumn{1}{c|}{Gain} \\
      \hline 
Training
&       Original Parser & 96.25 & 96.31 & 96.28 & NA    & 64.7 & NA
\\
&       Initial & 93.54 & 93.61 & 93.58 &  0.00 & 54.8  &  0.0
\\
&       BestF(15)       & 96.21 & 95.79 & 96.00 &  2.42 & 57.3  &  2.5
\\
&       Final(15)       & 96.21 & 95.79 & 96.00 &  2.42 & 57.3  &  2.5
\\
      \hline 
Test
& Original Parser &   88.73 &   88.54 &   88.63  & NA &34.9 &NA
\\
&   Initial & 88.05 & 88.09 & 88.07 &  0.00 & 33.3  &  0.0
\\
&       TrainBestF(15)  & 89.37 & 88.32 & 88.84 &  0.77 & 33.0  & -0.3
\\
&       TestBestF(14)   & 89.39 & 88.41 & 88.90 &  0.83 & 33.4  &  0.1
\\
&       Final(15)       & 89.37 & 88.32 & 88.84 &  0.77 & 33.0  & -0.3
\\
\hline
\end{tabular}


  \caption{Boosting the Treebank} 
  \label{table:varying:boost.sent}
\end{table}

In Figure \ref{fig:varying:boost.sent} and Table
\ref{table:varying:boost.sent} we see the results for boosting.  The
first thing to notice is that the notch in all the graphs at iteration
13 comes from the boosting algorithm backing off to bagging on that
iteration.  Secondly, we see a large plateau in performance from
iterations 5 through 12.  Because of their low accuracy and high
degree of specialization, the parsers produced in these iterations had
little weight during voting and had little effect on the cumulative
decision making.

As in the bagging experiment, it appears that there would be more
precision and recall gain to be had by creating a larger ensemble.
Again, and more than in the bagging experiment, time and resource
constraints dictated our ensemble size.

In the table we see that the boosting algorithm equaled bagging's
test set gains in precision and recall.  The initial performance for
boosting was lower, though.  We cannot explain this, and expect it is
due to unfortunate resampling of the data during the first iteration
of boosting.  Exact sentence accuracy, though, was not significantly
improved on the test set.

Overall, we prefer bagging to boosting for this problem when raw
performance is the goal.  There are some side effects of boosting that
are useful in other respects, though, which we explore in Section
\ref{section:varying:qualitycontrol}.

\subsection{Effects of Varied Dataset Size}

To put the gains from bagging and boosting in perspective, as well as
to give a better understanding of the problem, we examined the effect
that varying training corpus size has on the performance of the
induced parser.  Since labelling a corpus is the largest source of
human labor (and consequently cost) required for building a supervised
stochastic parser, this is an important issue when porting parsing
techniques to new languages and new annotation schemes.

\subsubsection{Single Parser Training Curves}

We suspect our parser diversification techniques are better than just
adding more data to the training set.  While we cannot test this
fairly without hiring more annotators we can extrapolate from how well
the parser performs using various-sized training sets to decide how
well it will perform on new data.  If the effect of training size was
unpredictable or created training curves that are not smooth, then
this technique can say nothing either way about the question.  That is
not the case, however.  We see that performance increases toward an
asymptote that is far less than the performance increase we see from
bagging and boosting.

\begin{sidewaysfigure}[htbp]
  \centering
\fbox{
\begin{tabular}{rl}
    \epsfig{file=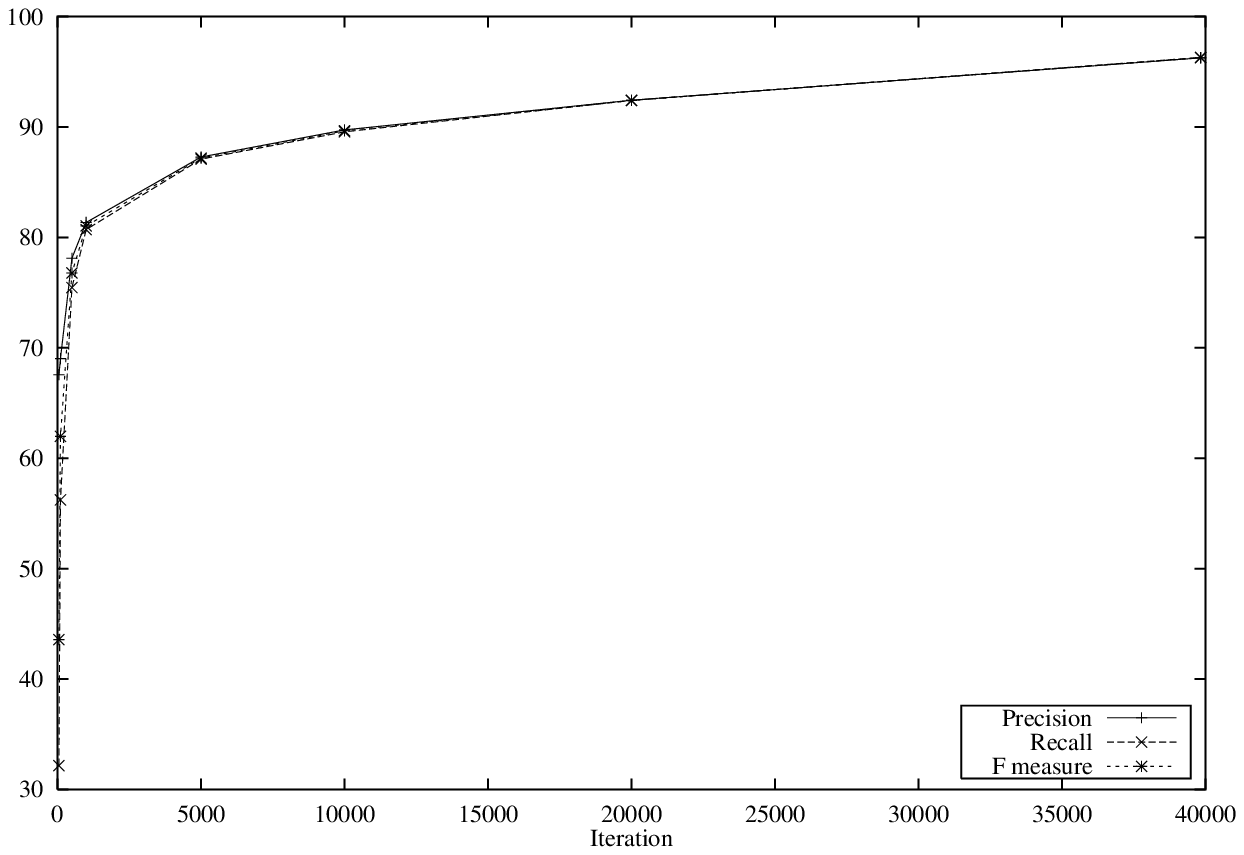, width= 0.49\textwidth}
&
    \epsfig{file=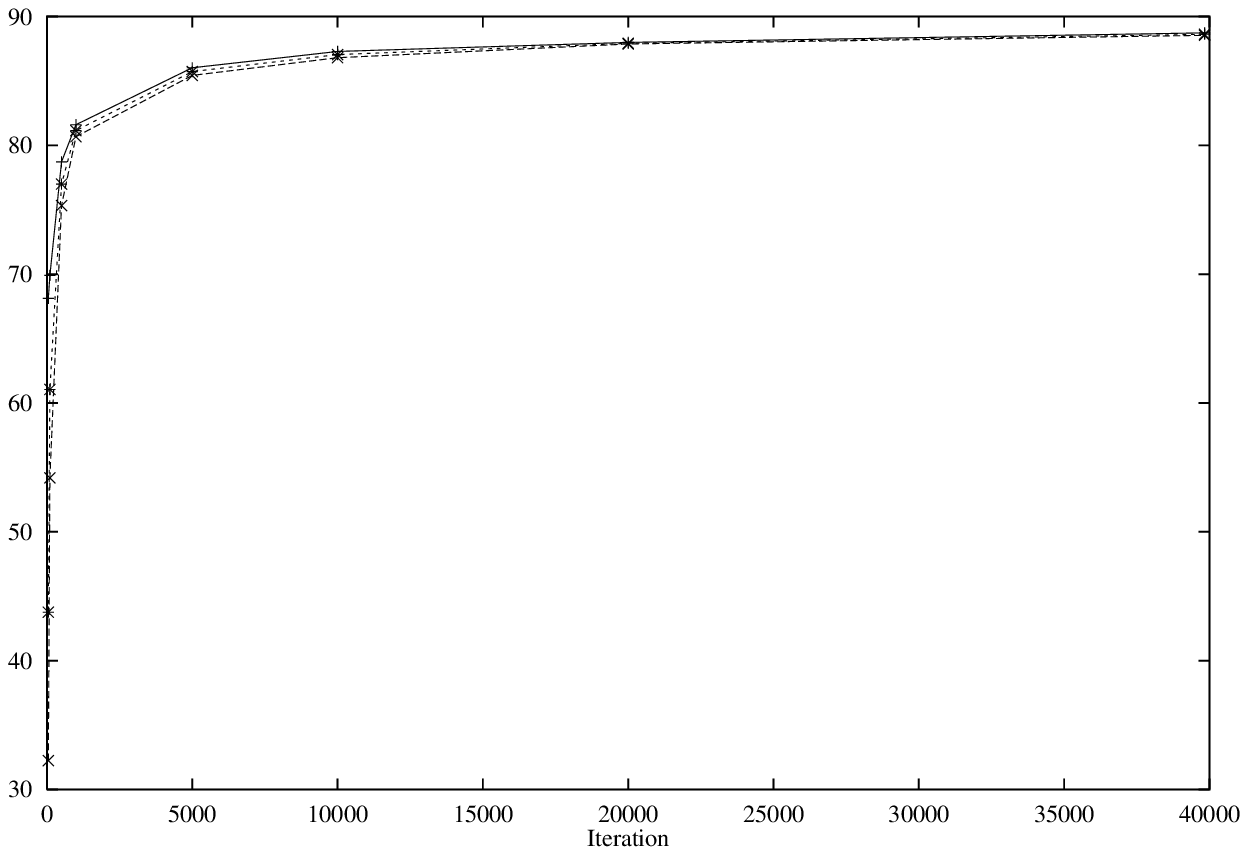, width= 0.49\textwidth}
\\
    \epsfig{file=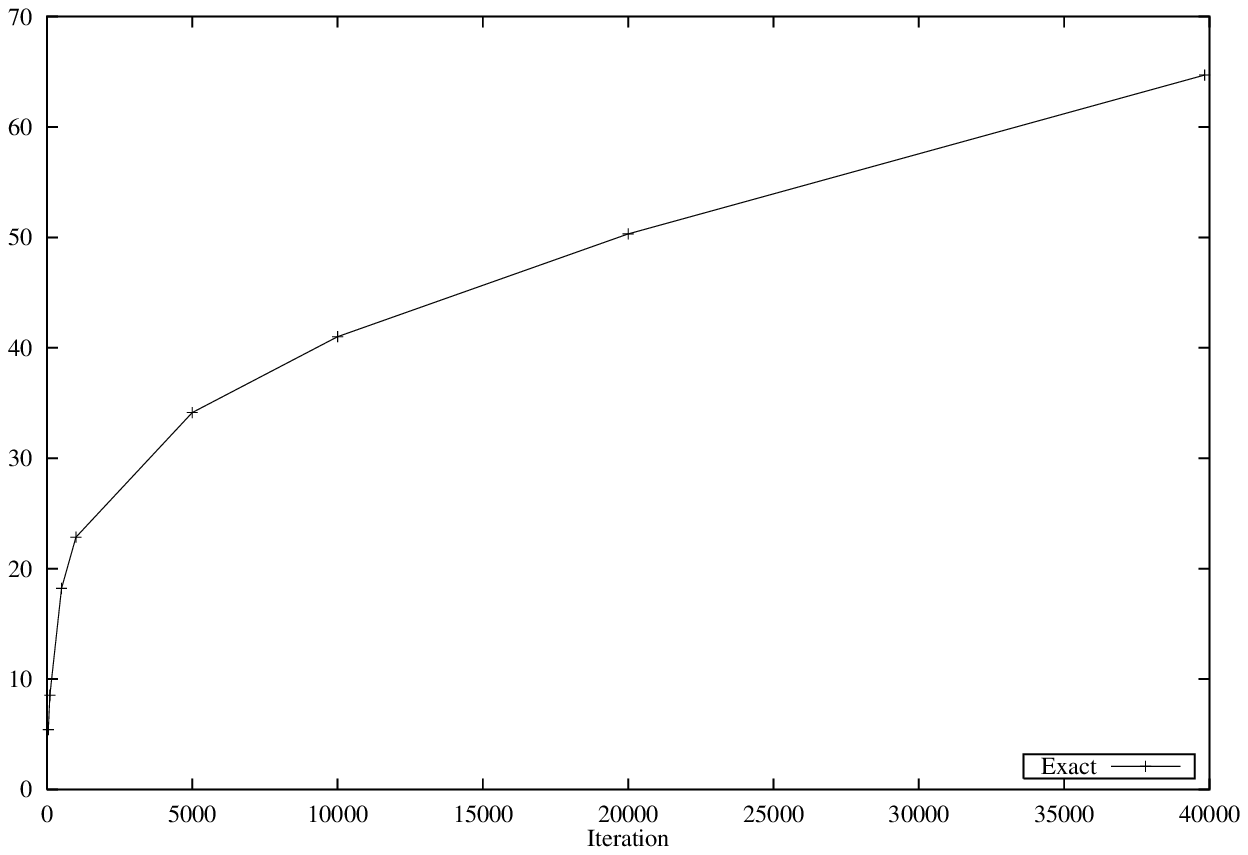, width= 0.49\textwidth}
&
    \epsfig{file=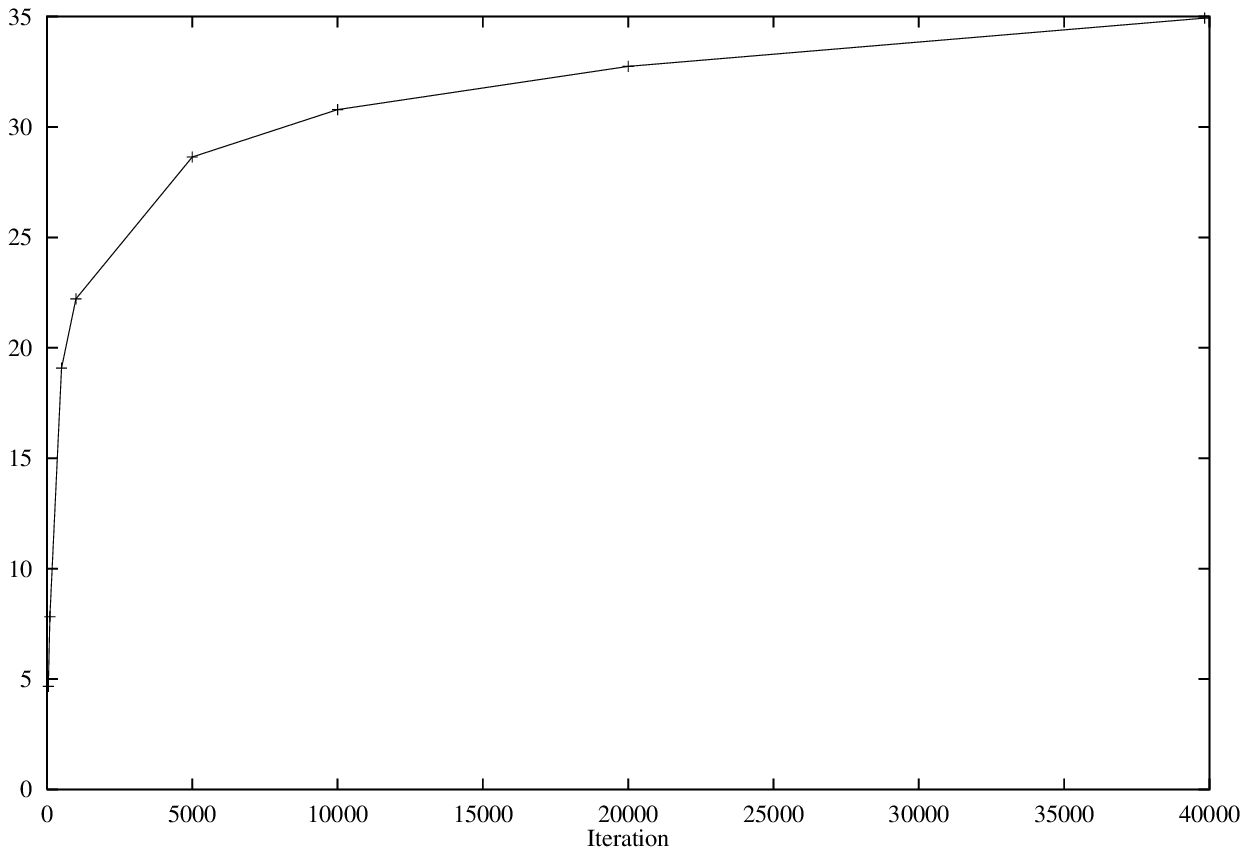, width= 0.49\textwidth}
\end{tabular}
}


  \caption{Effects of Varying Training Corpus Size} 
  \label{fig:varying:sizes}
\end{sidewaysfigure}

The training curves we present in Figure \ref{fig:varying:sizes} and
Table \ref{table:varying:sizes.train} suggest that roughly doubling
the corpus size in the quantity we are working with (10000-40000
sentences) gives a test set F-measure gain of approximately 0.70.
Bagging  achieved significant gains of approximately 0.60 over the best
reported previous F-measure without adding any new data.  In this
respect, these techniques show promise for making accuracy gains on
large corpora without adding more data or new parsers.  Boosting gave
a significant gain as well, but as we discussed it was subject to the
problems caused by a noisy corpus.

\begin{sidewaysfigure}[htbp]
  \centering
\fbox{
\begin{tabular}{rl}
    \epsfig{file=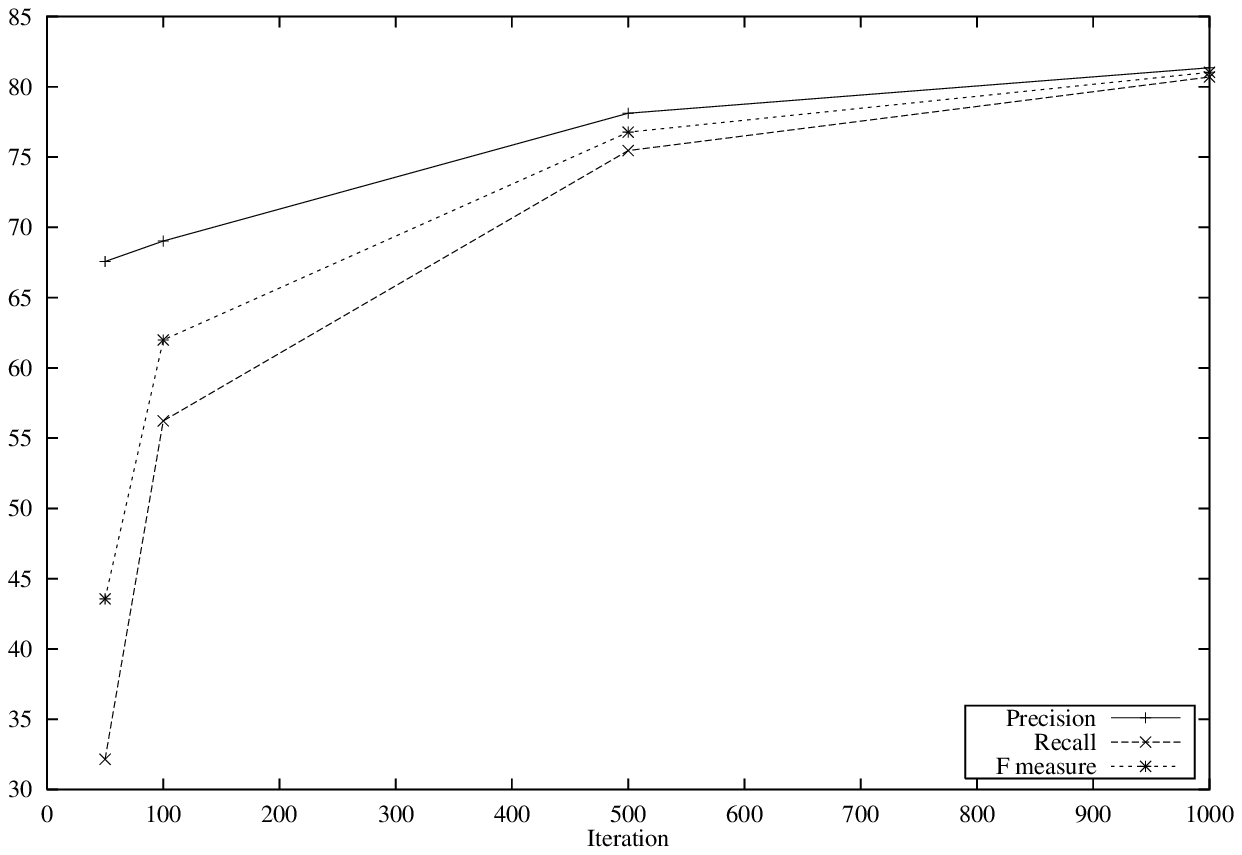, width= 0.49\textwidth}
&
    \epsfig{file=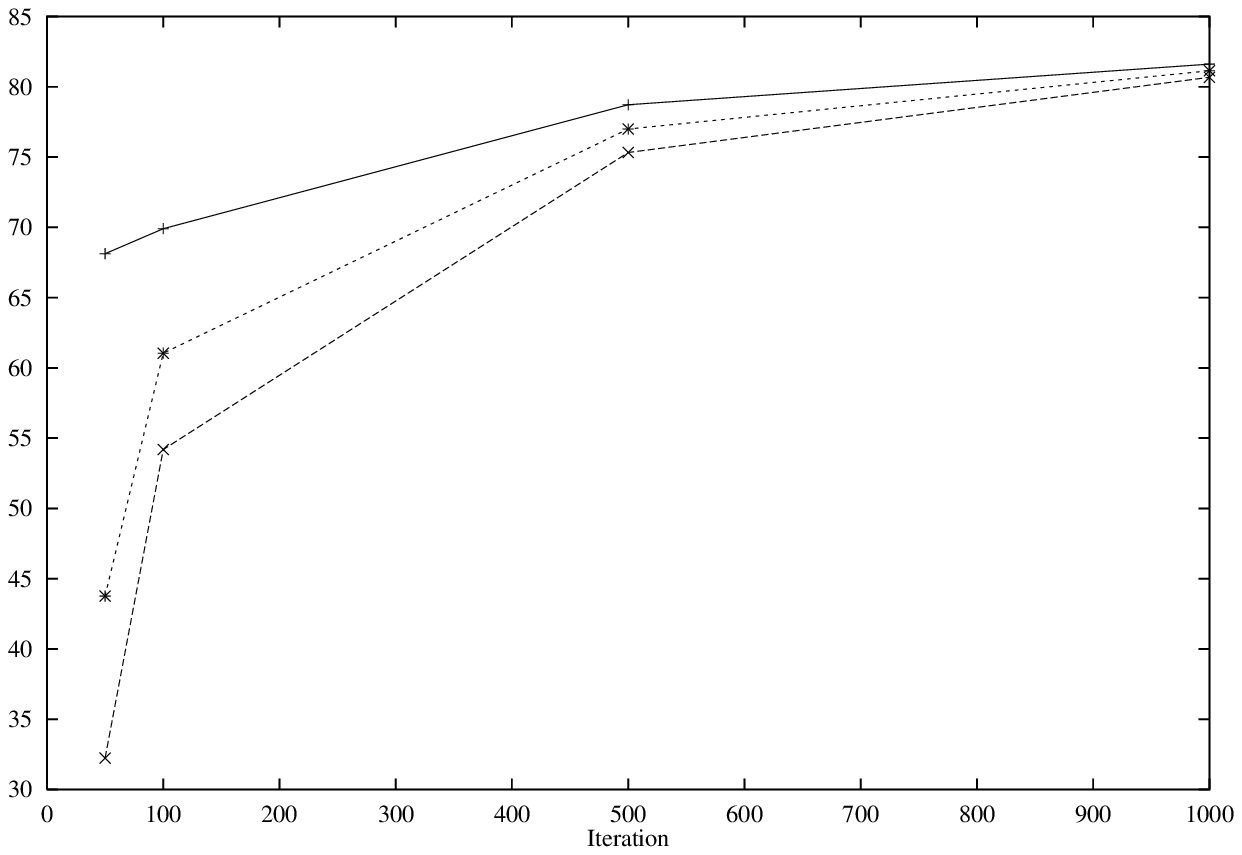, width= 0.49\textwidth}
\\
    \epsfig{file=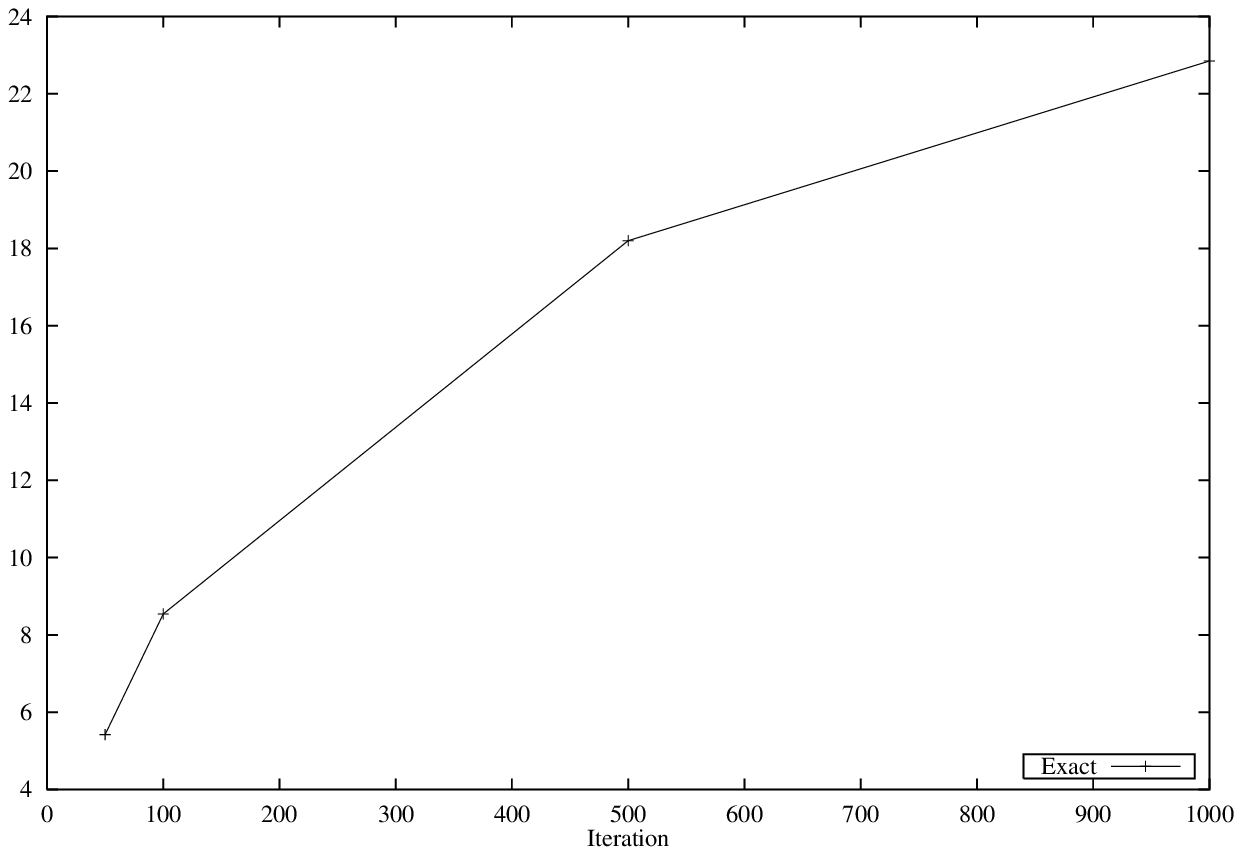, width= 0.49\textwidth}
&
    \epsfig{file=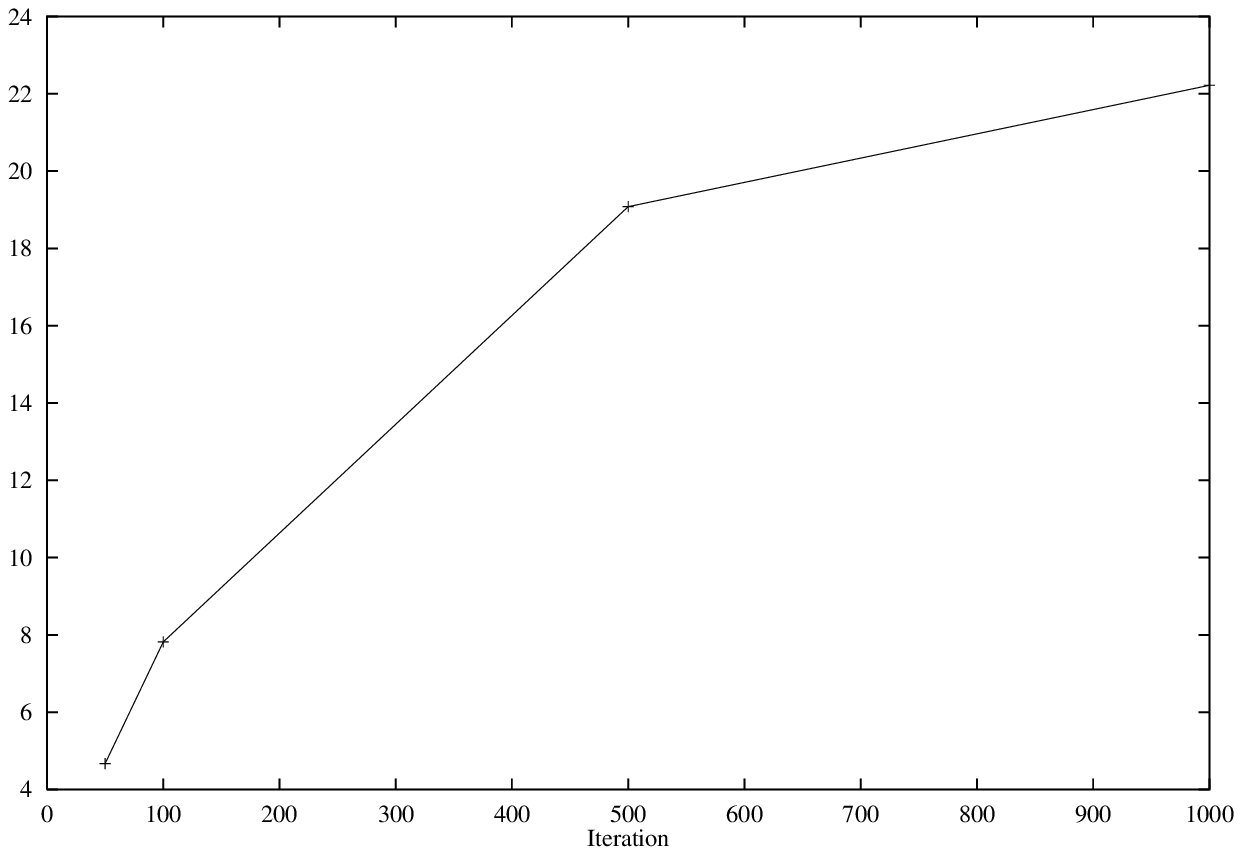, width= 0.49\textwidth}
\end{tabular}
}


  \caption{Effects of Varying Training Corpus Size (1000 Sentences and
    Less)} 
  \label{fig:varying:sizes.4}
\end{sidewaysfigure}

\begin{table}[htbp]
  \begin{center}
    \begin{tabular}{|c|l|rr|r|r|}
      \hline 
      Set
      &\multicolumn{1}{|c|}{Sentences}
      &\multicolumn{1}{c}{P} 
      & \multicolumn{1}{c|}{R} 
      & \multicolumn{1}{c|}{F}  
      & \multicolumn{1}{c|}{Exact}
      \\
      \hline 
      &50    & 67.57 & 32.15 & 43.57 &  5.4 \\
      &100   & 69.03 & 56.23 & 61.98 &  8.5 \\
      &500   & 78.12 & 75.46 & 76.77 & 18.2 \\
      &1000  & 81.36 & 80.70 & 81.03 & 22.9 \\
      &5000  & 87.28 & 87.09 & 87.19 & 34.1 \\
      &10000 & 89.74 & 89.56 & 89.65 & 41.0 \\
      &20000 & 92.42 & 92.40 & 92.41 & 50.3 \\
      \begin{rotate}{90}Training\end{rotate}
      &39832 & 96.25 & 96.31 & 96.28 & 64.7 \\
      \hline
      &50    & 68.13 & 32.24 & 43.76  &  4.7 \\
      &100   & 69.90 & 54.19 & 61.05  &  7.8 \\
      &500   & 78.72 & 75.33 & 76.99  & 19.1 \\
      &1000  & 81.61 & 80.68 & 81.14  & 22.2 \\
      &5000  & 86.03 & 85.43 & 85.73  & 28.6 \\
      &10000 & 87.29 & 86.81 & 87.05  & 30.8 \\
      &20000 & 87.99 & 87.87 & 87.93  & 32.7 \\
      \begin{rotate}{90}Testing\end{rotate}
      &39832 & 88.73 &   88.54 &   88.63  & 34.9 \\
      \hline 
    \end{tabular}
  \caption{Effects of Varying Training Corpus Size} 
  \label{table:varying:sizes.train}
  \end{center}
\end{table}


A second observation can be made about how well parsers can perform
based on small amounts of data.  They perform surprisingly well.
Looking at training set sizes of 1000 sentences and less gives us the
curves in \ref{fig:varying:sizes.4}.  These offer a suggestion as to
why the extremely complex parsing model of Hermjakob and Mooney
\cite{mooneyparse1997} can perform so well with such a small quantity
of data.  It was not previously known that Collins's parser (which
uses an extremely knowledge-impoverished model by comparison) performs
this well on such small amounts of training data.

\subsubsection{Data Loss from Selective Resampling}

We validated that the observations of Maclin
\cite{maclin98:boostregion} held in our boosting system.  The corpora
resampled during boosting contained many fewer types of sentences.  By
this we mean unique sentences.  Since the resampled version has no way
to suggest weights to the parser induction algorithm other than by
repeated insertion into the training set, the more skewed a
distribution becomes the fewer types of sentences are seen in it.

The importance of this is in the evaluation of novel events such as
novel words.  Head-passing parsers naturally rely on rare words seen
in the training corpus to predict how the words interact in novel
situations.  With few types of sentences in our training set, the
parsers will not have adequately informed models for novel words.


\begin{figure}[htbp]
  \centering
  \epsfig{file=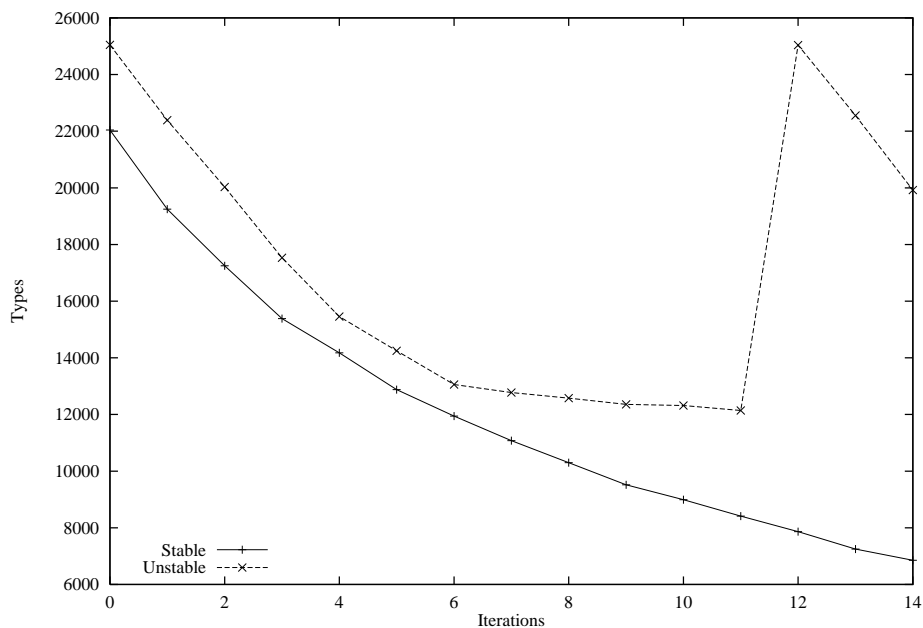, height= .4  \textheight}
  \caption{Unique Sentence Types During Boosting} 
  \label{fig:varying:types}
\end{figure}

In Figure \ref{fig:varying:types} we see the number of unique
sentences in the resampled corpora during boosting.  The upper curve
is the number of types in the corpora during boosting with back-off to
bagging.  We see that when boosting backs off at iteration 12 the
number of types returns to the original high value.  Prior to that the
value had been cut in half.  As we've seen earlier this suggests a hit
of approximately 0.70 units of F-measure based on corpus size
reduction alone.  The lower curve represents boosting operating on the
stable corpus, the one in which each sentence could be memorized.
The weak learner criterion holds much better here and the boosting
algorithm does not back off to restarting with a new bag.
Unfortunately though, in this case the types of sentences in the
training samples continues to dwindle as the parsers become
exceedingly specific.

As a comparison, the bagging corpora will consistently have the same
number of sentence types, and that number will match the leftmost
point on the curves shown.

There is no known general solution to this problem.

\section{Corpus Quality Control}
\label{section:varying:qualitycontrol}

In Appendix \ref{chapter:inconsistencies} we show some selected trees
from among the top 100 most heavily weighted trees at the end of
15 iterations of boosting the stable corpus.  In isolation, Collins's
parser is able to learn any one of these structures, but the presence
of conflicting information prevents it from getting 100\% accuracy on
the set.  

Boosting's sensitivity to noise is exemplified by these sentences, but
the side effect is that we can use the algorithm to perform quality
control.  We can weed out these sentences as we tried in one of our
experiments or suggest to the annotators that they be corrected.

There is one way in which these annotations might not be errors.
Parsers typically act by treating each sentence independently.  In
some cases differing interpretations can be assigned to one sentence
based on surrounding sentences alone.  Consider:

\vspace{1em}
\begin{tabular}{r|l}
  A &
 She left her binoculars at home.
\\
&
 She saw the boy with binoculars.
\\
\hline
  B &

 She saw the boy with binoculars.
\\
& He asked to borrow her binoculars, as he had none.
\end{tabular}
\vspace{1em}

\noindent
In situation A, we prefer to think that the boy must have the
binoculars, whereas in situation B we prefer to think that she used
the binoculars for seeing him.

Within the top 100 sentences, there were also some trees that did not
appear to have obvious problems.  There may be several causes for
these:

\begin{itemize}

\item
  
  The boosting algorithm was prematurely stopped.  It had not yet
  found a distribution that would suggest the parser does not possess
  the weak learner criterion.  These sentences could have fallen down
  in weight during the remaining boosting iterations.

\item

  The sentences may expose further inadequacies of the parser.

\item

  There could be an overabundance (majority) of incorrectly annotated
  trees further down in the ranking by weight that would prevent the
  parser from annotating the sentences correctly.

\end{itemize}

\section{Conclusions}

We have shown two method for automatically create ensembles of parsers
that produce better parses than any individual in the ensemble,
bagging and boosting.  We have studied several alternative
specializations of these algorithms as well.  None of the algorithms
exploit any specialized knowledge of the underlying parser induction
algorithm (weak learner), and we have restricted the data used in
creating the ensembles to a single training set to avoid issues of
training data quantity affecting the outcome.

Our best bagging system achieved consistently good performance on all
metrics, including exact sentence accuracy.  It resulted in a
statistically significant F-measure gain of 0.6 units in comparison to
the best previously known individual parsing result.

To put the gain in perspective, we studied the effect of training set
size on our underlying weak learner.  Reducing the corpus to one half
of it current size caused a loss of approximately 0.7 units of
F-measure, and reducing it to one quarter of its original size
resulted in a loss of 1.6 units of F-measure.  This allowed us to claim
that given the amount of training data we have, the bagging algorithm
is as effective at increasing F-measure as doubling the corpus size.
Clearly this is an incentive for utilizing this method, because even
though it is computationally expensive to create many parsers, the
cost is far outweighed by the opportunity cost of hiring humans to
annotate 40000 sentences.  With the ever increasing performance of
modern hardware, we expect the economic basis for using ensemble will
continue to improve.

The study of the effect of training set size on parser accuracy is in
itself a contribution of this chapter.  It is surprising (and
previously unknown) that 90\% of the accuracy of the parser we used
can be achieved by using a training set of only 1000 sentences.

Our boosting system fared well.  It also performed significantly
better than the best previously known individual parsing result.
However, it did not match the performance of bagging.  We suspect
several reasons for this, which we have explored and described.

We have shown how to exploit the distribution used in the boosting
algorithm to uncover inconsistencies in the corpus we are using.  We
have presented a semi-automated technique for doing this, as well as
many examples from the Treebank that are suspiciously inconsistent.
This can be used in many settings for cleaning corpora that are
suspected of having inconsistent annotations, especially when the
underlying phenomena in the corpus are not directly observable.  In
our case, the individual decisions made in parsing were not directly
observable in the corpus because matching their patterns is a
combinatorially expensive task.





\chapter{Conclusions}

In this thesis we have studied combination techniques for the task of
natural language parsing.  

The incentives for pursuing this work were threefold.  We wanted to
determine new bounds on the achievable performance for parsers trained
on the Penn Treebank corpus.  We wanted to compare how well automated
methods for varying parsers could fare compared to independent
research efforts. Finally, we wanted to explore the issues involved in
developing combination techniques for structured data instead of
simple classification.

Facilitation of this thesis came from the existence of multiple systems
which were all created to address the same task.   Technological
development naturally produces many systems in this way.  This was not
a unique situation.

We have shown that independent human research provides us with systems
that can be readily combined for large reductions in error.  We have
given supervised and unsupervised algorithms for the task, and
characterized the situations in which each are useful.

\section{Humans v. Machine}

In Chapter \ref{chapter:combining} we presented results on how well
the products of human research can be combined, and in Chapter
\ref{chapter:combining} we determined how well automated techniques
can produce diverse systems for combination.  A summary of our
findings are in Table \ref{table:conclusion:comparison}.  All of the
techniques we show performed significantly better than the best
individual parser.  The non-parametric systems for combining the
results of the human researchers gave far better results than the
automated methods for diversifying a single parser.  This is good news
for researchers, and generally not surprising.  Independent research
efforts produce more diverse systems than current automated
diversification algorithms can produce.

Remember that the boosting and bagging results were trying to produce
complementary systems, whereas the humans were simply producing
independent ones, only driving toward the goal of higher accuracy.
This shows that humans are still better at producing systems to be
combined \emph{by coincidence} than automated diversification tasks
can do by design.  Clearly there is plenty of room for progress on
producing better automated diversification algorithms.

\begin{table}[htbp]
  \begin{center}
    \begin{tabular}{|l|cc|c|c|}
      \hline 
      Reference / System 
      &\multicolumn{1}{c}{P} 
      & \multicolumn{1}{c|}{R} 
      & \multicolumn{1}{c|}{F} 
      & \multicolumn{1}{c|}{Exact} \\
      \hline 
      Average Individual Parser & 87.14 & 86.91 & 87.02 & 30.8\\
      Best Individual Parser    & 88.73 & 88.54 & 88.63 & 35.0\\
      \hline
      Distance Switching      & 90.24 & 89.58 & 89.91 & 38.0\\
      Constituent Voting      & 92.09 & 89.18 & 90.61 & 37.0\\
      \hline 
      \hline
      Boosting Initial        & 88.05 & 88.09 & 88.07 & 33.3 \\
      Boosting                & 89.37 & 88.32 & 88.84 & 33.0\\
      \hline
      Bagging Initial         & 88.43 & 88.34 & 88.38 & 33.3 \\
      Bagging                 & 89.54 & 88.80 & 89.17 & 34.6\\
      \hline
    \end{tabular}
    \caption{Comparison of Human and Automated Systems}
    \label{table:conclusion:comparison}
  \end{center}
\end{table}

\section{Future Work}

There are a few directions for future work based on this thesis.  The
three aspects: bound on achievable result for parsing, combining
techniques for independently produced systems, and automated
diversification of a single system are each tasks that can be extended
in useful ways.

It would be good service to the community to keep the bounds that were
derived in this work current.  As new parsers appear the bound should
be reevaluated to determine how much progress is being made on parsing
as a task.  Also, parsers are starting to produce more complicated
linguistic structure such as traces.  They should be incorporated into
the combining task as the become prevalent.  Maintenance of this sort
is a service that can be provided at little cost if the parsers can be
trained from data and the individual experimenters are careful not to
train on different parts of the corpus.

There are other natural language processing tasks that could
potentially benefit from more attempts at combining systems addressing
them. Anaphora resolution and coreference are two tasks that are
maturing to the point where there are multiple independent systems
addressing them.  Tasks introduced in the Message Understanding
Conferences (MUC) evaluations all have multiple systems available, and
they could be combined for better bounds for those tasks.  Data
sparseness is an issue there, though.  The systems used in those tasks
have not had controlled training and test datasets, though.  Some of
the systems have been trained on an order of magnitude more data than
others.  This is an issue that will need to be addressed.

The structural difference of those systems will present unique
challenges as well.  Parsing and machine translation
\cite{mt:threeheads} have both required some effort to deal with the
interdependence of predictions, and determining what substructures can
be combined with voting.  Work in natural language processing is
moving toward more structural data such as these.

The result given in this chapter is that algorithms are not currently
available for building diverse systems that are as independent as the
systems created by humans.  This points at what is probably the most
open problem: creating machine learning algorithms that can produce
systems as diverse as human creativity.  This may look like an
AI-complete problem, but working on a specific task and leveraging on
previous work in the field could make it possible.

Automating the process of creating diverse induction systems is a goal
along the path in pursuit of the larger task we described,
automating the process of scientific inference, experimentation, and
discovery.  Continuing to produce (and understand) systems that are
capable of utilizing other systems will produce results that will be
directly applicable to the larger task.

A continuing problem in experimentation in automating diversification
of state of the art systems is that computational resources (time and
space) are in more demand than in developing or utilizing a single
system.  In particular, the iterative process of boosting was
expensive and required the creation of a parallel distributed
implementation of the underlying parser in order to make it feasible.
There are systems issues here that could be addressed, and 
practical concerns will limit how complex the automated
diversification algorithms can become.

\bigskip

We hope this thesis has shown that there are interesting issues
involved in combining systems that induce structural linguistic
annotation.  Furthermore, we have shown that the task is useful in
providing bounds on achievable performance as well as achieving better
performance.  We have shown also that side effects of diversification
can be used to find questionable corpus annotation.  We hope we have
started a line of research that will continue to be pursued and
continue to prove itself fruitful.


\begin{appendix}
\chapter{Treebank Inconsistencies}
\label{chapter:inconsistencies}


\begin{parse}
\hspace*{-\fill}
\Tree
[.TOP [.NP [.NNS Fees ]  [.CD 1 ]  [.CD 7/8 ]  ]  ] 
\Tree
[.TOP [.NP [.NP [.NNS Fees ]  ]  [.NP [.QP [.CD 1 ]  [.CD 7/8 ]  ]  ]  ]  ] 
\Tree
[.TOP [.NP [.NP [.NNS Fees ]  ]  [.NP [.CD 1 ]  [.CD 7/8 ]  ]  ]  ] 
\Tree
[.TOP [.NP [.NNS Fees ]  [.CD 1 ]  [.CD 3/8 ]  ]  ] 
\Tree
[.TOP [.NP [.NNS Fees ]  [.QP [.CD 1 ]  [.CD 3/8 ]  ]  ]  ] 
\Tree
[.TOP [.NP [.NP [.NNS Fees ]  ]  [.NP [.CD 1 ]  [.CD 3/8 ]  ]  ]  ] 
\Tree
[.TOP{\footnotemark}
 [.NP [.NNS Fees ]  [.QP [.CD 1 ]  [.CD 7/8 ]  ]  ]  ] 
\footnotetext{4 copies of this tree appeared in the set.}
\end{parse} 

\newpage

\begin{parse}
\Tree
[.TOP [.PP [.IN In ]  [.NP [.JJ other ]  [.NN commodity ]  [.NNS
markets ]  [.NN yesterday ]  ]  ]  ]  
\end{parse} 

\begin{parse}
\Tree
[.TOP [.UCP [.PP [.IN In ]  [.NP [.JJ other ]  [.NN commodity ]  [.NNS
markets ]  ]  ]  [.NP [.NN yesterday ]  ]  ]  ]  
\end{parse} 

\begin{parse}
\Tree
[.TOP{\footnotemark}
 [.FRAG [.PP [.IN In ]  [.NP [.JJ other ]  [.NN commodity ]
[.NNS markets ]  ]  ]  [.NP [.NN yesterday ]  ]  ]  ]  
\footnotetext{5 copies.}
\end{parse} 

\newpage

\begin{parse}
\Tree
[.TOP [.NP [.NP [.NN Source ]  ]  [.NP [.NNP Fulton ]  [.NNP Prebon ]
[.PRN [.( ( ]  [.NAC [.NNP U.S.A ]  ]   [.) ) ]  ]
[.NNP Inc ]  ]  ]  ]  
\end{parse} 

\begin{parse}
\Tree
[.TOP [.NP [.NP [.NN Source ]  ]  [.NP [.NNP Fulton ]  [.NNP Prebon ]
[.PRN [.( ( ]  [.NP [.NN U.S.A ]  ]  [.) ) ]  ]  [.NNP
Inc ]  ]  ]  ]  
\end{parse} 

\begin{parse}
\Tree
[.TOP [.NP [.NP [.NN Source ]  ]  [.NP [.NNP Fulton ]  [.NNP Prebon ]
[.PRN [.( ( ]  [.NP [.NNP U.S.A ]  ]  [.) ) ]  ]
[.NNP Inc ]  ]  ]  ]  
\end{parse}

\newpage

\begin{sideways}
\begin{parse}
\scalebox{.8}{
\Tree [.TOP [.NP [.NP [.NP [.NP [.JJ
Annualized ]  [.JJ average ]  [.NN rate
]  ]  [.PP [.IN of ]  [.NP [.NN return ]  ]  ]  ]  [.PP [.IN after ]
[.NP [.NNS expenses ]  ]  ]  [.PP [.IN for ]  [.NP [.DT the ]  [.JJ
past ]  [.CD 30 ]  [.NNS days ]  ]  ]  ]  [.NP [.NP [.RB not ]  [.DT a
]  [.NN forecast ]  ]  [.PP [.IN of ]  [.NP [.JJ future ]  [.NNS
returns ]  ]  ]  ]  ]  ] 
}
\end{parse}
\end{sideways}

\begin{sideways}
\begin{parse}
\scalebox{.8}{
\Tree
[.TOP [.NP [.NP [.NP [.JJ Annualized ]  [.JJ average ]  [.NN rate ]  ]
[.PP [.IN of ]  [.NP [.NN return ]  ]  ]  [.PP [.IN after ]  [.NP
[.NNS expenses ]  ]  ]  [.PP [.IN for ]  [.NP [.DT the ]  [.JJ past ]
[.CD 30 ]  [.NNS days ]  ]  ]  ]  [.NP [.NP [.RB not ]  [.DT a ]  [.NN
forecast ]  ]  [.PP [.IN of ]  [.NP [.JJ future ]  [.NNS returns ]  ]
]  ]  ]  ]  
}
\end{parse}
\end{sideways}

\begin{sideways}
\begin{parse}
\scalebox{.8}{
\Tree
[.TOP [.NP [.NP [.NP [.VBN Annualized ]  [.JJ average ]  [.NN rate ]
]  [.PP [.IN of ]  [.NP [.NN return ]  ]  ]  [.PP [.IN after ]  [.NP
[.NNS expenses ]  ]  ]  [.PP [.IN for ]  [.NP [.DT the ]  [.JJ past ]
[.CD 30 ]  [.NNS days ]  ]  ]  ]  [.NP [.NP [.RB not ]  [.DT a ]  [.NN
forecast ]  ]  [.PP [.IN of ]  [.NP [.JJ future ]  [.NNS returns ]  ]
]  ]  ]  ]  
}
\end{parse}
\end{sideways}

\begin{sideways}
\begin{parse}
\scalebox{.7}{
\Tree
[.TOP [.FRAG [.NP [.NP [.JJ Annualized ]  [.JJ average ]  [.NN rate ]
]  [.PP [.IN of ]  [.NP [.NP [.NN return ]  ]  [.PP [.IN after ]  [.NP
[.NNS expenses ]  ]  ]  ]  ]  [.PP [.IN for ]  [.NP [.DT the ]  [.JJ
past ]  [.CD 30 ]  [.NNS days ]  ]  ]  ]  [.RB not ]  [.NP [.NP [.DT a
]  [.NN forecast ]  ]  [.PP [.IN of ]  [.NP [.JJ future ]  [.NNS
returns ]  ]  ]  ]  ]  ]  
}
\end{parse}
\end{sideways}

\begin{sideways}
\begin{parse}
\scalebox{.8}{
\Tree
[.TOP [.NP [.NP [.NP [.JJ Annualized ]  [.JJ average ]  [.NN rate ]  ]
[.PP [.IN of ]  [.NP [.NN return ]  ]  ]  [.PP [.IN after ]  [.NP [.NP
[.NNS expenses ]  ]  [.PP [.IN for ]  [.NP [.DT the ]  [.JJ past ]
[.CD 30 ]  [.NNS days ]  ]  ]  ]  ]  ]  [.NP [.NP [.RB not ]  [.DT a ]
[.NN forecast ]  ]  [.PP [.IN of ]  [.NP [.JJ future ]  [.NNS returns
]  ]  ]  ]  ]  ]  
}
\end{parse}
\end{sideways}

\begin{sideways}
\begin{parse}
\scalebox{.8}{
\Tree
[.TOP [.NP [.NP [.NP [.VBN Annualized ]  [.JJ average ]  [.NN rate ]
]  [.PP [.IN of ]  [.NP [.NN return ]  ]  ]  [.PP [.IN after ]  [.NP
[.NP [.NNS expenses ]  ]  [.PP [.IN for ]  [.NP [.DT the ]  [.JJ past
]  [.CD 30 ]  [.NNS days ]  ]  ]  ]  ]  ]  [.NP [.NP [.RB not ]  [.DT
a ]  [.NN forecast ]  ]  [.PP [.IN of ]  [.NP [.JJ future ]  [.NNS
returns ]  ]  ]  ]  ]  ]  
}
\end{parse}
\end{sideways}

\newpage

\begin{sideways}
\begin{parse}
\scalebox{.7}{
\Tree
[.TOP [.NP [.NP [.JJ FEDERAL ]  [.NNS FUNDS ]  ]  [.NP [.NP [.ADJP [.CD 9 ]  [.NN (\%) ]  ]  [.JJ high ]  ]  [.NP [.ADJP [.QP [.CD 8 ]  [.CD 13/16 ]  ]  [.NN (\%) ]  ]  [.JJ low ]  ]  [.NP [.ADJP [.QP [.CD 8 ]  [.CD 7/8 ]  ]  [.NN (\%) ]  ]  [.ADJP [.IN near ]  [.NN closing ]  ]  [.NN bid ]  ]  [.NP [.NP [.QP [.CD 8 ]  [.CD 15/16 ]  ]  [.NN (\%) ]  ]  [.VP [.VBN offered ]  ]  ]  ]  ]  ] 
}
\end{parse}
\end{sideways}

\begin{sideways}
\begin{parse}
\scalebox{.7}{
\Tree
[.TOP [.NP [.NP [.JJ FEDERAL ]  [.NNS FUNDS ]  ]  [.NP [.NP [.NP [.QP [.CD 8 ]  [.CD 3/4 ]  ]  [.NN (\%) ]  ]  [.ADJP [.JJ high ]  ]  ]  [.NP [.NP [.QP [.CD 8 ]  [.CD 5/8 ]  ]  [.NN (\%) ]  ]  [.ADJP [.JJ low ]  ]  ]  [.NP [.NP [.QP [.CD 8 ]  [.CD 11/16 ]  ]  [.NN (\%) ]  ]  [.PP [.IN near ]  [.NP [.NN closing ]  ]  ]  [.VP [.NN bid ]  ]  ]  [.NP [.NP [.QP [.CD 8 ]  [.CD 11/16 ]  ]  [.NN (\%) ]  ]  [.VP [.VBN offered ]  ]  ]  ]  ]  ] 
}
\end{parse}
\end{sideways}

\begin{sideways}
\begin{parse}
\scalebox{.7}{
\Tree
[.TOP [.NP [.NP [.JJ FEDERAL ]  [.NNS FUNDS ]  ]  [.NP [.NP [.NP [.CD 8 ]  [.CD 3/4 ]  [.NN (\%) ]  ]  [.ADJP [.JJ high ]  ]  ]  [.NP [.NP [.CD 8 ]  [.CD 5/8 ]  [.NN (\%) ]  ]  [.ADJP [.JJ low ]  ]  ]  [.NP [.NP [.CD 8 ]  [.CD 11/16 ]  [.NN (\%) ]  ]  [.PP [.IN near ]  [.NP [.NN closing ]  [.NN bid ]  ]  ]  ]  [.NP [.NP [.CD 8 ]  [.CD 3/4 ]  [.NN (\%) ]  ]  [.VP [.VBN offered ]  ]  ]  ]  ]  ] 
}
\end{parse}
\end{sideways}

\newpage

\begin{parse}
\hspace{-1.25 in}
\scalebox{.9}{
\Tree
[.TOP [.S [.NP [.NNP Freddie ]  [.NNP Mac ]  [.ADJP [.CD 9 ]  [.NN
(\%) ]  ]  [.NNS securities ]  ]  [.VP [.VBD were ]  [.PP [.IN at ]
[.NP [.CD 97 ]  [.CD 21/32 ]  ]  ]  [.ADVP [.IN up ]  [.NP [.CD 5/32 ]
]  ]  ]  ]  ]  
}
\end{parse}

\begin{parse}
\hspace{-1.25 in}
\scalebox{.9}{
\Tree
[.TOP [.S [.NP [.NNP Freddie ]  [.NNP Mac ]  [.ADJP [.CD 9 ]  [.NN
(\%) ]  ]  [.NNS securities ]  ]  [.VP [.VBD were ]  [.PP [.IN at ]
[.NP [.QP [.CD 97 ]  [.CD 4/32 ]  ]  ]  ]  [.ADVP [.RB down ]  [.NP
[.CD 1/32 ]  ]  ]  ]  ]  ]
}
\end{parse}


\end{appendix}
\bibliographystyle{abbrv}
\bibliography{thesis}


\begin{vita} 

John Henderson was born and raised near Mechanicsburg, Pennsylvania, a
small town in the U.S.A.  He received a B.S. in Mathematics/Computer
Science from Carnegie Mellon University in 1994, and an M.S.E. in
Computer Science from the Johns Hopkins University in 1997. After
finishing his Ph.D. dissertation in 1999, he joined the Intelligent
Information Access group at the MITRE Corporation in Bedford,
Massachusetts to continue research in data-driven natural language
processing.

\end{vita}


\end{document}